\documentclass{article}

\usepackage[numbers]{natbib}
\usepackage[preprint]{nips_2018}

\usepackage{microtype}
\usepackage{graphicx}
\usepackage{booktabs} 

\usepackage{hyperref}
\usepackage{algorithm}
\usepackage{algorithmic}

\usepackage[utf8]{inputenc} 
\usepackage[T1]{fontenc}    
\usepackage{hyperref}       
\usepackage{url}            
\usepackage{breakurl}
\usepackage{booktabs}       
\usepackage{amsfonts}       
\usepackage{nicefrac}       
\usepackage{microtype}      

\usepackage{epsfig}
\usepackage{epstopdf}
\usepackage{amssymb}
\usepackage{amsmath}
\usepackage{amsthm}
\usepackage{amsfonts}
\usepackage{bbding}
\usepackage{array}
\usepackage{caption,tabularx,booktabs}
\usepackage{arydshln}
\usepackage{xargs}                      
\usepackage[pdftex,dvipsnames]{xcolor}  
\usepackage{enumitem}

\usepackage{thmtools} 
\usepackage{thm-restate}

\usepackage{enumitem}
\newcolumntype{C}[1]{>{\centering\let\newline\\\arraybackslash\hspace{0pt}}m{#1}}

\usepackage{subcaption}
\usepackage[noabbrev]{cleveref}
\usepackage{wrapfig}

\newcommand\tagthis{\addtocounter{equation}{1}\tag{\theequation}}
\DeclareMathOperator{\Exp}{\mathbb{E}}           
\DeclareMathOperator{\R}{\mathbb{R}} 
\DeclareMathOperator{\Ocal}{\mathcal{O}} 
\newcommand{\eqdef}{\stackrel{\text{def}}{=}}
\newcommand{\setn}{[n]}

\newtheorem{thm}{Theorem}
\newtheorem{lem}{Lemma}

\newtheorem{remark}{Remark}

\newcommand{\x}{w}
\renewcommand{\a}{x}
\renewcommand{\b}{z}

\usepackage{titlesec}
\titlespacing\section{0pt}{2pt plus 1pt minus 1pt}{0pt plus 1pt minus 1pt}
\titlespacing\subsection{0pt}{2pt plus 1pt minus 1pt}{0pt plus 1pt minus 1pt}
\titlespacing\subsubsection{0pt}{2pt plus 1pt minus 1pt}{0pt plus 1pt minus 1pt}

\usepackage{newfloat}
\DeclareFloatingEnvironment[
    fileext=los,
    listname={List of Schemes},
    name=Algorithm,
    placement=tbhp,
]{scheme}

\usepackage[colorinlistoftodos,prependcaption,textsize=tiny]{todonotes}

\title{On the Acceleration of L-BFGS with Second-Order Information 
and Stochastic Batches}

\author{Jie Liu$^{1,2}$\quad Yu Rong$^1$\quad Martin Tak\'{a}\v{c}$^2$\quad Junzhou Huang$^1$ \\
  $^1$Tencent AI Lab\quad 
  $^2$Lehigh University\\
  \texttt{jild13@lehigh.edu,
  royrong@tencent.com,}
   \texttt{Takac.MT@gmail.com,
    joehhuang@tencent.com} \\
}

\begin{document}

\maketitle

\begin{abstract}
This paper proposes a framework of L-BFGS based on the (approximate) second-order information with stochastic batches, as a novel approach to the finite-sum minimization problems. Different from the classical L-BFGS where stochastic batches lead to instability, we use a smooth estimate for the evaluations of the gradient differences while achieving acceleration by well-scaling the initial Hessians. We provide theoretical analyses for both convex and nonconvex cases. 
In addition, we demonstrate that within the popular applications of least-square and cross-entropy losses, the algorithm admits a simple implementation in the distributed environment. Numerical experiments support the efficiency of our algorithms.
\end{abstract}

\section{Introduction}
\label{submission}

We consider the finite-sum minimization problem of the form
\begin{equation}\label{eq:problem}
\min_{\x\in\R^d} \Big\{F(\x):= \tfrac{1}{n} {\textstyle\sum}_{i\in\setn}  f(\x;\a_i,\b_i)\eqdef\tfrac{1}{n} {\textstyle\sum}_{i\in\setn} f_i(\x) \Big\},
\end{equation}
where $i \in \setn\eqdef\{1,\dots,n\}$, and $\{(\a_i, \b_i)\}_{i=1}^{i=n}$ are the data pairs. Throughout the paper, we assume there exists a global optimal solution $\x^{*}$ of \eqref{eq:problem}; in other words, we have a lower bound $F(\x^{*})$ of \eqref{eq:problem}. 

In general, the problem of form~\eqref{eq:problem} covers a wide range of convex and nonconvex problems including logistic regression~\cite{LR}, multi-kernel learning\cite{MKL, MKL1}, conditional random fields~\cite{CRF}, neural networks~\cite{ESL}, etc. Classical first-order methods to solve \eqref{eq:problem} are gradient descent (GD)~\cite{Nocedal2006NO} and stochastic gradient descent (SGD)~\cite{RM1951,pegasos}. A large class of optimization methods can be used to solve \eqref{eq:problem}, where the iterative updates can be generalized as follows,
\begin{equation}\label{eq:update}
\x_{k+1} = \x_k + \alpha_k p_k, \text{with } p_k = - H_k g_k,
\end{equation}
where $p_k$ is some descent direction, $H_k$ is an inverse Hessian approximation of $F$ at $\x_k$, and $g_k$ is an estimate of $\nabla F(\x_k)$.

When $H_k$ is an identity matrix, the update is considered a first-order method. Numerous work has focused on the choice of $g_k$ such as SAG/SAGA~\cite{SAGjournal,SAGA}, MISO/FINITO~\cite{MISO,FINITO}, SDCA~\cite{SDCA}, SVRG/S2GD~\cite{SVRG,konecny2015mini}, SARAH~\cite{nguyen2017sarah, SARAHnonconvex}. 
Nevertheless, with the importance of second-order optimization providing potential curvature around local optima and thus promoting fast convergence, the choice of non-identity $H_k$ is crucial to the development of modern optimization algorithms.

Within the framework of second-order optimization, a popular choice for $H_k$ is the inverse Hessian; however, we lack an efficient way to invert matrices, leading to increases in computation and communication costs to a problem for the distributed setting. Motivated by this, quasi-Newton methods, among which BFGS is one of the most popular, were developed, including a practical variant named limited-memory BFGS (L-BFGS)~\cite{Nocedal2006NO}. It has been widely known that batch methods have been successfully applied in first-order algorithms and provide effective improvements, but it remains a problem for L-BFGS due to the instability caused by randomness between different gradient evaluations. Therefore, the development of an efficient and stable L-BFGS is necessary. 

\textbf{Our contributions.} In this paper, we analyze L-BFGS with stochastic batches for both convex and nonconvex optimization, as well as its distributed implementation. LBFGS-H originates from the idea of L-BFGS and uses Hessian information to approximate the differences of gradients. LBFGS-F combines L-BFGS with Fisher information matrix from the natural gradient algorithm~\cite{Amari:1998:NGW:287476.287477, martens2014new, PascanuNatural14}. We show that they are efficient for minimizing finite-sum problems both in theory and in practice. The key contributions of our paper are summarized as follows.
\begin{itemize}[noitemsep,nolistsep]
\item We propose a framework for approximating the differences of gradients in the L-BFGS algorithm that ensures stability for the general finite-sum problems. We show it converges linearly to a neighborhood of the optimal solution for convex and nonconvex finite-sum problems under standard assumptions~\cite{MB-LBFGS}.
\item In a distributed environment, we introduce a variant LBFGS-F where the Hessian matrix for approximating gradient differences is replaced by the Fisher information matrix~\cite{martens2014new}. 
\item With a potential acceleration in practice using ADAM techniques~\cite{ADAM}, we verify the competitive performances of both LBFGS-H and LBFGS-F against mainstream optimization methods in both convex and nonconvex applications.
\end{itemize}

\section{The Algorithm}
In this section, we propose a new stochastic L-BFGS framework, as well as its distributed implementation with Fisher information matrix. Before proceeding to the new algorithm, let us revisit the procedure for the classical L-BFGS.

\subsection{Limited-memory BFGS}

The classical L-BFGS algorithm~\cite{Nocedal2006NO} is presented as below.

\begin{scheme}[H]
\caption{L-BFGS }
   \fbox{\begin{subfigure}[t]{0.62\columnwidth}
   \caption{Algorithm LBFGS}
   \label{LBFGS}
\begin{algorithmic}
   \STATE {\bfseries Initialize:} $x_0$, integer $m>0$
   \FOR{$k=1,2,\dots$}
   \STATE Choose $H_k^0$
   \STATE Compute a direction $p_k = -H_k \nabla f(\x_k)$ by Algorithm~\ref{LBFGSloop}
   \STATE Choose a learning rate $\alpha_k>0$
   \STATE Update the iterate: $\x_{k+1} = \x_k + \alpha_k p_k$ 
   \STATE Update the curvature pairs: 
   \STATE $\ \qquad s_k = \x_{k+1} - \x_k, y_k = \nabla F(\x_{k+1}) - \nabla F(\x_k)$
   \IF{$k\geq m$}
   \STATE Replace the oldest pair $(s_{k-m}, y_{k-m})$ by $(s_{k}, y_k)$
   \ELSE
   \STATE Store the vector pair $(s_{k}, y_k)$
   \ENDIF
   \ENDFOR
\end{algorithmic}
\end{subfigure}}
%
%
   \fbox{\begin{subfigure}[t]{0.38\columnwidth}
    \caption{Two-loop Recursion}
    \label{LBFGSloop}
\begin{algorithmic}
   \STATE $q = -g_k, \rho_i: = \tfrac{1}{y_i^Ts_i}$
   \FOR{$i=k-1$ to $k-m$}
   \STATE $\alpha_i = \rho_is_i^Tq$
   \STATE $q = q - \alpha_i y_i$
   \ENDFOR
   \STATE $r = H_k^0q$
   \FOR{$i=k-m$ to $k-1$}
   \STATE $\beta = \rho_i y_i^T r$
   \STATE $r = r +s_i(\alpha_i - \beta)$
   \ENDFOR
   \STATE \textbf{stop} with result $r=-H_k\nabla F(\x_k) $
\end{algorithmic}
\end{subfigure}}
\end{scheme}

In each iteration, first, we estimate the direction by using curvature pairs $\{(s_i, y_i)\}_{k-m\leq i\leq k-1}$. Then, the learning rate is chosen such that certain condition (e.g. line search) is satisfied, and we make an update. Last, we evaluate the curvature pairs $(s_k, y_k)$ and replace the pairs stored in the memory while keeping the number of curvature pairs no larger than $m$. The key step in this procedure is the evaluation of the search direction $p_k$ using the curvature pairs, which appears as the well-known two-loop recursion (Algorithm~\ref{LBFGSloop}). Note that in the classical L-BFGS, the main algorithm usually applies a line-search technique for choosing the learning rate $\alpha_k>0$.

The intrinsic idea within L-BFGS is to utilize the curvature information implied by the vector pairs $(s_k, y_k)$ to help regularize the gradient direction. However, within the setting of stochastic batches, the update of $y_{k} = \nabla F^{S_{k+1}}(\x_{k+1}) - \nabla F^{S_k}(\x_k)$, where the batch gradient is defined as
\begin{equation*}
\nabla F^{S_k} : = \tfrac{1}{|S_k|} {\textstyle\sum}_{i\in S_k} \nabla f_i(\x_k),
\end{equation*}
makes it difficult to stabilize the behavior of the algorithm. One of the remedies is to assume that there is an overlap between the samples $S_k$ and $S_{k+1}$, i.e., $S_k\cap S_{k+1}  = O_k \neq \emptyset$, and replace the $S_k, S_{k+1}$ with $O_k$ in $y_k$~\cite{MB-LBFGS}. However, this idea requires the batch size to be large enough.

Recall from the Taylor expansion for a multivariate vector-valued function $\mathbf{g}(x) = [g_1(x), \dots, g_d],$
\begin{equation*}
\mathbf{g}(\x_{k+1}) = \mathbf{g}(\x_{k}) + J_g(\x_k) (x_{k+1}-x_k) + o(\|\x_{k+1}-\x_k\|^2) \mathbf{1_d},
\end{equation*}
where $J_g$ is the Jacobian matrix with respect to $\x$, and $\mathbf{1_d}\in\R^d$ has all elements to be $1$. Hence, we can conclude that: when $\x_{k+1}$ is close to $\x_k$,
\begin{align*}
y_{k} &= \nabla F(\x_{k+1}) - \nabla F(\x_k)
= B_k (\x_{k+1}-\x_k) + o(\|\x_{k+1}-\x_k\|^2) \mathbf{1_d} \approx B_k (\x_{k+1}-\x_k),
\end{align*}
where $B_k$ is the Hessian at $\x_k$, which is exactly the secant equation in BFGS. Therefore, another possible remedy to stabilize L-BFGS is to approximate the differences of gradient using (approximate) second-order information, i.e.,
\begin{equation*}
y_{k} = B_k (\x_{k+1}-\x_k),
\end{equation*}
as this allows smooth and stable evaluation of $y_k$. Meanwhile, the Hessian-vector product can be easily computed and is not expensive~\cite{Martens:2010,DBLP:conf/icml/MartensG15}.

\subsection{Stochastic L-BFGS with Hessian Information and Vector-free Two-loop Recursion}

The proposed algorithm of stochastic L-BFGS with Hessian information (LBFGS-H) is formulated by replacing $y_k$ with the stochastic version of $B_k (\x_{k+1}-\x_{k})$ in Algorithm~\ref{LBFGS}, i.e.,
\begin{equation}\label{eq:LBFGSH}
y_k = B_k^{S_k} (\x_{k+1}-\x_{k}),
\end{equation}
where $S_k$ is the stochastic batch picked at iteration $k$ and $B_k^{S_k}\eqdef\tfrac{1}{|S_k|} {\textstyle\sum}_{i\in S_k} \nabla^2 f_i(\x_k)$. 


For an efficient implementation in a map-reduce environment (e.g. Hadoop, Spark), we use a vector-free L-BFGS (VL-BFGS) update in Algorithm~\ref{VF-LBFGS} originated from~\cite{VFLBFGS} for the two-loop recursion. \cite{VFLBFGS} proposes a vector-free L-BFGS based on the classical L-BFGS where they set $H_k^0 = \tfrac{y_{k-1}^Ts_{k-1}}{y_{k-1}^Ty_{k-1}} I $. However, the choice of $H_k^0$ is very important, therefore we propose the vector-free L-BFGS algorithm applicable to any feasible $H_k^0$ as follows.

In details, if we observe the direction generated by the two-loop recursion in Algorithm~\ref{LBFGSloop}, we are able to figure out that we can represent the output direction using the $(2m+1)$ invariable base vectors, i.e.,
\begin{equation}\label{eq:bs}
\begin{aligned}
&b_1=s_{k-m+1}, \dots, b_m = s_{k}, b_{m+1} = y_{k-m+1}, \dots, b_{2m} = y_{k}, b_{2m+1} = g_k.
\end{aligned}
\end{equation}
The direction after the first loop can be written as $q = {\textstyle\sum}_{l=m+1}^{2m+1} \delta_l b_l$, and after we scale the direction with $H_k^0$ we obtain $r_0 = H_k^0 q$, so the final result of the two-loop recursion can be written as
\begin{equation*}
-H_k\nabla F^{S_k}(\x_k) = \delta_0 r_0 + {\textstyle\sum}_{l=1}^{m} \delta_l b_l.
\end{equation*}
Note that the coefficients are evaluated with only dot-products which are defined in the matrix $M\in\R^{(m+1)\times m}$ of the following form:
\begin{equation}\label{eq:Ms}
M_{pq} = 
\begin{cases}
 y_{k-m+p}^T s_{k-m+q}, &\text{if } p, q\in\{1,2,\dots,m\},\\
 g_k^T s_{k-m+q}, &\text{if } p=m+1, q\in\{1,2,\dots,m\}.
\end{cases}
\end{equation}

Let us denote $j= i-(k-m)+1$ as in Algorithm~\ref{VF-LBFGS}. In the first loop, the evaluations of $\delta_1, \dots, \delta_m, \delta_{2m+1}$ are the same as \cite{VFLBFGS}, where $q$ is a linear combination of $y_{k-m+1}, \dots, y_k, g_k$ with the same corresponding coefficients $\delta_{m+1}, \dots, \delta_{2m+1}$, and from $i=k-1$ to $k-m$,
\begin{equation*}
\alpha_i = \rho_is_i^Tq = \rho_is_i^T\Big({\textstyle\sum}_{l=m+1}^{2m} \delta_l y_{k+l-2m} +\delta_{2m+1} g_k\Big) = \tfrac{1}{M_{(j,j)}} {\textstyle\sum}_{l=1}^{m+1} \delta_{l+m} M_{(l, j)}.
\end{equation*}
However, in the second loop which contributes to the coefficient evaluations of $s_{k-m+1}, \dots, s_k$, from $i=k-m$ to $i=k-1$, 
\begin{align*}
\beta &= \rho_i y_i^T r
= \rho_i y_i^T \Big(\delta_0 r + {\textstyle\sum}_{l=1}^{m} \delta_l b_l\Big)
= \tfrac{1}{M_{(j,j)}} \Big( \delta_0 Y_j + {\textstyle\sum}_{l=1}^{m} \delta_l M_{(j, l)}\Big),
\end{align*}
when we define a vector $Y\in\R^m$ with the elements $Y_i = y_i^T r_0, \forall i=1,\dots,m$.

Therefore, we can conclude that Algorithm~\ref{VF-LBFGS} is mathematically equivalent to Algorithm~\ref{LBFGSloop}. It is trivial to verify that with $H_k^0 = \frac{y_{k-1}^Ts_{k-1}}{y_{k-1}^Ty_{k-1}} I $, Algorithm~\ref{VF-LBFGS} recovers the vector-free L-BFGS in \cite{VFLBFGS}.

\begin{scheme}[H]
   \caption{Vector-free L-BFGS (Two-loop Recursion)}
   \label{VF-LBFGS}
   \fbox{\begin{subfigure}[t]{0.92\columnwidth}
\begin{algorithmic}
\STATE Compute the $(m+1)$ by $(m)$ matrix $M$ by \eqref{eq:Ms}
   \FOR{$i=1$ to $2m$}
   \STATE{$\delta_i = 0$}
   \ENDFOR
   \STATE{$\delta_0 = 1, \delta_{2m+1} = -1$}
   \FOR{$i=k-1$ to $k-m$}
   \STATE $j= i-(k-m)+1$
   \STATE $\alpha_i  =   \frac{1}{M_{(j,j)}} {\textstyle\sum}_{l=1}^{m+1} \delta_{l+m} M_{(l, j)}$
   \STATE $\delta_{m+j} = \delta_{m+j} - \alpha_i$
   \ENDFOR
   \STATE Compute $r_0 = H_k^0q$, where $q = {\textstyle\sum}_{l=m+1}^{2m+1} \delta_l b_l$, and broadcast $r = r_0$ to the workers
   \STATE Update vector $Y_i$s on the workers and send them to the server
   \FOR{$i=k-m$ to $k-1$}
   \STATE $j= i-(k-m)+1$
   \STATE $\beta =\frac{1}{M_{(j,j)}} \left( \delta_0 Y_j + {\textstyle\sum}_{l=1}^{m} \delta_l M_{(j, l)}\right)$
   \STATE $\delta_j = \delta_j + (\alpha_i - \beta)$
   \ENDFOR
   \STATE \textbf{return} with direction $p = \delta_0 r_0 + {\textstyle\sum}_{l=1}^{m} \delta_l b_l$
\end{algorithmic}
\end{subfigure}}
\end{scheme}

\subsection{Fisher Information Matrix as a Hessian Approximation and Distributed Optimization}

When we have no access to the second-order information, instead of utilizing $B_k$, we are still able to use approximations of $B_k$. Recently, numerous research has been conducted on the natural gradient algorithm, where in the update~\eqref{eq:update}, the inverse Fisher information matrix serves as $H_k$~\cite{Amari:1998:NGW:287476.287477, DBLP:conf/icml/MartensG15}. 

If we further consider the cost function in \eqref{eq:problem} as $F(\x) = L(h(\x); \b)$, where $h_i(\x) = h(\x;\a_i), i \in \setn$, $L$ is a convex loss and $h$ is some network structure, then an element of the Hessian matrix $B$ can be written as:
\begin{equation}\label{eq:Hessian}
B_{ij} = {\textstyle\sum}_{k=0}^d {\textstyle\sum}_{l=0}^d \tfrac{\partial L^2 (h(\x))}{\partial h_l(\x) \partial h_k(\x)} \tfrac{\partial h_l(\x)}{\partial \x_j} \tfrac{\partial h_k(\x)}{\partial \x_i} +{\textstyle\sum}_{k=0}^d \tfrac{\partial L (h(\x))}{\partial h_k(\x)} \tfrac{\partial h_k^2(\x)}{\partial \x_j \partial \x_i},
\end{equation}
where the first term is the component of the Hessian due to variation in $h_k$; since we are only looking at variation in $\x$, we do effectively a change of basis using the Jacobian of $h_k$. The second term, on the other hand, is the component that is due to variation in $\x$, which is why we see the Hessian of $h_k$. As it goes to the neighborhood of the minimum of the cost $L$, the first derivatives $\tfrac{\partial L(h(\x))}{\partial h_k(\x)}$ are approaching zero, which indicates that the second term is negligible. However, the first term, as an approximation of Hessian, which can be written as the following in the matrix form, is no different but identical to the Generalized Gauss-Newton matrix (GGN)~\cite{martens2014new},
\begin{equation}\label{eq:Fisher}
B_k \approx [J(\x_k)]^T L_{hh}^{(k)} J(\x_k) \eqdef \mathcal{B}_k,
\end{equation}
where $J(\x_k)$ is the Jacobian matrix of $h$ with respect to $\x$ at $\x_k$, $L_{hh}^{(k)}$ is the Hessian matrix of $L$ with respect to $h$ at $h = h(\x_k)$ and we use $\mathcal{B}_k$ to denote the Hessian or Hessian approximation used to smoothen $y_k$, i.e., $y_k = \mathcal{B}_k s_k$.

It has been verified that in the cases of popular loss functions such as cross-entropy and least-squares, the GGN matrix is exactly the Fisher information matrix (FIM)~\cite{martens2014new}. 
Note that here $h$ can be nonconvex which covers the applications of neural networks. 
Under the framework of stochastic L-BFGS we propose, we introduce the stochastic L-BFGS with Fisher information (LBFGS-F) by replacing $B_k^{S_k}$ in~\eqref{eq:LBFGSH} with the FIM. Note that when the predictor is linear, i.e. $h(\x;\a_i) = \a_i^T\x$, with the loss function $L$ as either the cross-entropy or the least-squares, LBFGS-F is identical to LBFGS-H (GGN = FIM =  Hessian). Similarly, this also applies to the batch version of the FIM.

As a map-reduce implementation of L-BFGS, the VL-BFGS update is praised for the parallelizable and distributed updates, and the possible communication cost in a distributed environment is $\mathcal{O}(m^2)$ in each iteration~\cite{VFLBFGS}, where $m$ is a small constant among the choices of $5, 10 ,20$. The classical L-BFGS needs an update on the gradient to calculate $y_k$ and this can be implemented by calculating local gradients from different workers and then the local gradients being aggregated on the server. We can still apply similar tricks to LBFGS-F but it requires more strict assumptions.

\begin{restatable}{ass}{Hdiagonal}[Diagonal Hessian of $L$]
\label{Hdiagonal}
The Hessian of the loss function $L$ with respect to the $h$ -- $L_{hh}$, is always diagonal.
\end{restatable}
\begin{remark}
This condition is not always true throughout all applications; however, in the cases of least-square loss and cross-entropy loss, where the prior case has $L_{hh}^S = \tfrac1{|S|} I$ with $S$ as the stochastic batch, and in the later case $L^S_{hh}=\text{diag}\Big(\big[\b_{i}/\left(h(\x_k;\a_i)\right)^2\big]_{i\in S}\Big)$, the Hessian is obviously always diagonal.
\end{remark}

Consider a specific batch $S_k$. If we split it into $\tau$ blocks, where the blocks are denoted as $S_{k_1}, S_{k_2}, \dots, S_{k_\tau}$, and assume the corresponding Jacobian block matrices as $J_h^{S_{k_1}}, \dots, J_h^{S_{k_\tau}}$, then the Hessian vector product with any vector $v$ can be written as
\begin{align*}
\mathcal{B}_k^{S_k} = [J_h]^T  L_{hh}^{S_k}  J_h = [J_h^{S_{k_1}} \cdots, J_h^{S_{k_\tau}}]^T L_{hh}^{S_k} [J_h^{S_{k_1}} \cdots, J_h^{S_{k_\tau}}]
\end{align*}
and since $L_{hh}^{S_k}$ is diagonal, we can write $L_{hh}^{S_k}$ in the form of diagonal blocks with the sizes to be $|S_{k_1}|, \dots, |S_{k_\tau}|$, thus the above is equivalent to
\begin{align*}
\mathcal{B}_k^{S_k} v ={\textstyle\sum}_{i=1}^\tau [J_h^{S_{k_i}}]^T L_{hh}^{S_{k_i}} J_h^{S_{k_i} }v = {\textstyle\sum}_{i=1}^\tau \mathcal{B}_k^{S_{k_i}} v, \forall\ v,
\end{align*}
which means that we can evaluate the FIM-vector products with data distributed on different workers and then aggregate them on the server. The communication cost in each round can be $\mathcal{O}(d)$.

\begin{restatable}{thm}{thmDis}[Distributed Optimization and Communication Cost]\label{communication}
Suppose that Assumption \ref{Hdiagonal} holds. Then Algorithm~LBFGS-F can be implemented in a distributed fashion, with a total communication cost of $\Ocal\big(d\log(\tau) + m^2\big)$ in each round, where $\tau\geq(m^2+m)$ is the number of workers.
\end{restatable}

\subsection{Implementation Details}\label{sec:details}

In this part, we cover important techniques for our stochastic LBFGS framework. The initialization and momentum are crucial in accelerating the algorithm. Meanwhile, keeping $H_k$ positive semi-definiteness is significant for finding the correct direction $p_k$, especially in the nonconvex setting.

\paragraph{Initialization and Momentum}  The initialization is crucial in the L-BFGS algorithm. The original L-BFGS proposes to use $\gamma I$ as $H_k^0$ where $\gamma>0$ is a constant and a commonly great choice suggested is $\gamma= \tfrac{y_{k-1}^Ts_{k-1}}{y_{k-1}^Ty_{k-1}}$. However, this may not be the case in the stochastic setting where stochasticity can lead to considerable fluctuations in Hessian scalings over the iterations. Therefore, we consider to use a momentum technique where we combine the past first-order information with the current one. With the recent success of ADAM~\cite{ADAM}, the scaling of the ADAM stochastic gradient provides excellent and stable performance. The authors evaluate the momentum stochastic gradient: $m_k = \beta_1 m_{k-1} + (1-\beta_1)g_k$ with $g_k=\nabla F^{S_k}(\x_k)$ and the momentum of the second moment of stochastic gradient $v_k = \beta_1 v_{k-1} + (1-\beta_1)g_k^2,$ followed by a bias correction step, i.e., $\hat m_k = m_k/(1-\beta_1^k)$ and $\hat v_k = v_k/(1-\beta_2^k)$, where $\hat v_k$ is an approximation to the diagonal of the Fisher information matrix~\cite{PascanuNatural14}. Then ADAM makes a step with a direction $\hat m_t/(\sqrt{\hat v_t}+10^{-8})$.

Hence, in our experiments, we estimate $H_k^0$ with the ADAM preconditioner, i.e., $H_k^0 = \text{diag}\big(1/(\sqrt{\hat v_t}+10^{-8})\big)$, and apply momentum to update the stochastic gradient with $\hat m_k$. Note that when the memory $m=0$ in Algorithm~\ref{VF-LBFGS}, our algorithm completely recovers ADAM.
\paragraph{Guarantees of Positive Semi-definiteness} The standard BFGS updates can fail in handling non-convexity because of difficulty in approximating Hessian with a positive definite matrix~\cite{dai2002convergence, mascarenhas2004bfgs}. Even L-BFGS with limited updates over each iteration, cannot guarantee the eigenvalues of approximate Hessian bounded above and away from zero. One has to apply a cautious update where the curvature condition $y_k^Ts_k>0$ is satisfied in order to maintain the positive definiteness of Hessian approximations~\cite{Nocedal2006NO}. As a well-suited approach to our algorithm, we employ a cautious strategy~\cite{li2001modified}: we skip the update, i.e., set $H_{k+1}=H_k$, if 
\begin{equation}\label{eq:modification}
y_k^Ts_k\geq \epsilon\|s_k\|^2
\end{equation}
is violated, where $\epsilon>0$ is a predefined positive constant. With the stated condition guaranteed at each L-BFGS update, the eigenvalues of the Hessian approximations generated by our framework are bounded above and away from zero (Lemma~\ref{lemma:Hessiannonconv}).

\section{Convergence Analysis}
In this section, we study the convergence of our stochastic L-BFGS framework. Due to the stochastic batches of the LBFGS-F and LBFGS-H, by using a fixed learning rate, one cannot establish the convergence to the optimal solution (or first-order stationary point) but only to a neighborhood of it. We provide theoretical foundations for both strongly convex and nonconvex objectives. Throughout the analysis, we will assume that $\forall i$, the function $f_i$ is $\Lambda$-Lipschitz continuous or $\Lambda$-smooth. i.e.,
\begin{equation}\label{eq:smooth1}
\|\nabla f_i(\x') - \nabla f_i(\x)\| \leq \Lambda\|\x'-x\|, \forall \x, \x' \in\R^d,
\end{equation}
or equivalently,
\begin{equation}\label{eq:smooth2}
f_i(\x')\leq f_i(\x) + \nabla f_i(\x)^T(\x'-\x) + \tfrac{\Lambda}{2}\|\x'-\x\|^2, \forall \x, \x' \in\R^d.
\end{equation}
The above implies that $F$ is also $\Lambda$-smooth.

\subsection{Strongly Convex Case}

Now we are ready to present the theoretical results for strongly convex objectives. Under this circumstance, the global optimal points $\x_*$ is unique. Before proceeding, we need to make the following standard assumptions~\cite{MB-LBFGS} about the objective and the algorithm.
\begin{restatable}{ass}{assConv}
\label{ass:conv}
Assume that the following assumptions hold.
\setlist{nolistsep}
\begin{itemize}[noitemsep]
\item[A.] $F$ is twice continuously differentiable.
\item[B.] There exist positive constants $\hat\lambda$ and $\hat\Lambda$ such that $\hat\lambda I\preceq \mathcal{B}^{S} \preceq \hat\Lambda$ for all $\x\in\R^d$ and all batches $S\subseteq\{1,2,\dots,n\}$ of size $b$, where $\mathcal{B}$ refers to the Hessian (approximation) to stabilize $y_k$.
\item[C.] $H_k^0$ in Algorithm~\ref{VF-LBFGS} is symmetric and there exists $0<\sigma\leq\Sigma$ such that $\sigma I\preceq H_k^0 \preceq \Sigma I$.
\item[D.] The batches $S$ are drawn independently and $\nabla F^{S}(\x)$ is an unbiased estimator of the true gradient $\nabla F(\x)$ for all $\x\in\R^d$, i.e., $\Exp[\nabla F^S(\x)]= \nabla F(\x)$.
\end{itemize}
\end{restatable}

We should be aware that Assumption~\ref{ass:conv}B also suggests that there is some $0<\lambda\leq\Lambda$ such that $\lambda I \preceq \nabla^2 F(\x) \preceq \Lambda I$, i.e., F is strongly convex with $\lambda$ and $\Lambda$-smooth. Because of $\lambda$-strong convexity, $F$ satisfies:
\begin{equation}\label{eq:strongconv}
F(\x')\geq F(\x) + \nabla F(\x)^T(\x'-\x) + \tfrac\lambda2\|\x'-\x\|^2, \forall \x', \x\in\R^d.
\end{equation} 

In addition, we should remark here that Assumption~\ref{ass:conv} is different to the standard assumption in~\cite{MB-LBFGS} in the sense that we do not require a bounded stochastic gradient assumption since such assumption is barely correct in both theory and practice~\cite{nguyen2018}. We also remark that Assumption~\ref{ass:conv} is generalization to the corresponding assumption in~\cite{MB-LBFGS} where by setting $\mathcal{B} = \nabla F^S$, we recover the assumption in~\cite{MB-LBFGS} so Assumption~\ref{ass:conv}C is not a new assumption. Under the above assumptions, we are able to declare the following lemma that the Hessian approximation formulated by Algorithm~\ref{VF-LBFGS} are bounded above and away from zero.

\begin{restatable}{lem}{lemmaHessian}\label{lemma:Hessian}
If Assumptions~\ref{ass:conv}A-C hold, then there exist constants $0<\mu_1\leq \mu_2$ such that $\{H_k\}$ generated by Algorithm~\ref{VF-LBFGS} in the stochastic form satisfy:
\begin{equation*}
\mu_1 I \preceq H_k\preceq \mu_2 I, \text{ for }k=0,1,2,\dots
\end{equation*}
\end{restatable}

With the help of Lemma~\ref{lemma:Hessian}, but different from~\cite{MB-LBFGS}, we can prove the following theorem without bounded assumption for the stochastic gradient.

\begin{thm}\label{thm:convergence1}
Suppose that Assumptions~\ref{ass:conv}A-D hold, $f_i$s are convex, and let $F*=F(\x^*)$ where $\x^*$ is the minimizer of $F$. Let $\{\x_k\}$ be the iterates generated by the stochastic L-BFGS framework with a constant learning rate 
$\alpha_k = \alpha\in \left(0, \tfrac{\lambda\mu_1}{\mu_2^2(\lambda + \Lambda\beta(b))\Lambda}\right),$ 
 and with $H_k$ generated by Algorithm~\ref{VF-LBFGS}. Then for all $k\geq 0$,
 \begin{align*}
\mathbb{E}[F(\x_k)-&F^*] \leq  
\big\{1- [1-2\alpha(\lambda\mu_1 - \alpha\mu_2^2(\lambda+\Lambda\beta(b))\Lambda)] ^k \big\}\tfrac{\alpha\mu_2^2\Lambda N}{4(\lambda\mu_1 - \alpha\mu_2^2(\lambda+\Lambda\beta(b))\Lambda)} 
\\&
+ 
[1-2\alpha(\lambda\mu_1 - \alpha\mu_2^2(\lambda+\Lambda\beta(b))\Lambda)]^k[F(\x_0)-F^*] 
\xrightarrow{k\to\infty} \tfrac{\alpha\mu_2^2\Lambda N}{4(\lambda\mu_1 - \alpha\mu_2^2(\lambda+\Lambda\beta(b))\Lambda)},
\end{align*}
where $\beta(b) = \tfrac{n-b}{b(n-1)}$, and $N = 2\Exp[\|\nabla f_i(x_*)\|^2]$. (Check a complete version in Appendix~\ref{appendixA}.)

\end{thm}

\subsection{Nonconvex Case}

Under the following standard nonconvex assumptions~\cite{MB-LBFGS}, we can proceed with the convergence for nonconvex problems for the first-order stationary points.

\begin{restatable}{ass}{assNonConv}
\label{ass:nonconv}
Assume that the following assumptions hold.

\setlist{nolistsep}
\begin{itemize}[noitemsep]
\item[A.] $F$ is twice continuously differentiable.
\item[B.] There exists a positive constant $\hat\Lambda$ such that $\mathcal{B}^{S}\preceq\hat\Lambda$
 for all batches $S\subseteq\{1,2,\dots,n\}$ of size $b$. $F$ is $\Lambda$-smooth.
\item[C.] $H_k^0$ in Algorithm~\ref{VF-LBFGS} is symmetric and there exists $0<\sigma\leq\Sigma$ such that $\sigma I\preceq H_k^0 \preceq \Sigma I$.
\item[D.] The function $F(\x)$ is bounded below by a scalar $\hat F$.
\item[E.] There exist constants $\gamma\geq 0$ and $\eta>0$ such that $\Exp_S [\|\nabla F^S(\x)\|^2] \leq\gamma^2 + \eta\|\nabla F(\x)\|^2$ for all $\x\in\R^d$ and batches $S\subseteq\{1,2,\dots,n\}$ of size $b$.
\item[F.] The batches $S$ are drawn independently and $\nabla F^{S}(\x)$ is an unbiased estimator of the true gradient $\nabla F(\x)$ for all $\x\in\R^d$, i.e., $\Exp[\nabla F^S(\x)]= \nabla F(\x)$.
\end{itemize}
\end{restatable}
Similar as the strongly convex case, by setting $\mathcal{B}^S = \nabla F^S$, Assumption~\ref{ass:nonconv}B is equivalent to saying that $F^S$ is $\Lambda$-smooth or $\nabla F^S$ is $\Lambda$-Lipschitz continuous which recovers the corresponding assumption in~\cite{MB-LBFGS}. However, different from the strongly convex case, here we need the bounded gradient assumption (Assumption~\ref{ass:nonconv}E). Again, with the help of the above assumptions, we can conclude that $H_k$ bounded above and away from zero as follows.

\begin{restatable}{lem}{lemmaHessianNonConv}\label{lemma:Hessiannonconv}
If Assumptions~\ref{ass:nonconv}A-C hold, then there exist constants $0<\mu_1\leq \mu_2$ such that $\{H_k\}$ generated by Algorithm~\ref{VF-LBFGS} (we use a skipping scheme in Section~\ref{sec:details}, i.e., we skip the update by setting $H_{k+1}=H_k$ when \eqref{eq:modification} is violated) in the stochastic form satisfy:
\begin{equation*}
\mu_1 I \preceq H_k\preceq \mu_2 I, \text{ for }k=0,1,2,\dots
\end{equation*}
\end{restatable}

With Lemma~\ref{lemma:Hessiannonconv}, the convergence to a neighborhood can also be proven for nonconvex cases.

\begin{restatable}{thm}{thmConvergenceNonConv}\label{thm:convergence2}
Suppose that Assumptions~\ref{ass:nonconv}A-F hold. Let $\{\x_k\}$ be the iterates generated by the stochastic L-BFGS framework with a constant learning rate 
$\alpha_k = \alpha\in \left(0, \tfrac{\mu_1}{\mu_2^2\eta\Lambda}\right),$ 
 and starting from $\x_0$ by setting $H_{k+1}=H_k$ whenever \eqref{eq:modification} is violated. Then for all $L\geq 1$,
\begin{align*}
\mathbb{E}\Big[ \tfrac1L {\textstyle\sum}_{k=0}^{L-1} \|\nabla F(\x_k)\|^2\Big] &\leq \tfrac{\alpha\mu_2^2\gamma^2\Lambda}{\mu_1} + \tfrac{2[F(\x_0)-F^*]}{\alpha\mu_1L} 
\xrightarrow{L\to\infty} \tfrac{\alpha\mu_2^2\gamma^2\Lambda}{\mu_1}.
\end{align*}

\end{restatable}


\section{Numerical Experiments}\label{sec:experiments}

In this section, we present numerical results to illustrate the properties and performance of our proposed algorithms (LBFGS-H and LBFGS-F) on both convex and nonconvex applications. For comparison, we show performance of popular stochastic gradient algorithms, namely, ADAM~\cite{ADAM}, ADAGRAD~\cite{AdaGrad} and SGD (momentum SGD). Besides, we include the performance for classical L-BFGS where $H_k^0 = \frac{y_{k-1}^Ts_{k-1}}{y_{k-1}^Ty_{k-1}}I$, and a stochastic L-BFGS as LBFGS-S where we set $y_k = \nabla F^{S_k} (\x_k) - \nabla F^{S_{k-1}}(\x_{k-1}).$ In the convex setting, we test logistic regression problem on \emph{ijcnn1}~\footnote{Available at \burl{http://www.csie.ntu.edu.tw/~cjlin/libsvmtools/datasets/}.}. where LBFGS-H is identical to LBFGS-F because of the linear predictor, so we omit the results for LBFGS-F. On the other hand, we show performance of 1-hidden layer neural network (with 300 neurons) and LeNet-5 (a classical convolutional neural network)~\cite{Lecun98} on \emph{MNIST}. Across all the figures, each epoch refers to a full pass of the dataset, i.e., $n$ component gradient evaluations.

 \begin{figure}[H]
 \vspace*{-8pt}
\centering
 \epsfig{file=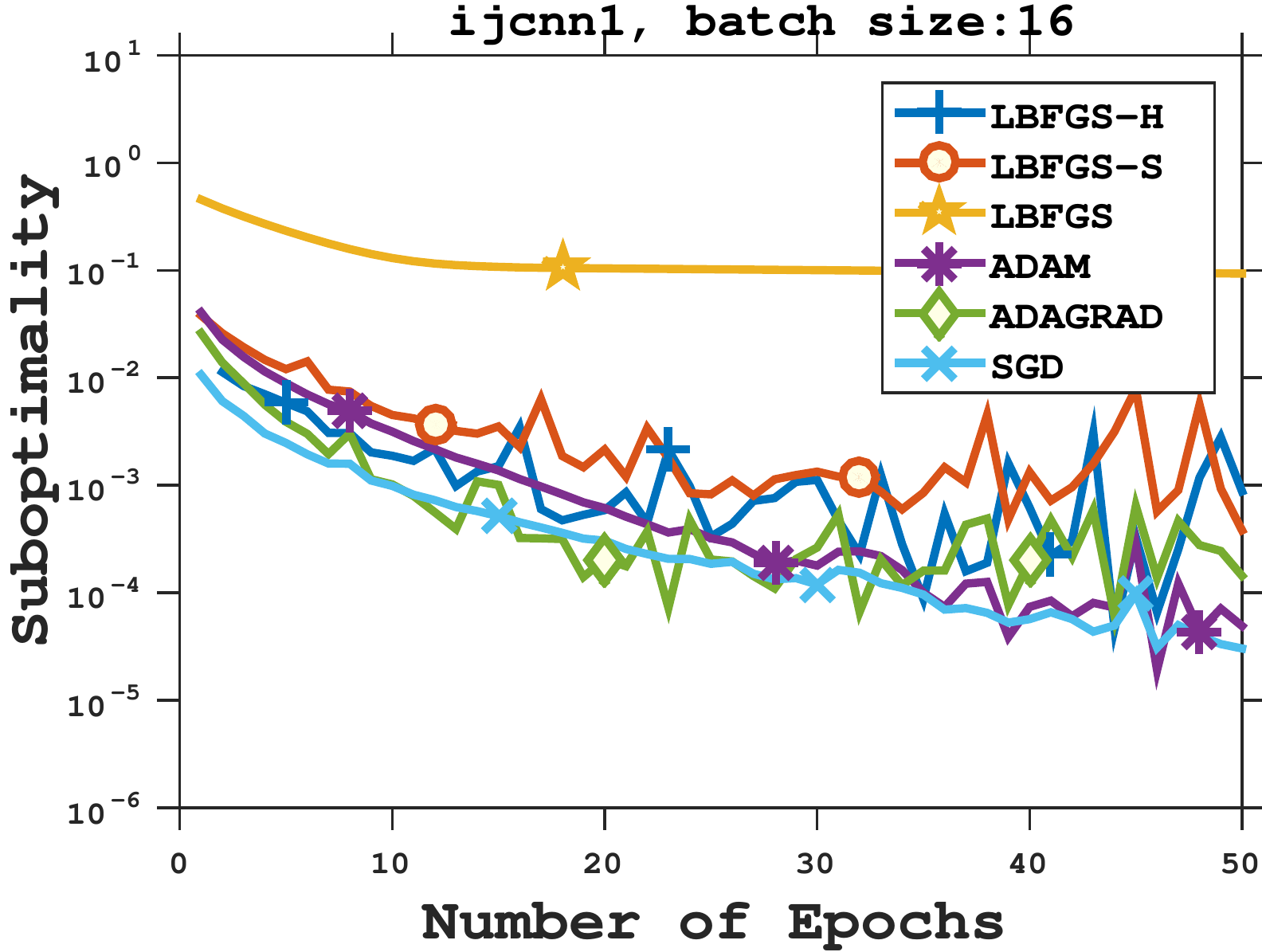,width=0.24\textwidth}
 \epsfig{file=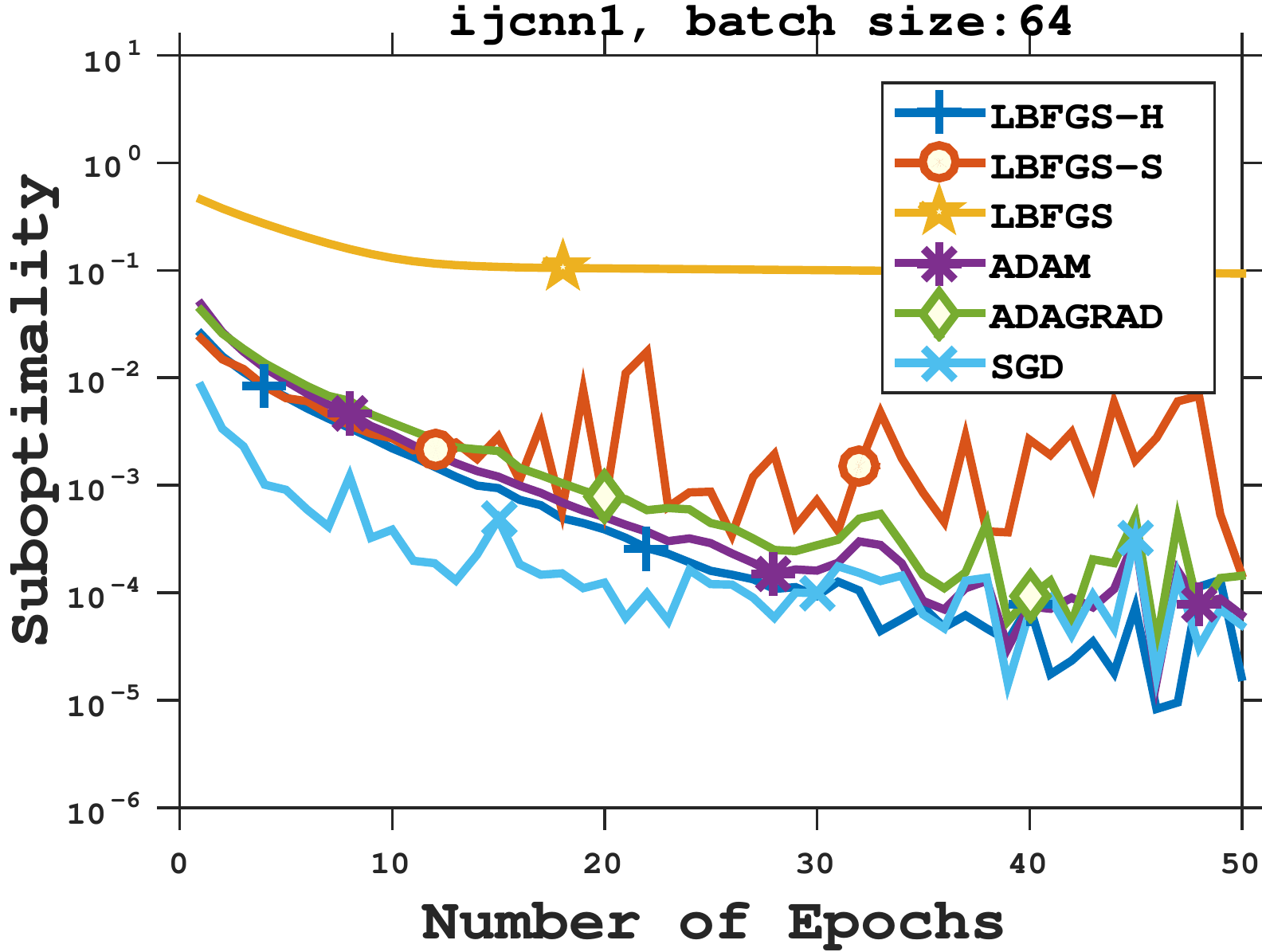,width=0.24\textwidth} 
 \epsfig{file=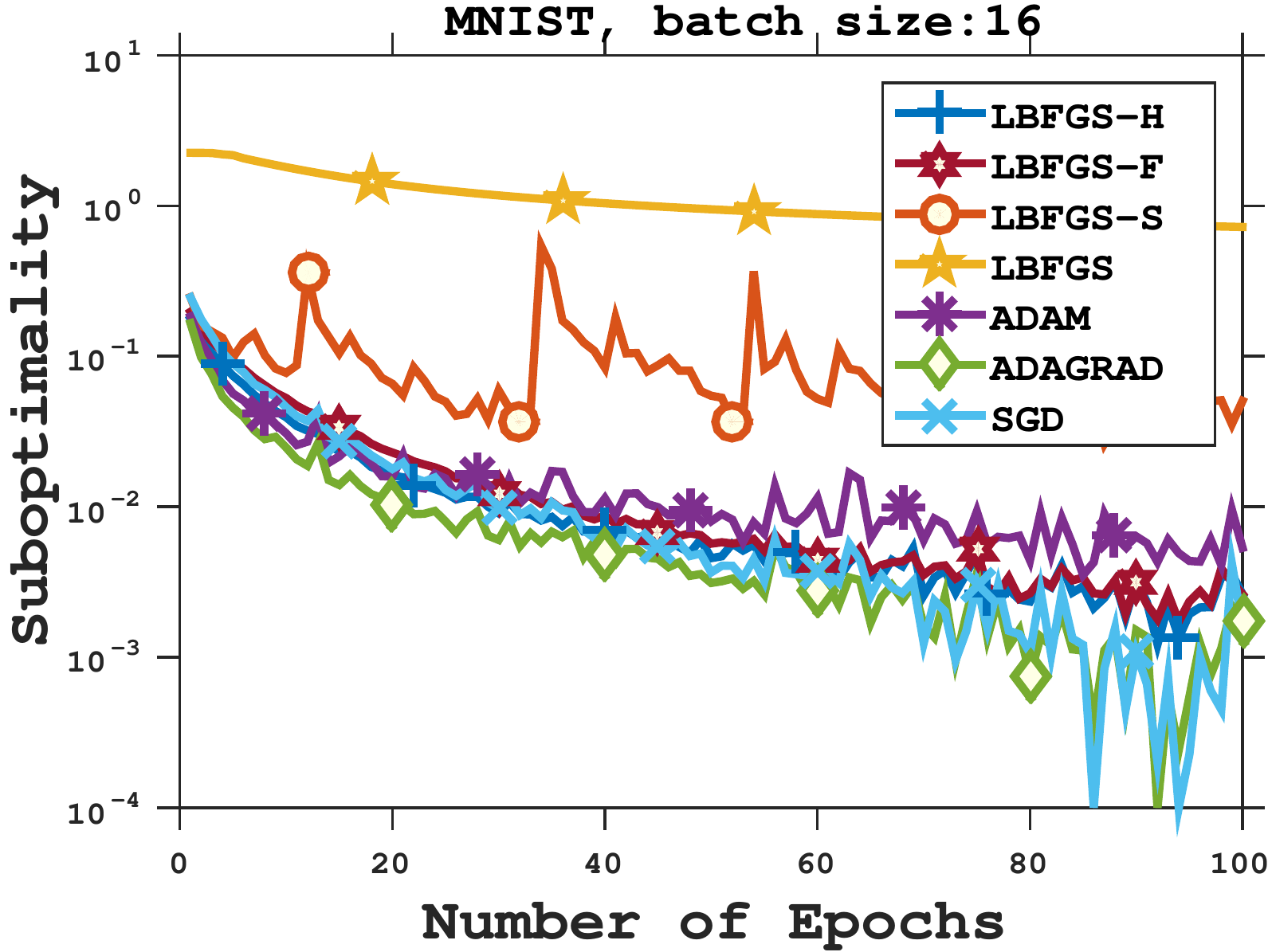,width=0.24\textwidth}
  \epsfig{file=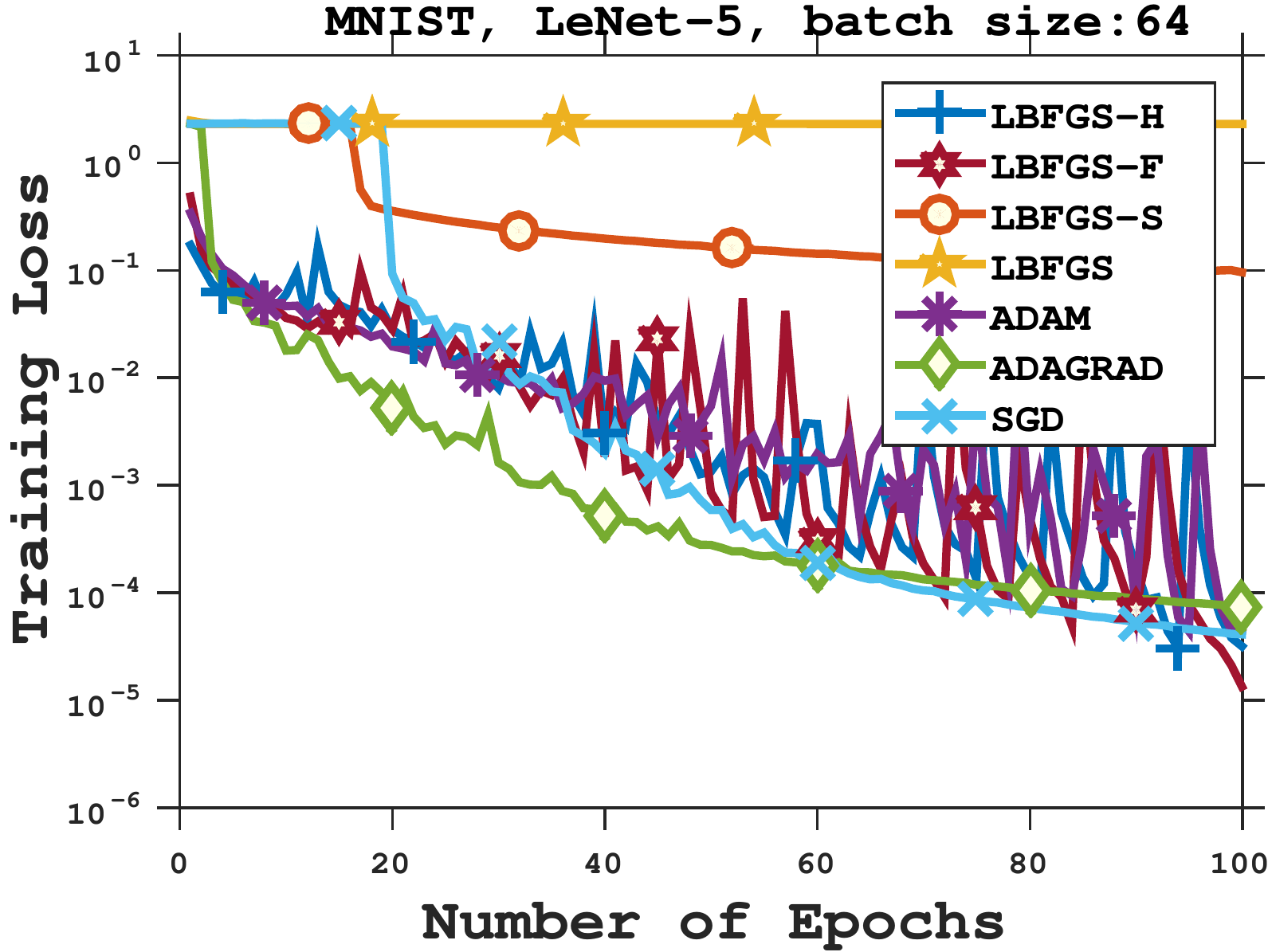,width=0.24\textwidth} 
    \epsfig{file=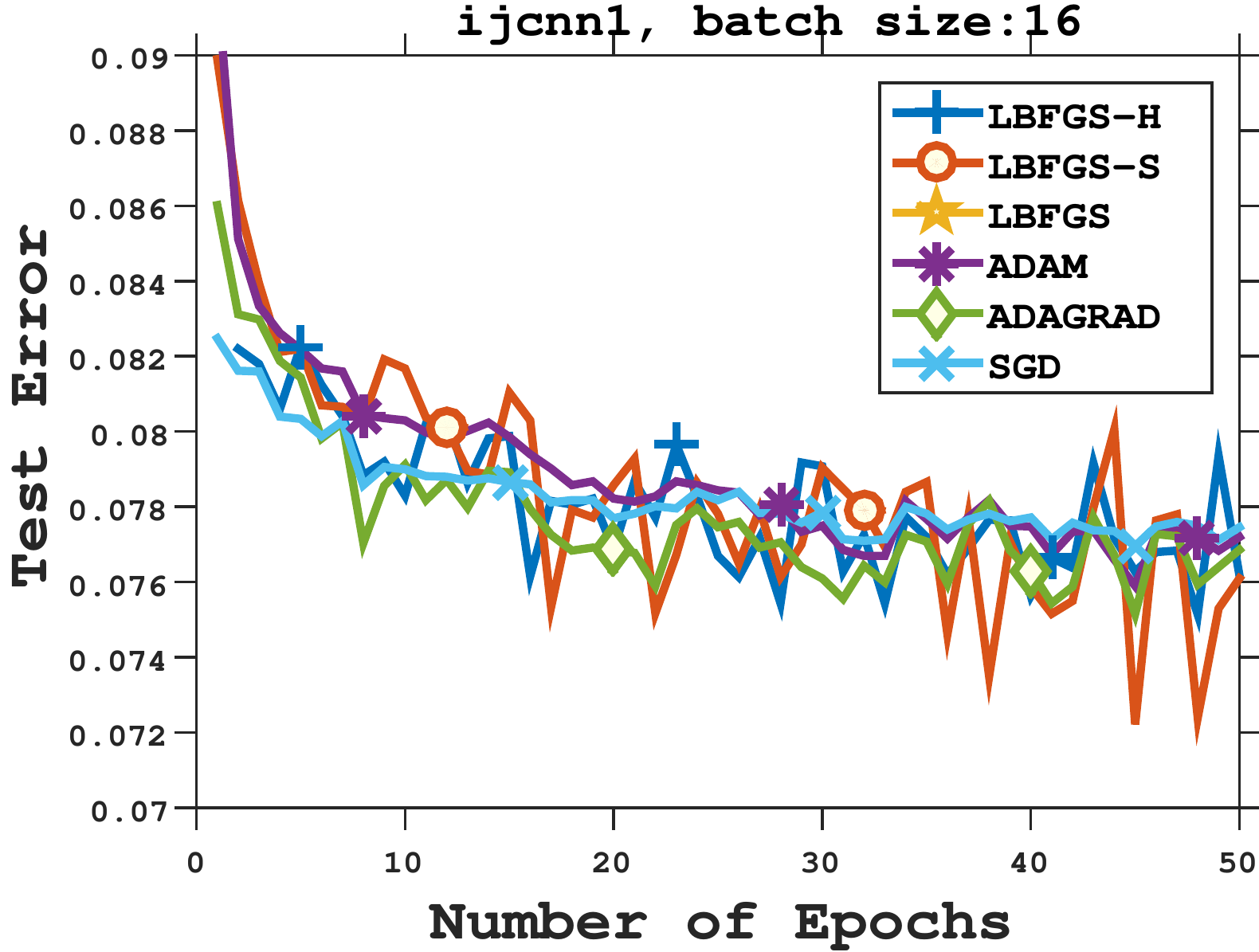,width=0.24\textwidth}
   \epsfig{file=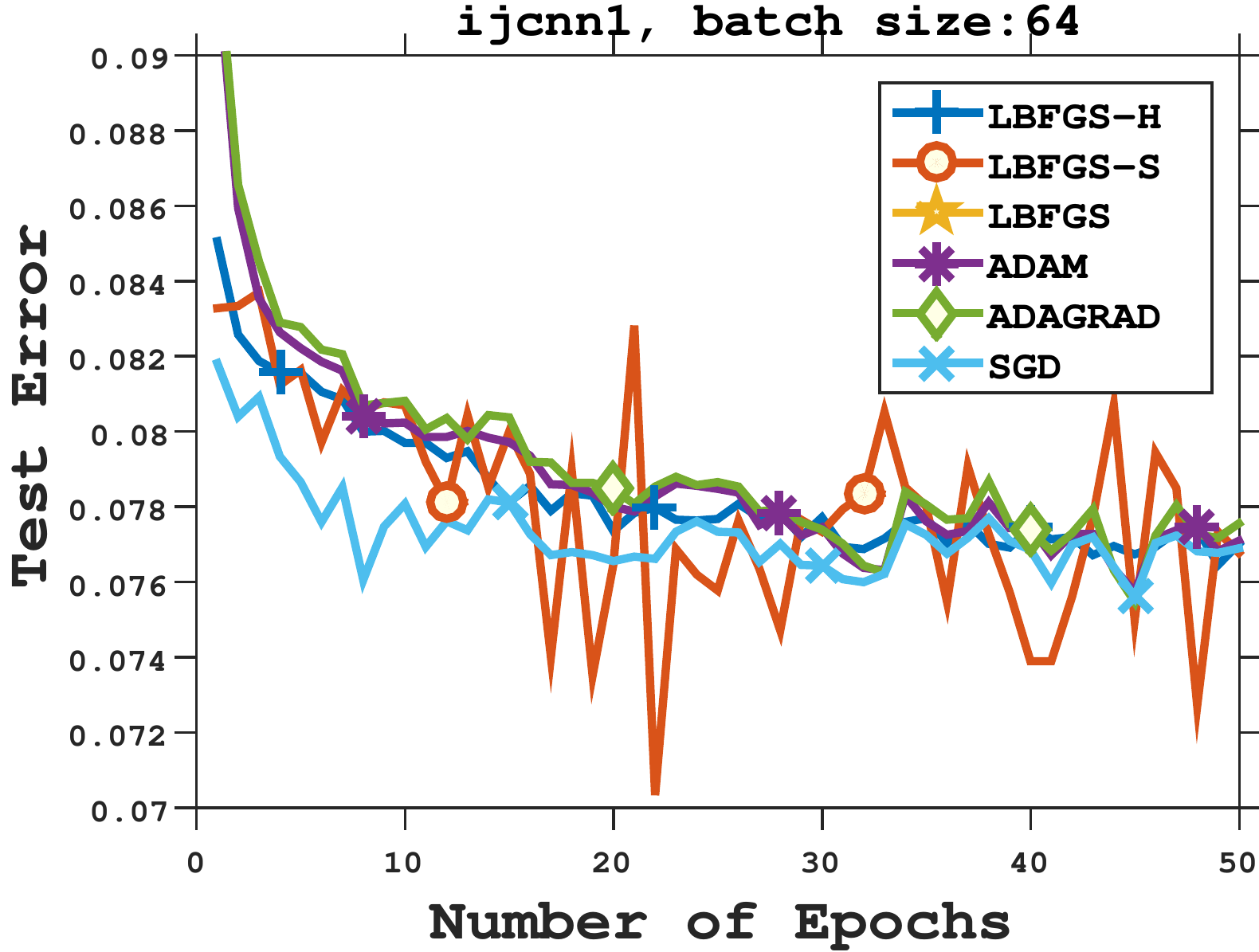,width=0.24\textwidth} 
     \epsfig{file=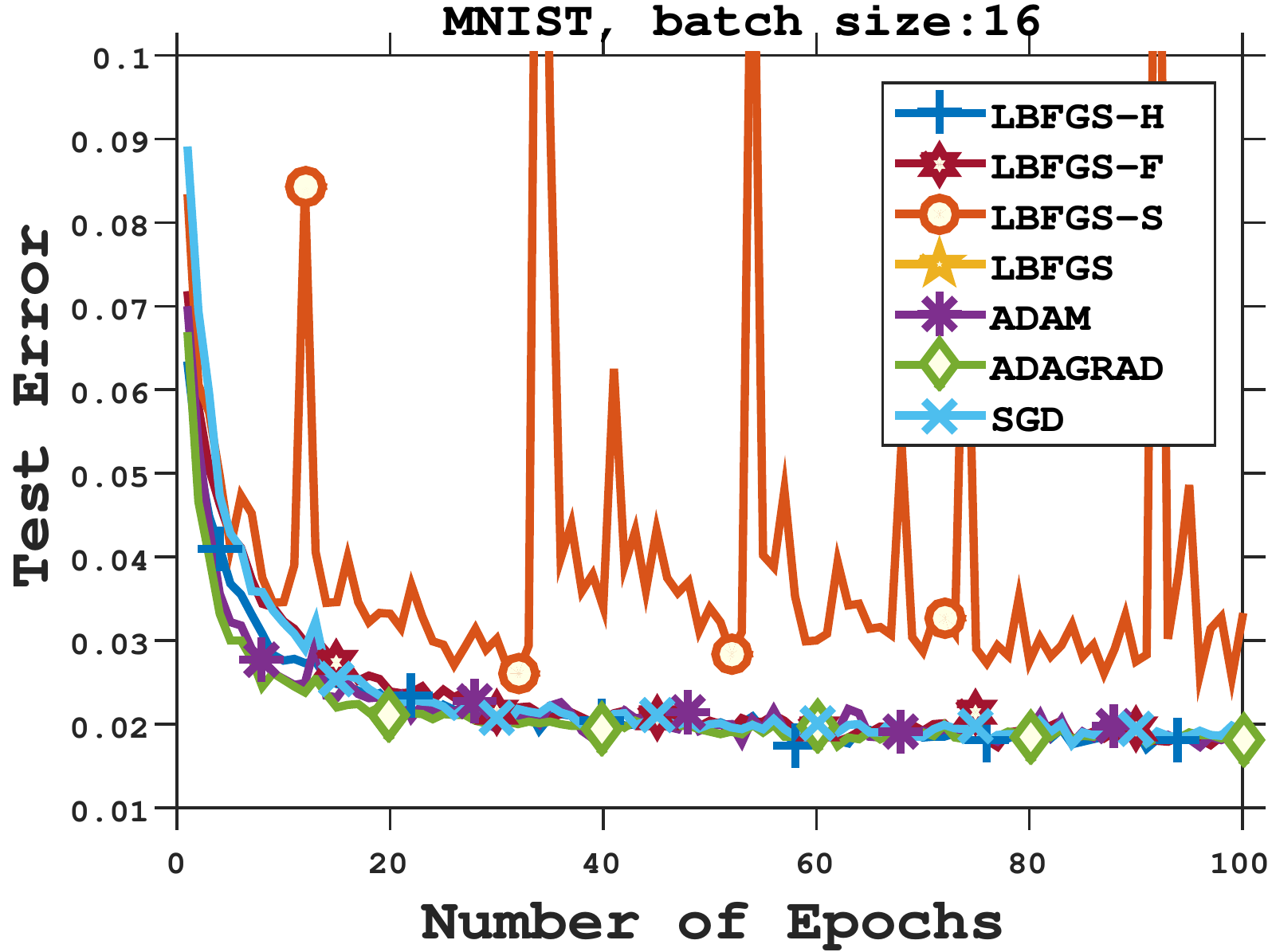,width=0.24\textwidth}
    \epsfig{file=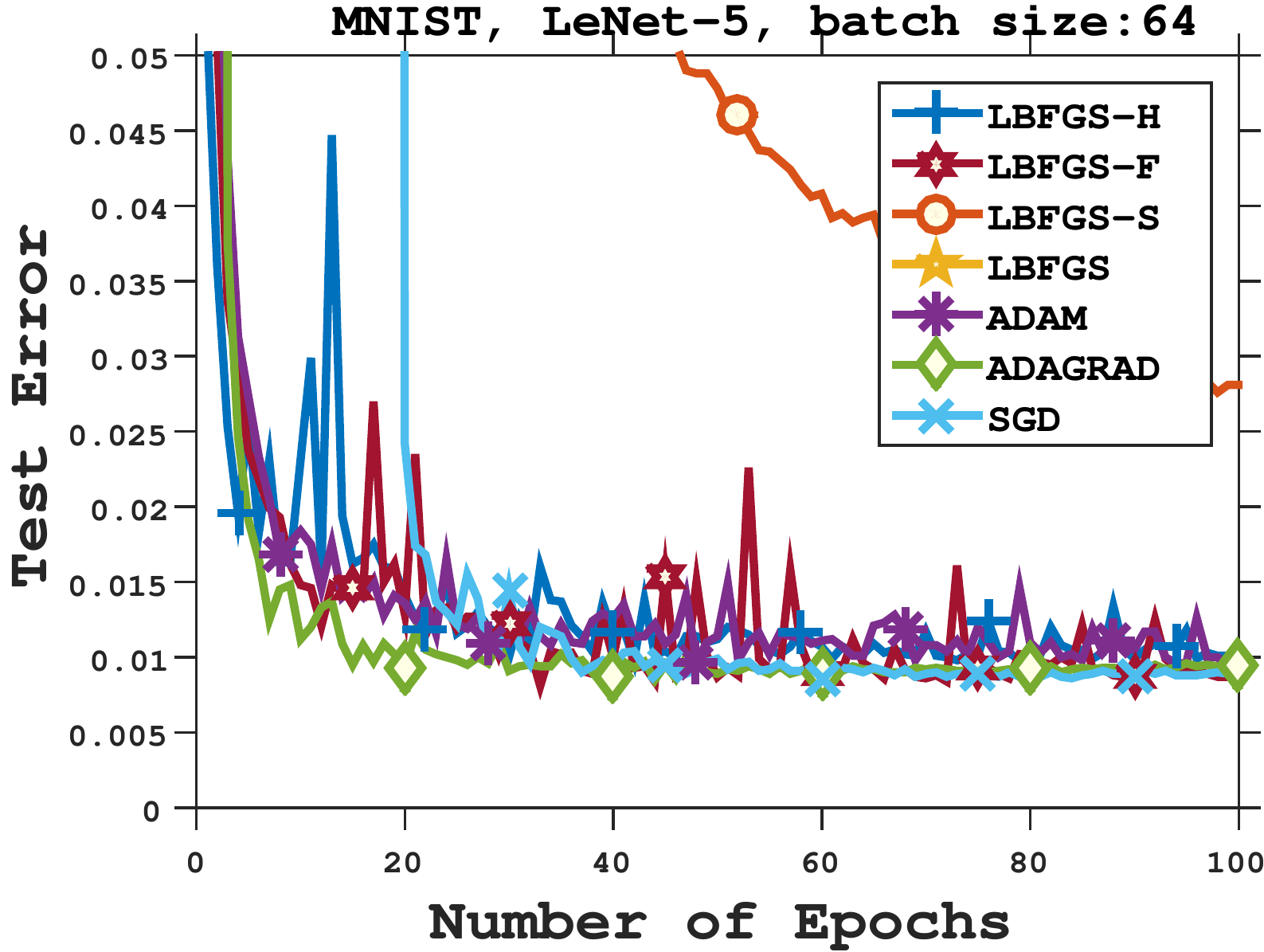,width=0.24\textwidth}
    \vspace*{-5pt}
 \caption{\footnotesize Comparisons of sub-optimality (top) and test errors (bottom) for different algorithms with batch sizes 16, 64 on \emph{ijcnn1} (logistic regression) and 16, 64 on \emph{MNIST} with 1 hidden layer neural network and LeNet-5.}
 \vspace*{-8pt}
   \label{fig:exps}
 \end{figure}
 Figure~\ref{fig:exps} shows sub-optimality $F(\x_k)-F(\x_*)$ (training loss $F(\x_k)$ for the last column) and test errors of various methods with batch sizes $16$ and $64$ on the logistic regression problem with \emph{ijcnn1} for the first two columns, and LBFGS-H exhibits competitive performance with ADAM, SGD and ADAGRAD while LBFGS-S seems highly unstable. On the nonconvex examples for the last two columns in the figure, similar results are presented with LBFGS-S to be extremely unstable and slow.
 
   \begin{figure}[h]
  \centering
 \epsfig{file=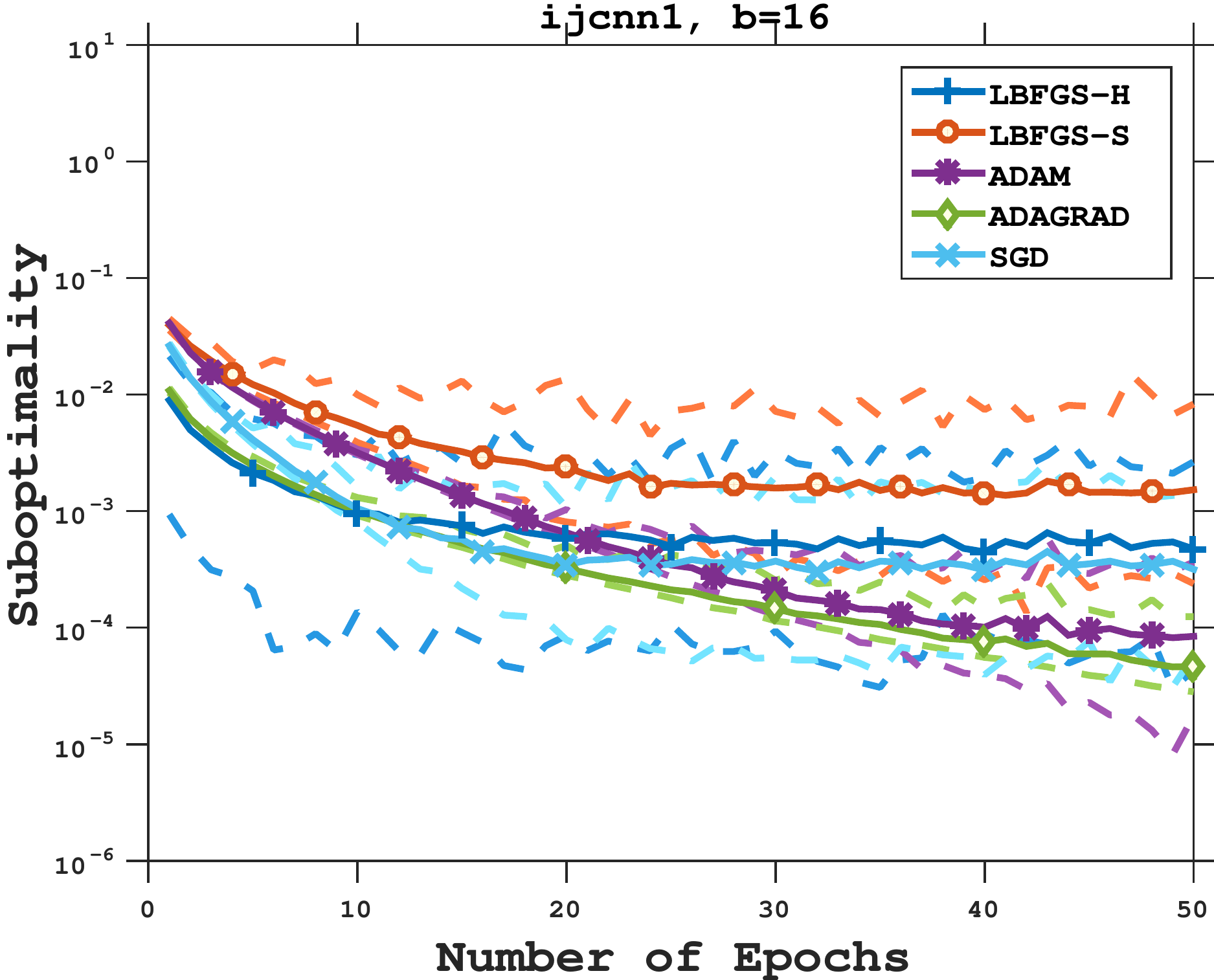,width=0.24\textwidth}
 \epsfig{file=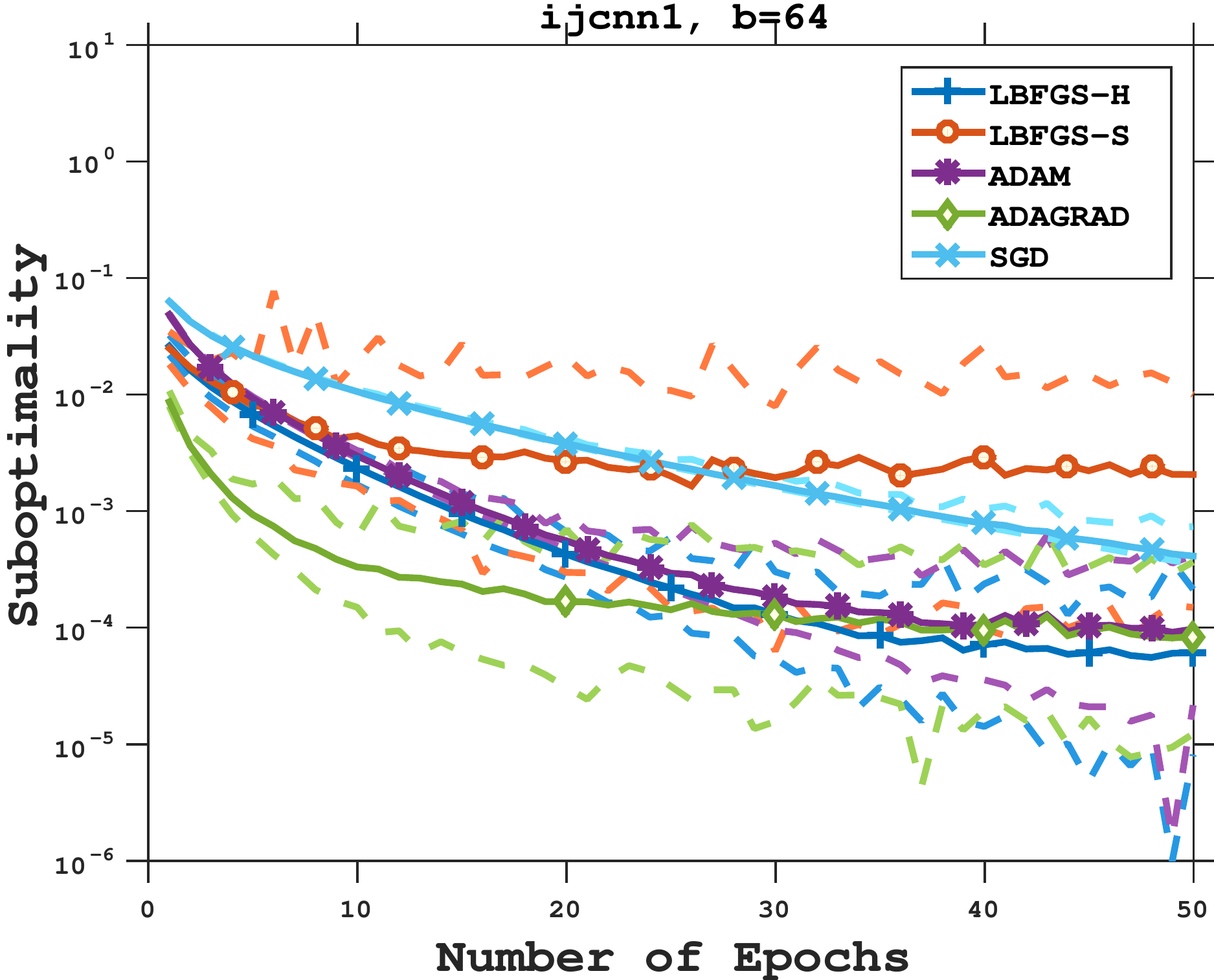,width=0.24\textwidth} 
    \epsfig{file=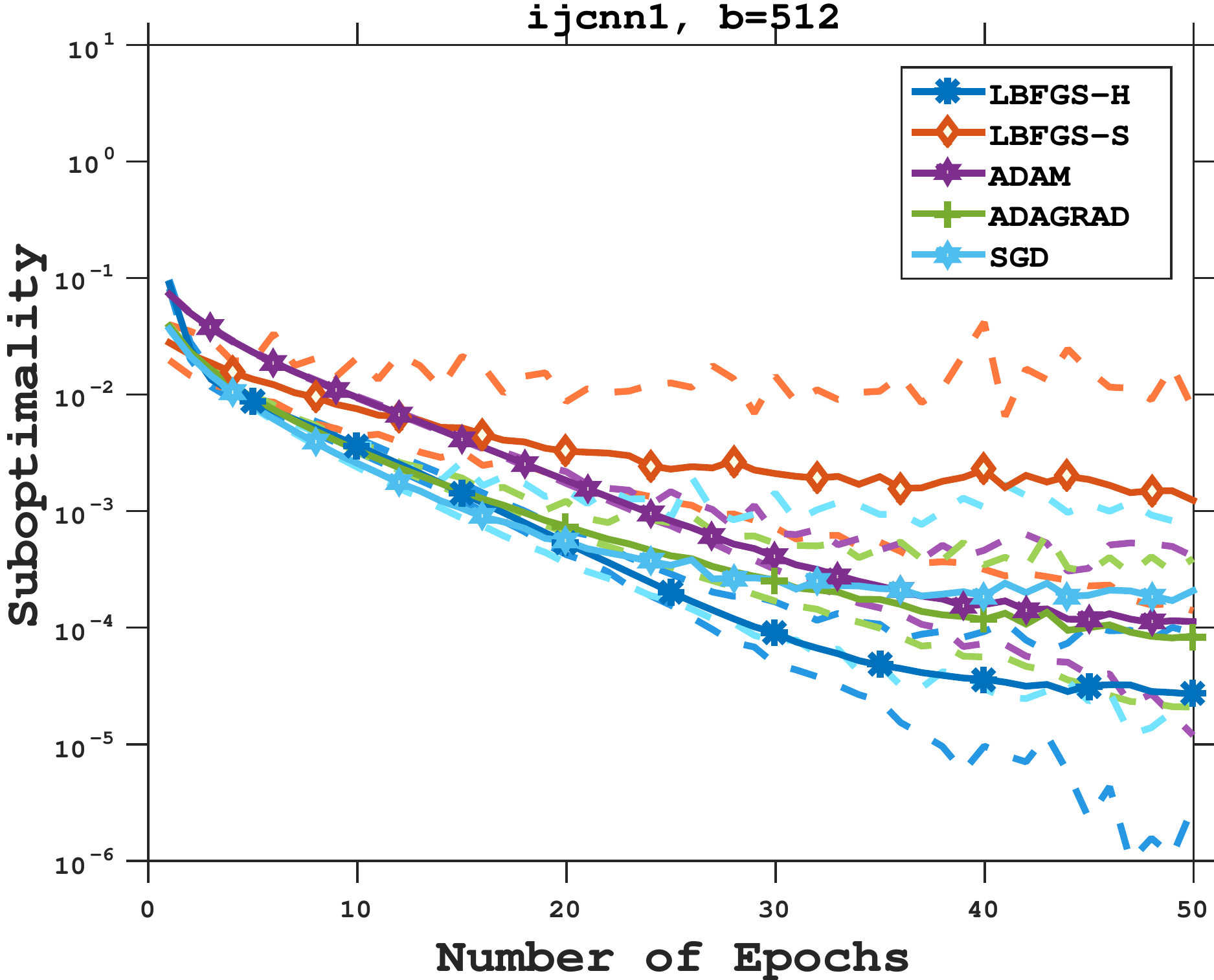,width=0.24\textwidth} 
    \epsfig{file=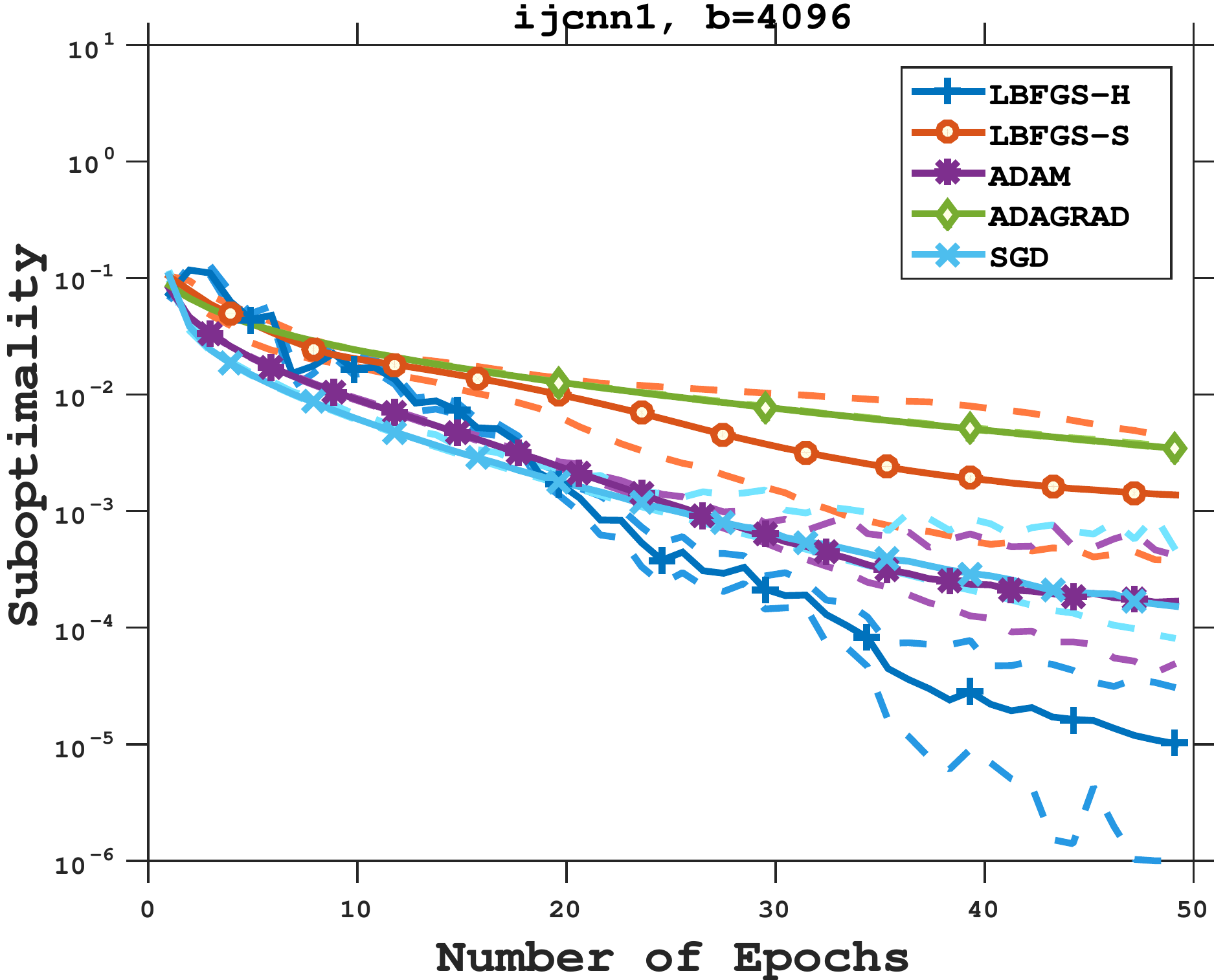,width=0.24\textwidth}
    \vspace*{-5pt}
 \caption{\footnotesize Comparisons of sub-optimality (top) and test errors (bottom) for different stochastic methods with batch sizes 16, 64, 512, 4096 on \emph{ijcnn1}, convex, logistic regression.}
   \label{fig:randseeds}
 \end{figure}
To further show the robustness of LBFGS-H (LBFGS-F), we run each method with different batch sizes and 100 different random seeds on the logistic regression problem with dataset \emph{ijcnn1} in Figure~\ref{fig:randseeds}, and report the results. The dotted lines represent the best and worst performance of the corresponding algorithm and the solid line shows the average performance. Obviously, with large batch sizes, the performance of ADAM, ADAGRAD and SGD worsen while LBFGS-H behaves steadily fast and outperforms the others in sub-optimality. This also conveys that to achieve the same accuracy, fewer epochs are needed, leading to fewer communications for our framework when the batch size is large.

\begin{wrapfigure}{L}{0.22\textwidth}
\vspace*{-7pt}
\centering
 \epsfig{file=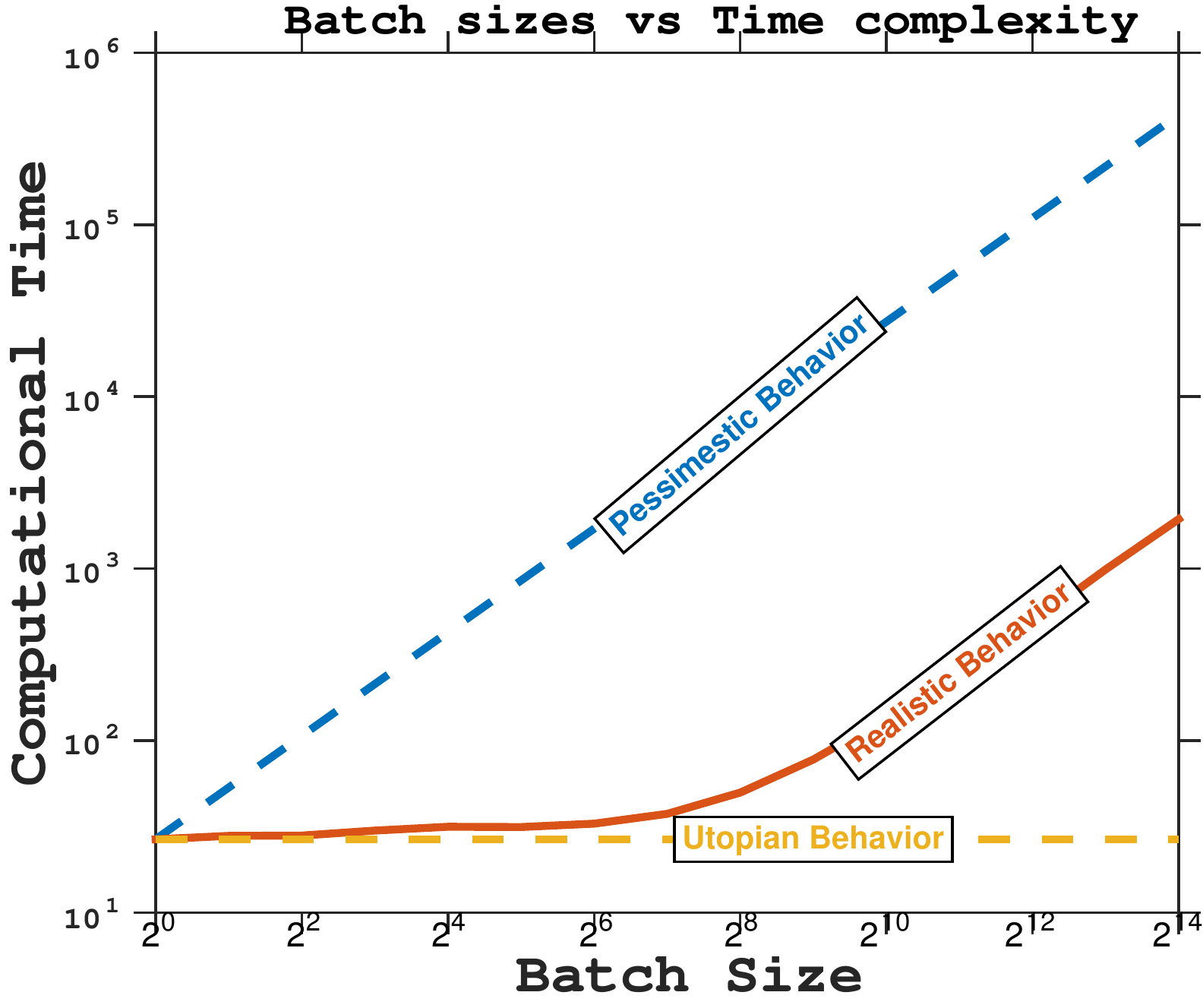,width=0.23\textwidth}
 \caption{Batch size vs time complexity.}\label{fig:timecomplexity}
 \vspace*{-1pt}
\end{wrapfigure}

The ability to use a large batch size is of particular interest in a distributed environment since it allows us to scale to multiple GPUs without reducing the per-GPU workload and without sacrificing model accuracy. In order to illustrate the benefit of large batch sizes, we evaluate the stochastic gradient $\nabla F^S(\x)$ on a neural network with different batch sizes ($b=2^0, 2^1, \dots, 2^{14}$) on a single GPU (Tesla K80), and compare the computational time against that of the pessimistic and utopian cases in Figure~\ref{fig:timecomplexity}. Up to $b = 2^6$, the computational time stays almost constant; nevertheless, with a sufficiently large batch size ($b>2^8$), the problem becomes computationally bounded and suffers from the computing resource limited by the single GPU, hence doubling batch size leads to doubling computational time. 
Therefore, the efficiency of our proposed algorithm shown in Algorithm~\ref{fig:randseeds} can benefit tremendously from a distributed environment.

%
%
%
%
\section{Conclusion}
We developed a novel framework for the L-BFGS method with stochastic batches that is stable and efficient. Based on the framework, we proposed two variants -- LBFGS-H and LBFGS-F, where the latter tries to employ Fisher information matrix instead of the Hessian to approximate the difference of gradients. LBFGS-F also admits a distributed implementation. We show that our framework converges linearly to a neighborhood of the optimal solution for convex and nonconvex settings under standard assumptions, and provide numerical experiments on both convex applications and nonconvex neural networks.


\subsubsection*{Acknowledgments}
Jie Liu was partially supported by the IBM PhD Fellowship. Martin Tak\'{a}\v{c} was partially supported by the U.S. National Science Foundation, under award number NSF:CCF:1618717, NSF:CMMI:1663256 and NSF:CCF:1740796. We would like to thank Courtney Paquette for her valuable advice on the paper.
%


\bibliography{reference}
\bibliographystyle{plain}


\newpage
\appendix

  \vbox{%
    \hsize\textwidth
    \linewidth\hsize
    \vskip 0.1in
  \hrule height 4pt
  \vskip 0.25in
  \vskip -5.5pt%
  \centering
    {\LARGE\bf{On the Acceleration of L-BFGS with Second-Order Information and Stochastic Batches\\
			Supplementary Material, NIPS 2018} \par}
      \vskip 0.29in
  \vskip -5.5pt
  \hrule height 1pt
  \vskip 0.09in%
    
  \vskip 0.2in
    
  }

\section{Assumptions, Lemmas and Theorems}\label{appendixA}
\Hdiagonal*
\assConv*
\assNonConv*

\lemmaHessian*
\lemmaHessianNonConv*

\begin{lem}[Lemma 4 in \cite{konecny2015mini}]\label{lemma:randvar}
Let $\{\xi_i\}_{i=1}^n$ be  vectors in $\R^d$
and $\bar{\xi} \eqdef \frac1n \sum_{i=1}^n \xi_i \in \R^d$.
Let $\hat S$ be  a random subset of $\setn$ of size $\tau$, chosen uniformly at random from all subsets of this cardinality. Taking expectation with respect to $\hat{S}$, we have
\begin{equation}
\label{eq:varianceBound}
 \textstyle 
\Exp \left[ \left\|\frac1\tau \sum_{i\in \hat S} \xi_i - \bar{\xi}  \right\|^2 \right]
\leq
\frac1{n\tau}
\frac{  n-\tau}{ (n-1)}
\sum_{i=1}^n \left\| \xi_i\right\|^2.
\end{equation}
\end{lem}

\begin{lem}[Lemma 3 in \cite{nguyen2018} and Equation (10) in \cite{SVRG}]\label{lemma:fibatch}
If $f_i$s are convex and $\Lambda$-smooth , then $\forall \x\in\R^d$,
\begin{equation}\label{eq:fibound}
 \Exp[\|\nabla f_i(\x) - \nabla f_i(\x_*)\|^2] \leq 2\Lambda[F(\x) - F(\x_*)],
\end{equation}
where $\x_*=\arg\min_\x F(\x)$.
\end{lem}

\begin{restatable}{lem}{lemmabatch}\label{lemma:batch}
If $F$ is strongly convex with $\lambda$ and $f_i$s are $\Lambda$-smooth , then $\forall \x\in\R^d$, the batch gradient $\nabla F^{S}(\x) = \tfrac1b \sum_{i\in\S}\nabla f_i (\x)$ has the following bound,
\begin{equation}\label{eq:sgdbound1}
\Exp[\|\nabla F^{S}(\x)\|^2] \leq 4\beta(b)\Lambda\kappa [F(\x) - F(\x_*)] + 2\|\nabla F(\x)\|^2 +N,
\end{equation}
where $\beta(b) = \frac{n-b}{b(n-1)}, \kappa=\Lambda/\lambda$ and $N = 2\Exp[\|\nabla f_i(\x_*)\|^2]$. If we further have $f_i$s convex, the bound shrinks to
\begin{equation}\label{eq:sgdbound2}
\Exp[\|\nabla F^{S}(\x)\|^2] \leq 4\beta(b)\Lambda [F(\x) - F(\x_*)] + 2\|\nabla F(\x)\|^2 +N.
\end{equation}
\end{restatable}

\thmDis*

\thmConvergenceNonConv*

\begin{thm}[A complete version of Theorem~\ref{thm:convergence1}]\label{thm:convergence1c}
Suppose that Assumptions~\ref{ass:conv}A-D hold, and let $F*=F(\x^*)$ where $\x^*$ is the minimizer of $F$. Let $\{\x_k\}$ be the iterates generated by the stochastic L-BFGS framework with a constant learning rate 
$\alpha_k = \alpha\in \left(0, \tfrac{\lambda\mu_1}{\mu_2^2(\lambda + \Lambda\beta(b)\kappa)\Lambda}\right),$ 
 and with $H_k$ generated by Algorithm~\ref{VF-LBFGS}. Then for all $k\geq 0$,
\begin{align*}
\mathbb{E}[F(\x_k)-F^*] &\leq  [1-2\alpha(\lambda\mu_1 - \alpha\mu_2^2(\lambda+\Lambda\beta(b))\Lambda)]^k[F(\x_0)-F^*] 
\\&
+ \{1- [1-2\alpha(\lambda\mu_1 - \alpha\mu_2^2(\lambda+\Lambda\beta(b)\kappa)\Lambda)]^k \}\frac{\alpha\mu_2^2\Lambda N}{4(\lambda\mu_1 - \alpha\mu_2^2(\lambda+\Lambda\beta(b)\kappa)\Lambda)} \\
&\longrightarrow \frac{\alpha\mu_2^2\Lambda N}{4(\lambda\mu_1 - \alpha\mu_2^2(\lambda+\Lambda\beta(b)\kappa)\Lambda)} \text{ as }k\to\infty,
\end{align*}
where $\beta(b) = \frac{n-b}{b(n-1)}, \kappa=\Lambda/\lambda$ and $N = 2\Exp[\|\nabla f_i(\x_*)\|^2]$. If we further have $f_i$s convex, then similarly, with a learning rate $\alpha\in\left(0, \tfrac{\lambda\mu_1}{\mu_2^2(\lambda + \Lambda\beta(b))\Lambda}\right)$, we have 
\begin{align*}
\mathbb{E}[F(\x_k)-F^*] &\leq  [1-2\alpha(\lambda\mu_1 - \alpha\mu_2^2(\lambda+\Lambda)\Lambda)]^k[F(\x_0)-F^*] 
\\&
+ \{1- [1-2\alpha(\lambda\mu_1 - \alpha\mu_2^2(\lambda+\Lambda\beta(b))\Lambda)]^k \}\frac{\alpha\mu_2^2\Lambda N}{4(\lambda\mu_1 - \alpha\mu_2^2(\lambda+\Lambda\beta(b))\Lambda)} \\
&\longrightarrow \frac{\alpha\mu_2^2\Lambda N}{4(\lambda\mu_1 - \alpha\mu_2^2(\lambda+\Lambda\beta(b))\Lambda)} \text{ as }k\to\infty.
\end{align*}
\end{thm}

\begin{restatable}{thm}{thmConvergenceConv}\label{thm:convergenceC}
Suppose that Assumptions~\ref{ass:conv}A-D hold, $f_i$s are convex and let $F*=F(\x^*)$ where $\x^*$ is the minimizer of $F$. Let $\{\x_k\}$ be the iterates generated by the stochastic L-BFGS framework with $\alpha_k = \tfrac{\alpha}{k+E}$, where $\alpha>0$ and $E$ satisfies
\begin{equation*}
E >  \tfrac{2\mu_2^2}{\mu_1^2} (\kappa+\beta\kappa^2), \quad 2\alpha\lambda\mu_1-1-\tfrac{2\alpha^2\mu_2^2\Lambda(\lambda+\Lambda\beta(b))}{E} >0.
\end{equation*}
Then starting from $\x_0$, for all $k\geq 0$, 
\begin{equation*}
\Exp[F(\x_k)-F^*]\leq \tfrac{G(\alpha, E)}{k+E},
\end{equation*}
where 
\begin{equation}\label{eq:G}
G(\alpha.E)\eqdef\max\left\{\tfrac{\alpha^2\mu_2^2\Lambda N}{4\alpha\lambda\mu_1-2-\tfrac{4\alpha^2\mu_2^2\Lambda(\lambda+\Lambda\beta(b))}{E}]}, E\Exp[F(\x_0)-F^*]\right\}.
\end{equation}

\end{restatable}

%
%

\newpage
\section{Proofs}

\subsection{Proof of Lemma~\ref{lemma:Hessian}}\label{sec:Hessianconv}
\begin{proof}
Instead of analyzing the algorithm in $H_k$, we study the Hessian approximation where $\mathcal{H}_k = H_k^{-1}$. In this case, the L-BFGS are updated as follows (note that the superscript $(i)$ of $\mathcal{H}_k$ denotes the iteration of $m$ Hessian updates in each iteration).
\begin{itemize}
\item[1] Set $\mathcal{H}_k^{(0)} = (H_k^0)^{-1}$ such that 
\begin{equation}\label{eq:Bkbound}
\Sigma^{-1}I\preceq \mathcal{H}_k^{(0)}\preceq \sigma^{-1} I.
\end{equation}
\item[2] For $i=0, \dots, m-1,$ set $j=k-m+1+i$ and compute
\begin{equation*}
\mathcal{H}_k^{(i+1)} = \mathcal{H}_k^{(i)} - \frac{\mathcal{H}_k^{(i)} s_js_j^T\mathcal{H}_k^{(i)} }{s_j^T\mathcal{H}_k^{(i)} s_j} + \frac{y_jy_j^T}{y_j^Ts_j}.\end{equation*}
\item[3] Set $\mathcal{H}_{k} = B_k^{(m)}.$
\end{itemize}

Note that following the above updates, the curvature pairs are 
\begin{equation*}
y_{k} = \mathcal{B}_k s_k, s_k = \x_{k+1}-\x_k.
\end{equation*}
It is also easy to know that for LBFGS-H, $\mathcal{B}_k = \nabla^2 F^{S_k}(\x_k)$ is symmetric, .i.e., $\mathcal{B}_k^T = \mathcal{B}_k$; therefore
\begin{equation*}
\|y_k\|^2 = y_k^Ty_k = s_k^T\mathcal{B}_k^T\mathcal{B}_ks_k,
\end{equation*}
and by Assumption~\ref{ass:conv}B, we have that $\mathcal{B}_k- \hat\lambda I \succeq 0,$ and since $\mathcal{B}_k\succeq 0, \mathcal{B}_k^T= \mathcal{B}_k$, thus $(\mathcal{B}_k^T - \hat\lambda I)\mathcal{B}_k = \mathcal{B}_k (\mathcal{B}_k^T - \hat\lambda I)$ is normal{\footnote{A matrix $A$ is normal if and only if $A^*A = AA^*$ where $A^*$ denotes the conjugate transpose of $A$; and in the case of matrix with real values, $A^*=A^T$.}}, therefore according to Theorem 3 in~\cite{LinearAlgebra}, we have $(\mathcal{B}_k^T - \hat\lambda I)\mathcal{B}_k\succeq 0$ and hence
\begin{equation*}
 s_k^T\mathcal{B}_k^T\mathcal{B}_k s_k\geq \hat\lambda s_k^T\mathcal{B}_ks_k = \hat\lambda y_k^Ts_k,
 \end{equation*}
and similarly we can also claim that
\begin{equation*}
s_k^T\mathcal{B}_k^T\mathcal{B}_k s_k\leq \hat\Lambda s_k^T\mathcal{B}_k s_k = \hat\Lambda y_k^Ts_k.
\end{equation*}
Therefore,
\begin{equation}\label{eq:yk2bound}
\hat\lambda\leq \frac{\|y_k\|^2}{y_k^Ts_k}\leq \hat\Lambda.
\end{equation}
In addition
\begin{equation}\label{eq:sk2bound}
\tfrac{y_k^T s_k}{\|s_k\|^2} = \tfrac{s_k^T\mathcal{B}_k^Ts_k}{\|s_k\|^2} \geq \tfrac{\hat\lambda\|s_k\|^2}{\|s_k\|^2} = \hat\lambda.
\end{equation}

Then following the proof of Lemma 3.1 in \cite{MB-LBFGS}, we should have the desired result. Here we provide the rest of the proof as follows for completeness.

Since $0<\sigma\leq\Sigma$, we now use the following Trace-Determinant argument to show that the egeinvalues of $B_k$ are bounded above and away from zero.

Denote $\text{tr}(\mathcal{H})$ and $\text{det}(\mathcal{H})$ as the trace and determinant of $\mathcal{H}$, respectively, and set $j_i = k-m+i$, then the trace of the matrix $\mathcal{H}_{k}$ can be written as:
\begin{align*}
\text{tr}(\mathcal{H}_{k}) &= \text{tr}(\mathcal{H}_k^{(0)}) -  \text{tr} \left(\sum_{i=1}^m \tfrac{\mathcal{H}_k^{(i)}s_{j_i} s_{j_i}^T\mathcal{H}_k^{(i)}}{s_{j_i}^T\mathcal{H}_k^{(i)} s_{j_i}}\right) + \text{tr}\left( \sum_{i=1}^m \tfrac{y_{j_i} y_{j_i}^T}{y_{j_i}^Ts_{j_i}}\right)
\\&\leq \text{tr}(\mathcal{H}_k^{(0)})  +  \sum_{i=1}^m \tfrac{\|y_{j_i}\|^2}{y_{j_i}^Ts_{j_i}}
\\&\overset{\eqref{eq:Bkbound},\eqref{eq:yk2bound}}{\leq} \text{tr}(\sigma^{-1}I) + m\hat\Lambda = d\sigma^{-1} + m\hat\Lambda \eqdef C_1,\tagthis\label{eq:boundC1}
\end{align*}
which implies that the largest eigenvalue of $B_{k+1}$ is no larger than $C_1$, i.e., $B_{k+1}\preceq C_1 I$.

Based on a result by Powell~\cite{powell1976some}, the determinant of the matrix $\mathcal{H}_{k}$ generated by our proposed stochastic L-BFGS framework can be written as,
\begin{align*}
\text{det}(\mathcal{H}_{k}) &= \text{det}(\mathcal{H}_k^{(0)}) \prod_{1}^m \tfrac{y_{j_i}^Ts_{j_i}}{s_{j_i}^T \mathcal{H}_k^{(i-1)}s_{j_i}}
\\&= \text{det}(\mathcal{H}_k^{(0)}) \prod_{1}^m \tfrac{y_{j_i}^Ts_{j_i}}{s_{j_i}^Ts_{j_i}}\tfrac{s_{j_i}^Ts_{j_i}}{s_{j_i}^T \mathcal{H}_k^{(i-1)}s_{j_i}}
\\&\overset{\eqref{eq:sk2bound},\eqref{eq:boundC1}}{\geq}  \text{det}(\mathcal{H}_k^{(0)})  \left(\tfrac{\hat\lambda}{C_1}\right)^m
\\&\overset{\eqref{eq:Bkbound}}{\geq} \left(\Sigma^{-1}\right)^d\left(\tfrac{\hat\lambda}{C_1}\right)^m,
\end{align*}
and this indicates that the eigenvalues of all matrices $\mathcal{H}_{k}$ is bounded away from zero, uniformly.

\end{proof}

\subsection{Proof of Lemma~\ref{lemma:Hessiannonconv}}
\begin{proof}
Following the proof of Lemma~\ref{lemma:Hessian}(Section~\ref{sec:Hessianconv}), we can obviously obtain
\begin{equation*}
\|y_k\|^2 = s_k^T\mathcal{B}_k^T\mathcal{B}_k s_k\leq \hat\Lambda s_k^T\mathcal{B}_k s_k = \hat\Lambda y_k^Ts_k \quad\Longrightarrow\quad \tfrac{\|y_k\|^2}{y_k^Ts_k}\leq \hat\Lambda.
\end{equation*}

Under the skipping scheme mentioned in the paper, we do not skip when \eqref{eq:modification} holds
\begin{equation*}
 \epsilon\|s_k\|^2 \leq y_k^Ts_k \leq \|y_k\|\|s_k\| \quad\Longrightarrow\quad \|s_k\|\leq \tfrac{1}{\epsilon}\|y_k\|;
\end{equation*}
hence, $\tfrac{y_k^Ts_k}{\|s_k\|^2}\geq\epsilon$, and
\begin{equation*}
y_k^Ts_k \leq \|y_k\|\|s_k\| \leq \tfrac{1}{\epsilon}\|y_k\|^2 \quad\Longrightarrow\quad\tfrac{\|y_k\|^2}{y_k^Ts_k}\geq \epsilon,
\end{equation*}

Therefore,
\begin{equation*}
\epsilon \leq \frac{\|y_k\|^2}{y_k^Ts_k}\leq \hat\Lambda.
\end{equation*}

Then following the proof of Lemma~\ref{lemma:Hessian} in Section~\ref{sec:Hessianconv}, we should have the desired result.
\end{proof}

\subsection{Proof of Lemma~\ref{lemma:batch}}
\begin{proof}

According to Lemma~\ref{lemma:randvar}, we have 
\begin{equation}
\label{eq:varianceBound2}
 \textstyle 
\Exp \left[ \left\|\frac1\tau \sum_{i\in \hat S} \xi_i   \right\|^2 \right] = \Exp \left[ \left\|\frac1\tau \sum_{i\in \hat S} \xi_i - \bar{\xi}  \right\|^2 \right] + \|\bar\xi\|^2
\leq
\frac1{n\tau}
\frac{  n-\tau}{ (n-1)}
\sum_{i=1}^n \left\| \xi_i\right\|^2 + \|\bar\xi\|^2.
\end{equation}
By defining $\beta(b) = \tfrac{n-b}{b(n-1)}$, the following holds,
\begin{align*}
\Exp[\|\nabla F^{S}(\x)\|^2] &= \Exp[\|\nabla F^{S}(\x) - \nabla F^S(\x_*) + \nabla F^S(\x_*)\|^2]
\\&\leq 2  \Exp[\|\nabla F^{S}(\x) - \nabla F^S(\x_*)\|^2] + 2\Exp[\|\nabla F^S(\x_*)\|^2]
\\&\overset{\eqref{eq:varianceBound2}}{\leq}
\frac{2\beta(b) }{n}\sum_{i=1}^n \|\nabla f_i(\x) - \nabla f_i(\x_*)\|^2 + 2\|\nabla F(\x)\|^2 + N\tagthis\label{eq:batchbound}
\\&\overset{\eqref{eq:smooth1}}{\leq}
2\beta(b) \Lambda^2\|\x - \x_*\|^2  + 2\|\nabla F(\x)\|^2 +N
\\&\overset{\eqref{eq:strongconv}}{\leq} \frac{4\beta(b) \Lambda^2}{\lambda}[F(\x) - F(\x_*)]  + 2\|\nabla F(\x)\|^2 +N
\\&=4\beta(b) \Lambda\kappa [F(\x) - F(\x_*)] + 2\|\nabla F(\x)\|^2+N,
\end{align*}
where $\kappa=\Lambda/\lambda$ and $N = 2\Exp[\|\nabla F_S(\x_*)\|^2] $.

If we further have $f_i$s convex, then we can possibly have a tighter bound,
\begin{align*}
\Exp[\|\nabla F^{S}(\x)\|^2] &\overset{\eqref{eq:batchbound}}{\leq}
2\beta(b) \Exp[\|\nabla f_i(\x) - \nabla f_i(\x_*)\|^2] +2\|\nabla F(\x)\|^2+N
\\&\overset{\eqref{eq:fibound}}{\leq}
4\beta(b)\Lambda [F(\x)-F(\x_*)]  + 2\|\nabla F(\x)\|^2+N.
\end{align*}
\end{proof}

\subsection{Proof of Theorem~\ref{communication}}
\begin{proof}
In the distributed setting of LBFGS-F, we assume that we have a unique server (master node) and $\tau$ workers. There are mainly three communication costs: the evaluations of $\mathcal{B}_k^{S_k}$ and $\nabla F^{S_k}(\x_k)$, and the Algorithm~\ref{VF-LBFGS}.

The communication cost for evaluating $g_k=\nabla F^{S_k}(\x_k)$ includes the broadcasting of $\x_k$ from the server with a cost of $\Ocal(d)$, and retrieving the sum of the local gradients from workers with a cost of $\Ocal\big(d\log(\tau)\big)$. We have $\log(\tau)$ instead of $\tau$ because for $\tau$ vectors, we can use a binary-tree structure for the workers and the server to sum the local gradients up in $\log(\tau)$ operations (e.g. by using MPI\_Reduce).

Every iteration, we store the pairs $\{(s_i, y_j)\}, \{(s_i, g_k)\}$ into $\tau$ workers without overlap ($\tau\geq m(m+1)$) and calculate every dot-products defined in~\eqref{eq:Ms} using a map-reduce step. First, the server need to broadcast the new $s_k$ to the workers with a cost of $\Ocal(d)$ and after the evaluations of local partitions of $\mathcal{B}_k^{S_k}s_k$, the server can receive the sum of the local partitions with a communication cost of $\Ocal\big(d \log(\tau) \big)$ so that it can evaluate $y_k =\mathcal{B}_k^{S_k} s_k = {\sum}_{i=1}^\tau \mathcal{B}_k^{S_{k_i}} s_k$. Then, the server again broadcasts $y_i$s to workers with a cost of $\Ocal(d)$. This whole procedure has a total communication cost of $\Ocal\big(d + d\log(\tau)  + d\big)= \Ocal\big(d\log(\tau)\big)$. 

After the calculation of each dot-product defined in~\eqref{eq:Ms} using a map-reduce step, we need to pass the dot-products from workers to the server to formulate $M$ in Algorithm~\ref{VF-LBFGS} with a communication cost of $(m+1)m = \Ocal(m^2)$. Next, after the first loop, the server evaluates $r_0 = H_k^0q$ and sends $r=r_0$ to workers to calculate $Y_i = y_i^Tr_0, \forall i=1.\dots,m$ with a communication cost of $d$, and then retrieves $Y_i$s with a cost of $m$. This process invokes a total communication cost of $(m^2 + d + m) = \Ocal(m^2+d)$.

Hence, the total communication cost in each iteration or each round is $\Ocal\big(d\log(\tau) +d\log(\tau) + m^d +d \big) = \Ocal\big(d\log(\tau) + m^2\big)$.
\end{proof}

\subsection{Proof of Theorem~\ref{thm:convergence2}}
This proof exactly follows from~\cite{MB-LBFGS} and we refer the readers to the reference.

\subsection{Proof of Theorem~\ref{thm:convergence1c}}
\begin{proof}
By Lemma~\ref{lemma:Hessian}, we have 
\begin{align*}
F(\x_{k+1}) = F(\x_k - \alpha_k H_k \nabla F^{S_k}(\x_k))
&\overset{\eqref{eq:smooth2}}{\leq} F(\x_k) - \alpha_k \nabla F(\x_k)^T H_k \nabla F^{S_k}(\x_k) + \frac{\Lambda}{2}\|\alpha_k H_k\nabla F^{S_k}(\x_k)\|^2
\\& \leq 
F(\x_k) - \alpha_k \nabla F(\x_k)^T H_k \nabla F^{S_k}(\x_k) + \frac{\alpha_k^2\mu_2^2\Lambda}{2}\|\nabla F^{S_k}(\x_k)\|^2.\tagthis\label{eq:F1}
\end{align*}
Define $\phi_{k} = [F(\x_{k})-F^*]$ and  take expectation of \eqref{eq:F1} with respect to $S_k$ gives us

\begin{align*}
\mathbb{E}[\phi_{k+1}] &\overset{\eqref{eq:F1}}{\leq} \phi_k - \alpha_k \nabla F(\x_k)^T H_k \Exp[ \nabla F^{S_k}(\x_k)]
+ \frac{\alpha_k^2\mu_2^2\Lambda}{2}\Exp^{S_k}\left[ \|\nabla F^{S_k} (\x_k)\|^2\right]
\\&\overset{\eqref{eq:sgdbound1}}{\leq} \phi_k - \alpha_k\mu_1 \|\nabla F(\x_k)\|^2
+ \frac{\alpha_k^2\mu_2^2\Lambda}{2}[4\beta(b)\Lambda\kappa[F(\x_k)-F^*]+N+2\|\nabla F(\x_k)\|^2]
\\&\leq (1-2\lambda\alpha_k(\mu_1 - \alpha_k\mu_2^2\Lambda)) \phi_k +2\alpha_k^2\mu_2^2\Lambda^2\beta(b)\kappa \phi_k + \frac{\alpha_k^2\mu_2^2\Lambda N}{2}
\\&= [1-2\alpha_k (\lambda\mu_1 - \alpha_k\mu_2^2(\lambda+\Lambda\beta(b)\kappa)\Lambda)] \phi_k + \frac{\alpha_k^2\mu_2^2\Lambda N}{2}, \tagthis\label{eq:phik}
\end{align*}
where the last inequality follows from the following property of strong convexity with $\x'=\x_k, \x=\x_*$, and optimality condition $\nabla F(\x_*) = 0$,
\begin{equation*}
F(\x') \leq F(\x) + \nabla F(x)^T(y-x) + \frac1{2\mu} \|\nabla F(y)- \nabla F(x)\|^2.
\end{equation*}

Therefore, by using a constant $\alpha_k = \alpha>0$, 

\begin{align*}
\mathbb{E}[\phi_{k+1}]  - &\frac{\alpha^2\mu_2^2\Lambda N}{4\alpha(\lambda\mu_1 - \alpha\mu_2^2(\lambda+\Lambda\beta(b)\kappa)\Lambda)}
\\&\overset{\eqref{eq:phik}}{\leq}  [1-2\alpha(\lambda\mu_1 - \alpha\mu_2^2(\lambda+\Lambda\beta(b)\kappa)\Lambda)] \phi_k + \frac{\alpha^2\mu_2^2\Lambda N}{2} - \frac{\alpha^2\mu_2^2\Lambda N}{4\alpha(\lambda\mu_1 - \alpha\mu_2^2(\lambda+\Lambda\beta(b)\kappa)\Lambda)}
\\&=  [1-2\alpha(\lambda\mu_1 - \alpha\mu_2^2(\lambda+\Lambda\beta(b)\kappa)\Lambda)]  [\phi_k - \frac{\alpha^2\mu_2^2\Lambda N}{4\alpha(\lambda\mu_1 - \alpha\mu_2^2(\lambda+\Lambda\beta(b)\kappa)\Lambda)}].
\end{align*}

Take the expectation and apply the above inequality recursively, we have
\begin{align*}
\Exp[\phi_k]
&\leq  [1-2\alpha(\lambda\mu_1 - \alpha\mu_2^2(\lambda+\Lambda\beta(b)\kappa)\Lambda)]^k [\phi_0- \frac{\alpha^2\mu_2^2\Lambda N}{4\alpha(\lambda\mu_1 - \alpha\mu_2^2(\lambda+\Lambda\beta(b)\kappa)\Lambda)}]  
\\&\qquad + \frac{\alpha^2\mu_2^2\Lambda N}{4\alpha(\lambda\mu_1 - \alpha\mu_2^2(\lambda+\Lambda\beta(b)\kappa)\Lambda)} 
\\&= [1-2\alpha(\lambda\mu_1 - \alpha\mu_2^2(\lambda+\Lambda\beta(b)\kappa)\Lambda)]^k \phi_0 
\\&\qquad+ \{1- [1-2\alpha(\lambda\mu_1 - \alpha\mu_2^2(\lambda+\Lambda\beta(b)\kappa)\Lambda)] ^k \}\frac{\alpha\mu_2^2\Lambda N}{4(\lambda\mu_1 - \alpha\mu_2^2(\lambda+\Lambda\beta(b)\kappa)\Lambda)}.
\end{align*}

We need the learning rate to satisfy 
\begin{equation*}
0<1-2\alpha(\lambda\mu_1 - \alpha\mu_2^2(\lambda+\Lambda\beta(b)\kappa)\Lambda)<1,
\end{equation*}
and thus,
\begin{equation*}
0<\alpha<\tfrac{\lambda\mu_1}{\mu_2^2(\lambda + \Lambda\beta(b)\kappa)\Lambda}.
\end{equation*}

If we further assume that $f_i$s are convex and use \eqref{eq:sgdbound2} instead of \eqref{eq:sgdbound1}, then similarly we have
\begin{align*}
\mathbb{E}[\phi_{k+1}]
&\leq [1-2\alpha_k (\lambda\mu_1 - \alpha_k\mu_2^2(\lambda+\Lambda\beta(b))\Lambda)] \phi_k + \frac{\alpha_k^2\mu_2^2\Lambda N}{2}. \tagthis\label{eq:phik2}
\end{align*}

Hence, following the steps above, the bound can be expressed as:
\begin{align*}
\Exp[\phi_k]
&\leq [1-2\alpha(\lambda\mu_1 - \alpha\mu_2^2(\lambda+\Lambda\beta(b))\Lambda)]^k \phi_0 
\\&\qquad+ \{1- [1-2\alpha(\lambda\mu_1 - \alpha\mu_2^2(\lambda+\Lambda\beta(b))\Lambda)] ^k \}\frac{\alpha\mu_2^2\Lambda N}{4(\lambda\mu_1 - \alpha\mu_2^2(\lambda+\Lambda\beta(b))\Lambda)},
\end{align*}
with 
\begin{equation*}
0<\alpha<\tfrac{\lambda\mu_1}{\mu_2^2(\lambda + \Lambda\beta(b))\Lambda}.
\end{equation*}

\end{proof}

\subsection{Proof of Theorem~\ref{thm:convergenceC}}
\begin{proof}

Let us prove the conclusion by induction. First, when $k=0$, we have
\begin{equation*}
\tfrac{G(\alpha, E)}{k+E} \overset{\eqref{eq:G}}{\geq} \tfrac{E\Exp[F(\x_0)-F(\x_*)]}{E} = \Exp[F(\x_0)-F(\x_*)].
\end{equation*}
Next, let us assume that with $\alpha_k = \tfrac{\alpha}{k+E}$, the following inequality holds,
\begin{equation}\label{eq:FkF*}
\Exp [ F(\x_k)-F(\x_*)] \leq \tfrac{G(\alpha, E)}{k+E}.
\end{equation}

Since $f_i$s are convex, take the total expectation of \eqref{eq:phik2} with $\phi_k = [F(\x_k) - F(\x_*)]$, and use the learning rate $\alpha_k = \tfrac{\alpha}{k+E}$, we have
\begin{align*}
\mathbb{E}[F(\x_{k+1} ) - F(\x_*)] &\overset{\eqref{eq:phik2}}{\leq} [1-2\alpha_k (\lambda\mu_1 - \alpha_k\mu_2^2(\lambda+\Lambda\beta(b))\Lambda)] \Exp [F(\x_k) - F(\x_*)] + \frac{\alpha_k^2\mu_2^2\Lambda N}{2}
\\&\overset{\eqref{eq:FkF*}}{\leq}
\left[1-\tfrac{2\alpha\lambda\mu_1}{k+E} + \tfrac{2\alpha^2\mu_2^2\Lambda(\lambda+\Lambda\beta(b))}{(k+E)^2}\right] \tfrac{G(\alpha, E)}{k+E} + \frac{\alpha^2\mu_2^2\Lambda N}{2(k+E)^2}
\\&\overset{k\geq 0}{\leq} 
\left[\tfrac{1}{k+E}-\tfrac{2\alpha\lambda\mu_1}{(k+E)^2} + \tfrac{2\alpha^2\mu_2^2\Lambda(\lambda+\Lambda\beta(b))}{(k+E)^2E}\right] G(\alpha, E) + \frac{\alpha^2\mu_2^2\Lambda N}{2(k+E)^2}.\tagthis\label{eq:proof1}
\end{align*}

Choose $E, \alpha>0$ such that the following holds,
\begin{align*}
\left[1-\tfrac{2\alpha\lambda\mu_1}{k+E} + \tfrac{2\alpha^2\mu_2^2\Lambda(\lambda+\Lambda\beta(b))}{(k+E)^2}\right] &\geq 0,\\
2\alpha\lambda\mu_1-1-\tfrac{2\alpha^2\mu_2^2\Lambda(\lambda+\Lambda\beta(b))}{E} &>0, \tagthis\label{eq:Ebound}
\end{align*}
then we obtain that
\begin{align*}
\mathbb{E}[F(\x_{k+1} ) - F(\x_*)] 
&\overset{\eqref{eq:proof1}}{\leq} 
\left[\tfrac{1}{k+E}-\tfrac{2\alpha\lambda\mu_1}{(k+E)^2} + \tfrac{2\alpha^2\mu_2^2\Lambda(\lambda+\Lambda\beta(b)\kappa)}{(k+E)^2E}\right] G(\alpha, E) + \frac{\alpha^2\mu_2^2\Lambda N}{2(k+E)^2}
\\&\overset{\eqref{eq:G}}{\leq}\tfrac{G(\alpha, E)}{k+E} + \tfrac{G(\alpha, E)}{(k+E)^2}\left[-2\alpha\lambda\mu_1 + \tfrac{2\alpha^2\mu_2^2\Lambda(\lambda+\Lambda\beta(b)\kappa)}{E}\right] + \left[2\alpha\lambda\mu_1-1-\tfrac{2\alpha^2\mu_2^2\Lambda(\lambda+\Lambda\beta(b)\kappa)}{E}\right] \tfrac{G(\alpha, E)}{(k+E)^2}
\\&= \tfrac{G(\alpha, E)}{k+E} \tfrac{k+E-1}{k+E} = \tfrac{(k+E)^2-1}{(k+E)^2(k+E+1)}G(\alpha, E)
\\&\leq \tfrac{G(\alpha, E)}{k+E+1}.
\end{align*}
Therefore the conclusion is proven by replacing $F(\x_*)$ with $F^*$ and enforce the following so that ~\eqref{eq:Ebound} has solutions,
\begin{equation*}
E> \tfrac{2\mu_2^2}{\mu_1^2} \tfrac{\Lambda(\lambda+\Lambda\beta)}{\lambda^2} =  \tfrac{2\mu_2^2}{\mu_1^2} (\kappa+\beta\kappa^2).
\end{equation*}
\end{proof}

\newpage

\section{Additional Experiments}
In this section, we provide additional numerical results with LBFGS-H (LBFGS-F), LBFGS-S, LBFGS, ADAM, ADAGRAD and SGD.
\subsection{Results on logistic regression (convex), \emph{ijcnn1}}
The first experiment is conducted for the logistic regression problem on \emph{ijcnn1}, which has been discussed in Section~\ref{sec:experiments} in details. Additionally, we present the figure of training loss, and the 2nd and 5th rows are the zoom-in versions of the 1st and 4th rows, respectively.

\subsubsection{Small Batch Sizes}

\begin{figure}[H]
\centering
 \epsfig{file=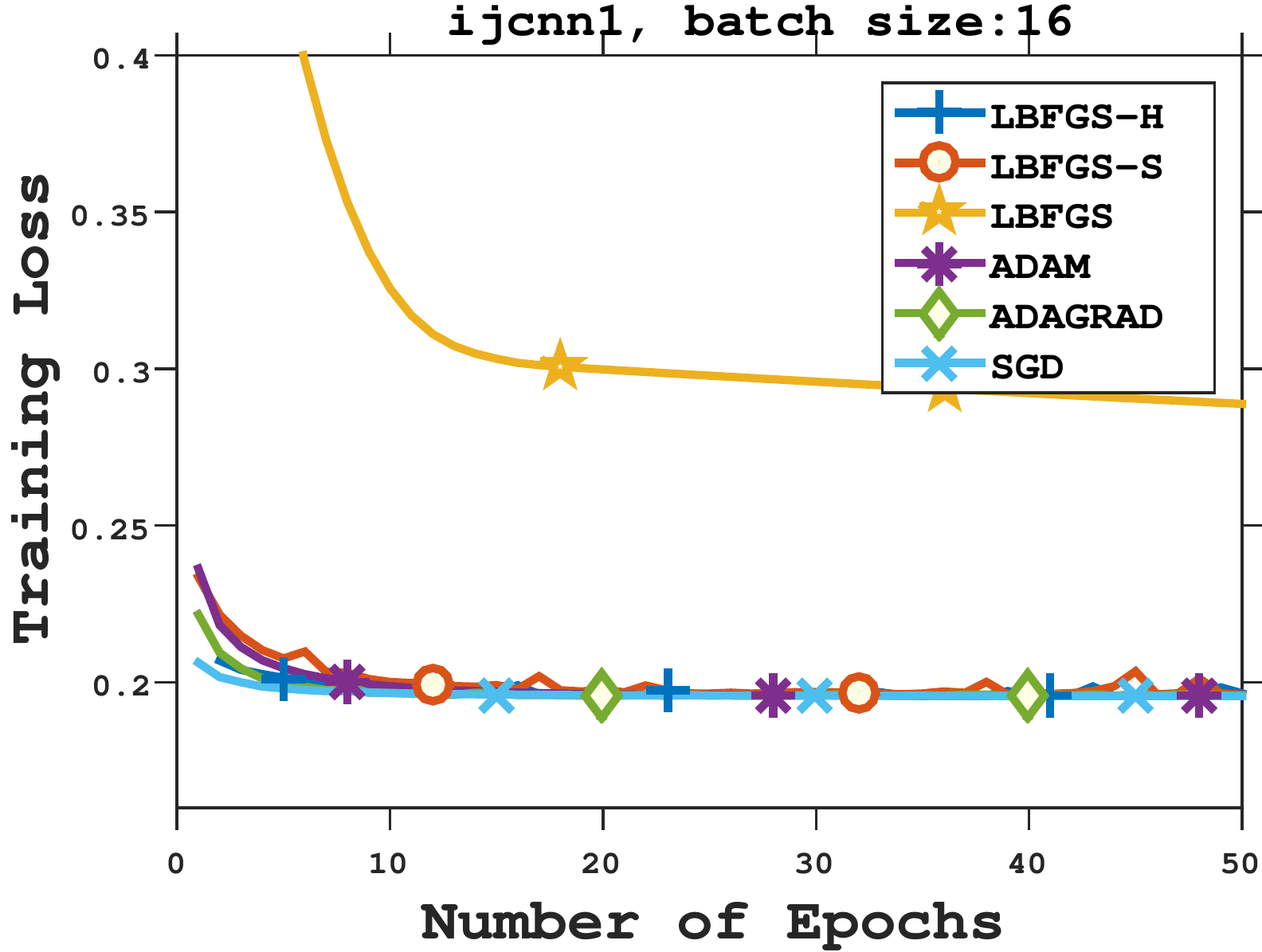,width=0.32\textwidth}
  \epsfig{file=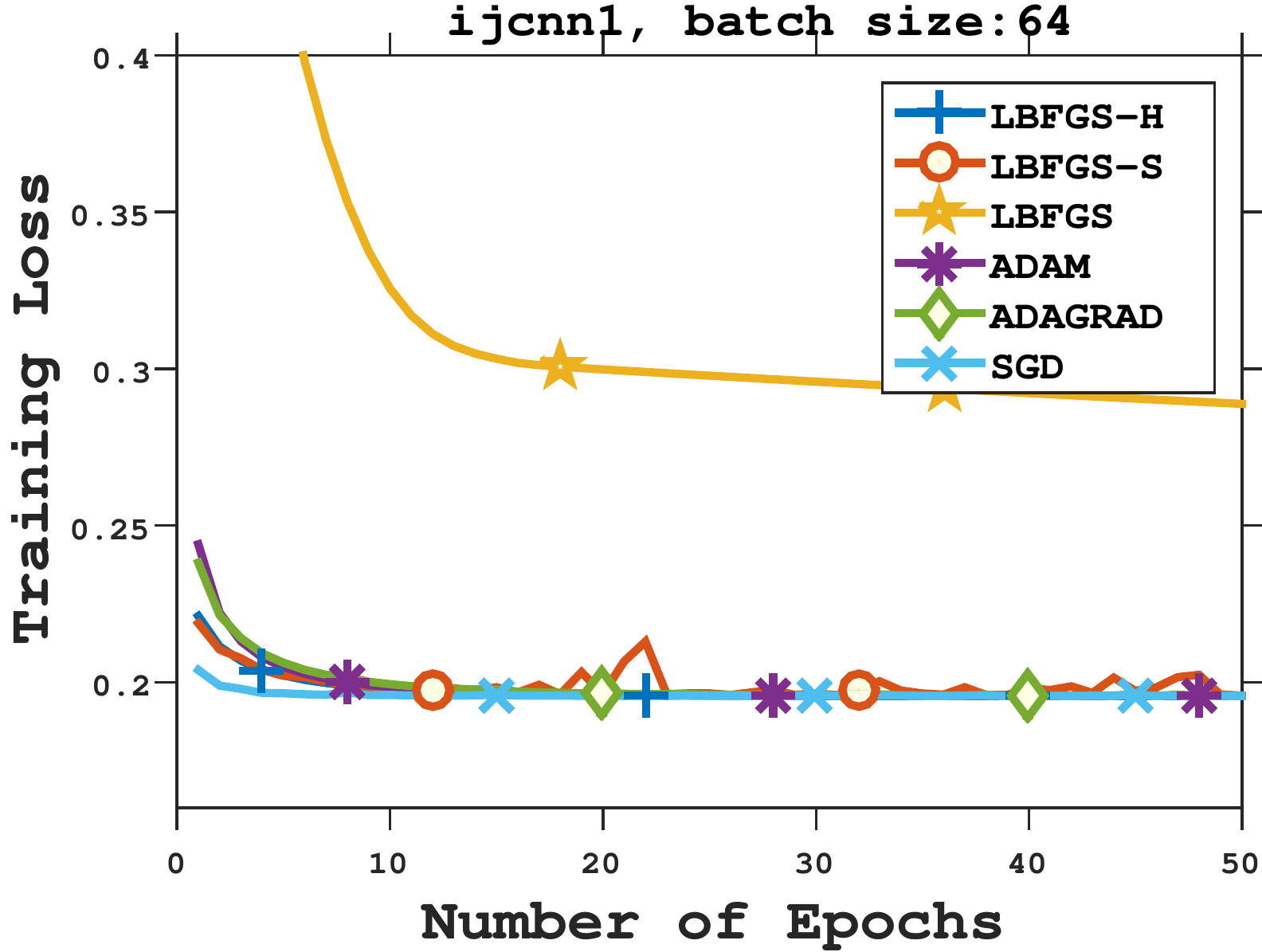,width=0.32\textwidth} 
  \epsfig{file=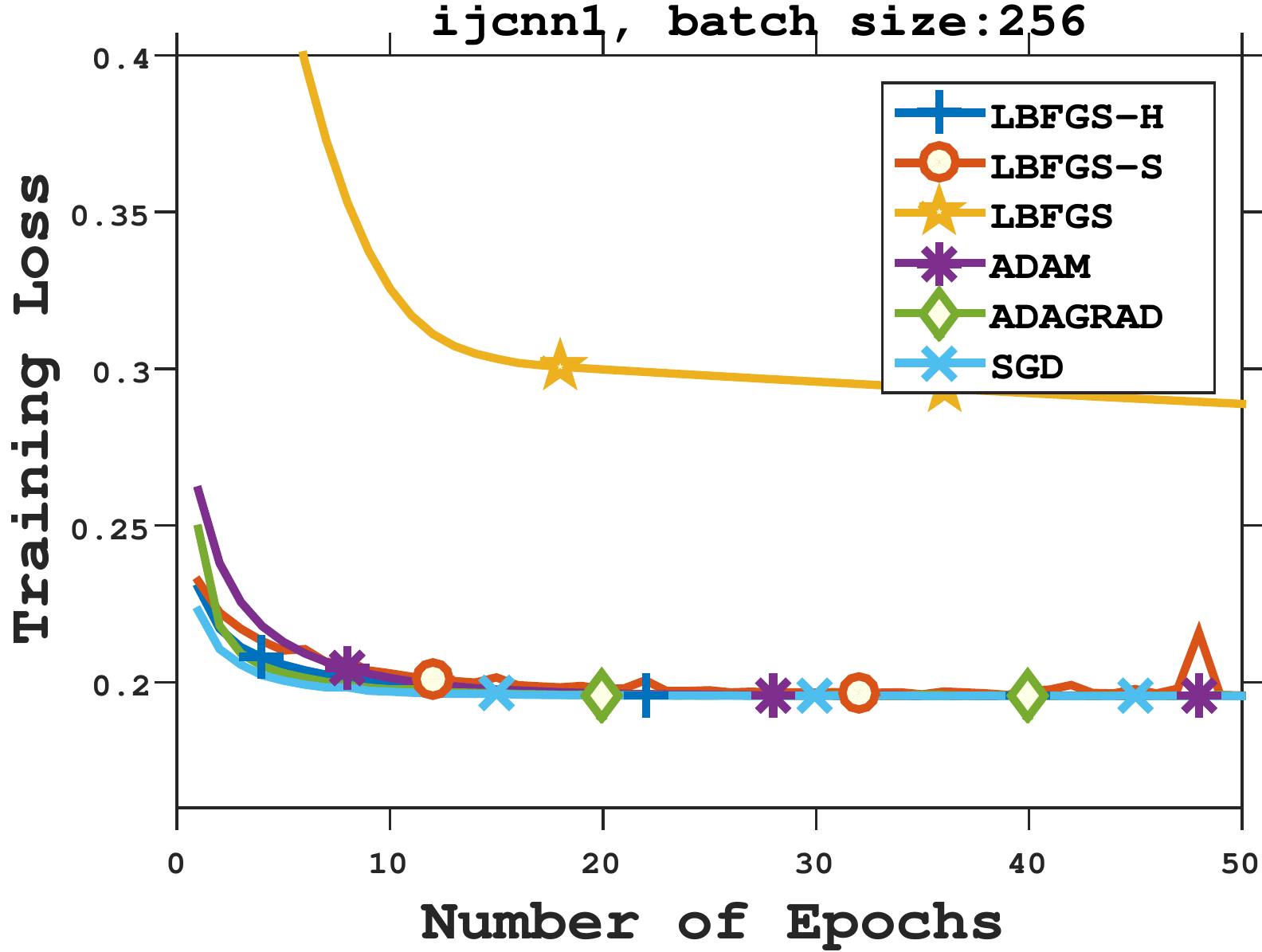,width=0.32\textwidth}

 \epsfig{file=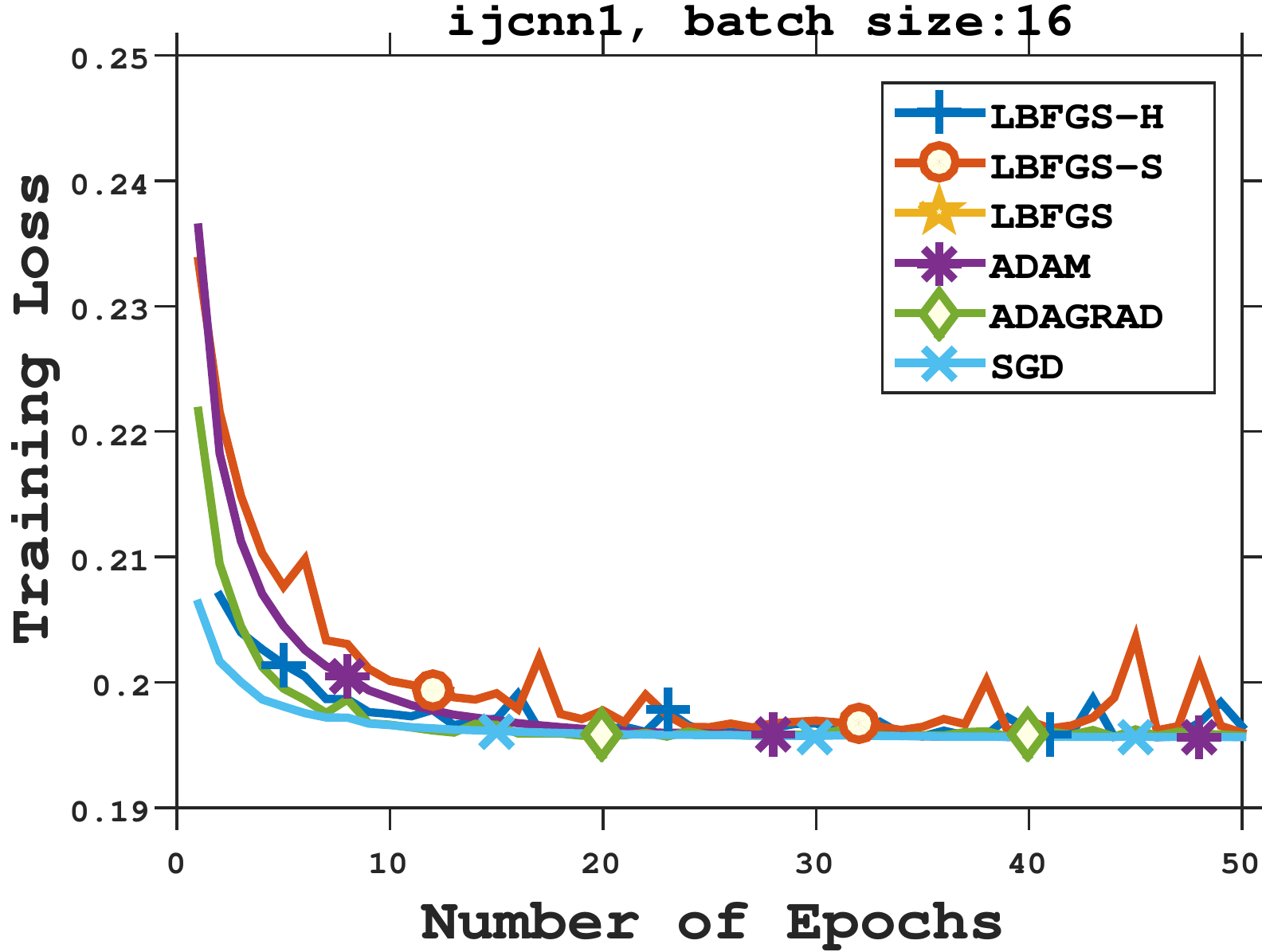,width=0.32\textwidth}
  \epsfig{file=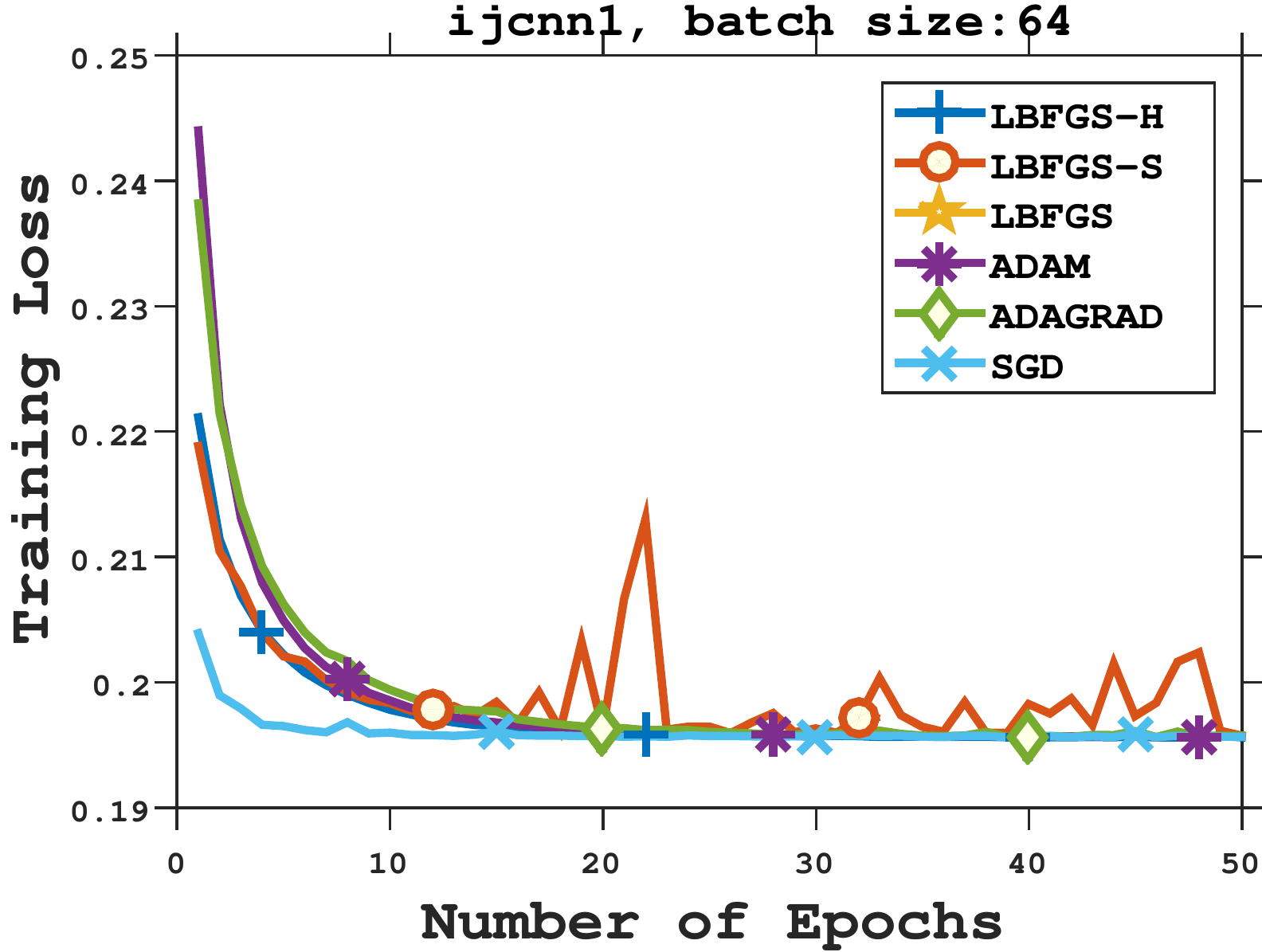,width=0.32\textwidth} 
  \epsfig{file=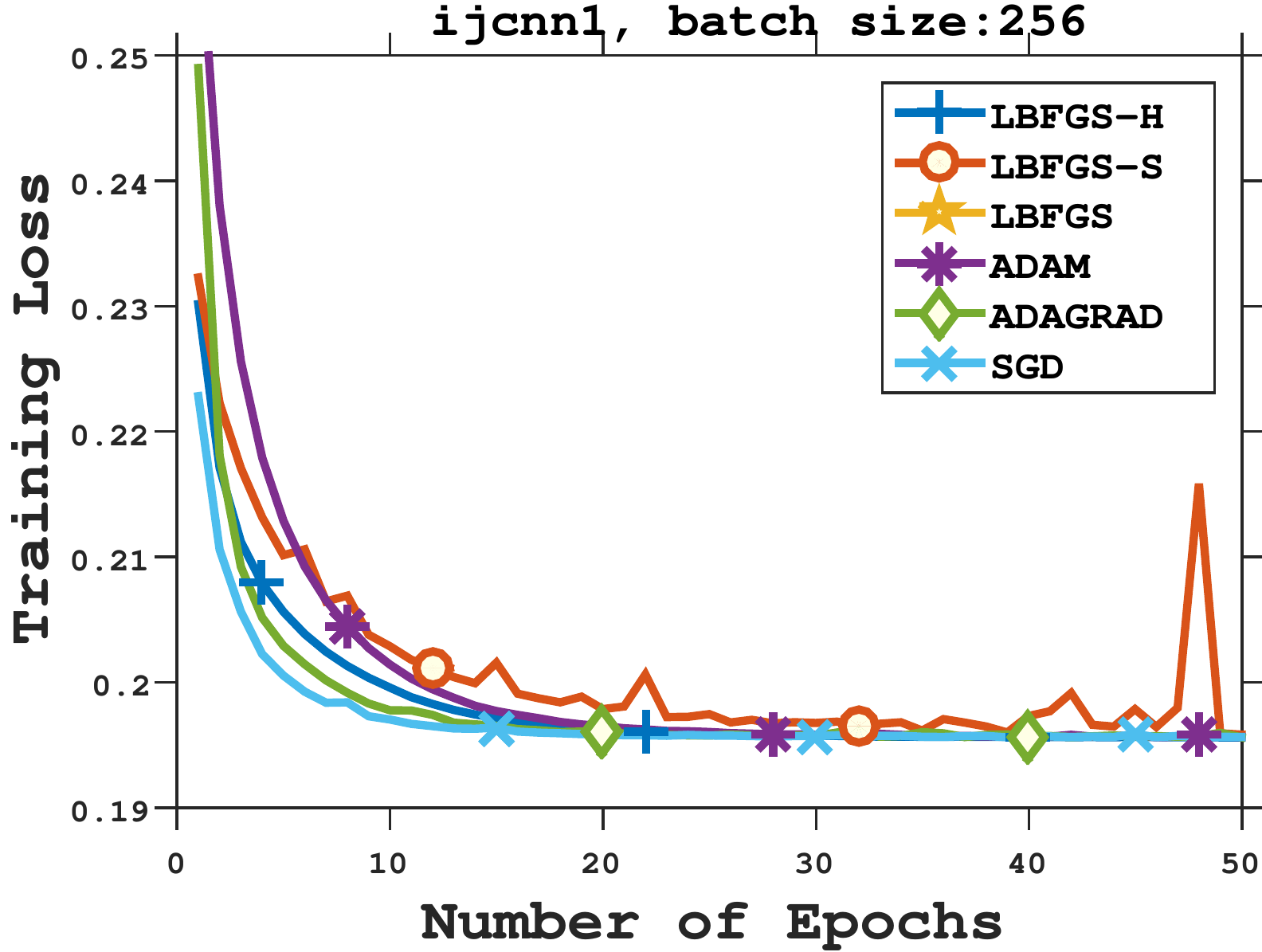,width=0.32\textwidth}

 \epsfig{file=Figs/ijcnn1_loss16_2.eps,width=0.32\textwidth}
 \epsfig{file=Figs/ijcnn1_loss64_2.eps,width=0.32\textwidth} 
  \epsfig{file=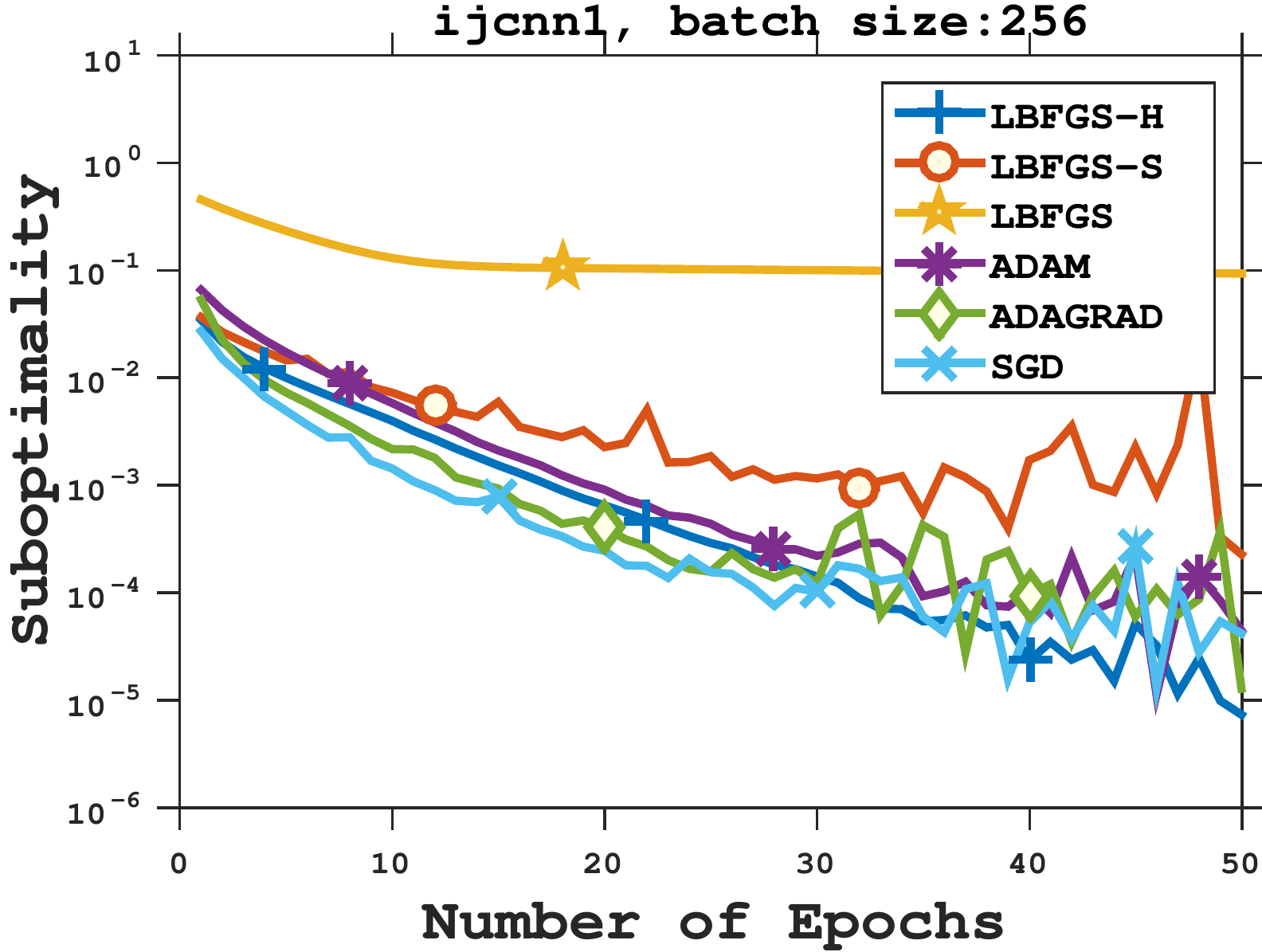,width=0.32\textwidth}

   \epsfig{file=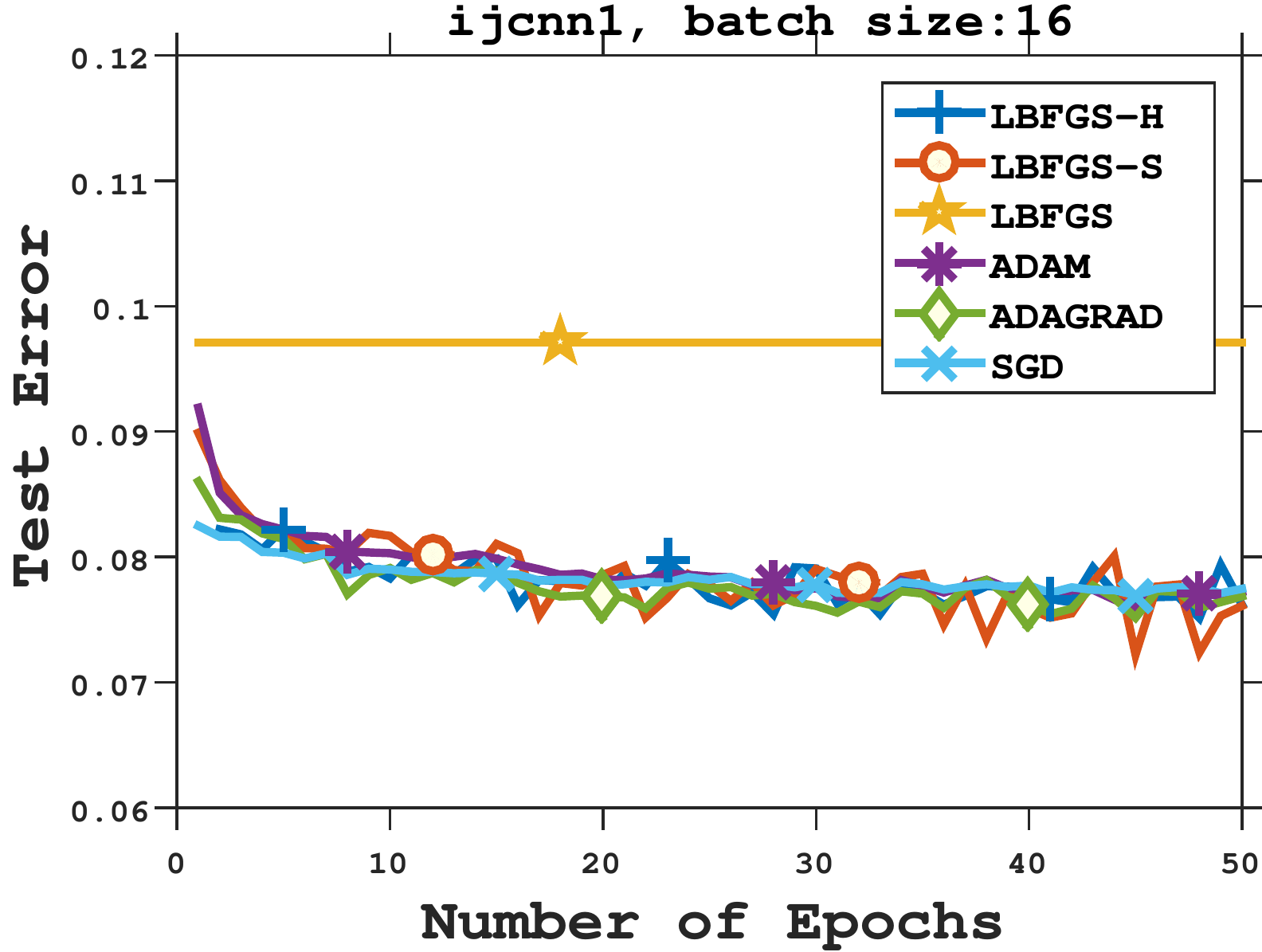,width=0.32\textwidth}
   \epsfig{file=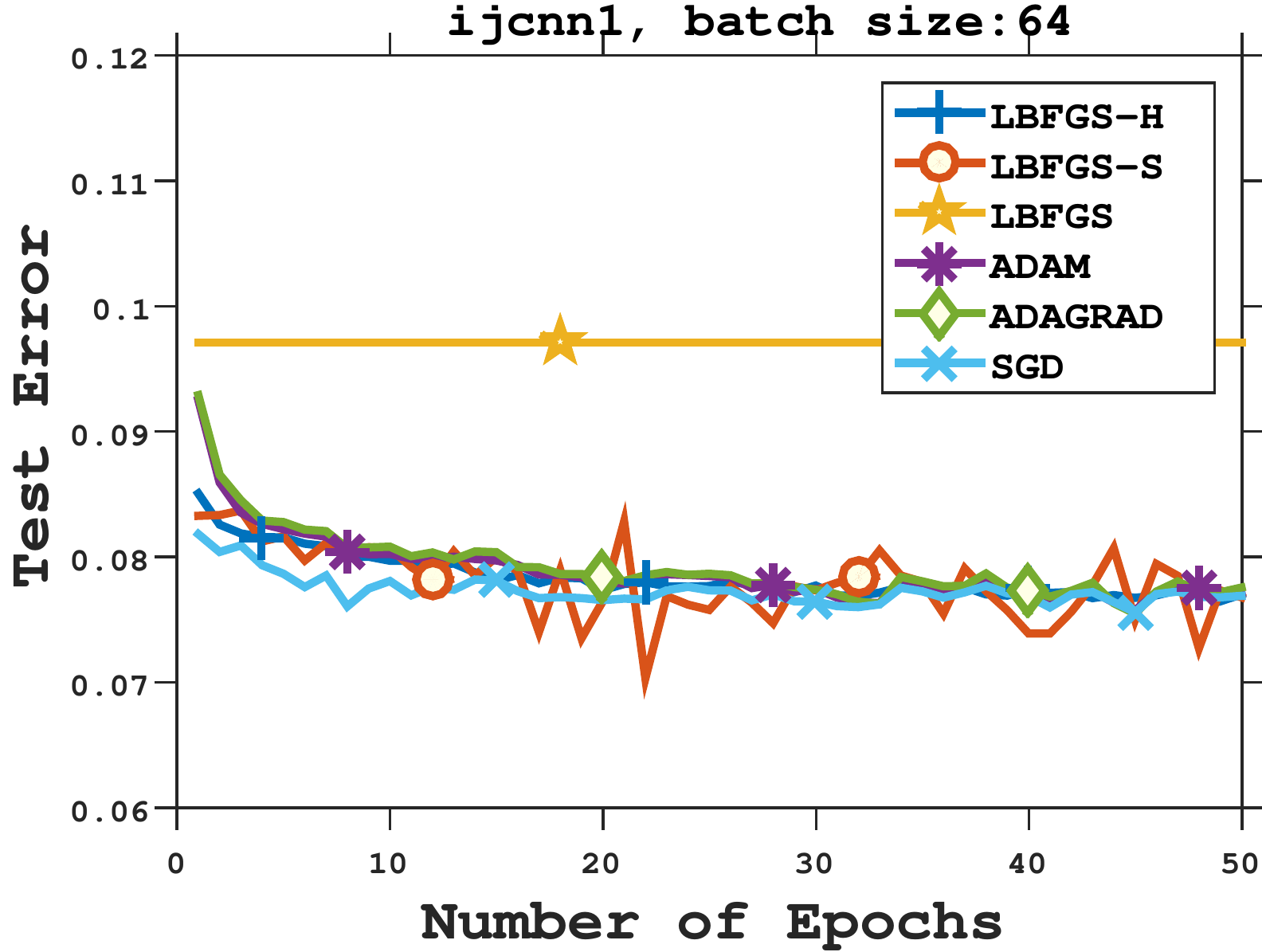,width=0.32\textwidth} 
    \epsfig{file=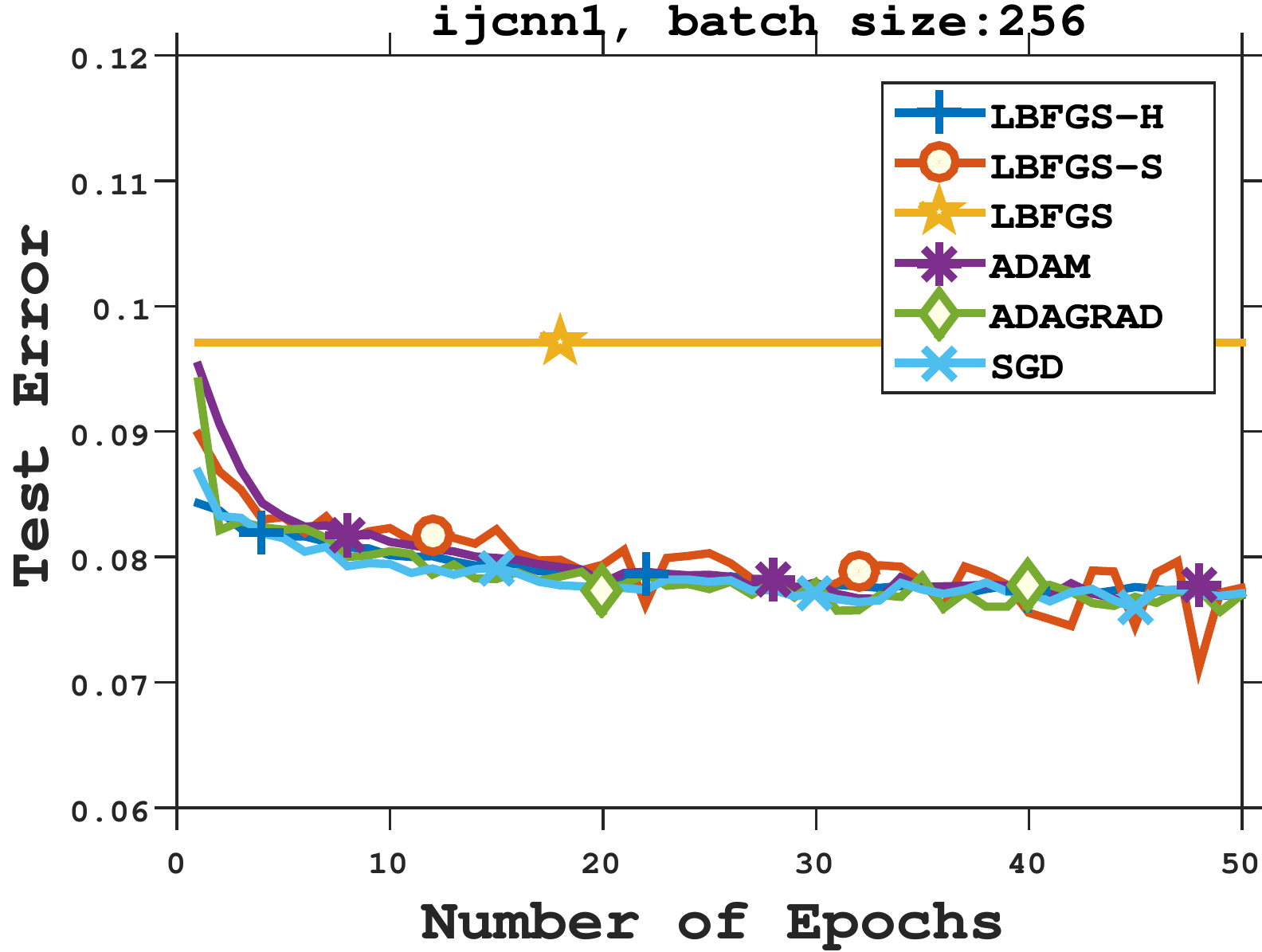,width=0.32\textwidth}
    \epsfig{file=Figs/ijcnn1_error16closer.eps,width=0.32\textwidth}
   \epsfig{file=Figs/ijcnn1_error64closer.eps,width=0.32\textwidth} 
    \epsfig{file=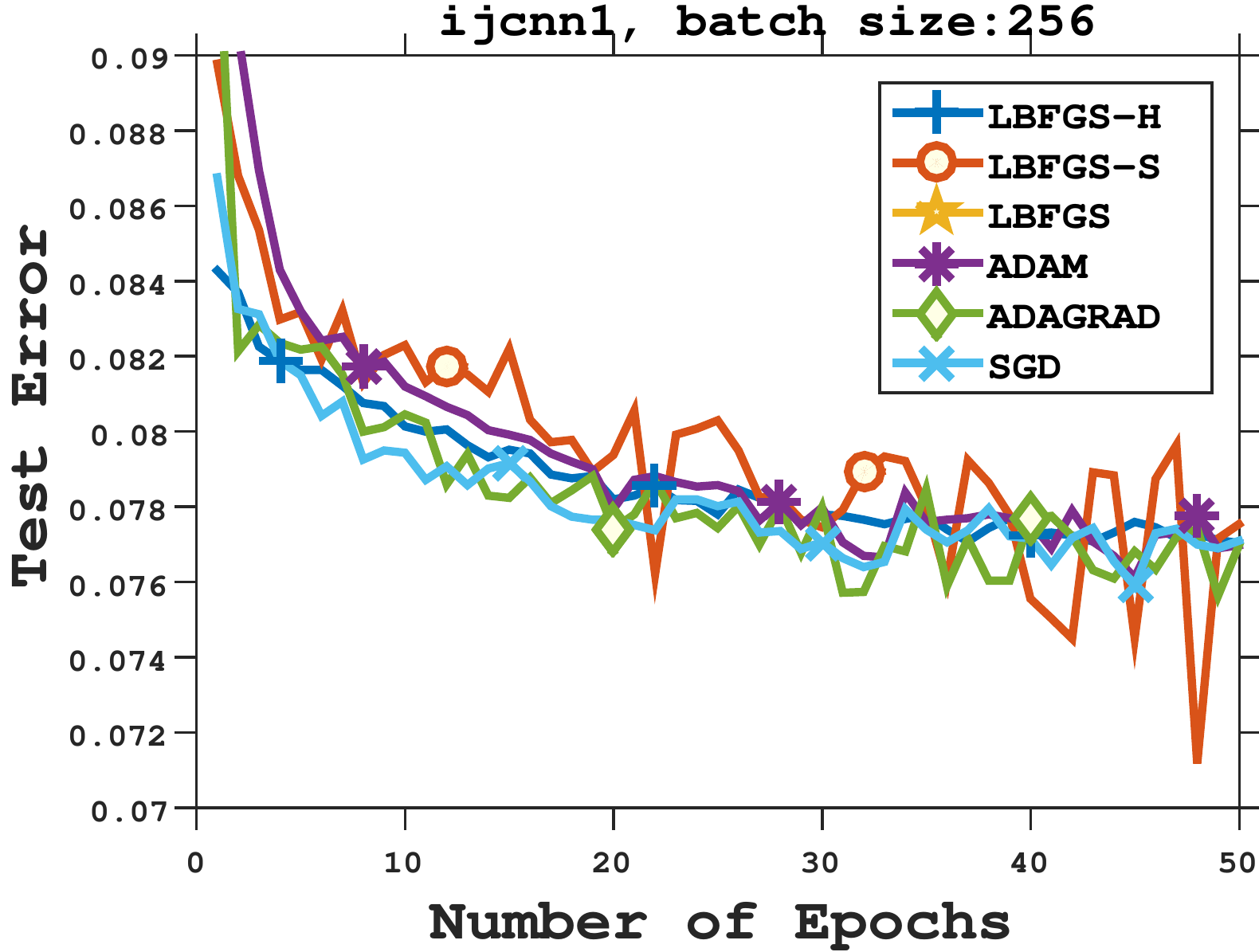,width=0.32\textwidth}
 
 \caption{\footnotesize Comparisons of training loss (top 2 rows), sub-optimality (middle row) and test errors (bottom 2 rows) for different algorithms with batch sizes 16, 64, 256 on \emph{ijcnn1}, convex, logistic regression.}
   \label{fig:add1}
 \end{figure}

Figure~\ref{fig:add1} presents results on the small batch sizes $b=16, 64, 256$, as discussed in Section~\ref{sec:experiments}.
 
  \newpage

 \subsubsection{Larger Batch Sizes}
\begin{figure}[h]
\centering
 \epsfig{file=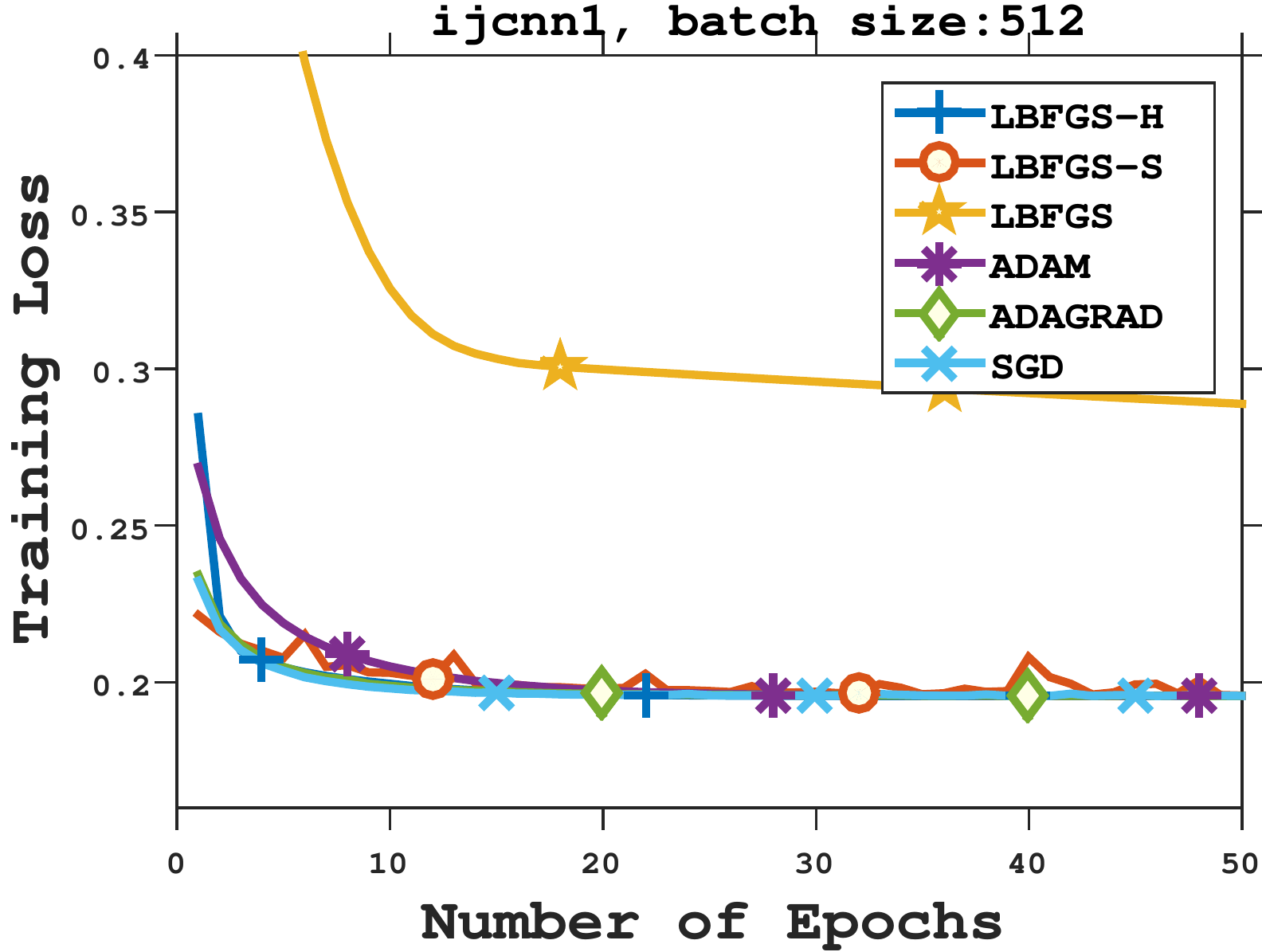,width=0.24\textwidth}
  \epsfig{file=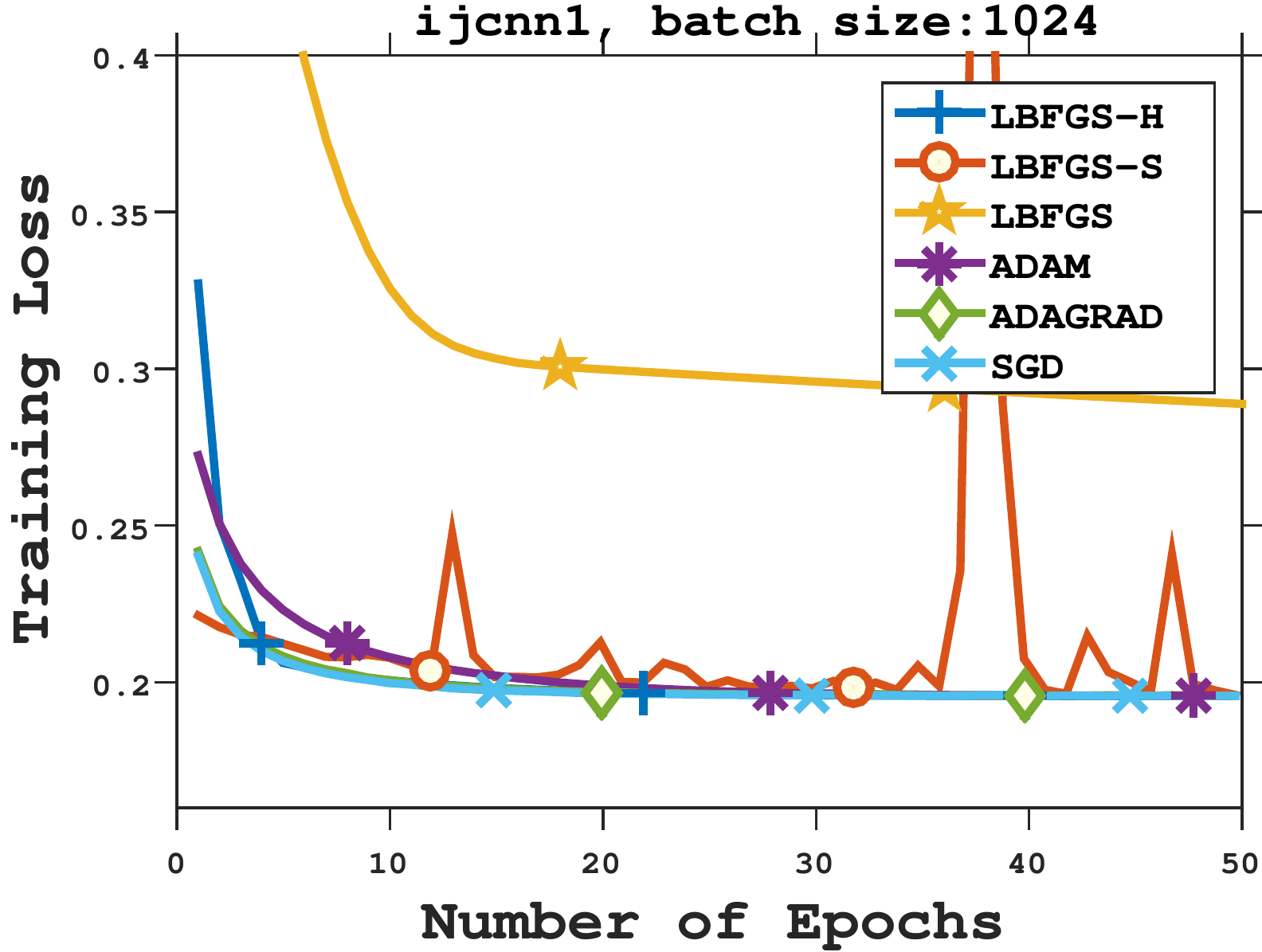,width=0.24\textwidth} 
  \epsfig{file=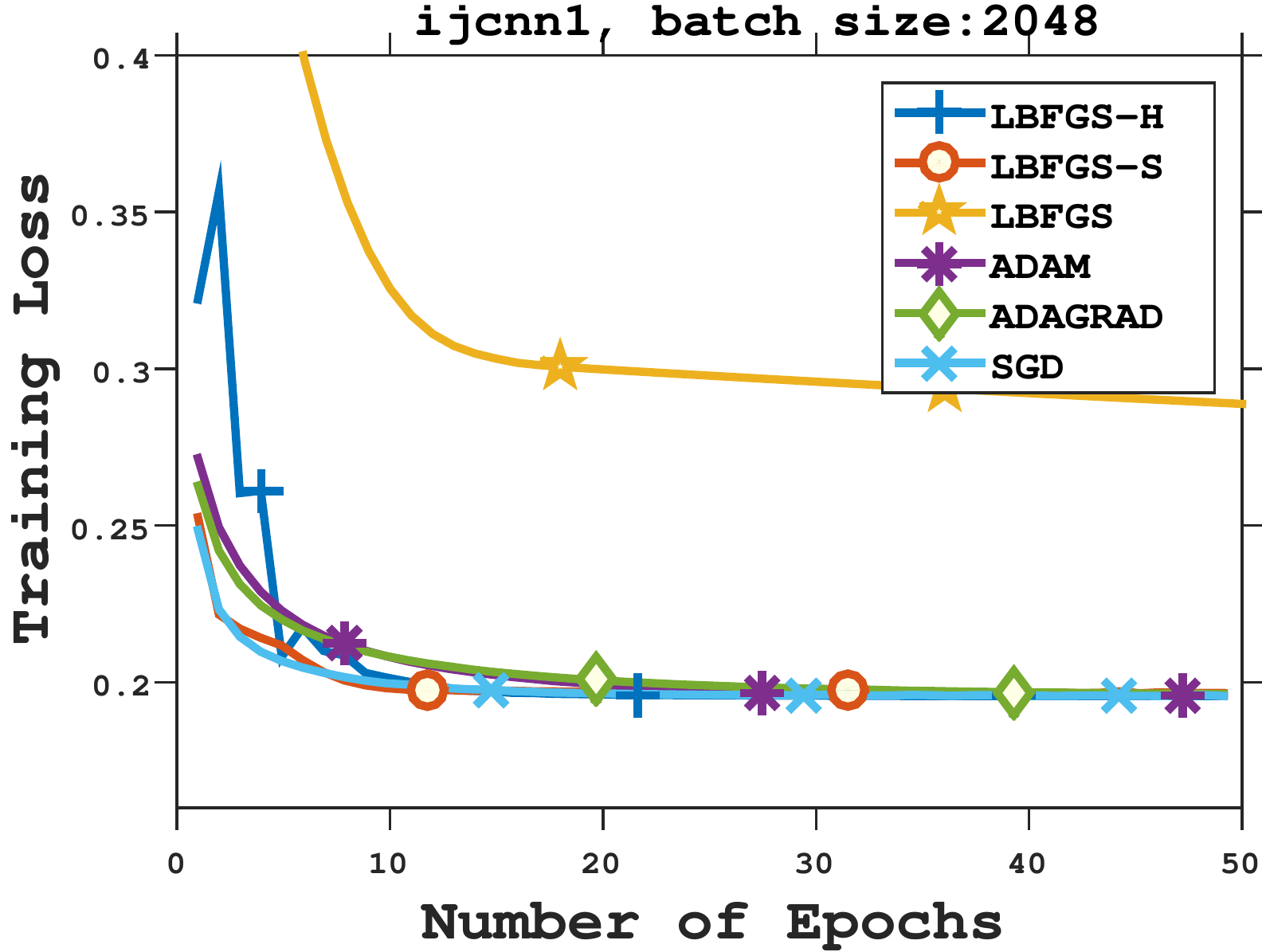,width=0.24\textwidth} 
  \epsfig{file=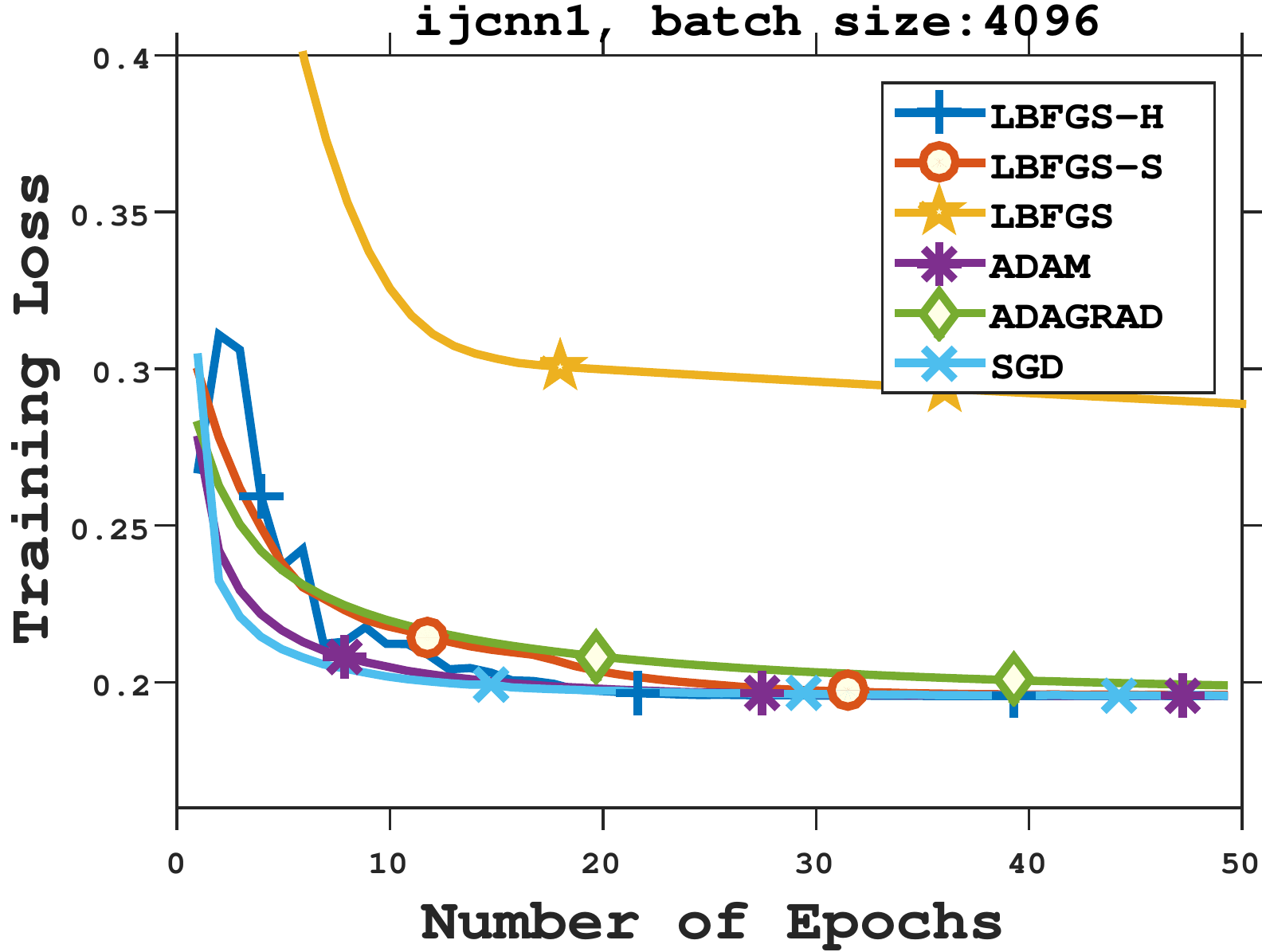,width=0.24\textwidth} 

 \epsfig{file=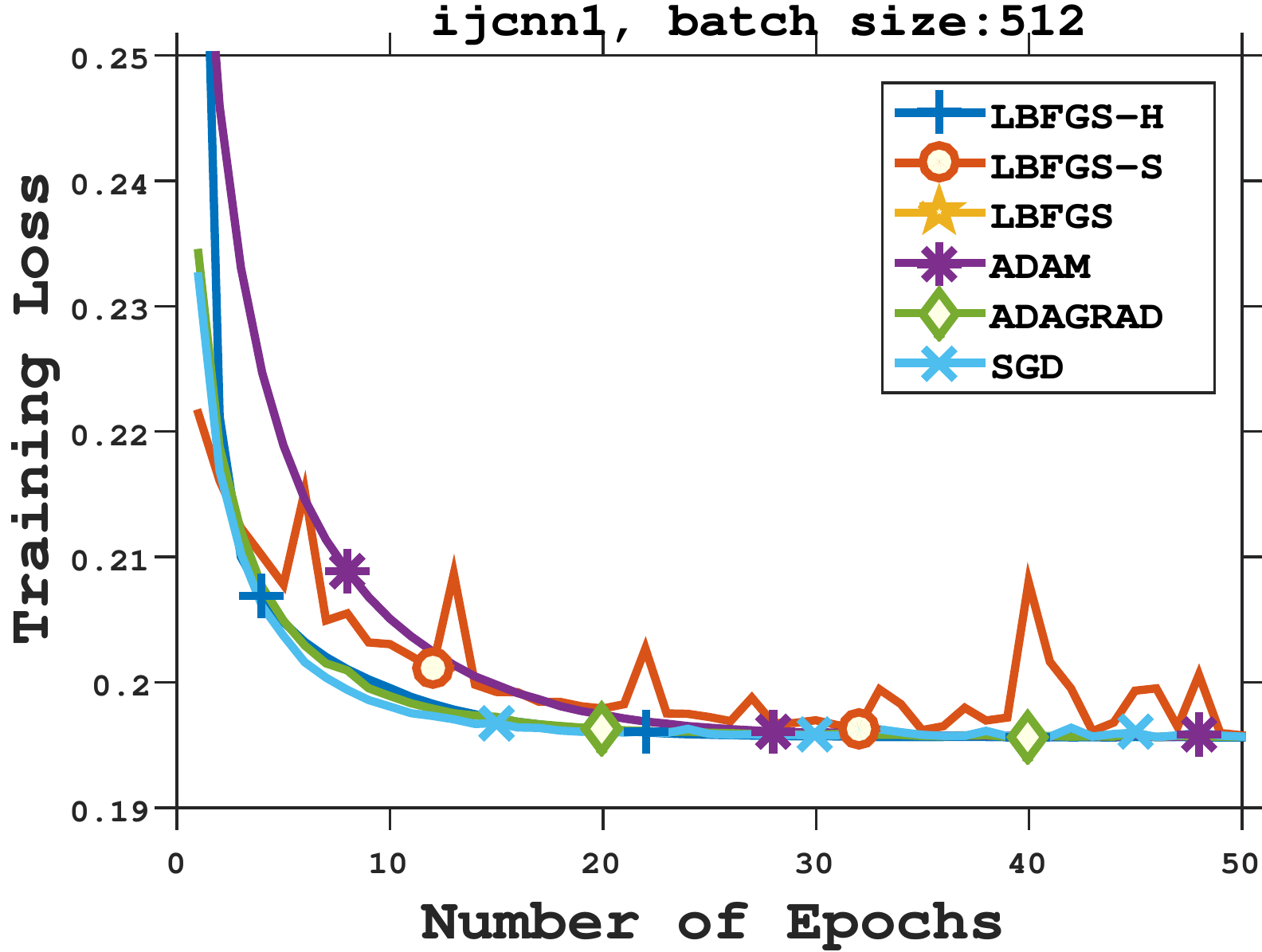,width=0.24\textwidth}
  \epsfig{file=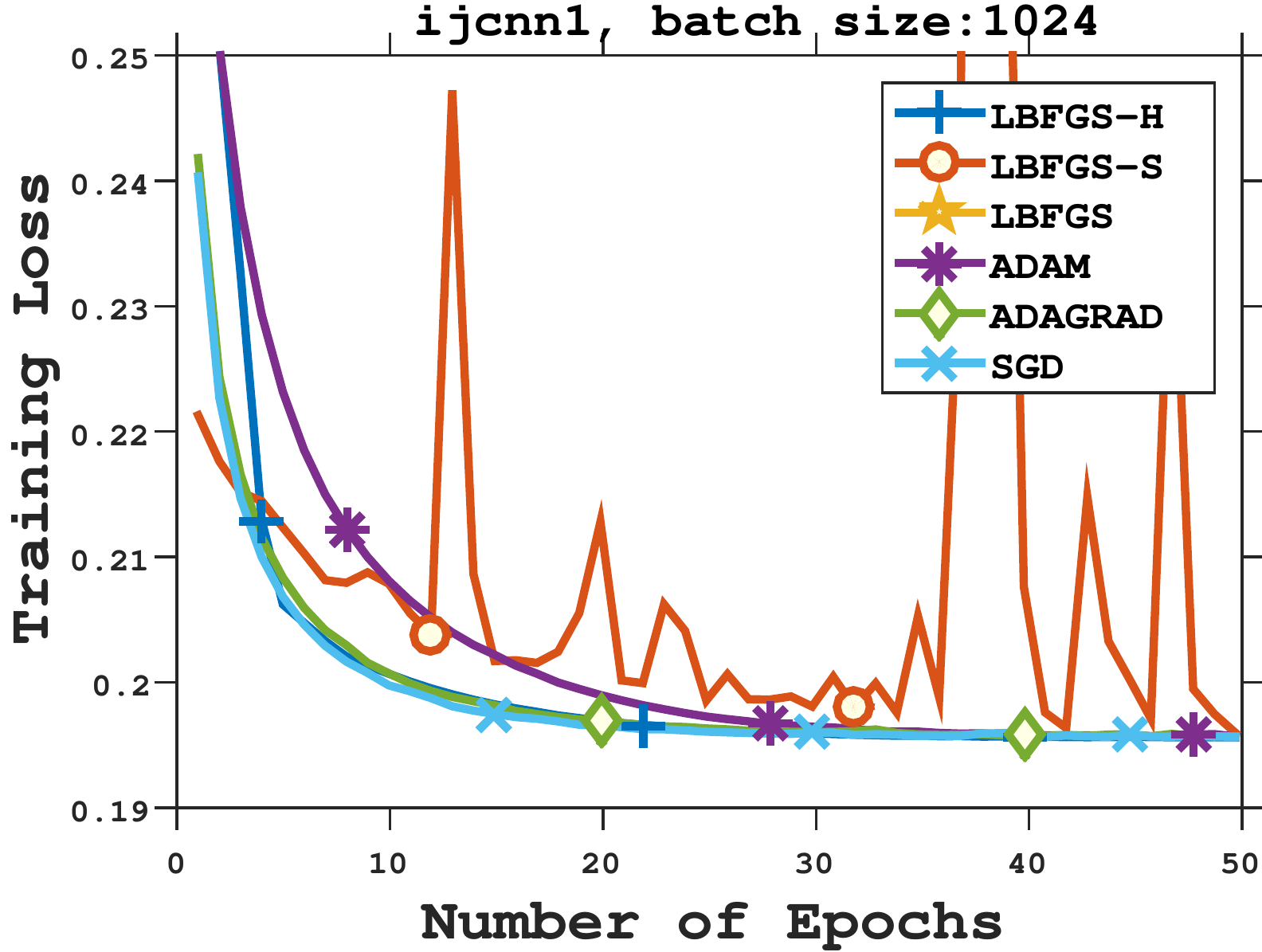,width=0.24\textwidth} 
  \epsfig{file=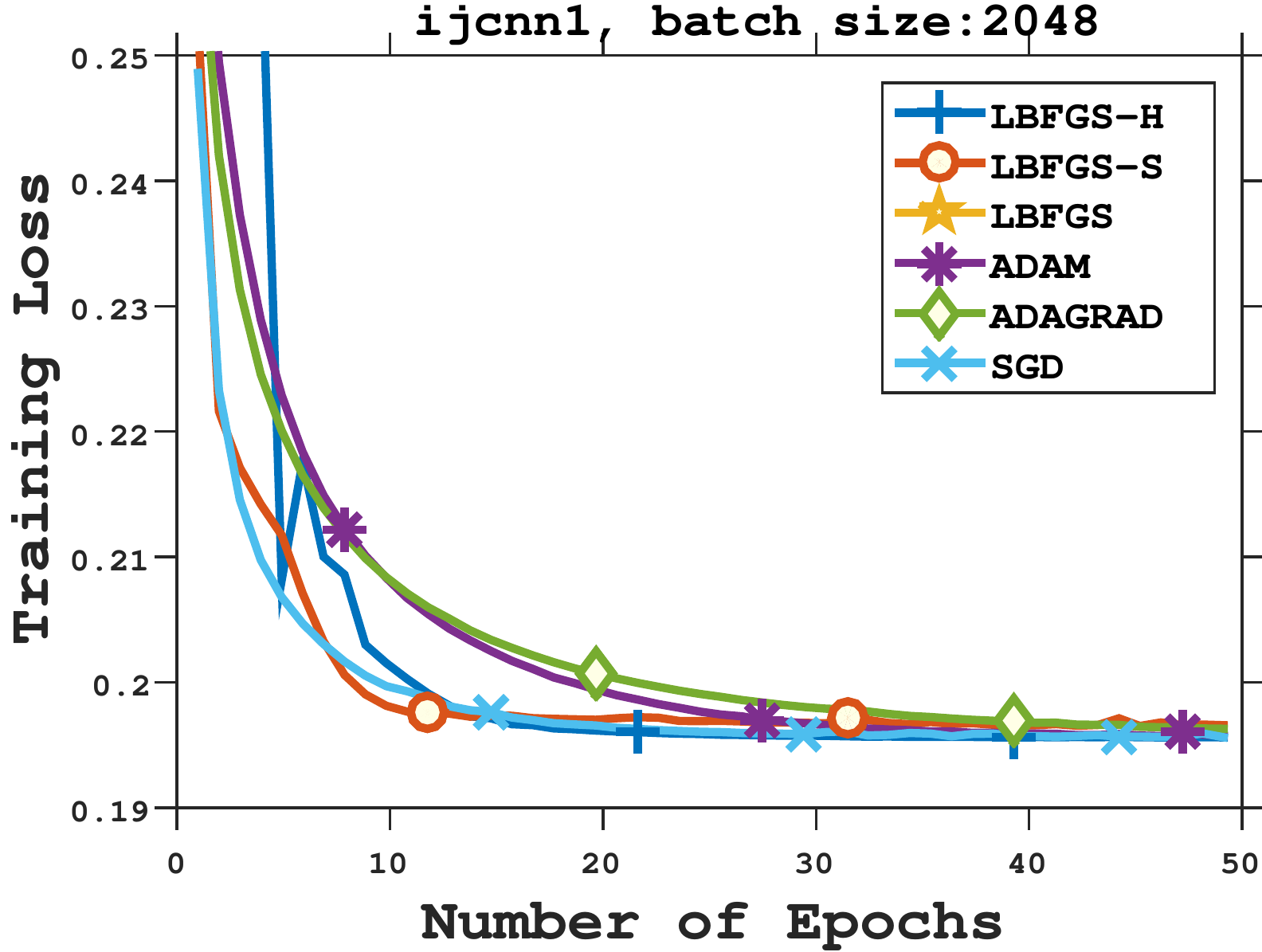,width=0.24\textwidth} 
  \epsfig{file=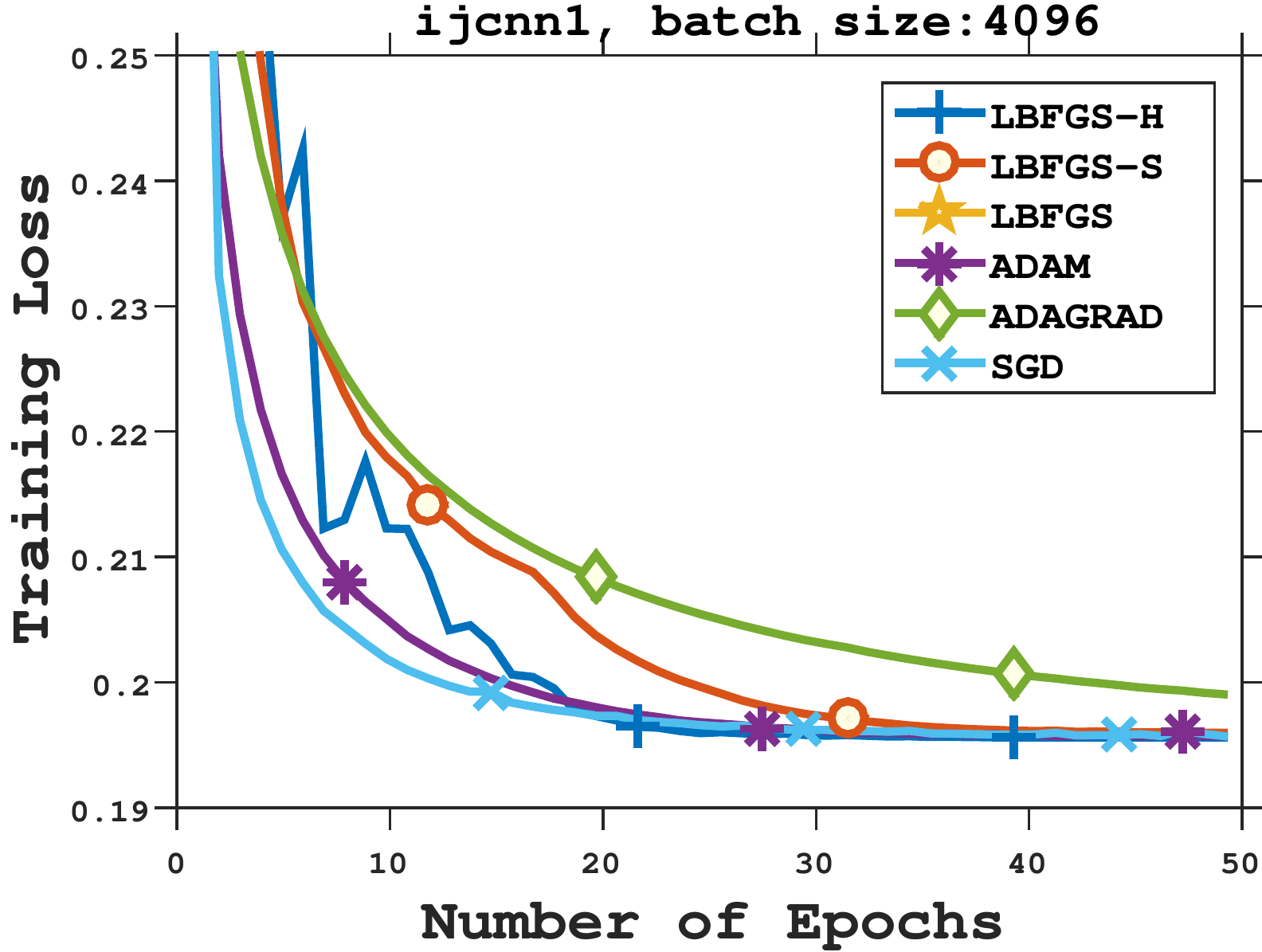,width=0.24\textwidth}

 \epsfig{file=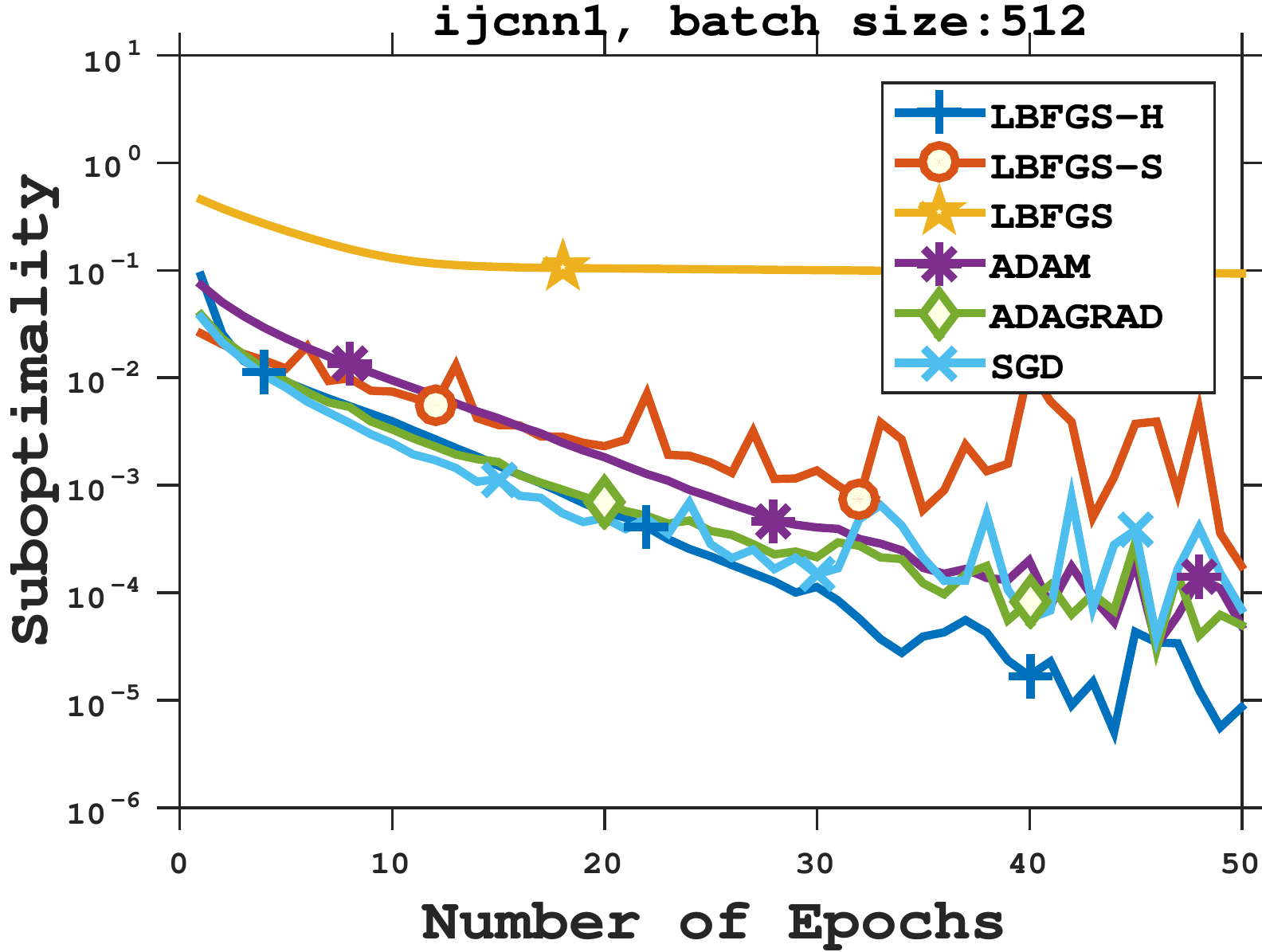,width=0.24\textwidth}
 \epsfig{file=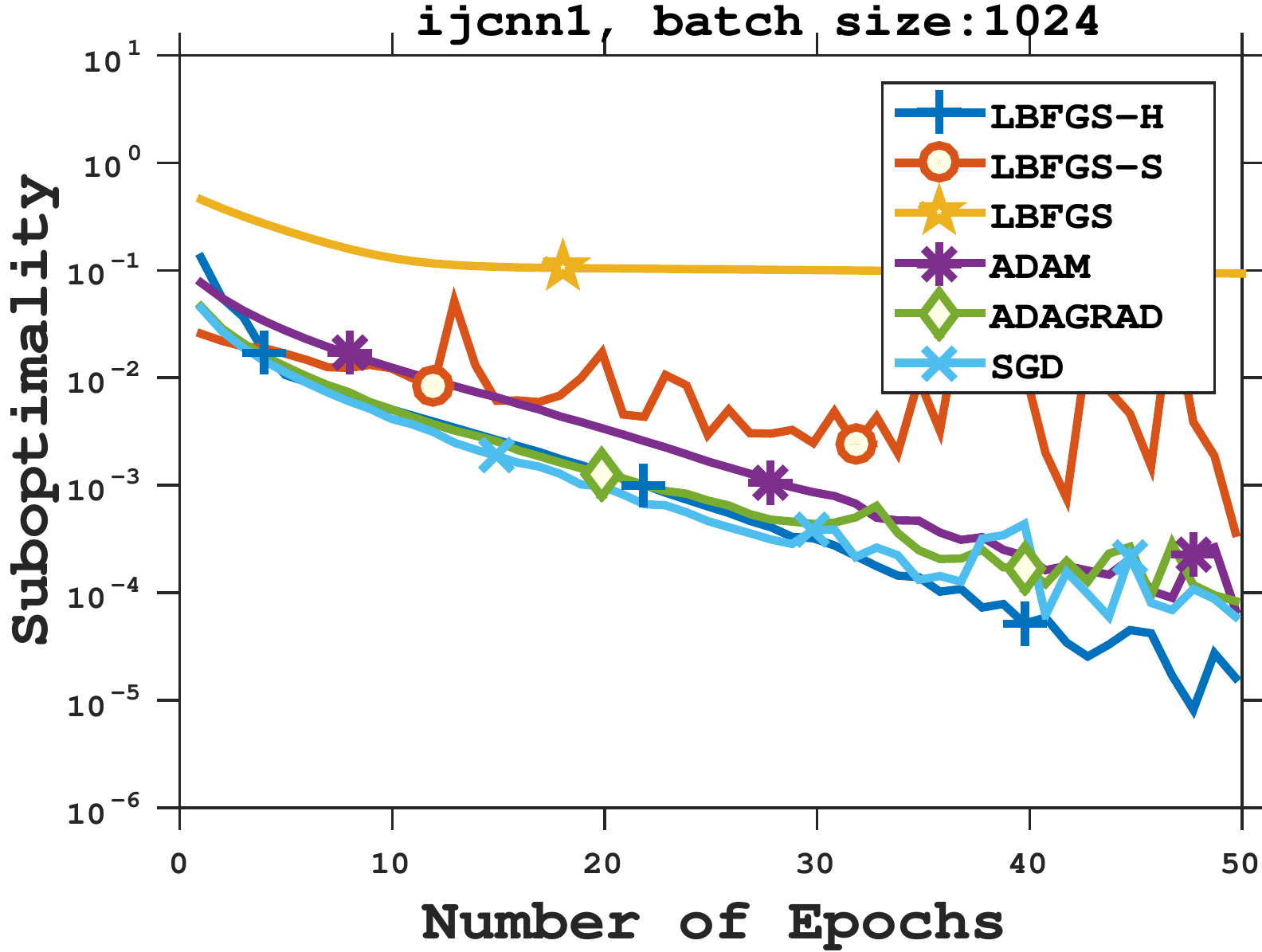,width=0.24\textwidth} 
  \epsfig{file=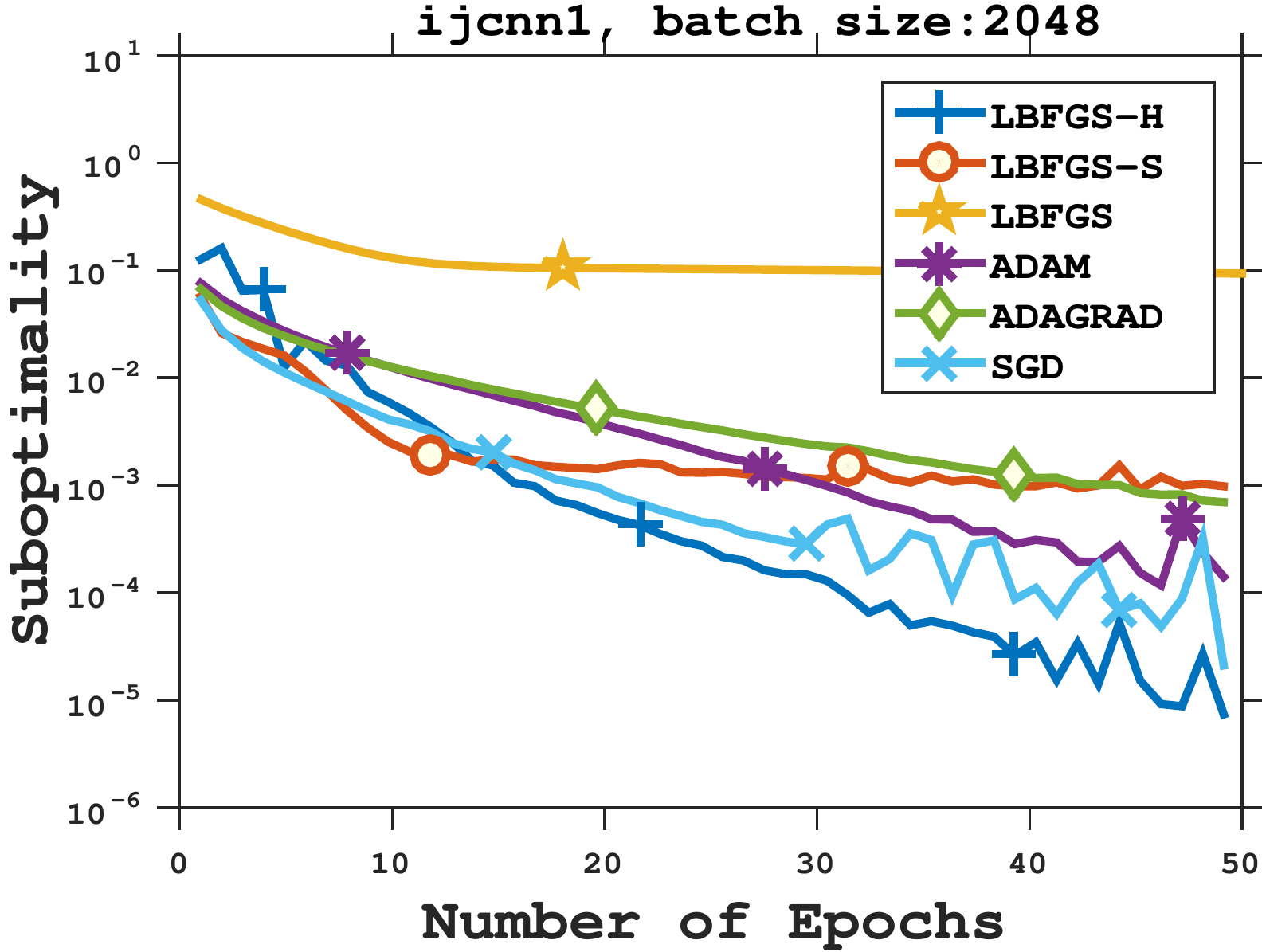,width=0.24\textwidth} 
   \epsfig{file=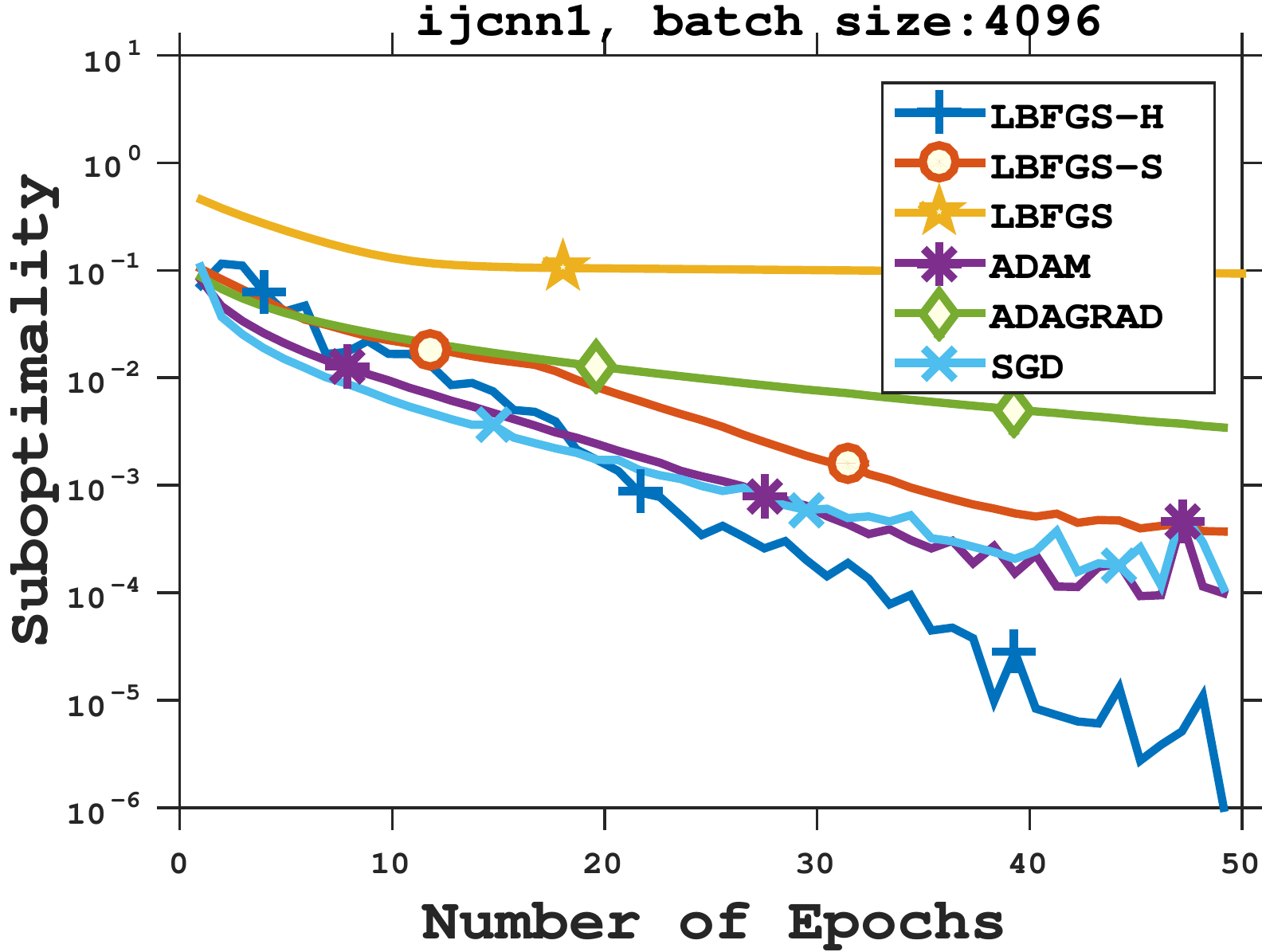,width=0.24\textwidth} 
  
    \epsfig{file=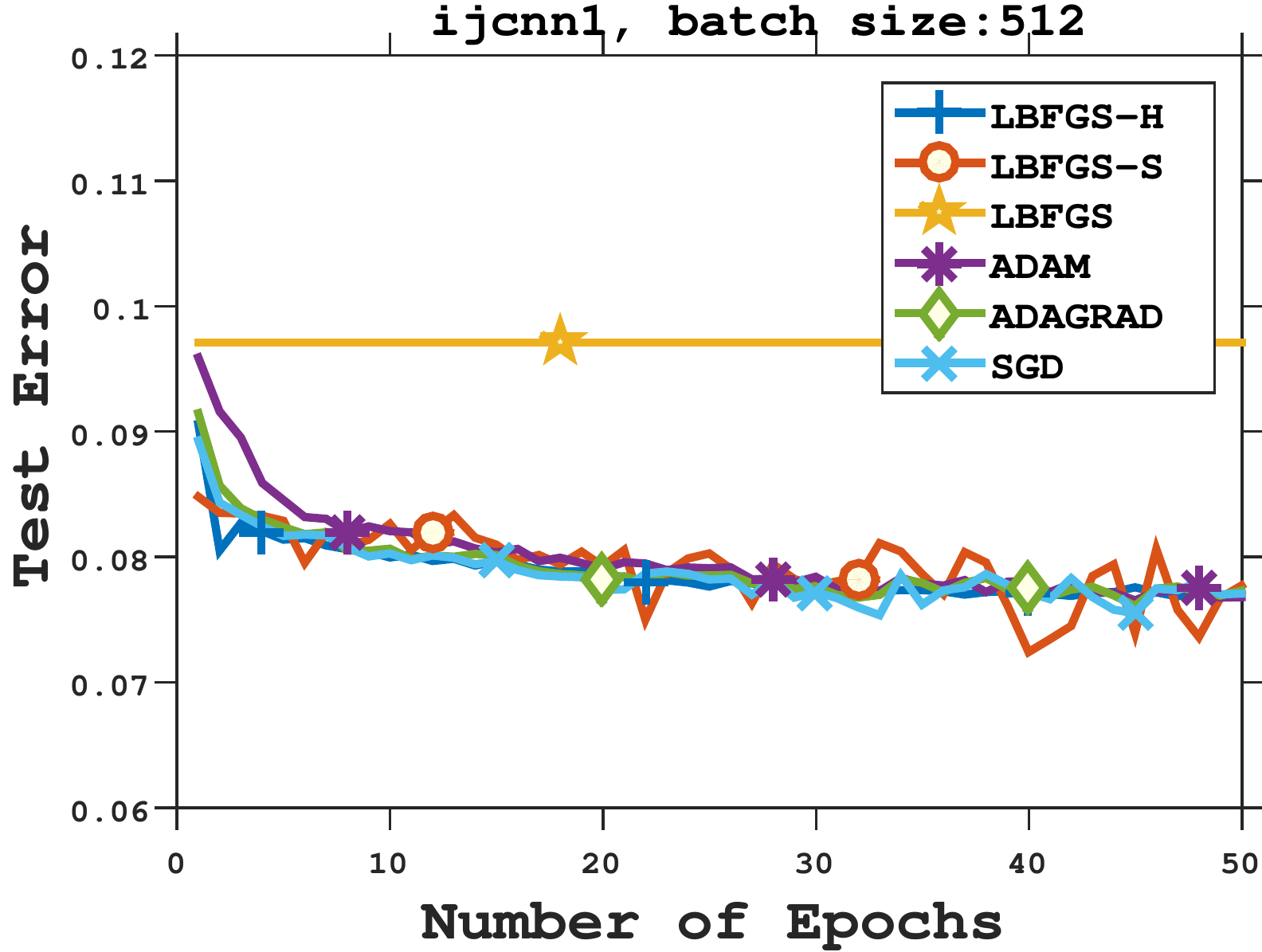,width=0.24\textwidth}
   \epsfig{file=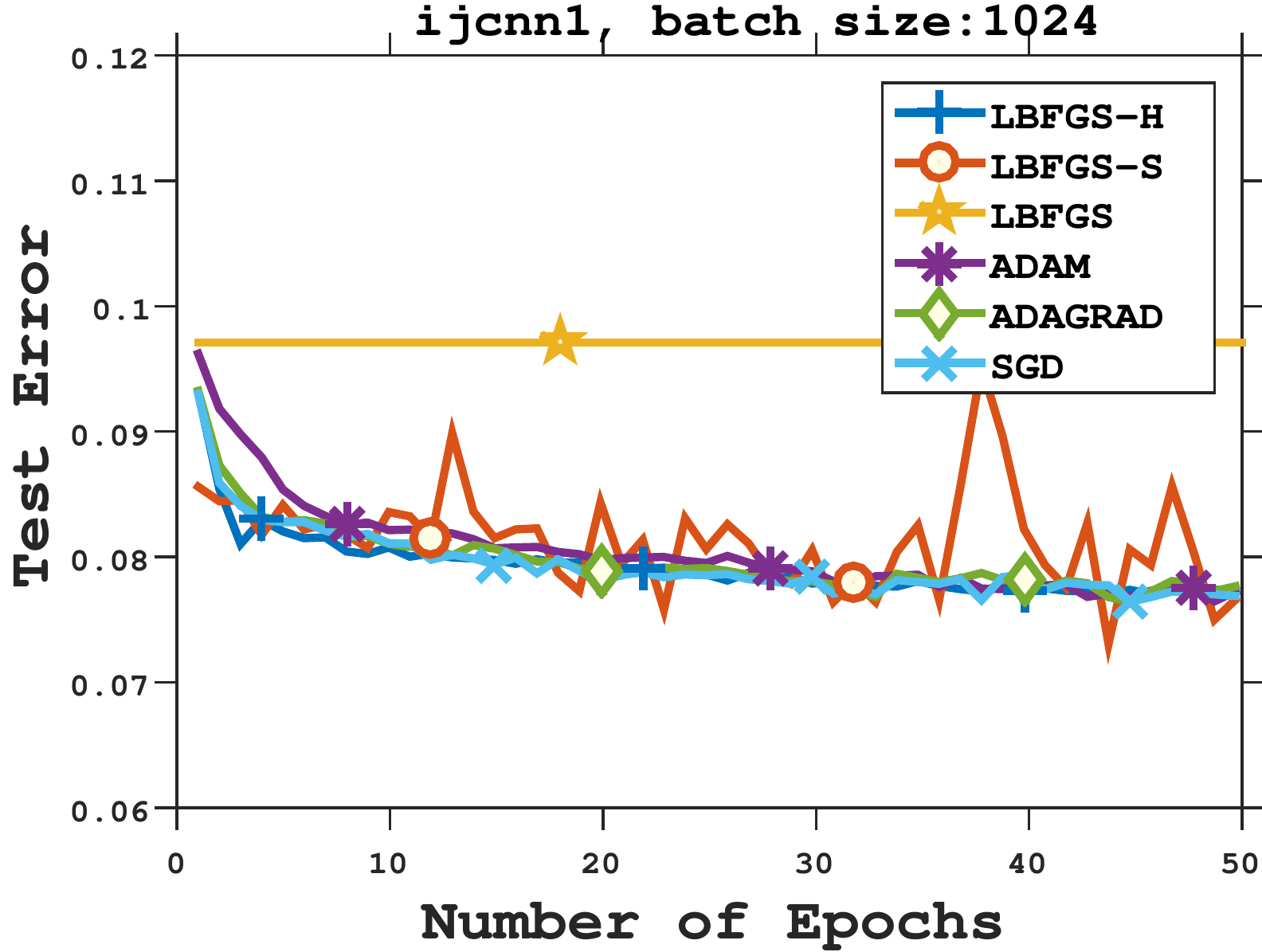,width=0.24\textwidth} 
    \epsfig{file=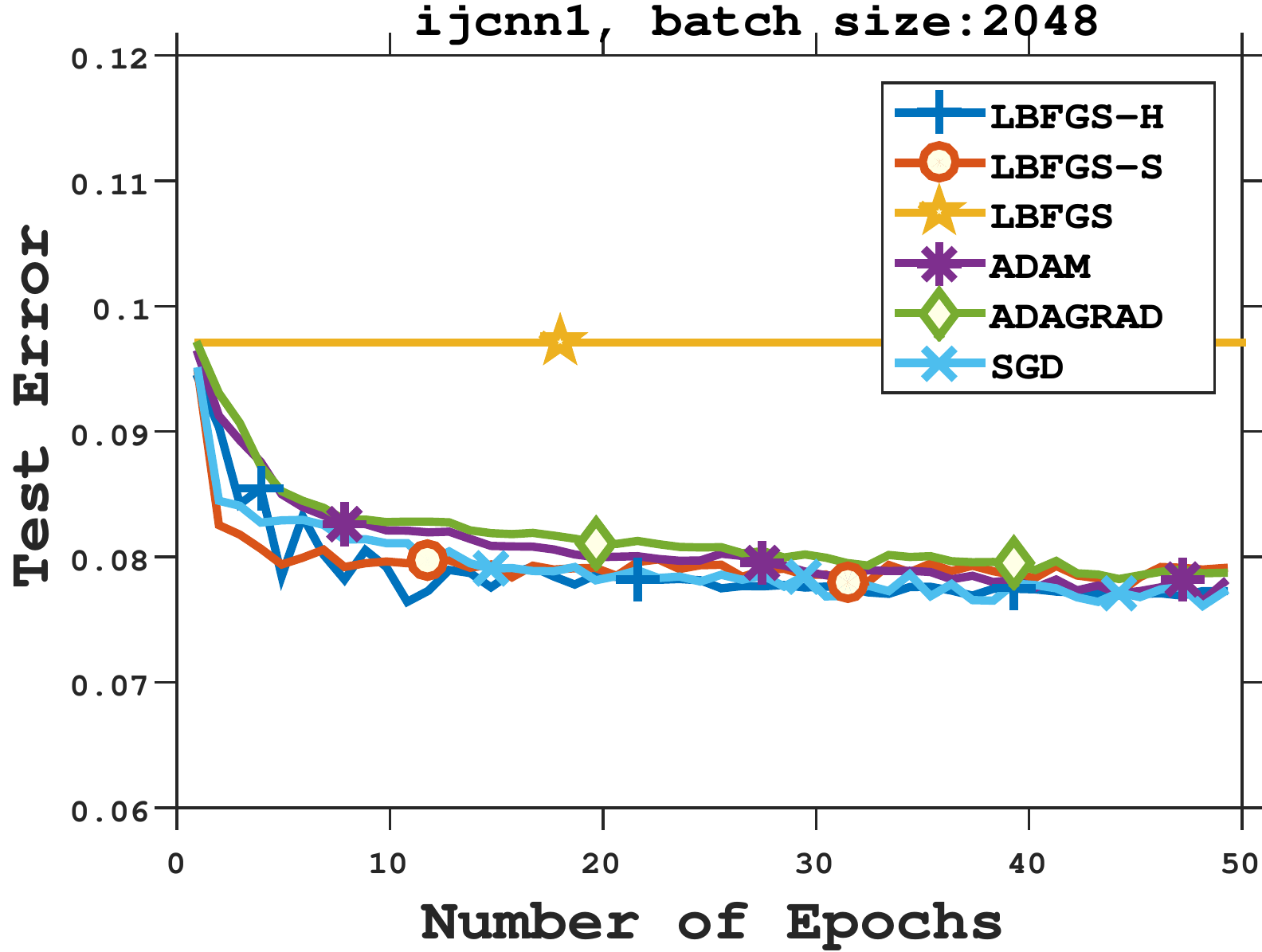,width=0.24\textwidth}
     \epsfig{file=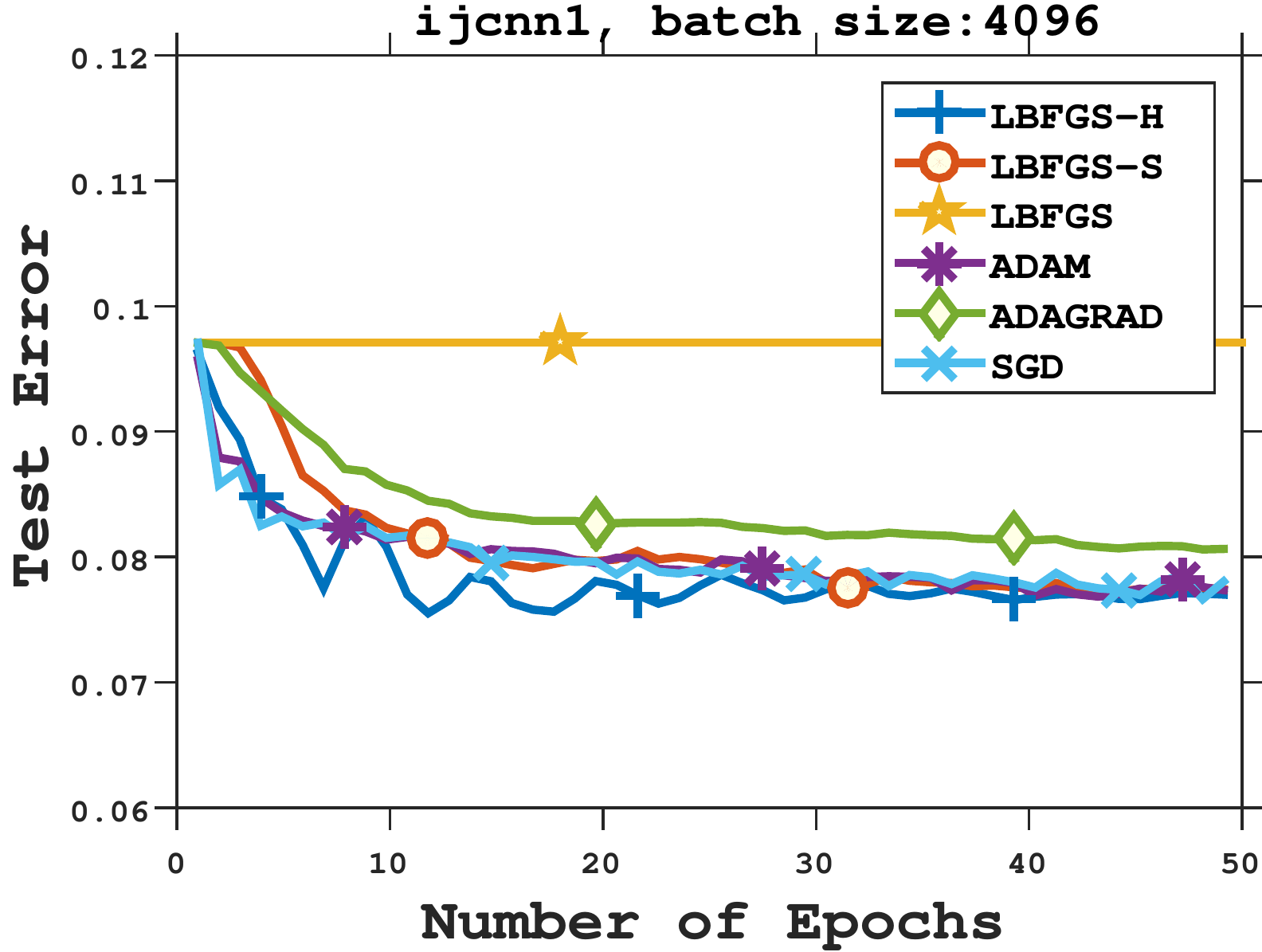,width=0.24\textwidth}
    \epsfig{file=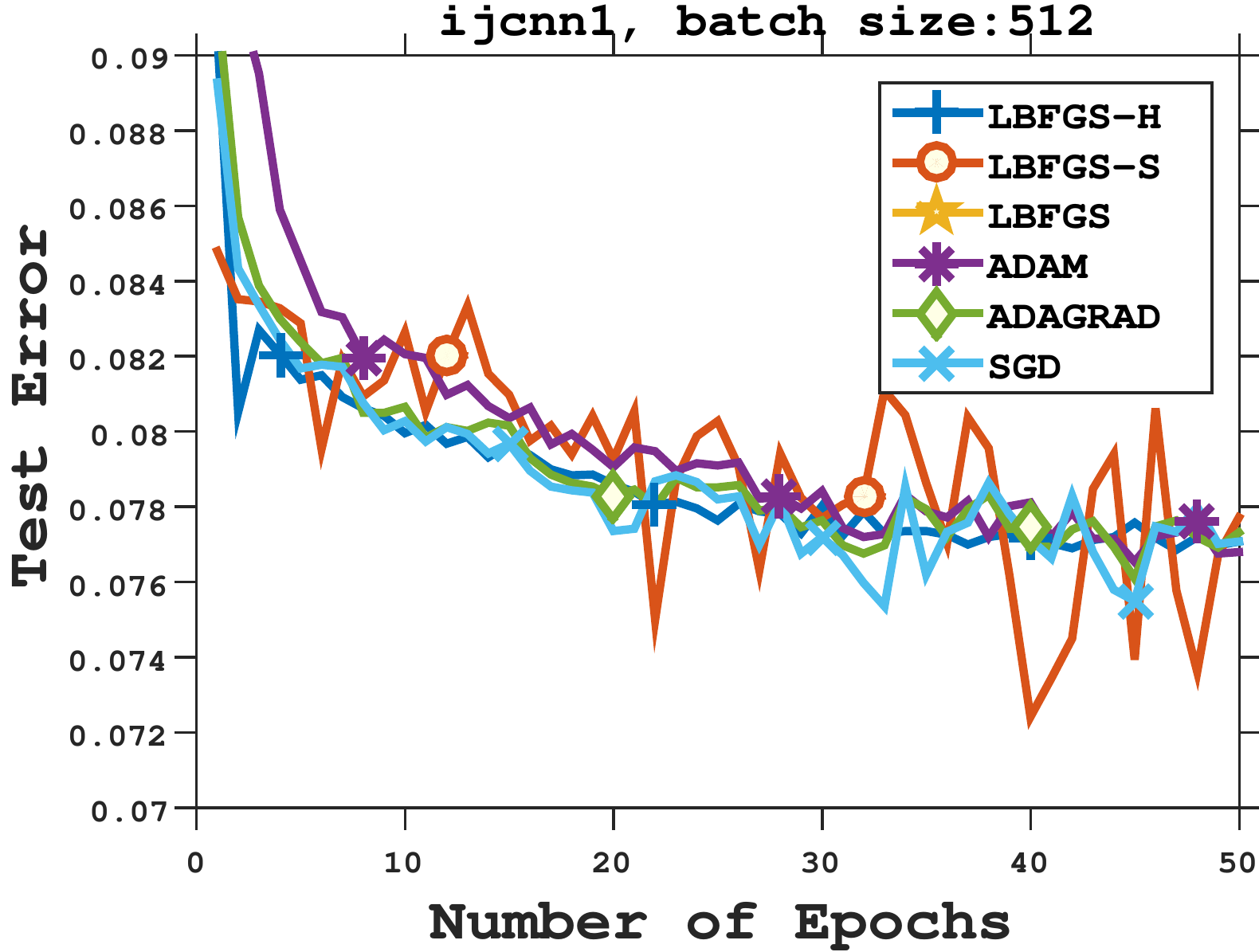,width=0.24\textwidth}
   \epsfig{file=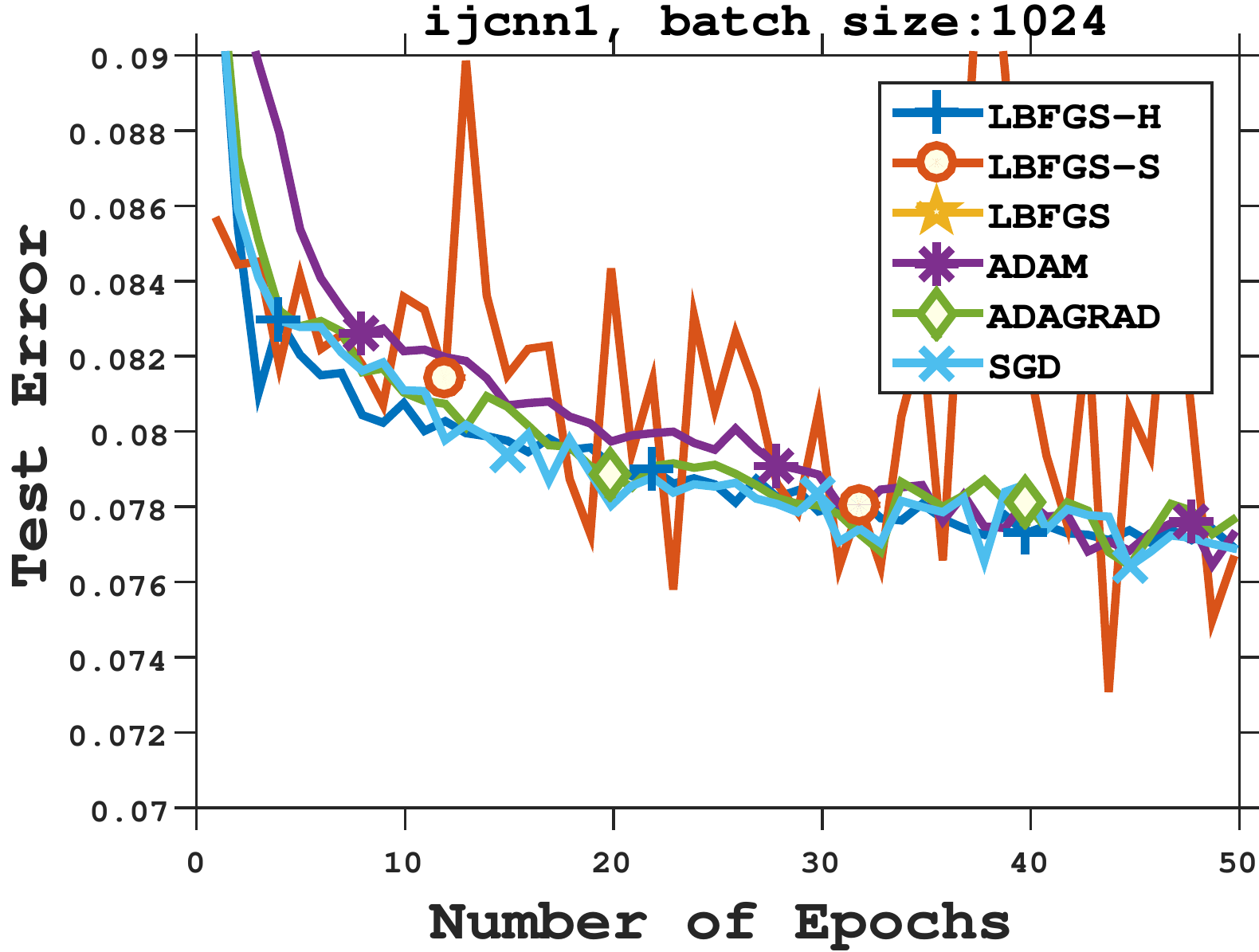,width=0.24\textwidth} 
    \epsfig{file=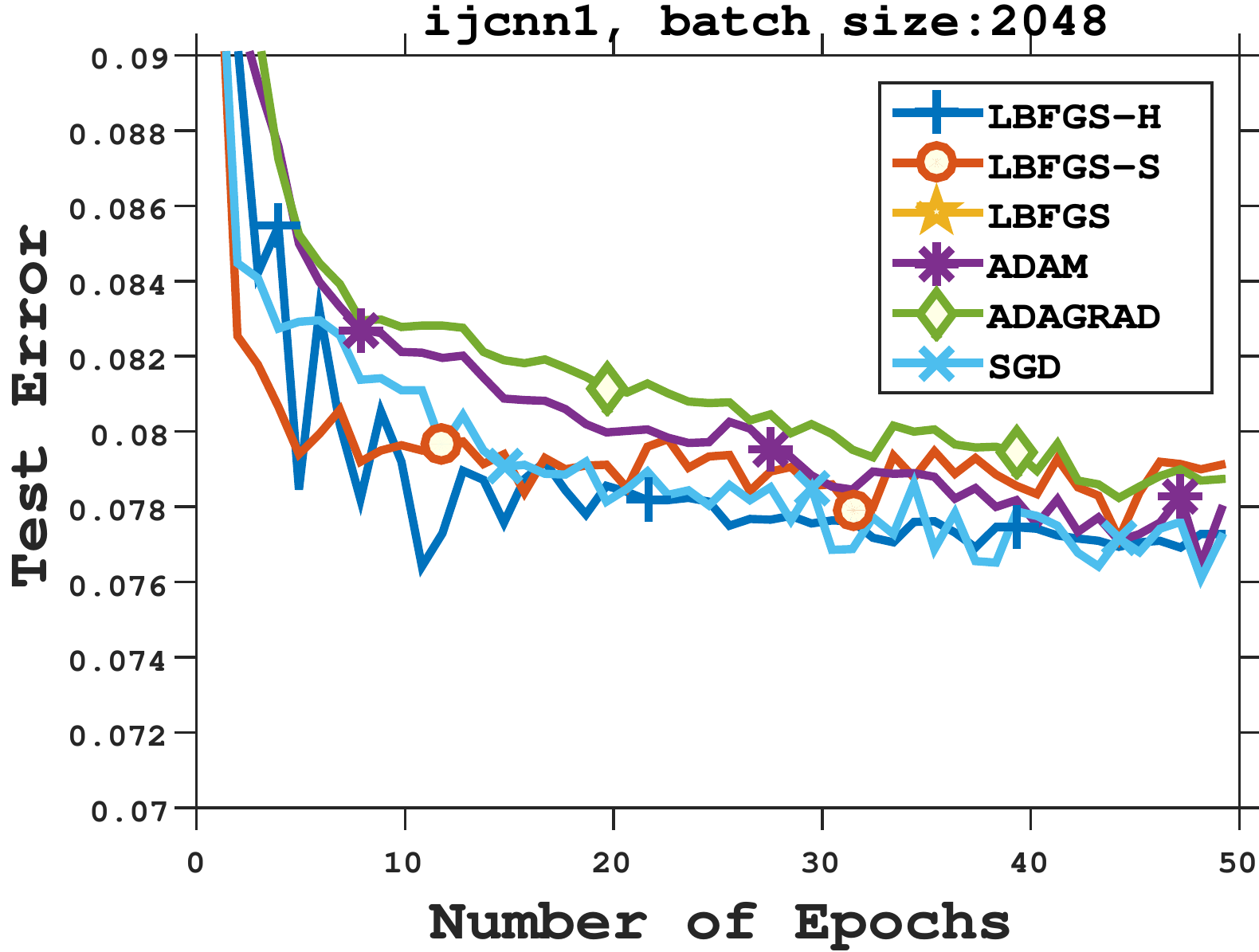,width=0.24\textwidth}
     \epsfig{file=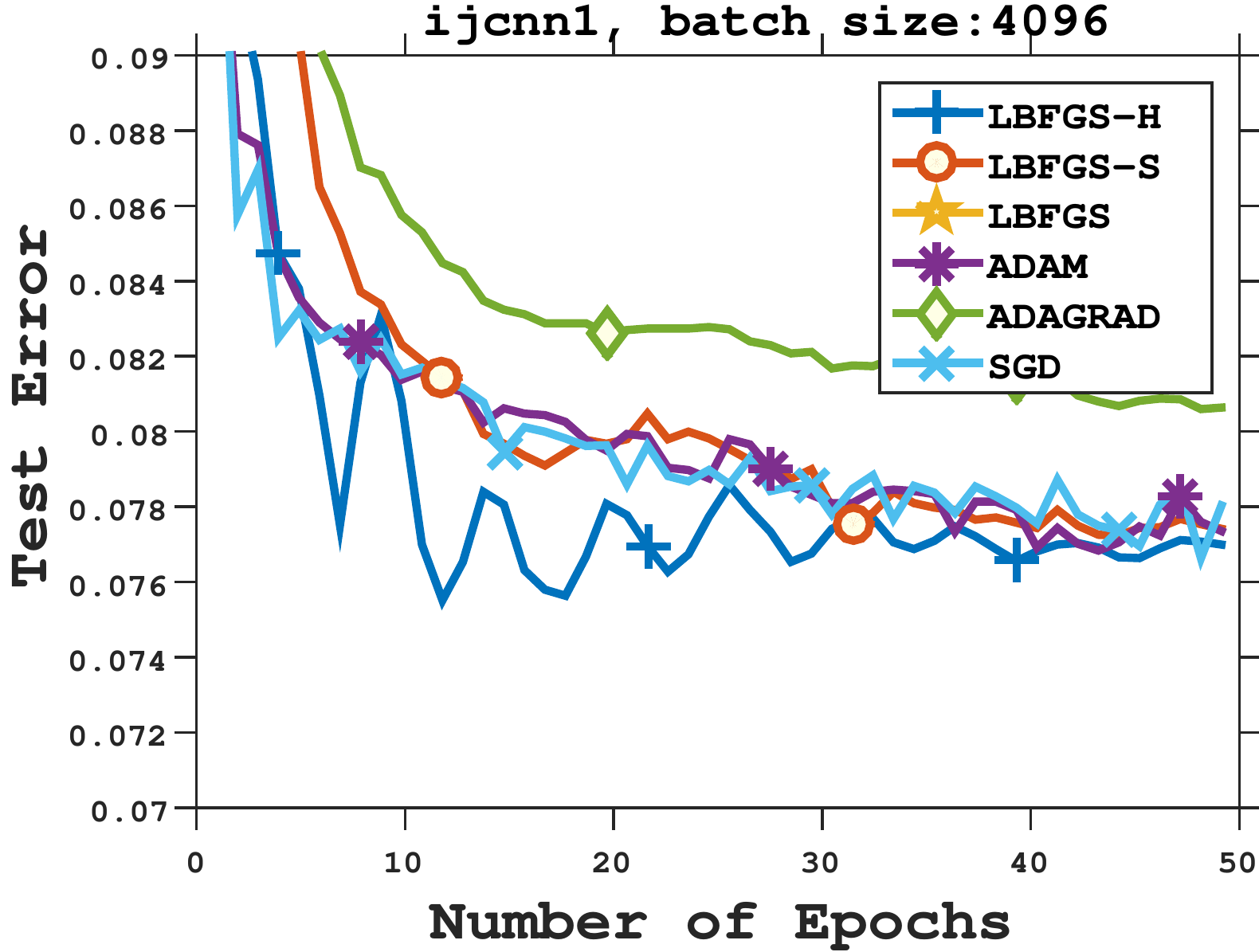,width=0.24\textwidth}
 
 \caption{\footnotesize Comparisons of training loss (top 2 rows), sub-optimality (middle row) and test errors (bottom 2 rows) for different algorithms with batch sizes 512, 1024, 2048, 4096 on \emph{ijcnn1}, convex, logistic regression.}
   \label{fig:add2}  
 \end{figure}

 Figure~\ref{fig:add2} exhibits results of the same experiment with larger batch sizes $b=512, 1024, 2048, 4096$. With larger batch sizes, LBFGS-H (LBFGS-F) outperforms other methods more and more, suggesting LBFGS-F as an excellent choice for the distributed setting. In addition, the figure presents the instability of LBFGS-S with large batch sizes, and it won't stabilize until $b>2048 = 2^{11}$ in this case. Moreover, the convergences of ADAM, ADAGRAD and SGD are slightly slowed down with the increasing of batch sizes.
 
\newpage
  \subsubsection{Randomization}
\begin{figure}[h]
\centering
 \epsfig{file=Figs/ijcnn1_lossrand16,width=0.32\textwidth}
 \epsfig{file=Figs/ijcnn1_lossrand64.eps,width=0.32\textwidth} 
  \epsfig{file=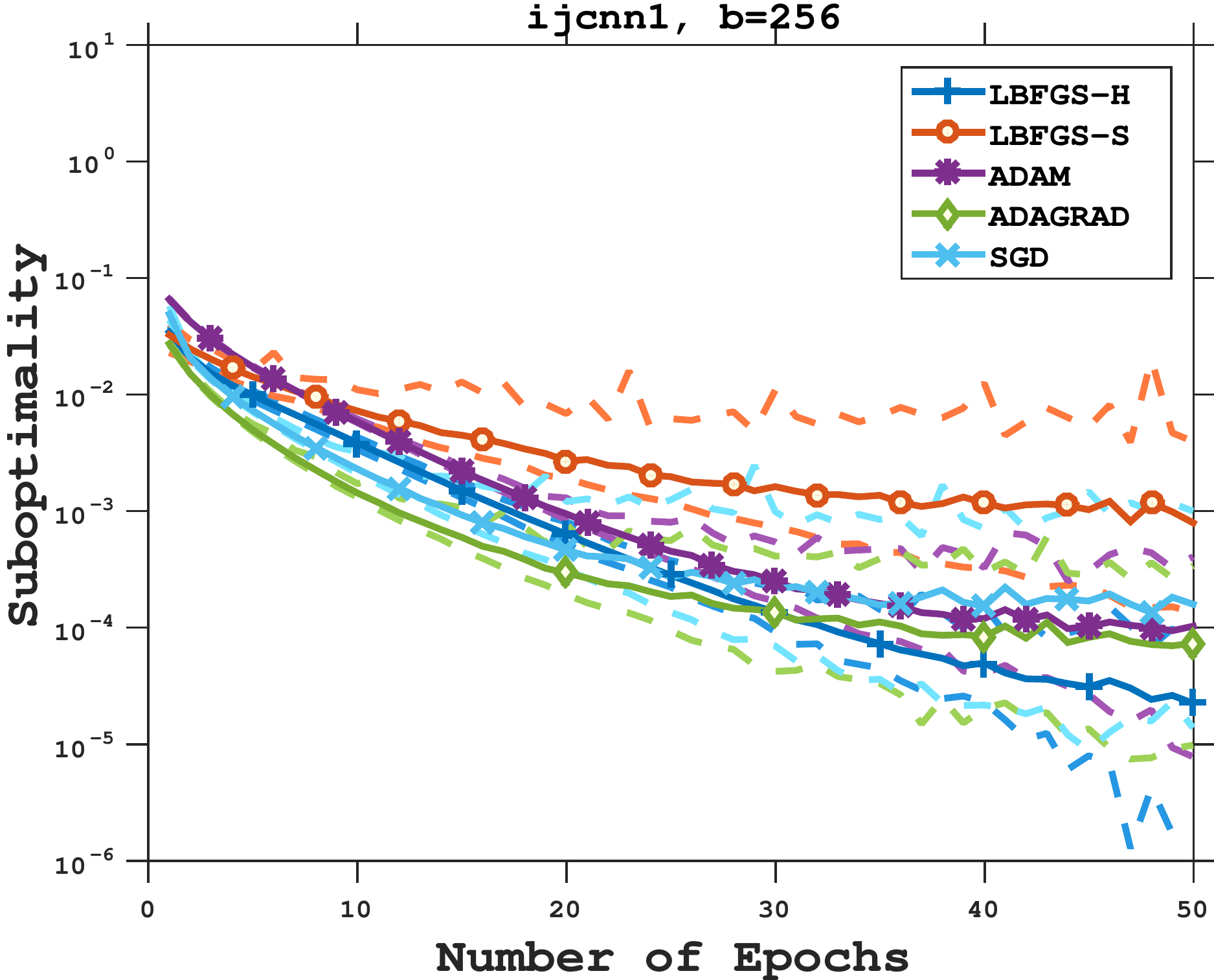,width=0.32\textwidth} 
 
    \epsfig{file=Figs/ijcnn1_lossrand512.eps,width=0.32\textwidth}
   \epsfig{file=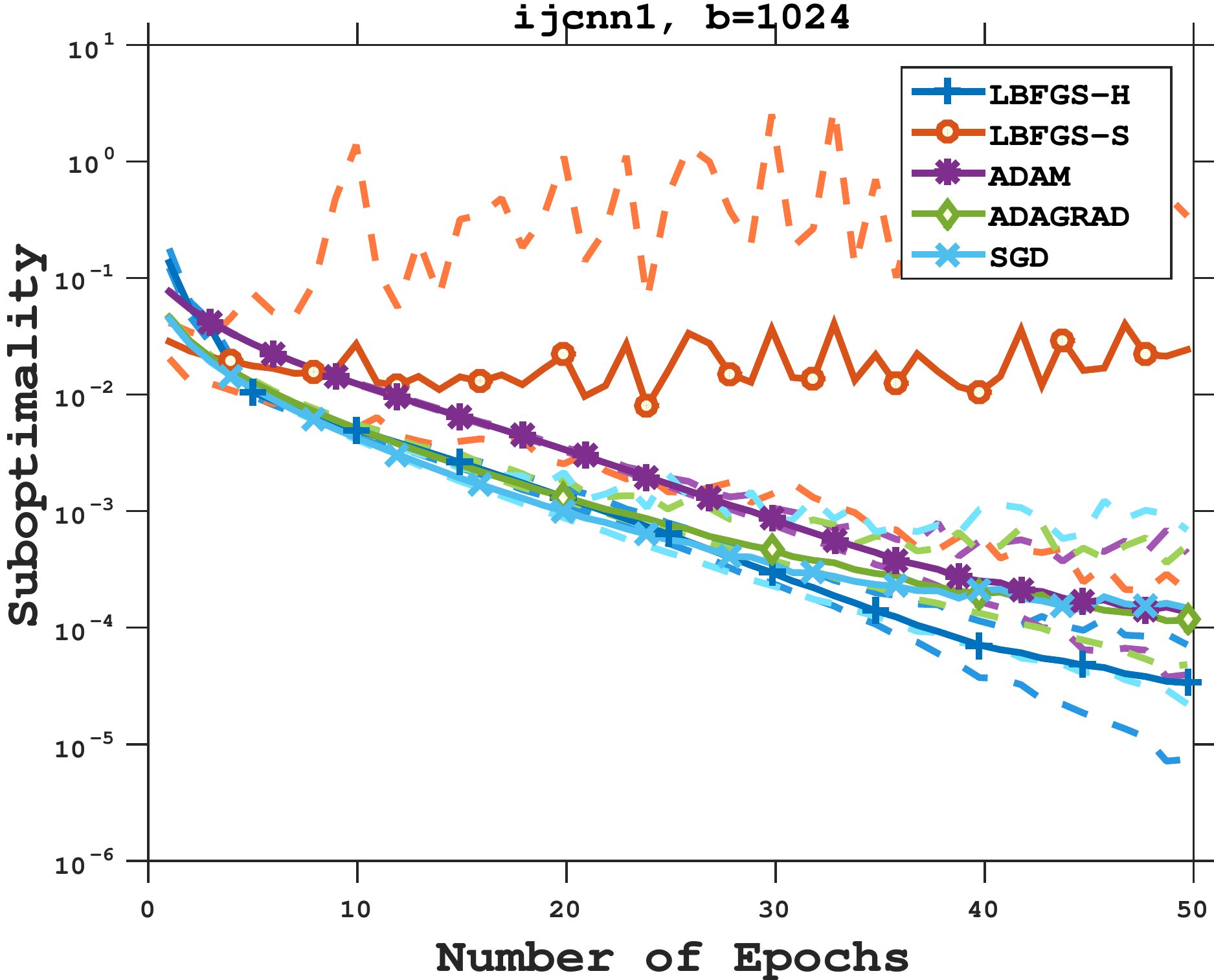,width=0.32\textwidth} 
    \epsfig{file=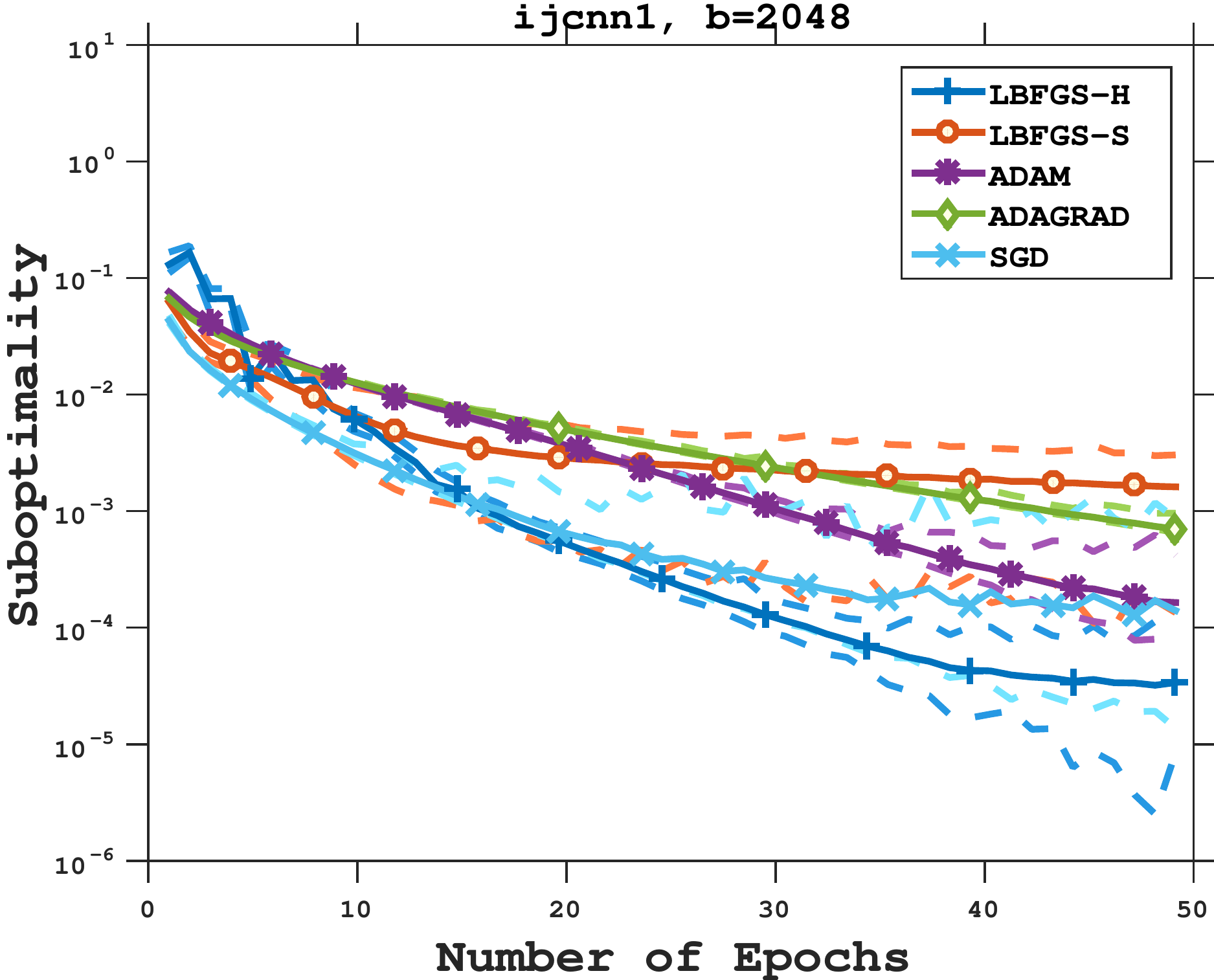,width=0.32\textwidth}

        \epsfig{file=Figs/ijcnn1_lossrand4096.eps,width=0.32\textwidth}
 
 \caption{\footnotesize Comparisons of sub-optimality for different stochastic methods with batch sizes 16, 64, 256, 512, 1024, 2048, 4096 on \emph{ijcnn1}, convex, logistic regression.}
   \label{fig:add3}
 \end{figure}

 In order to verify the stability with different random seeds, we conduct the same experiment with $100$ different random seeds and present the "reliable" areas enclosed by the dotted lines with the same colors for each algorithm in Figure~\ref{fig:add3}. As we discussed in Section~\ref{sec:experiments}, with large batch sizes, the performance of ADAM, ADAGRAD and SGD worsen while LBFGS-H outperforms the others in sub-optimality. To achieve the same accuracy, fewer epochs are needed and thus fewer communications for our framework when the batch size is large.
 
 \newpage
 \subsection{Results on cross-entropy (convex), \emph{MNIST}}
 
 The second experiment is conducted for the linear predictor with cross-entropy loss on \emph{MNIST}. Similarly, LBFGS-S is unstable while LBFGS-H (LBFGS-F) performs better and better with the increasing batch sizes and outperforms the others in the case when the batch size $b=64, 256$.

  \begin{figure}[H]
\centering
 \epsfig{file=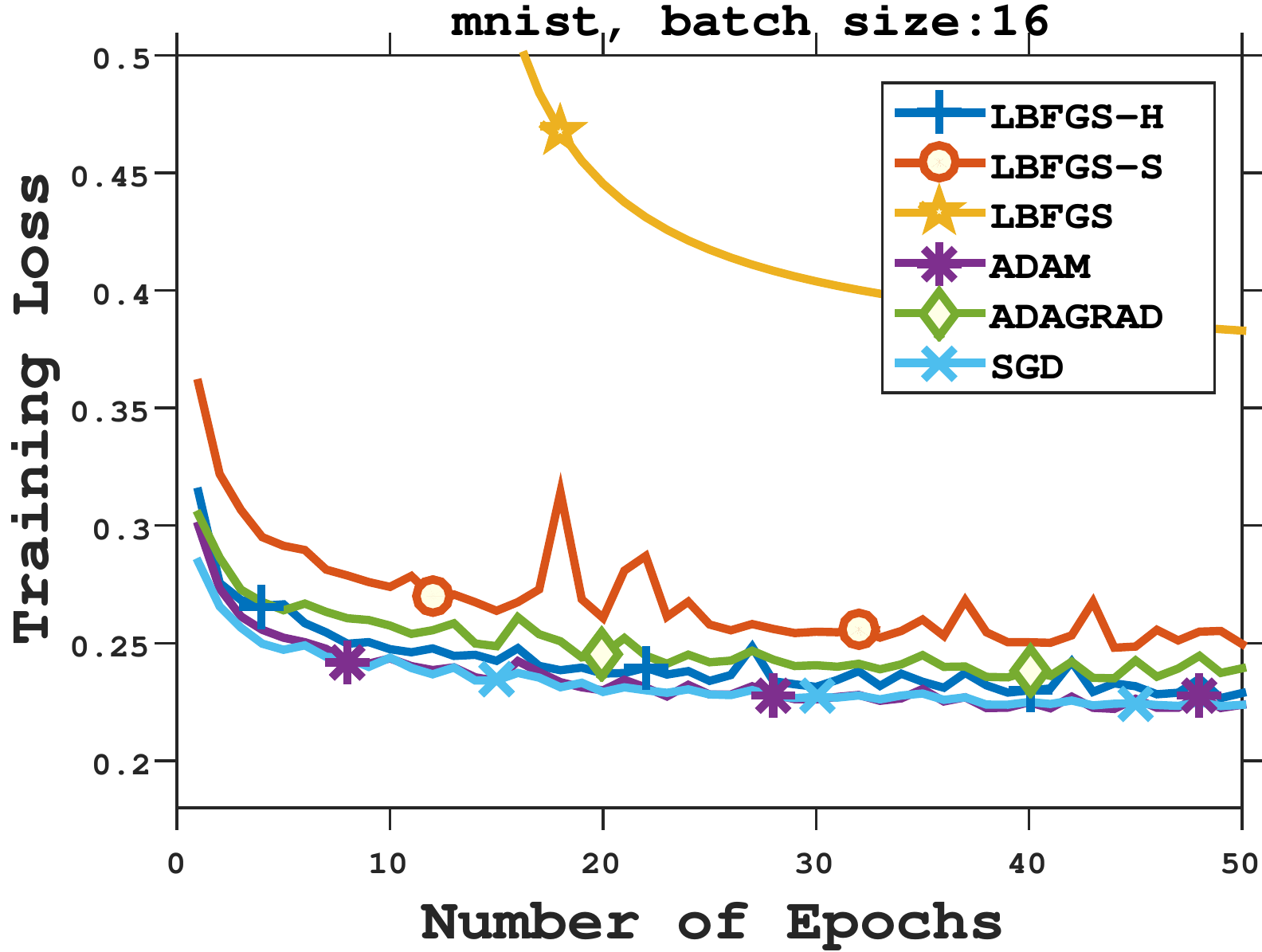,width=0.32\textwidth}
  \epsfig{file=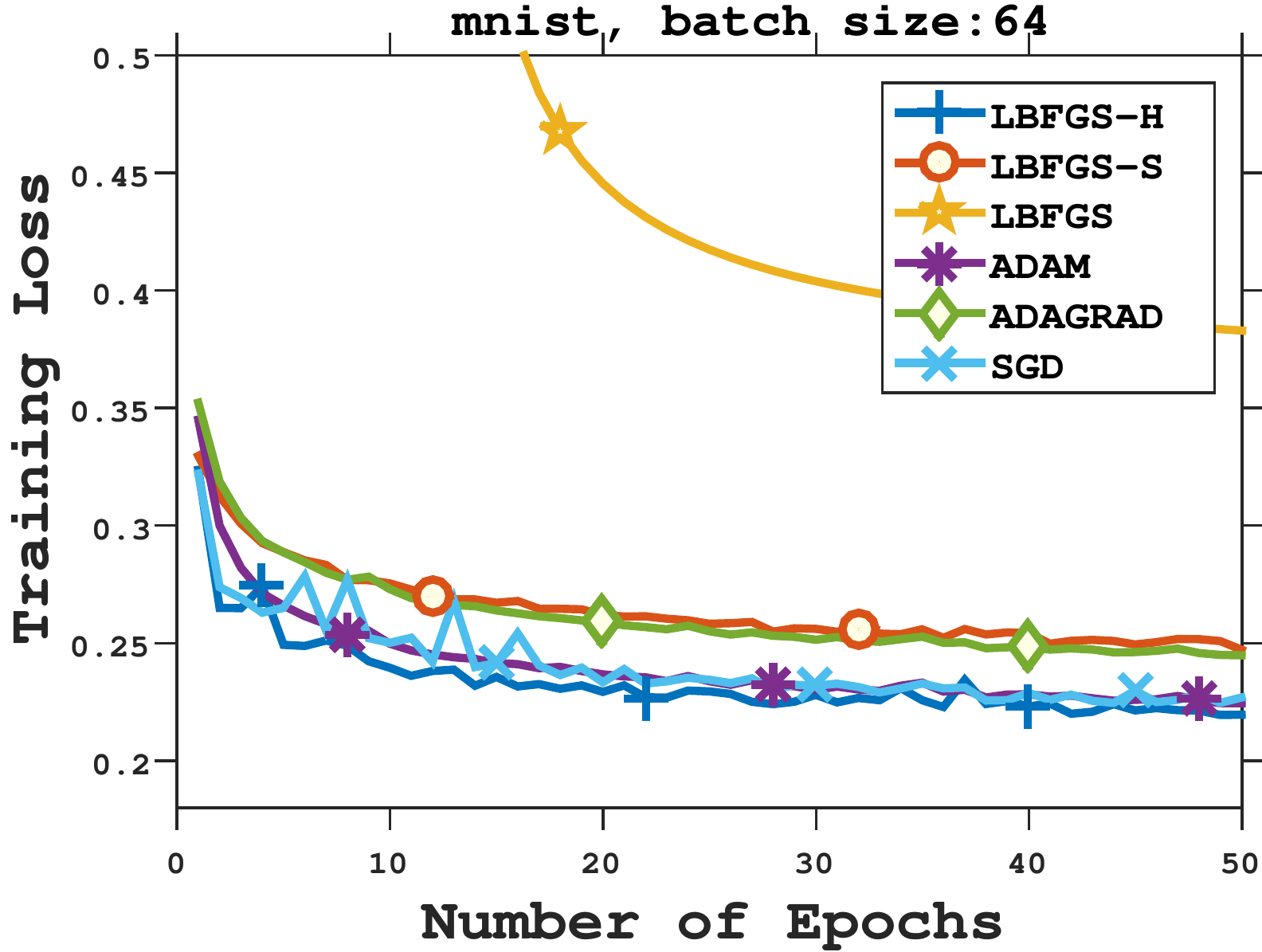,width=0.32\textwidth} 
  \epsfig{file=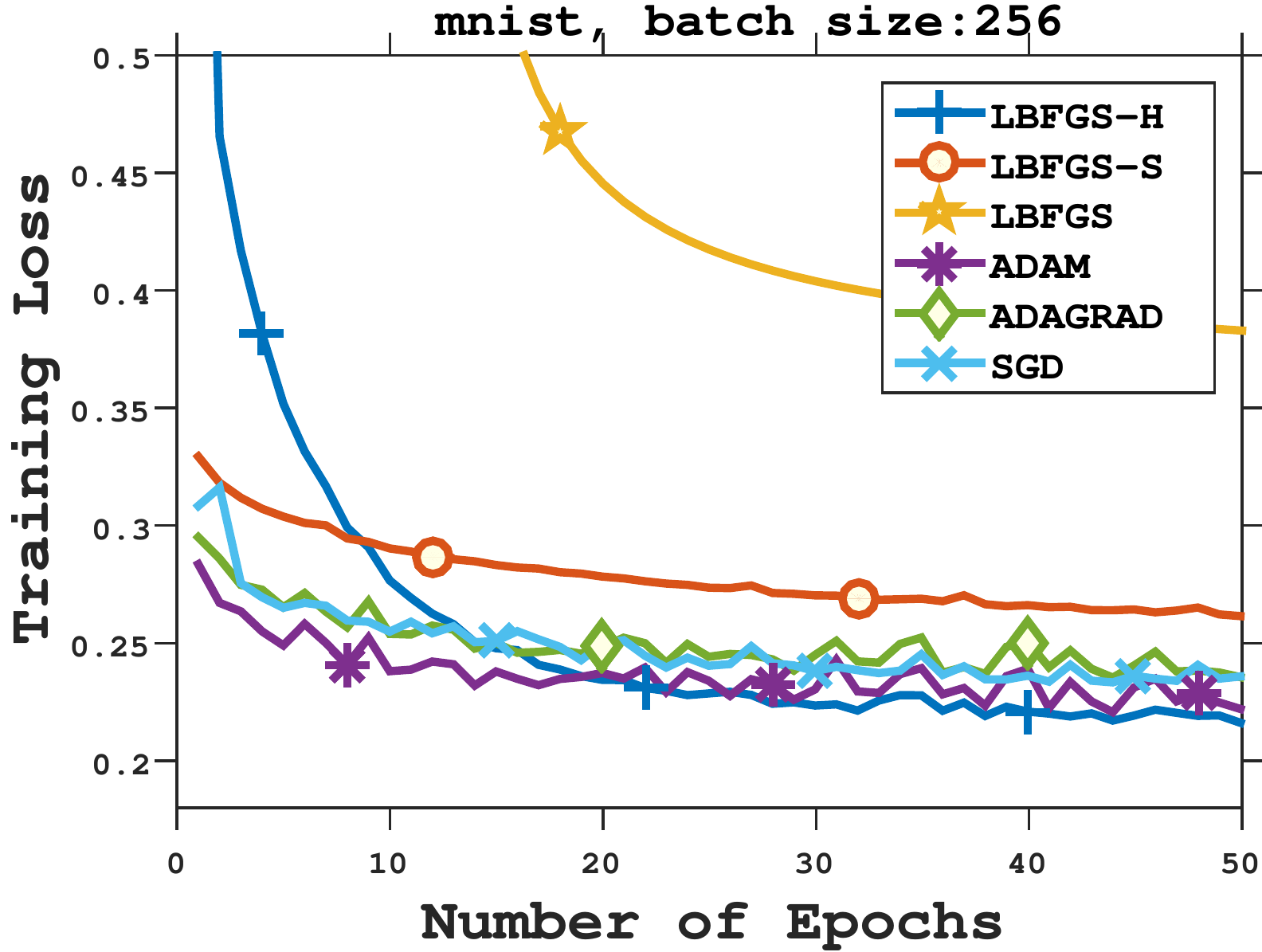,width=0.32\textwidth} 
  
 \epsfig{file=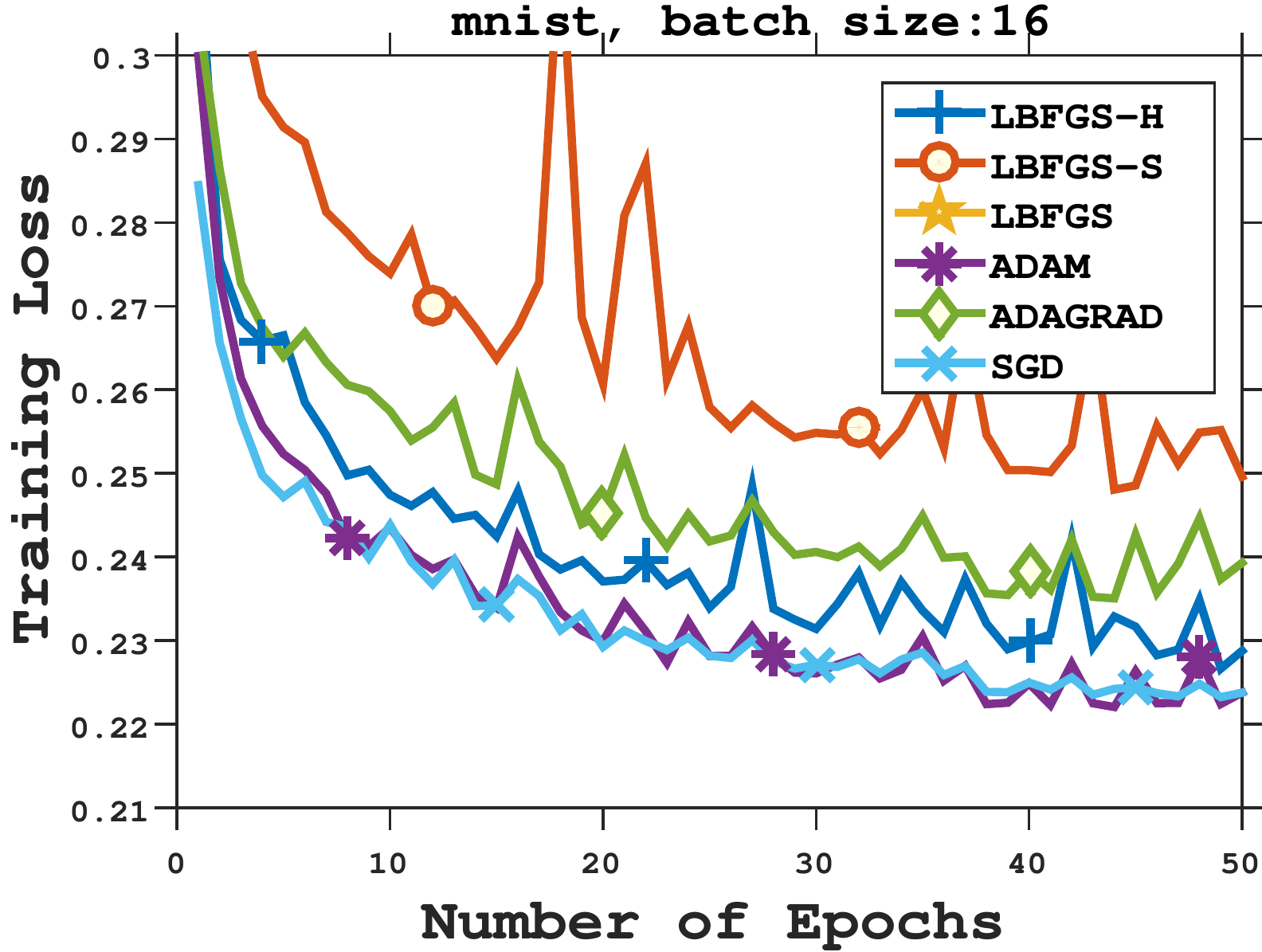,width=0.32\textwidth}
  \epsfig{file=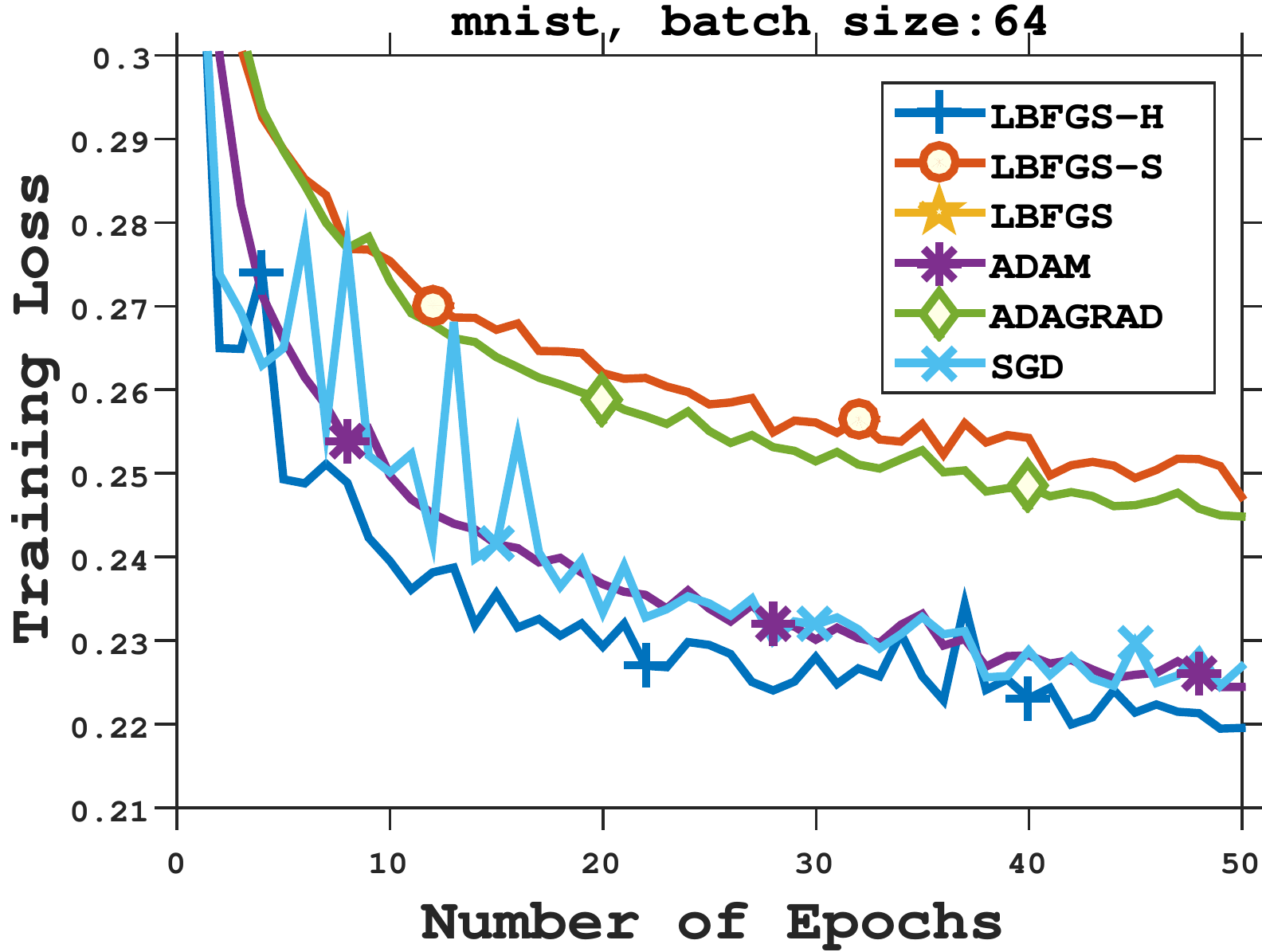,width=0.32\textwidth} 
  \epsfig{file=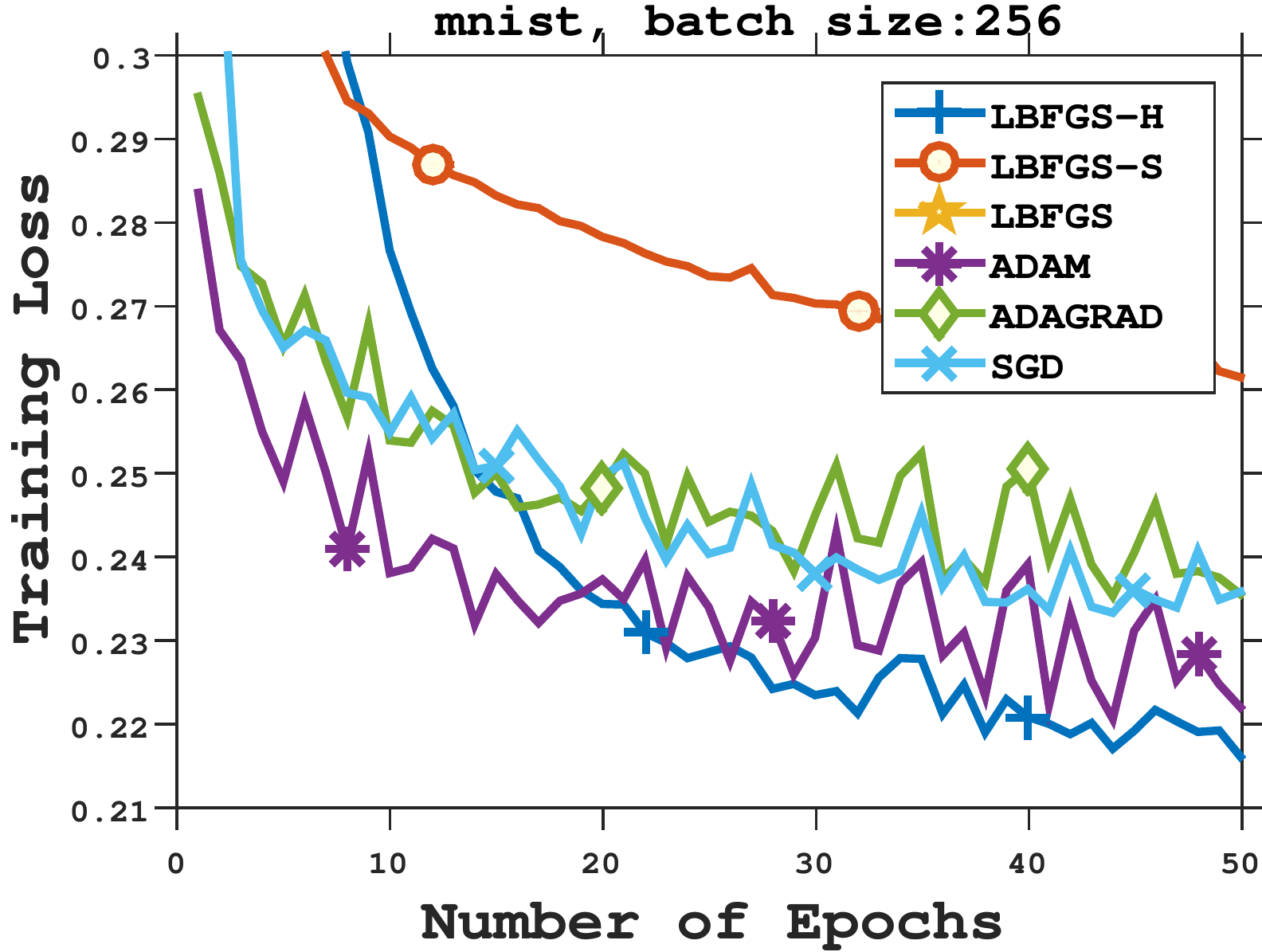,width=0.32\textwidth} 

 \epsfig{file=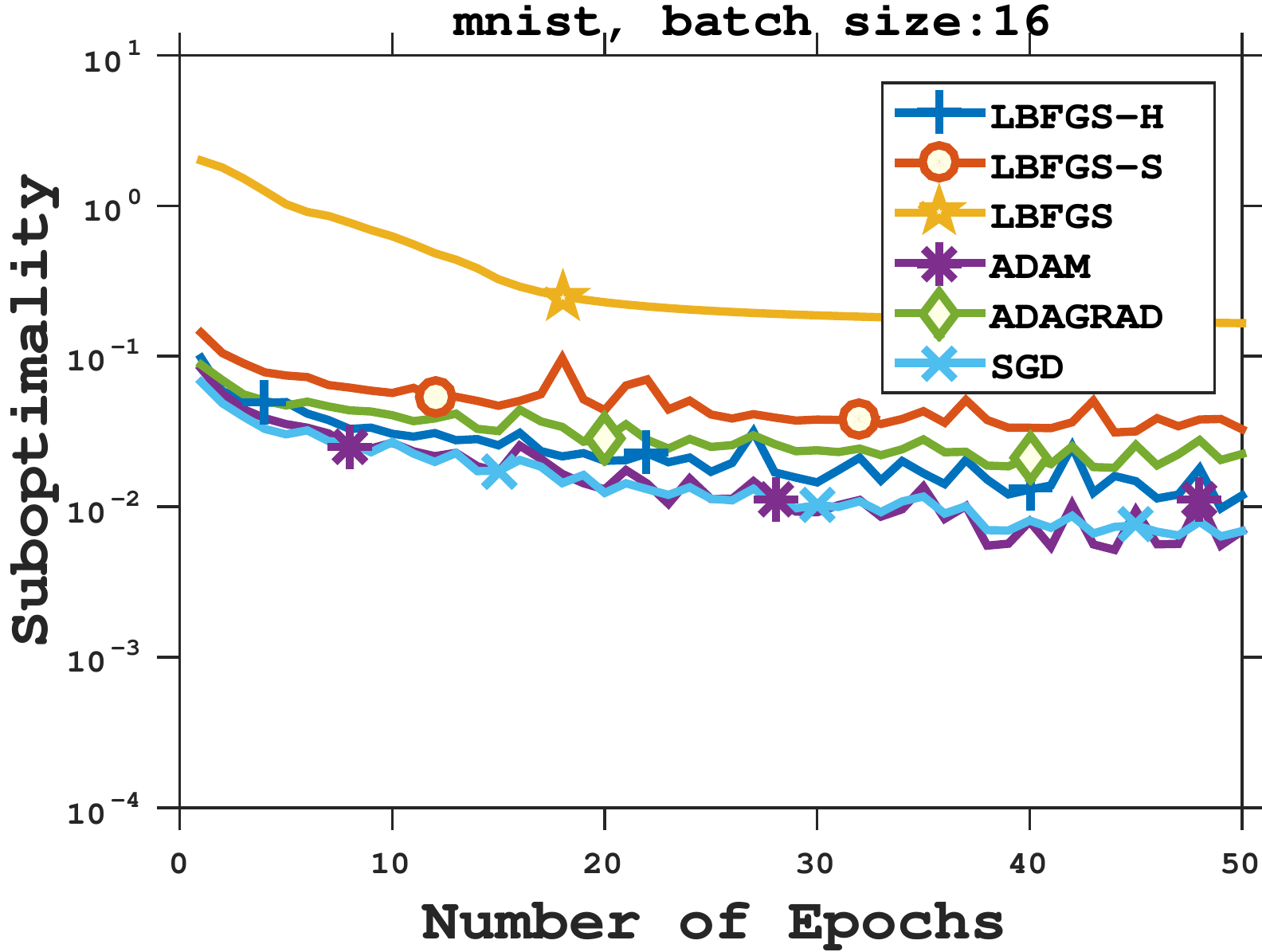,width=0.32\textwidth}
  \epsfig{file=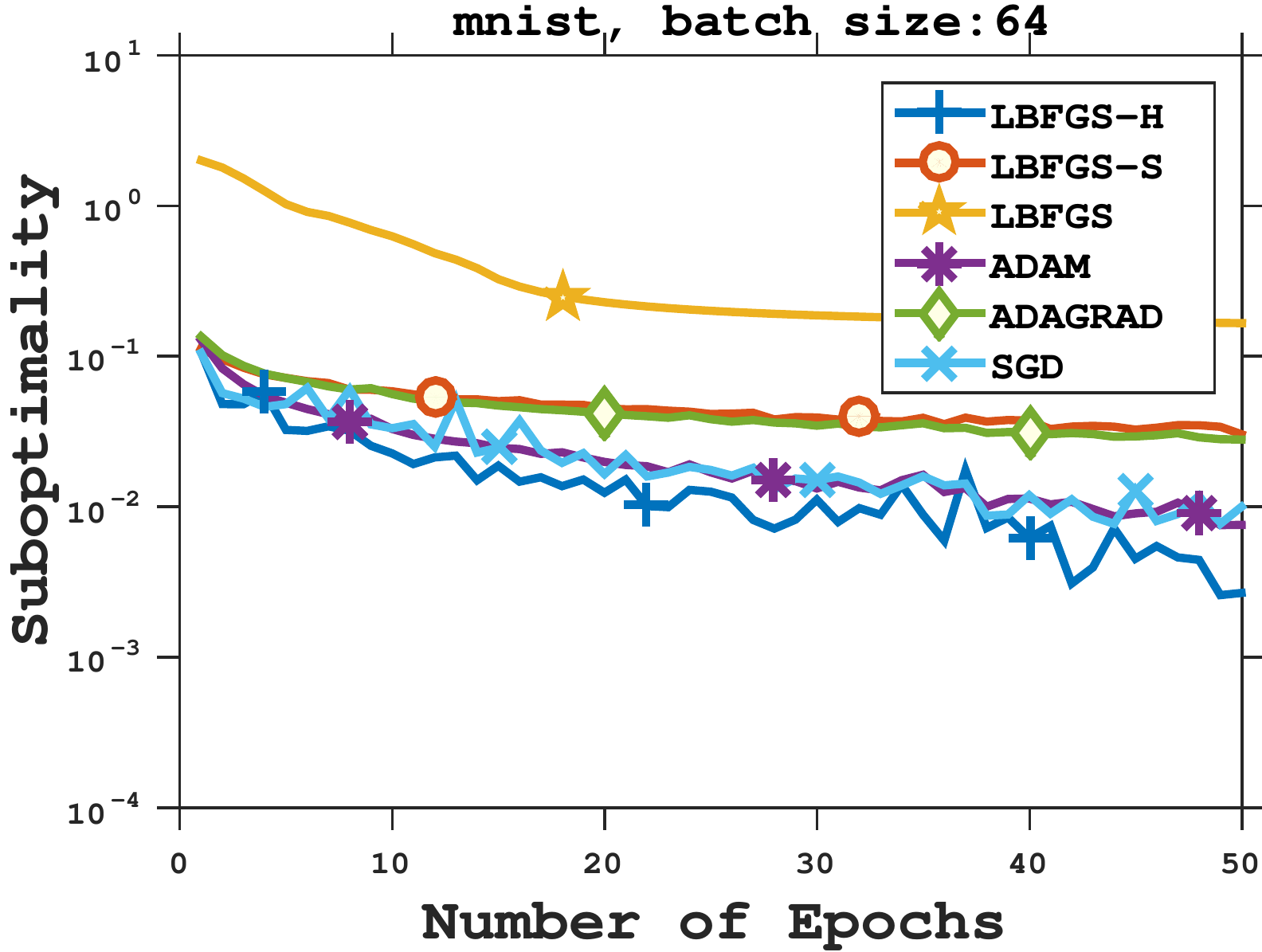,width=0.32\textwidth} 
  \epsfig{file=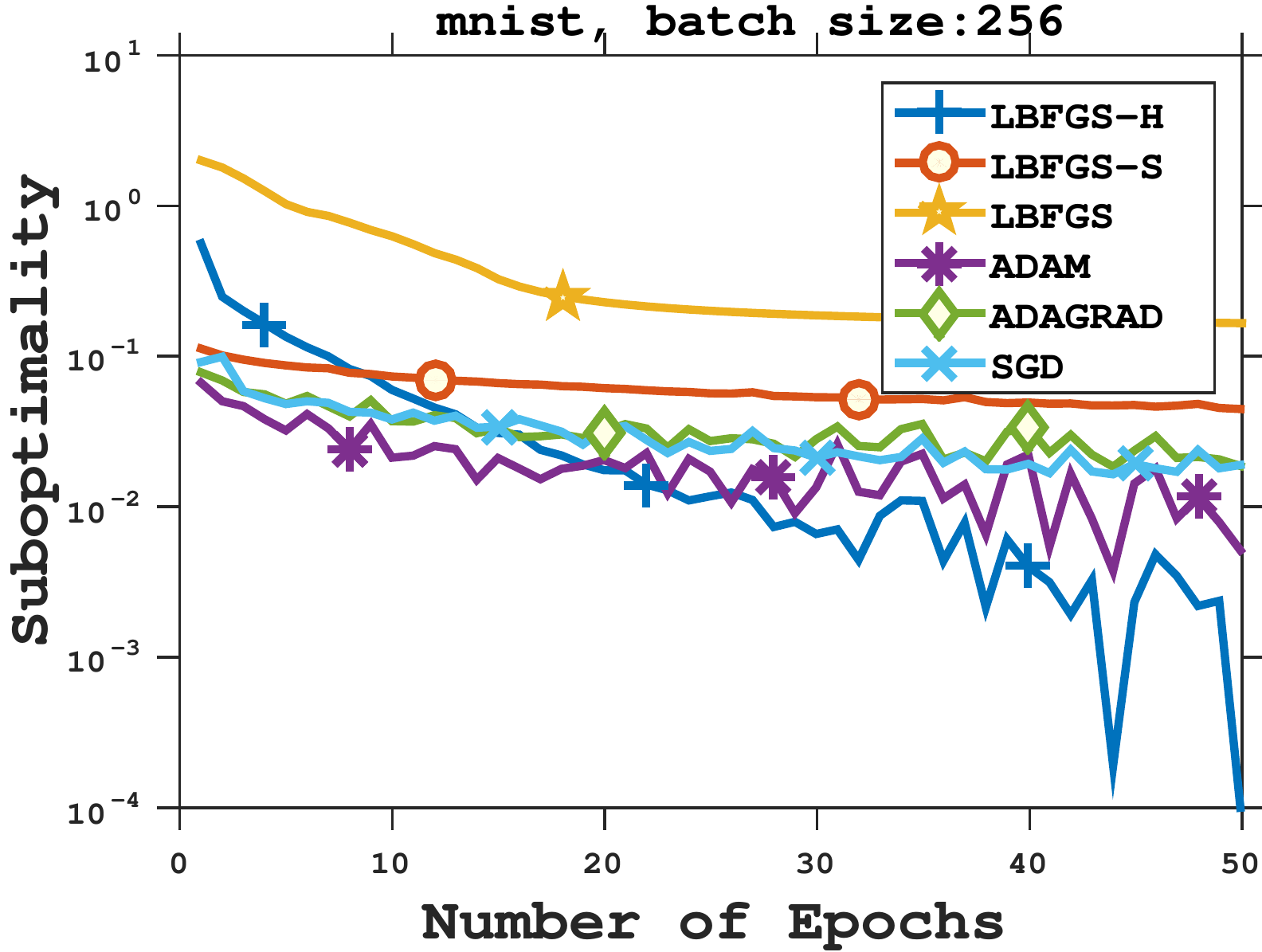,width=0.32\textwidth} 
   \epsfig{file=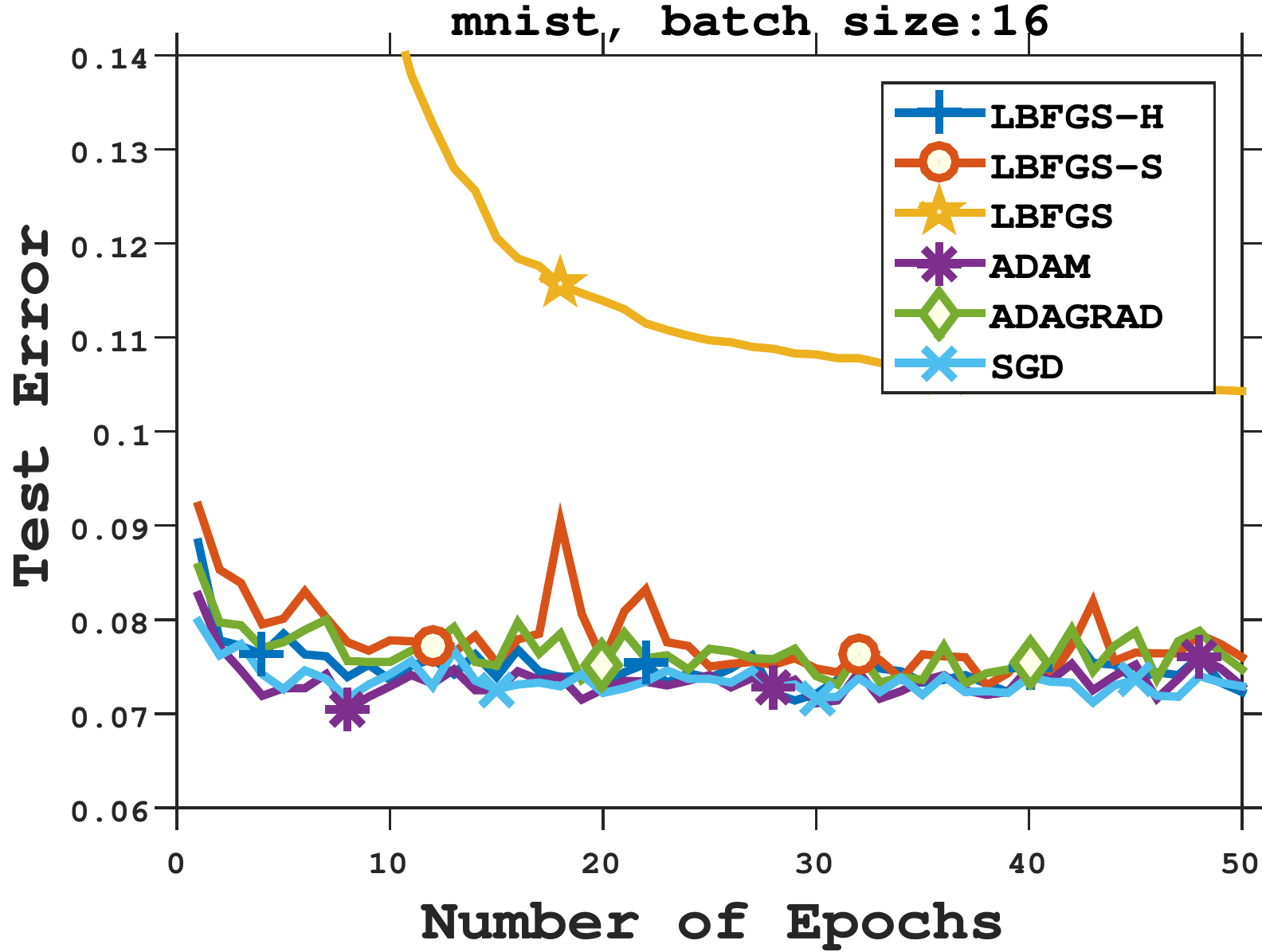,width=0.32\textwidth}
   \epsfig{file=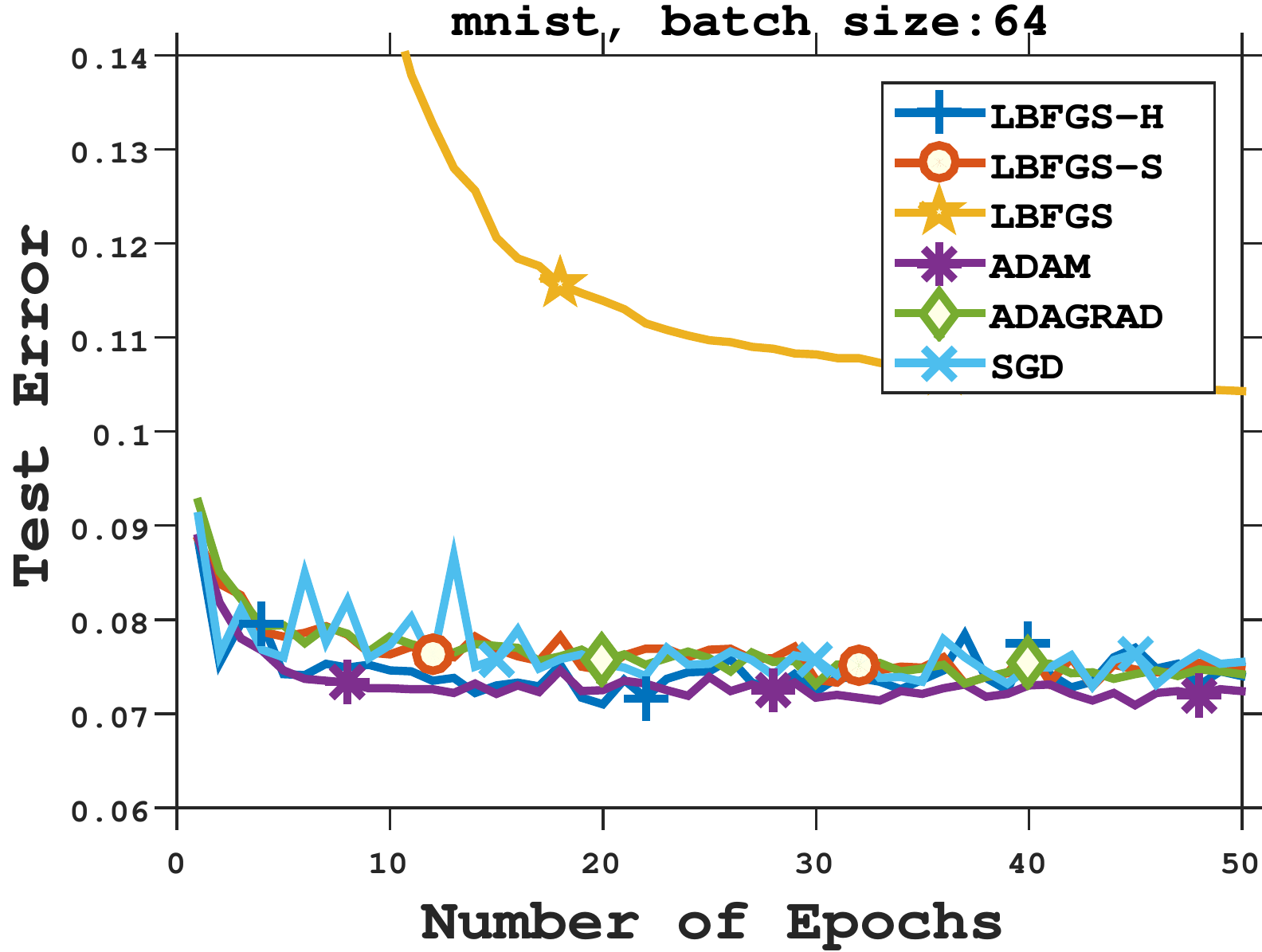,width=0.32\textwidth} 
    \epsfig{file=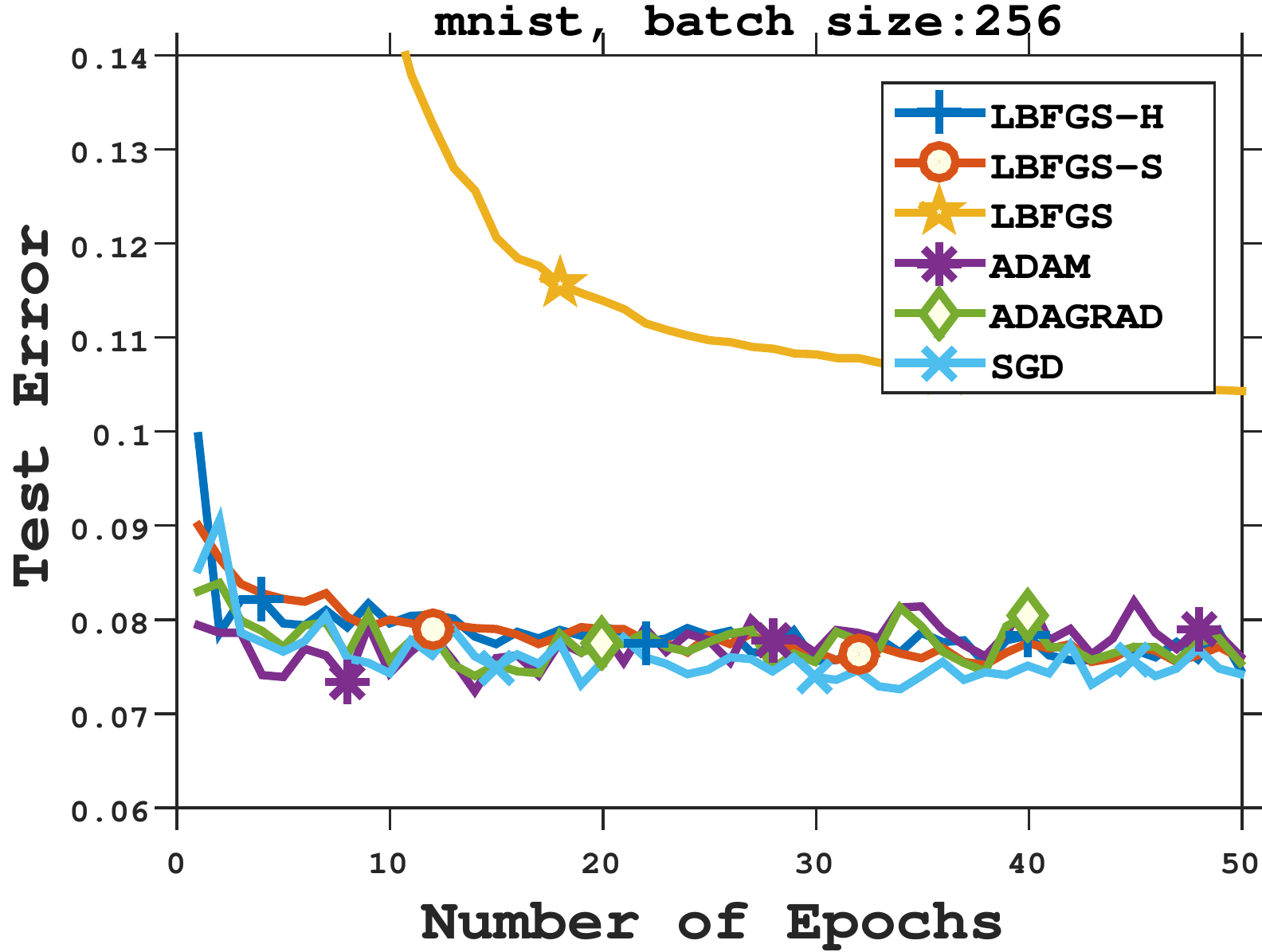,width=0.32\textwidth}
    \epsfig{file=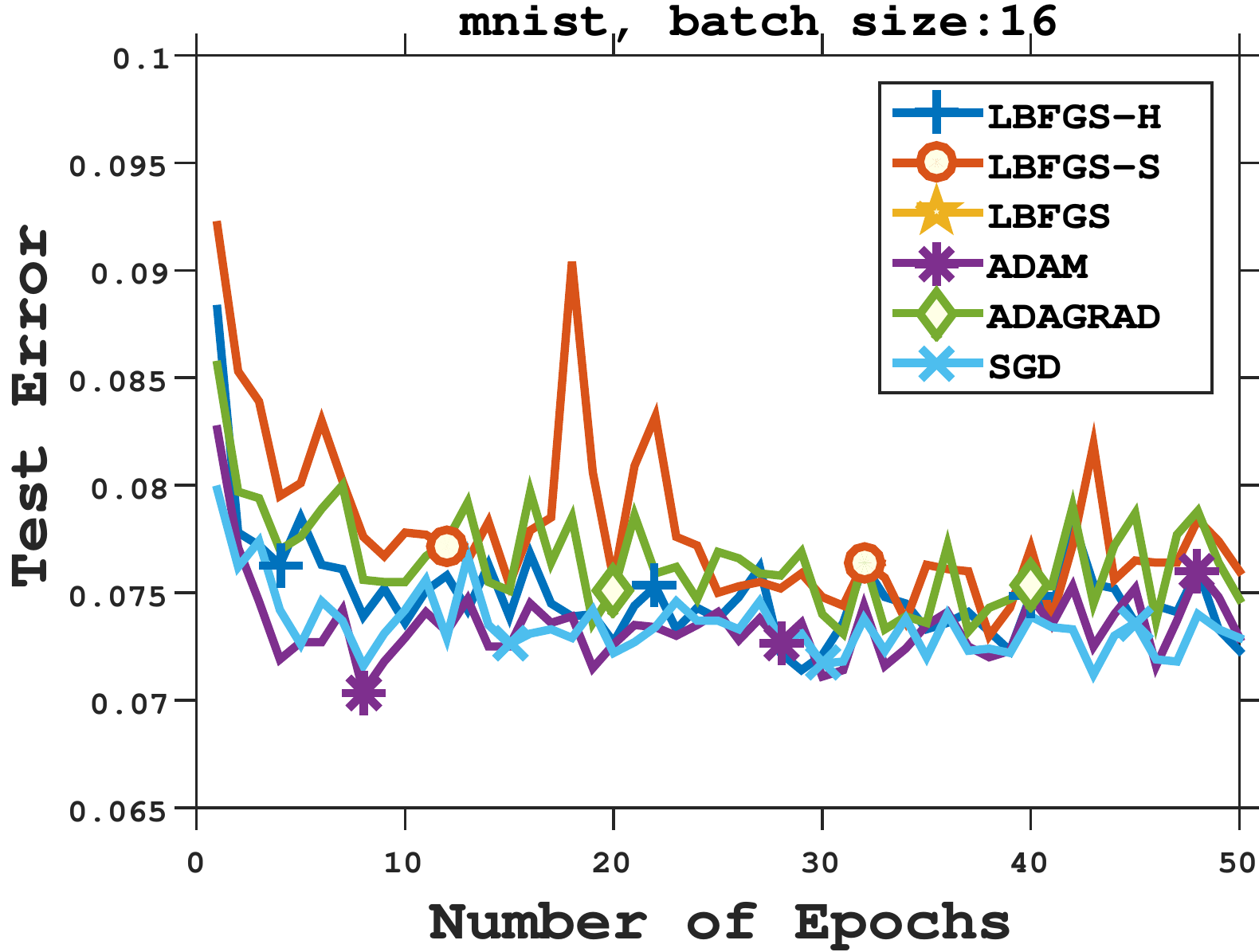,width=0.32\textwidth}
   \epsfig{file=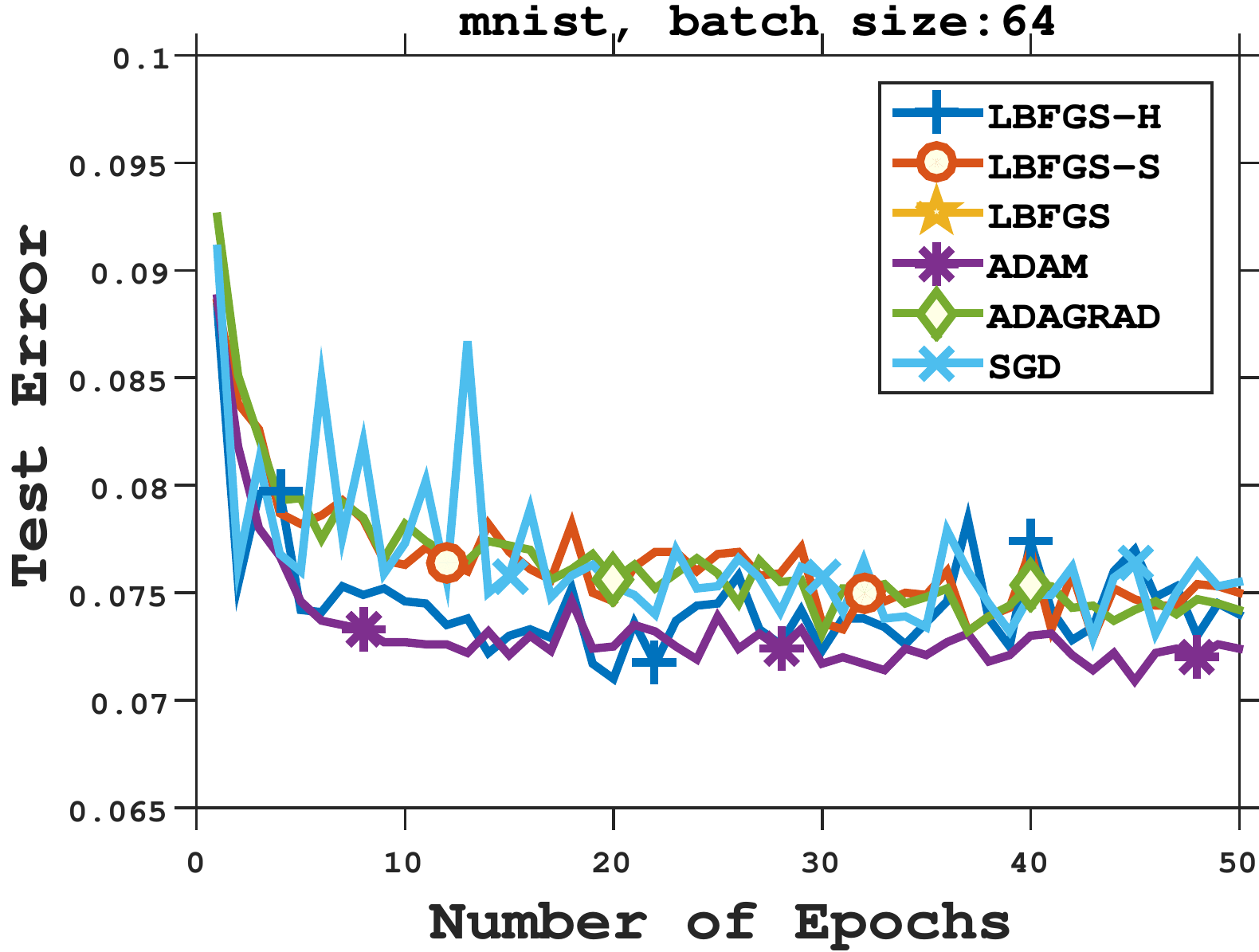,width=0.32\textwidth} 
    \epsfig{file=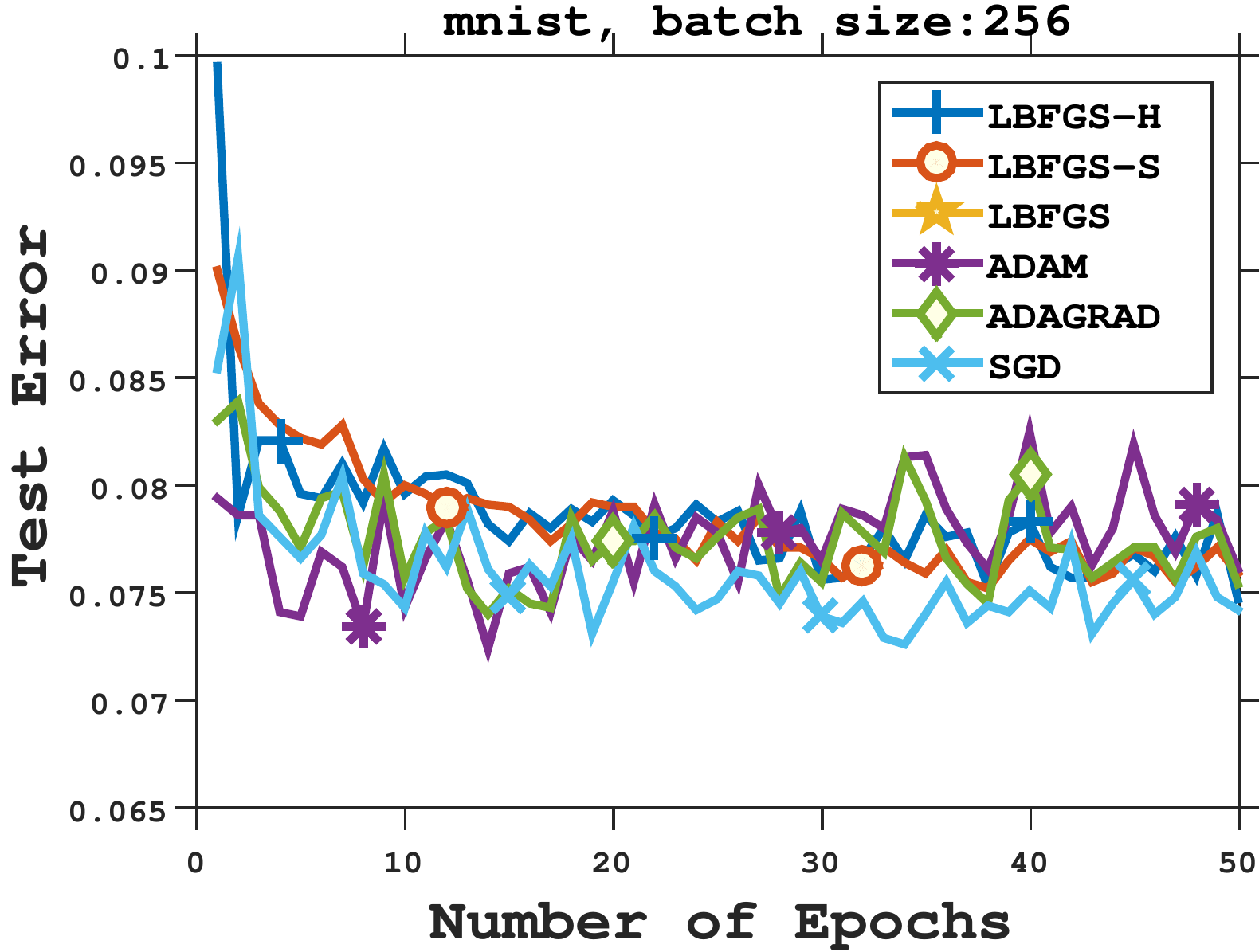,width=0.32\textwidth}
 
 \caption{\footnotesize Comparisons of training loss (top 2 rows), sub-optimality (middle row) and test errors (bottom 2 rows) from different algorithms with batch sizes 16, 64, 256 on \emph{MNIST}, convex, cross-entropy.}
   \label{fig:add4}
 \end{figure}

 \newpage
\subsection{Results on 1 hidden-layer neural network (nonconvex), \emph{MNIST}}

The thrid experiment is conducted for 1 hidden-layer neural network with cross-entropy loss on the dataset \emph{MNIST}. 
 
\subsubsection{Small Batch Sizes}
\begin{figure} [H]
\centering
 \epsfig{file=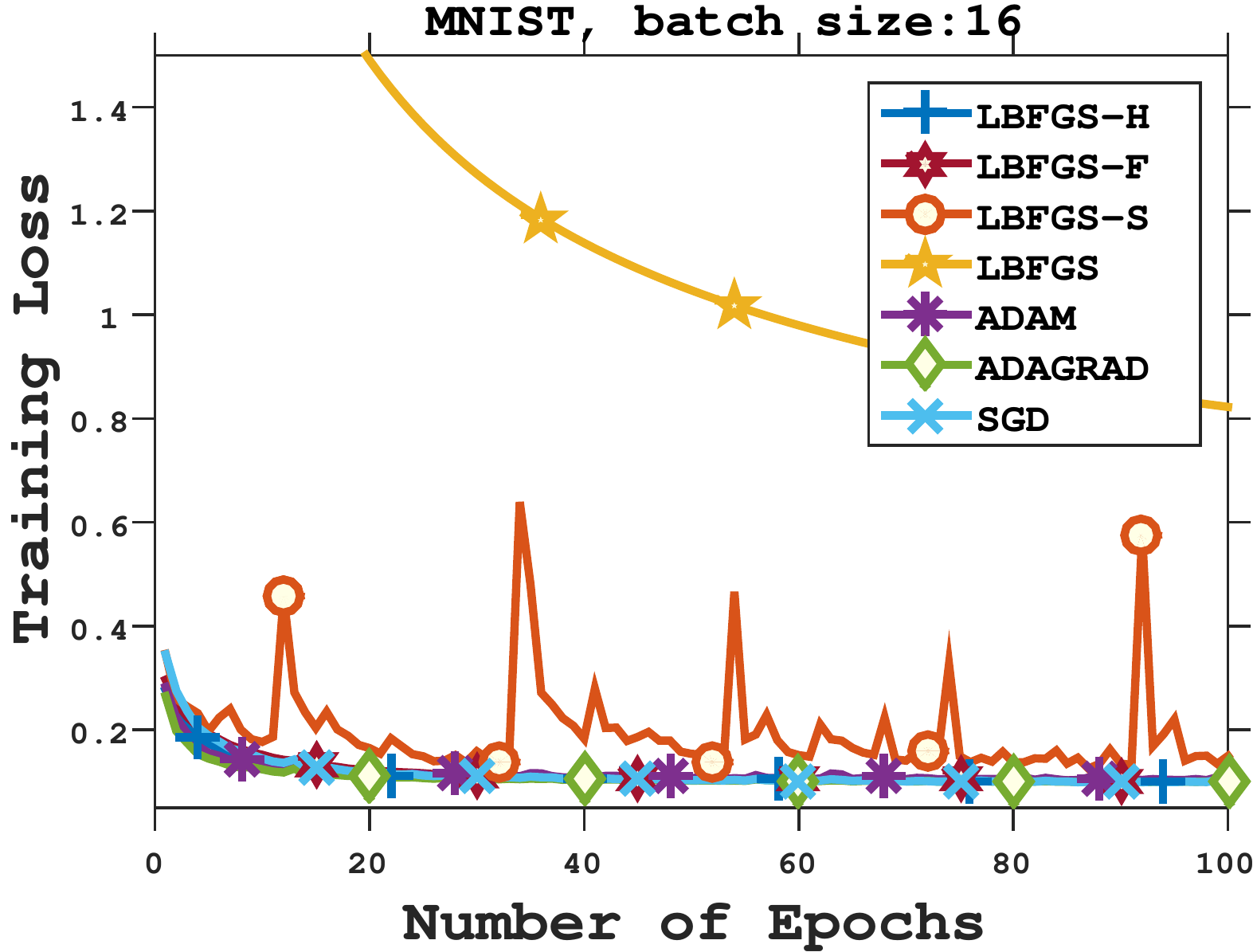,width=0.32\textwidth}
  \epsfig{file=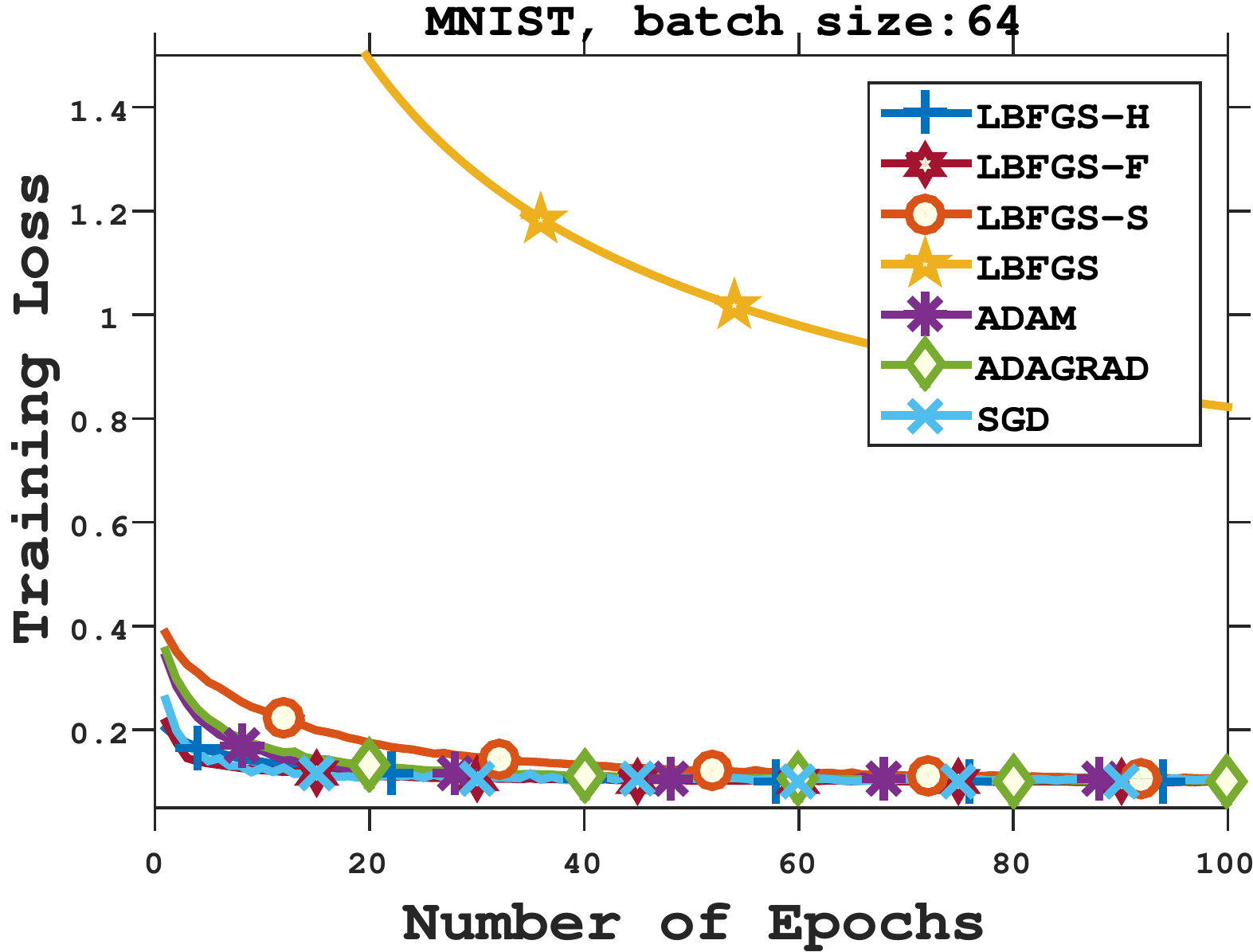,width=0.32\textwidth} 
  \epsfig{file=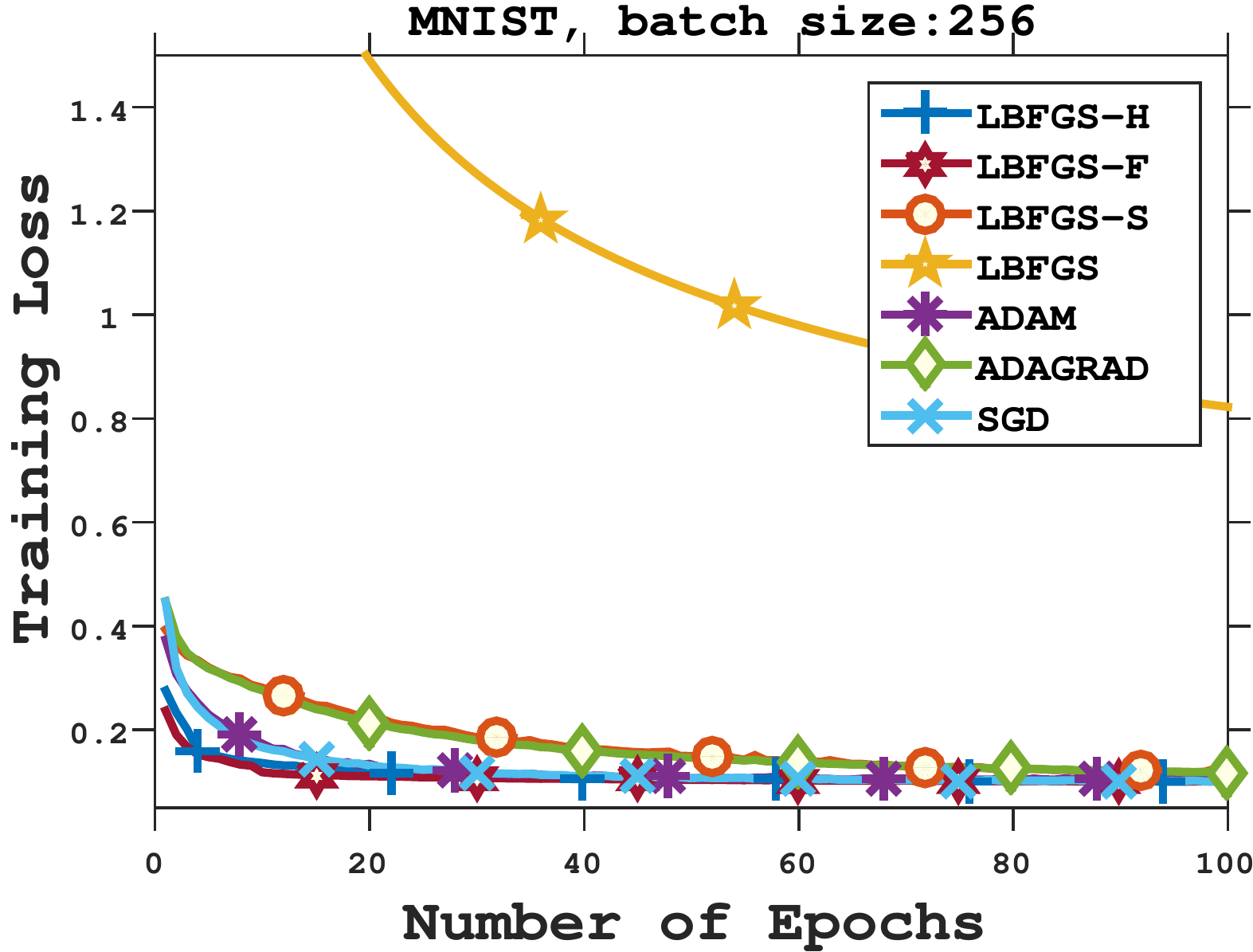,width=0.32\textwidth} 
  
 \epsfig{file=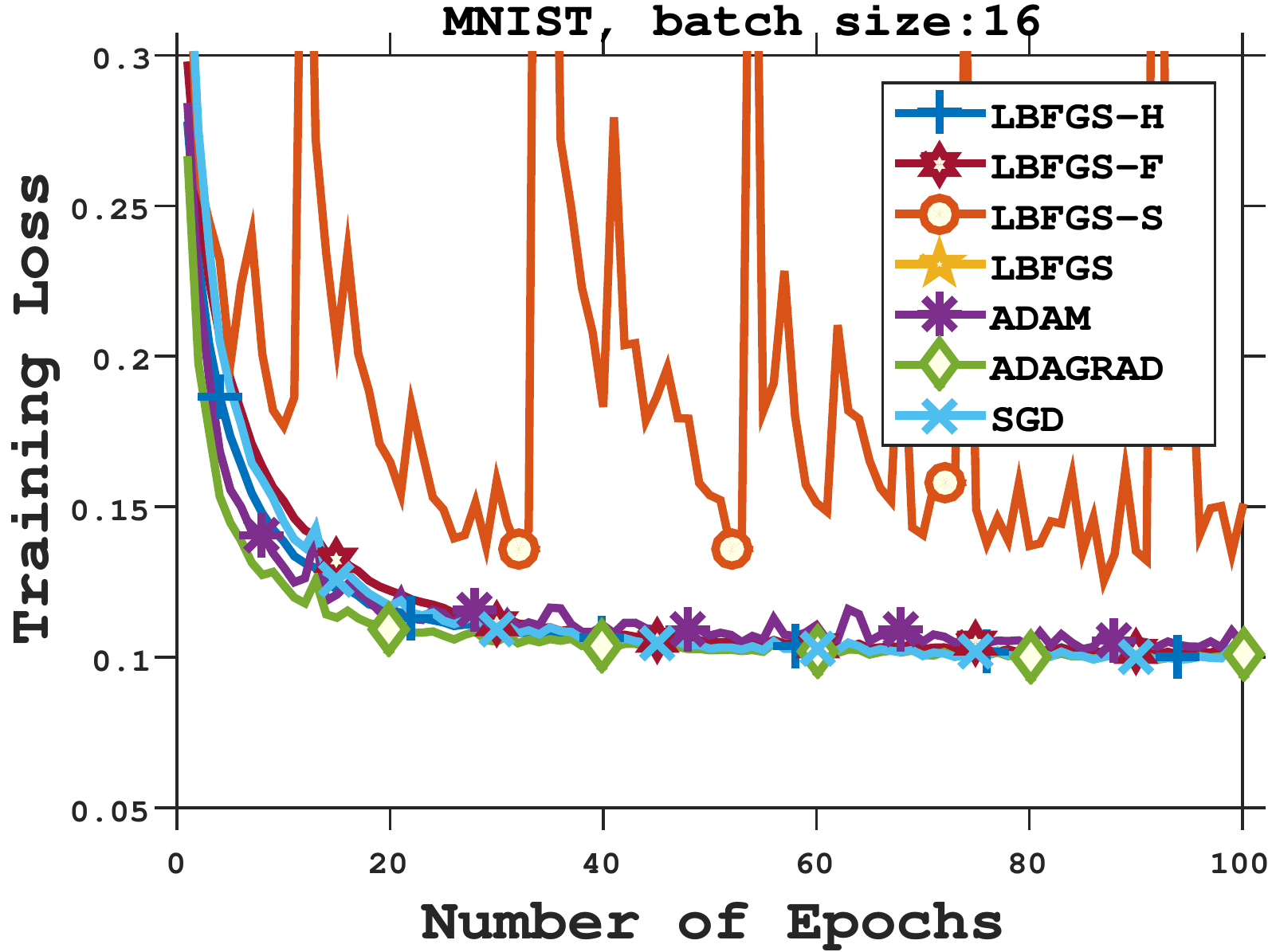,width=0.32\textwidth}
  \epsfig{file=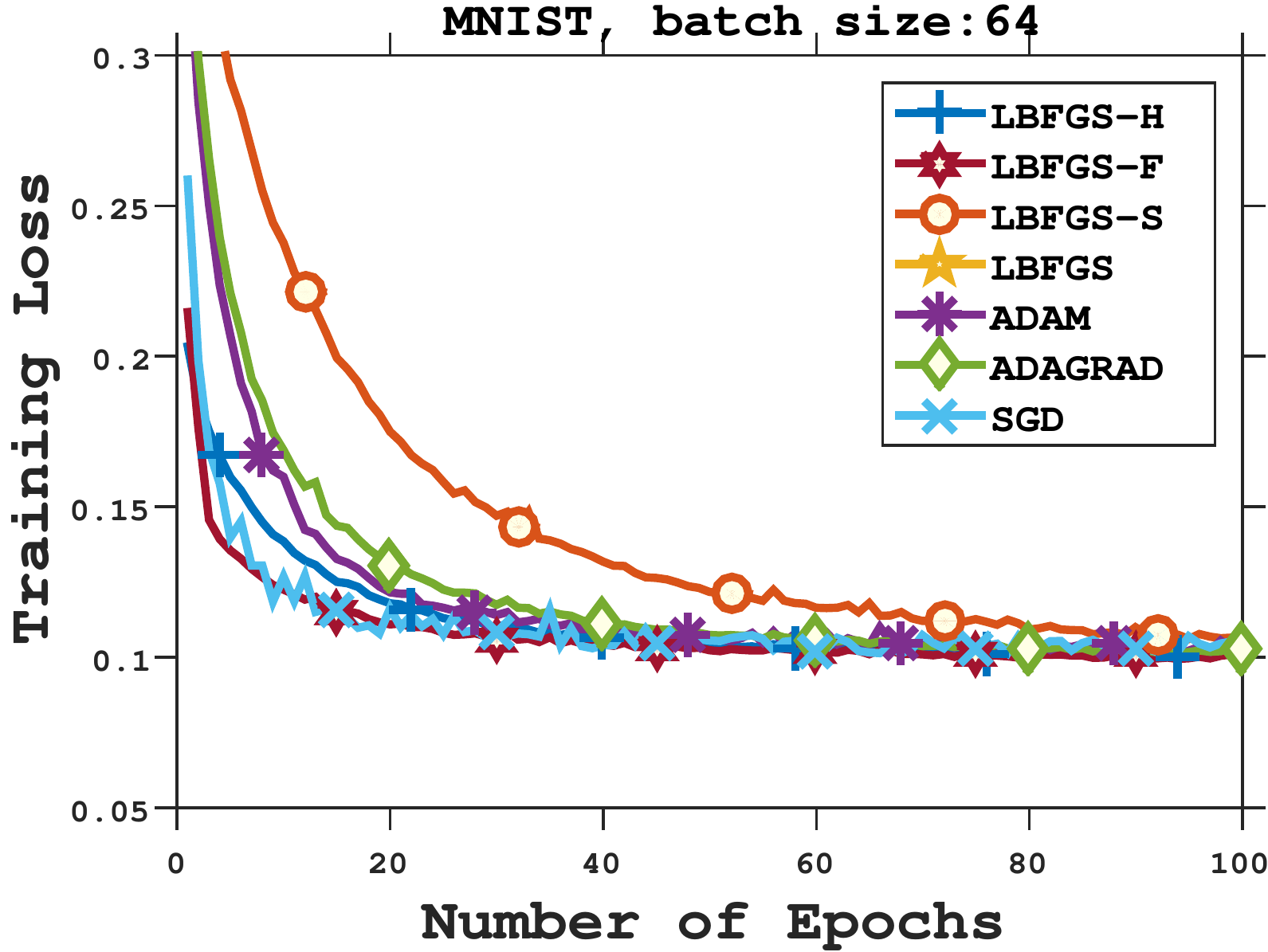,width=0.32\textwidth} 
  \epsfig{file=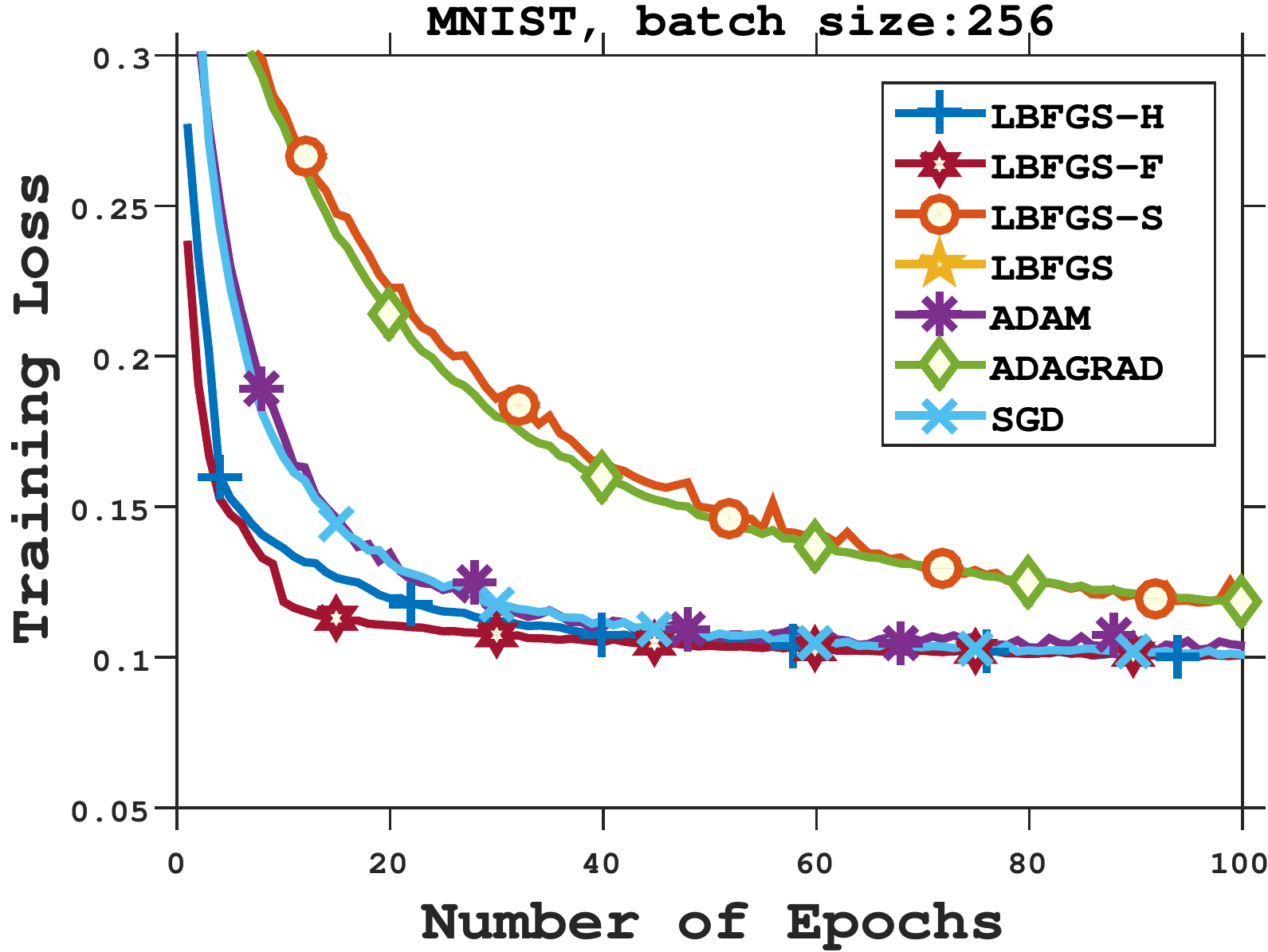,width=0.32\textwidth} 

 \epsfig{file=Figs/mnist_loss16_2_v2.eps,width=0.32\textwidth}
  \epsfig{file=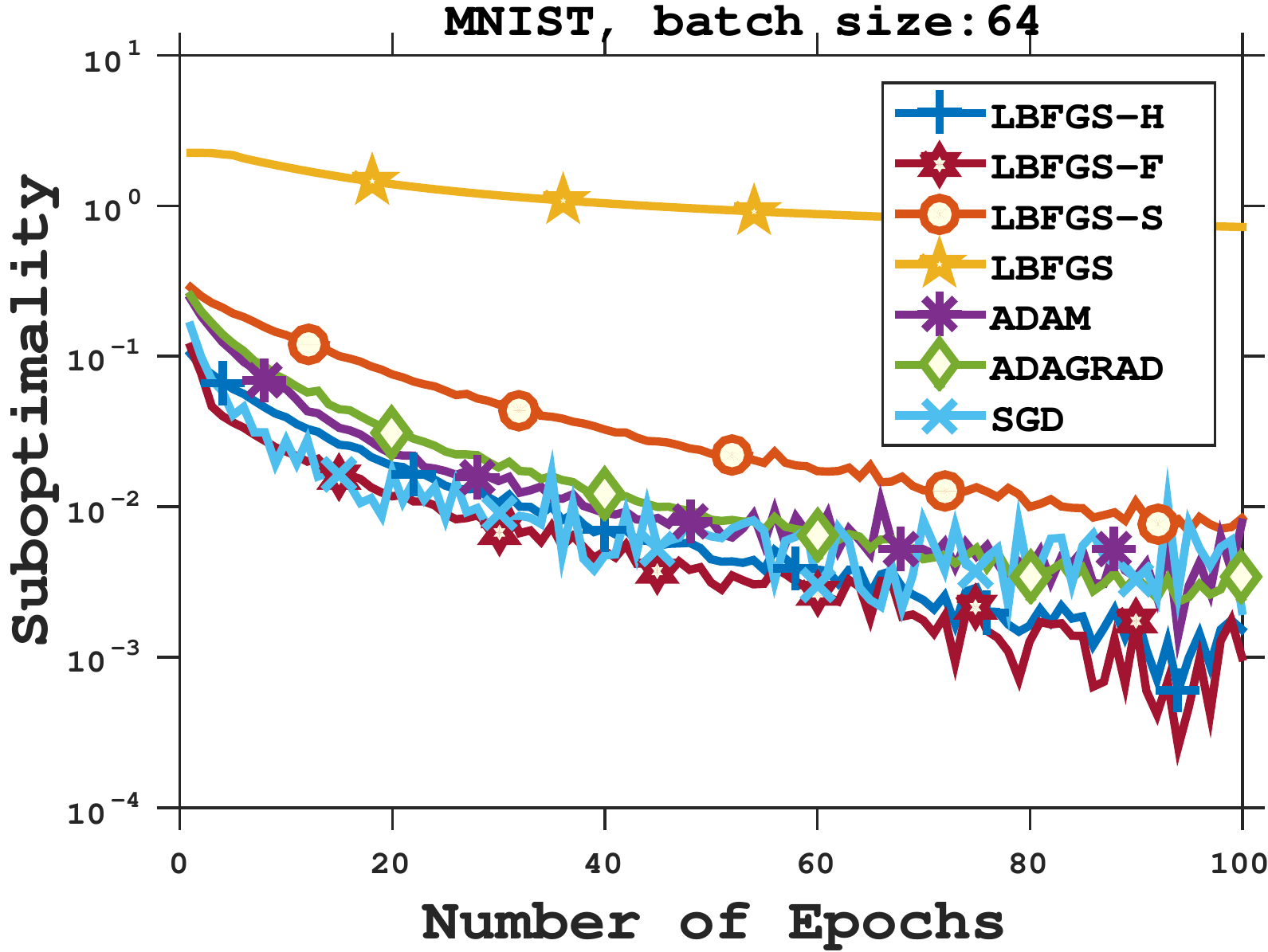,width=0.32\textwidth} 
  \epsfig{file=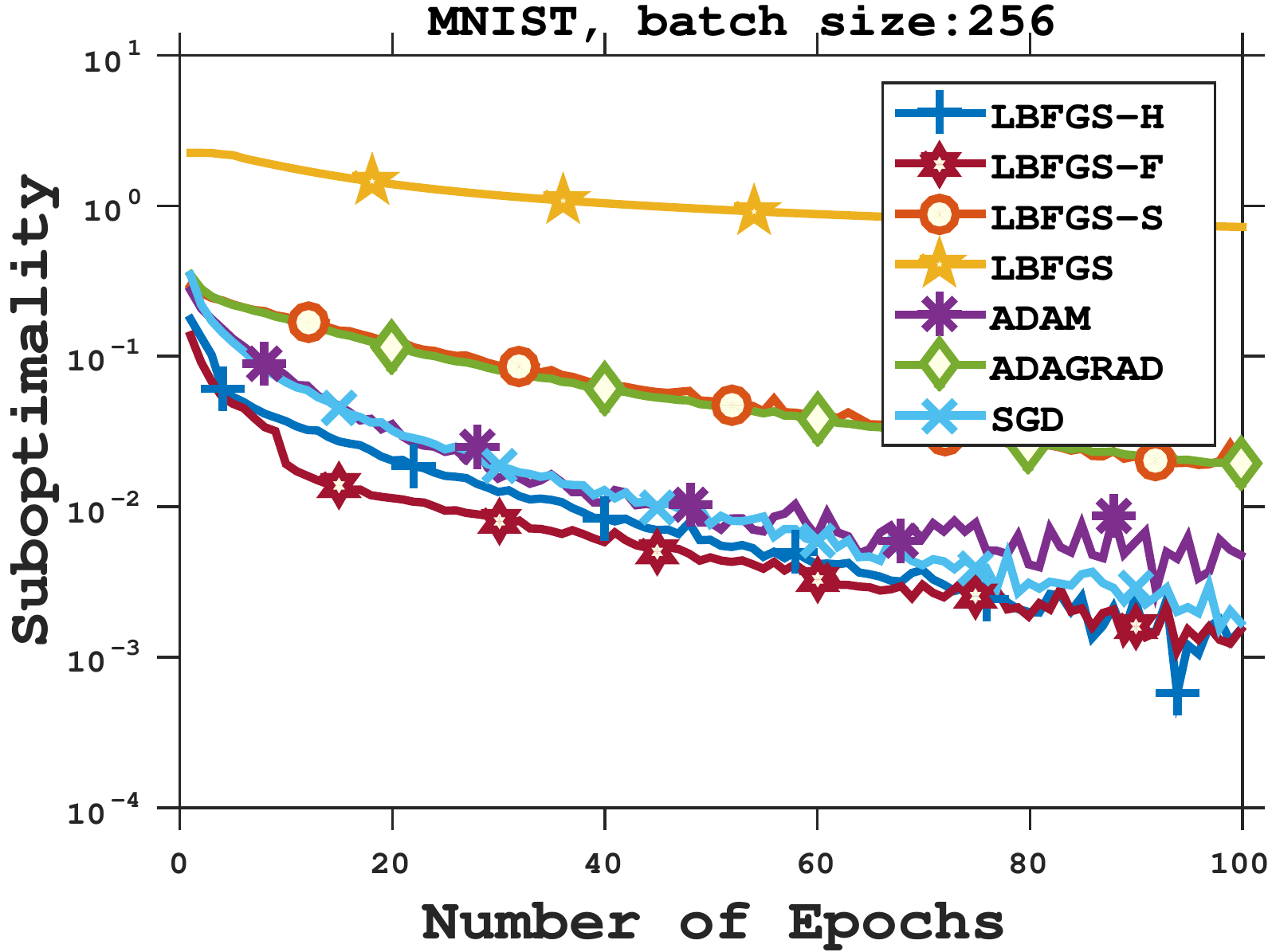,width=0.32\textwidth}

   \epsfig{file=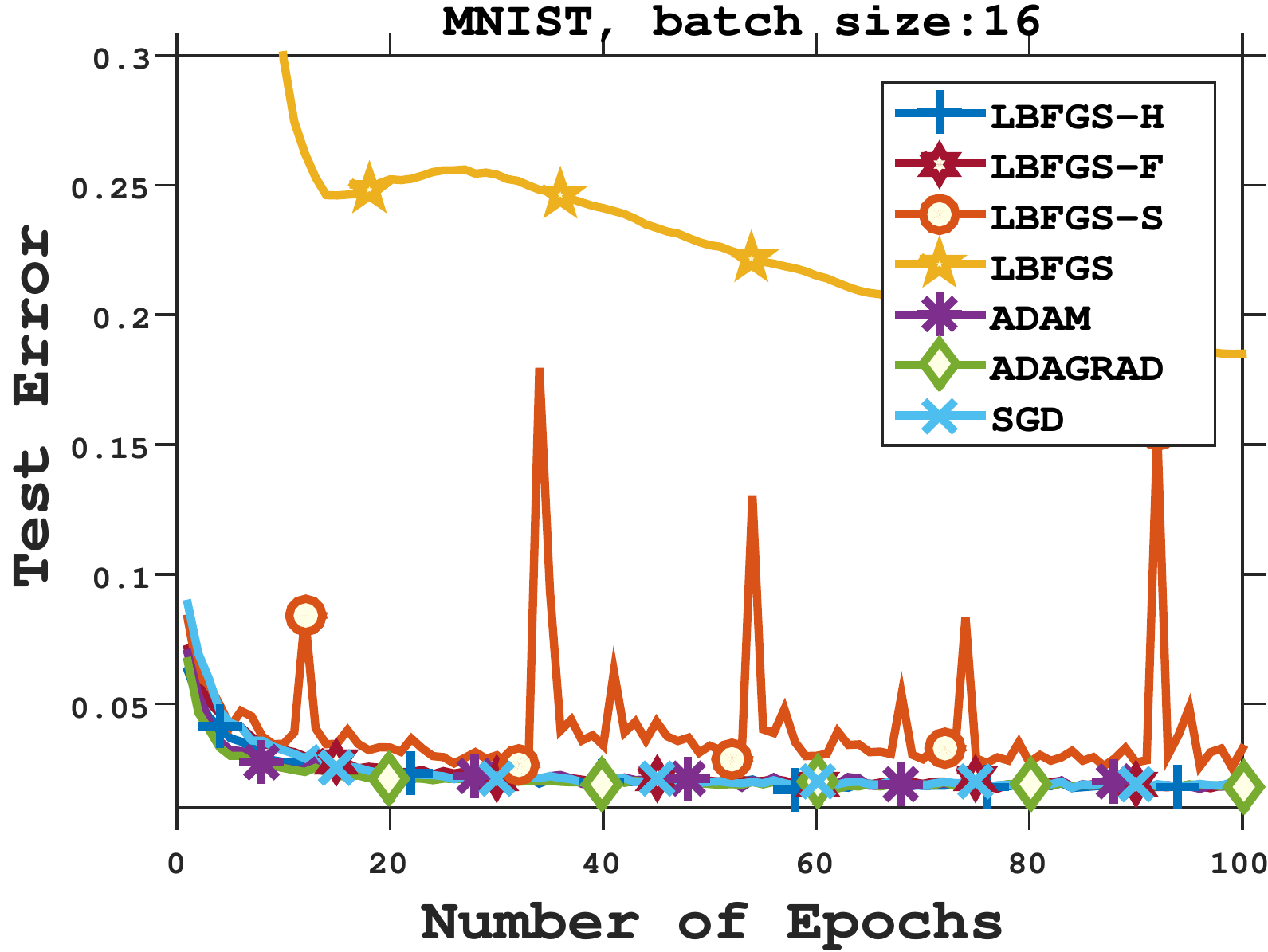,width=0.32\textwidth}
   \epsfig{file=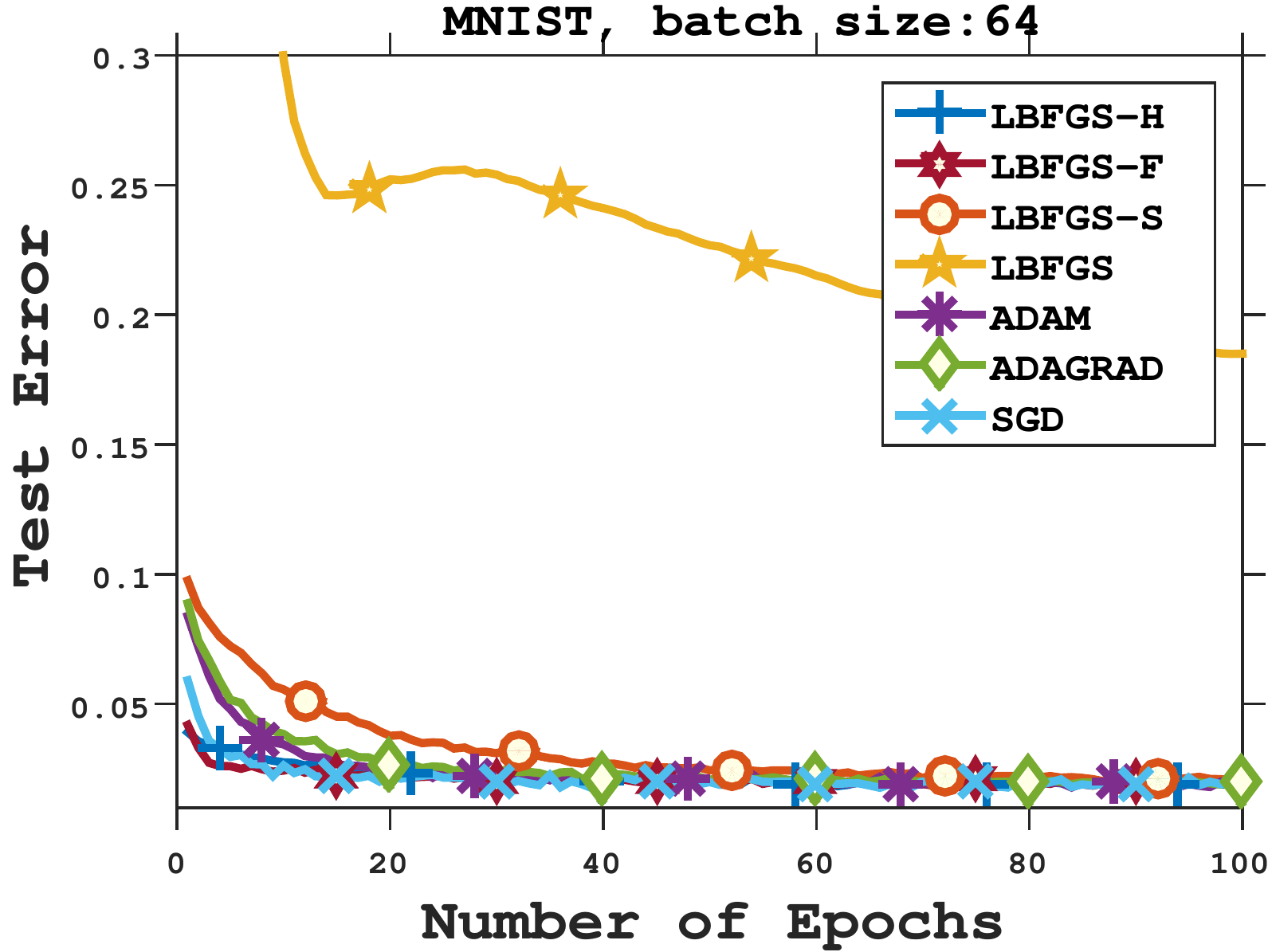,width=0.32\textwidth} 
    \epsfig{file=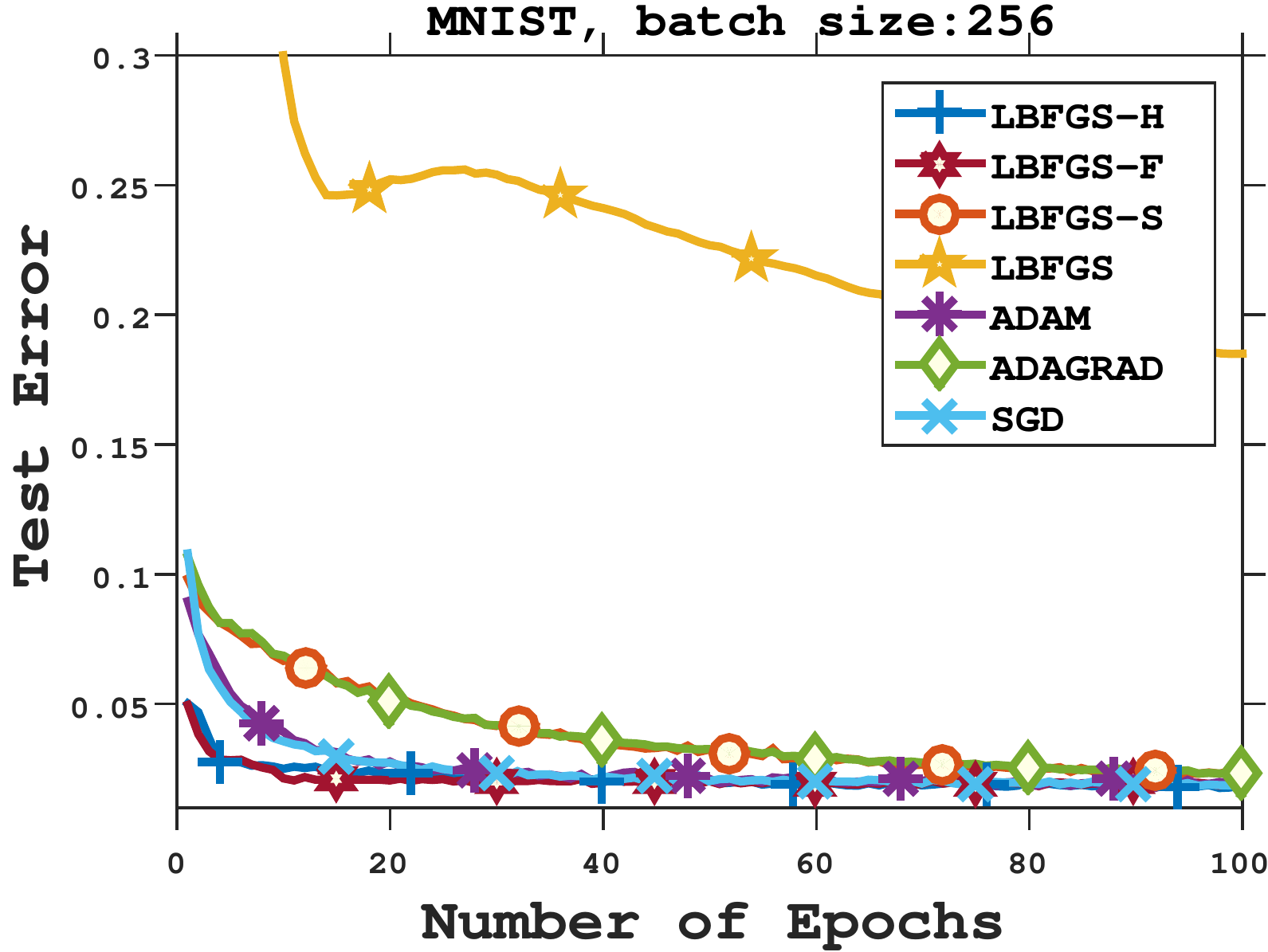,width=0.32\textwidth}
    \epsfig{file=Figs/mnist_error16closer_v2.eps,width=0.32\textwidth}
   \epsfig{file=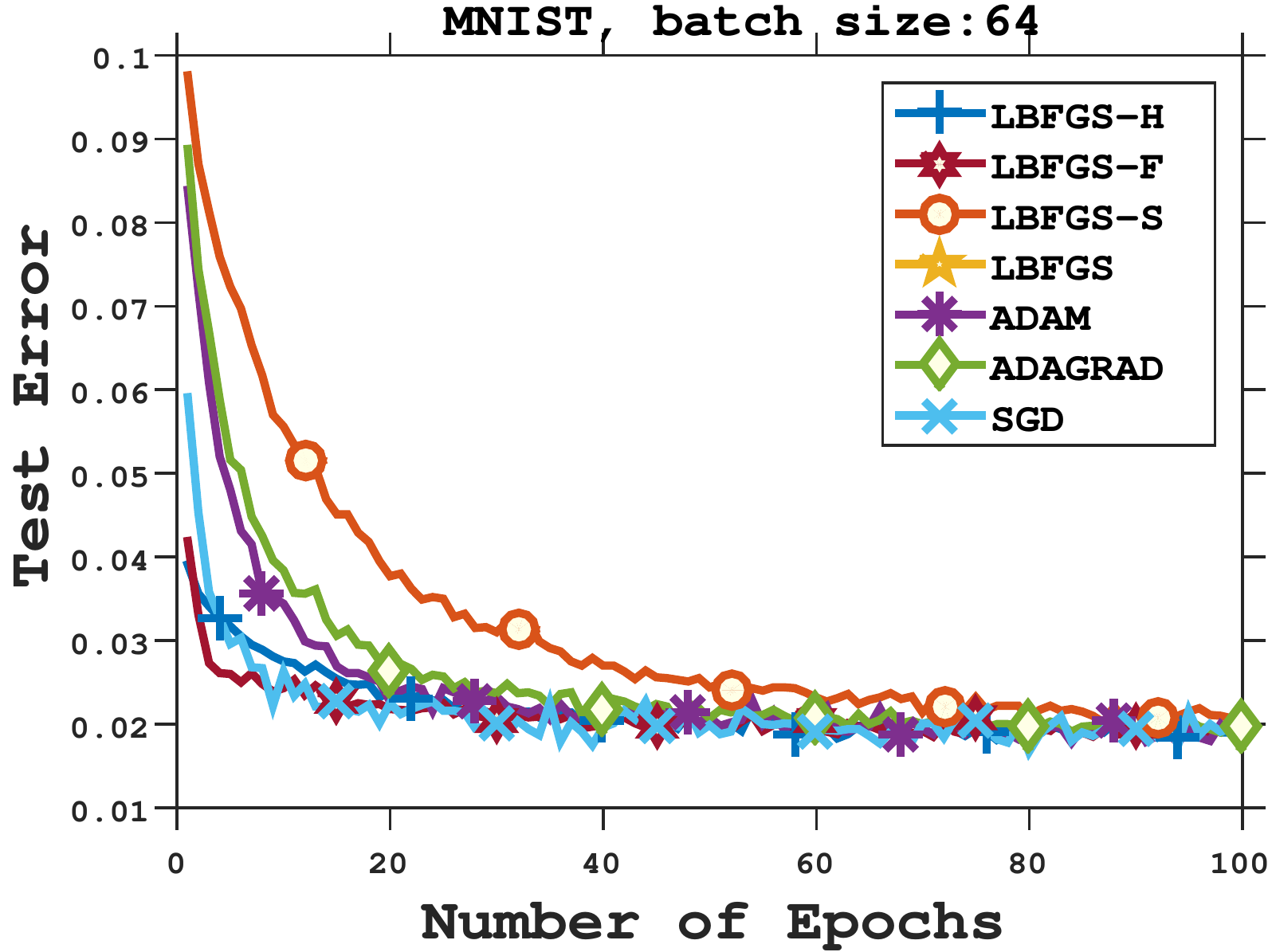,width=0.32\textwidth} 
    \epsfig{file=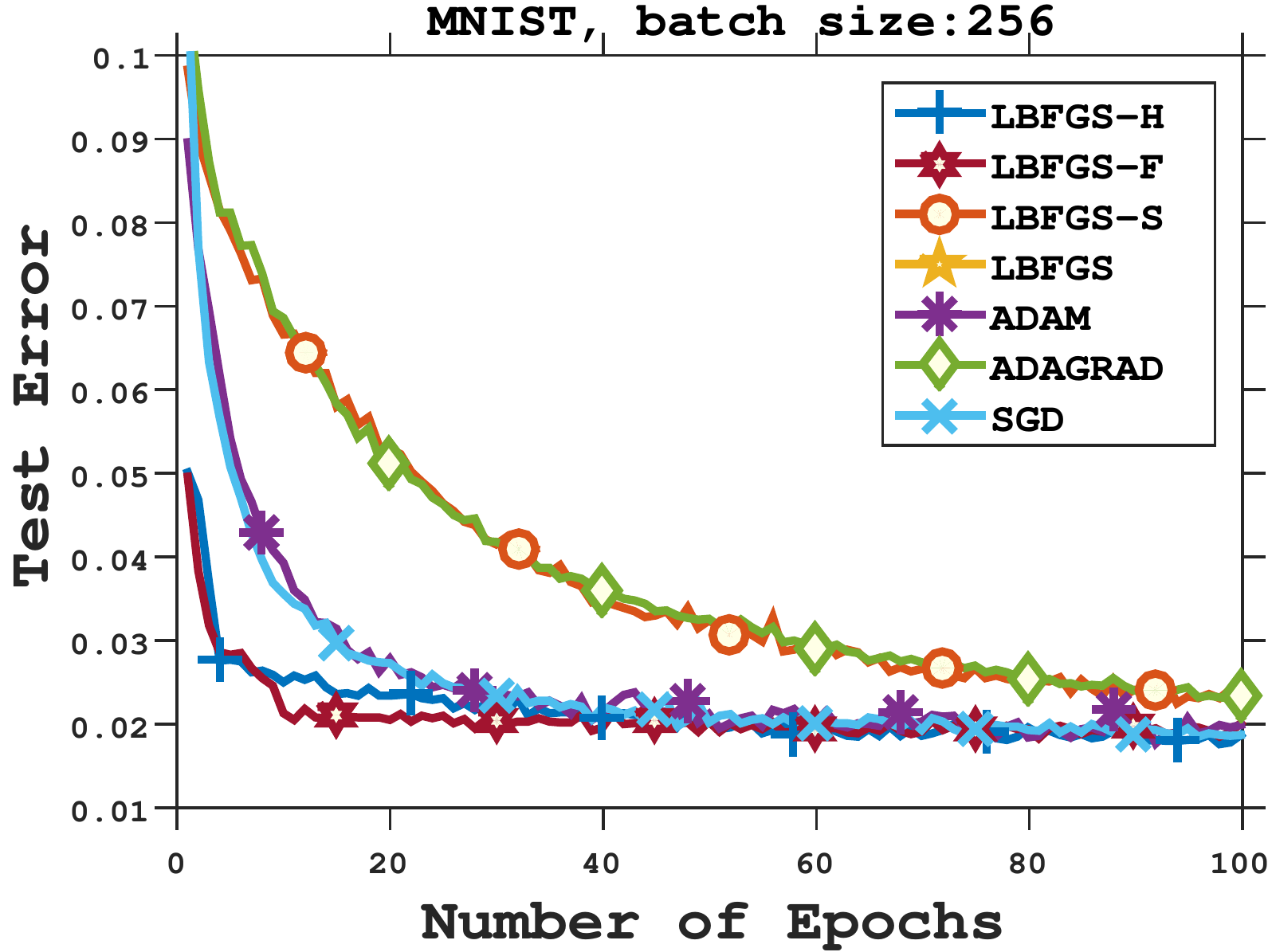,width=0.32\textwidth}
 
 \caption{\footnotesize Comparisons of training loss (top 2 rows), sub-optimality (middle row) and test errors (bottom 2 rows) from different algorithms with batch sizes 16, 64, 256 on \emph{MNIST}, nonconvex, neural network with 1 hidden layer of 300 hidden units.}
   \label{fig:add5}
 \end{figure}

 In Figure~\ref{fig:add5}, the instability of LBFGS-S is more severe on this nonconvex problem. LBFGS-H and LBFGS-F continues to be superior than the others in the case when the batch size $b=64, 256$ and ADAGRAD obviously slows down with the increase of the batch size.
 
 \newpage
  \subsubsection{Larger Batch Sizes}
\begin{figure}[H]
\centering

  \epsfig{file=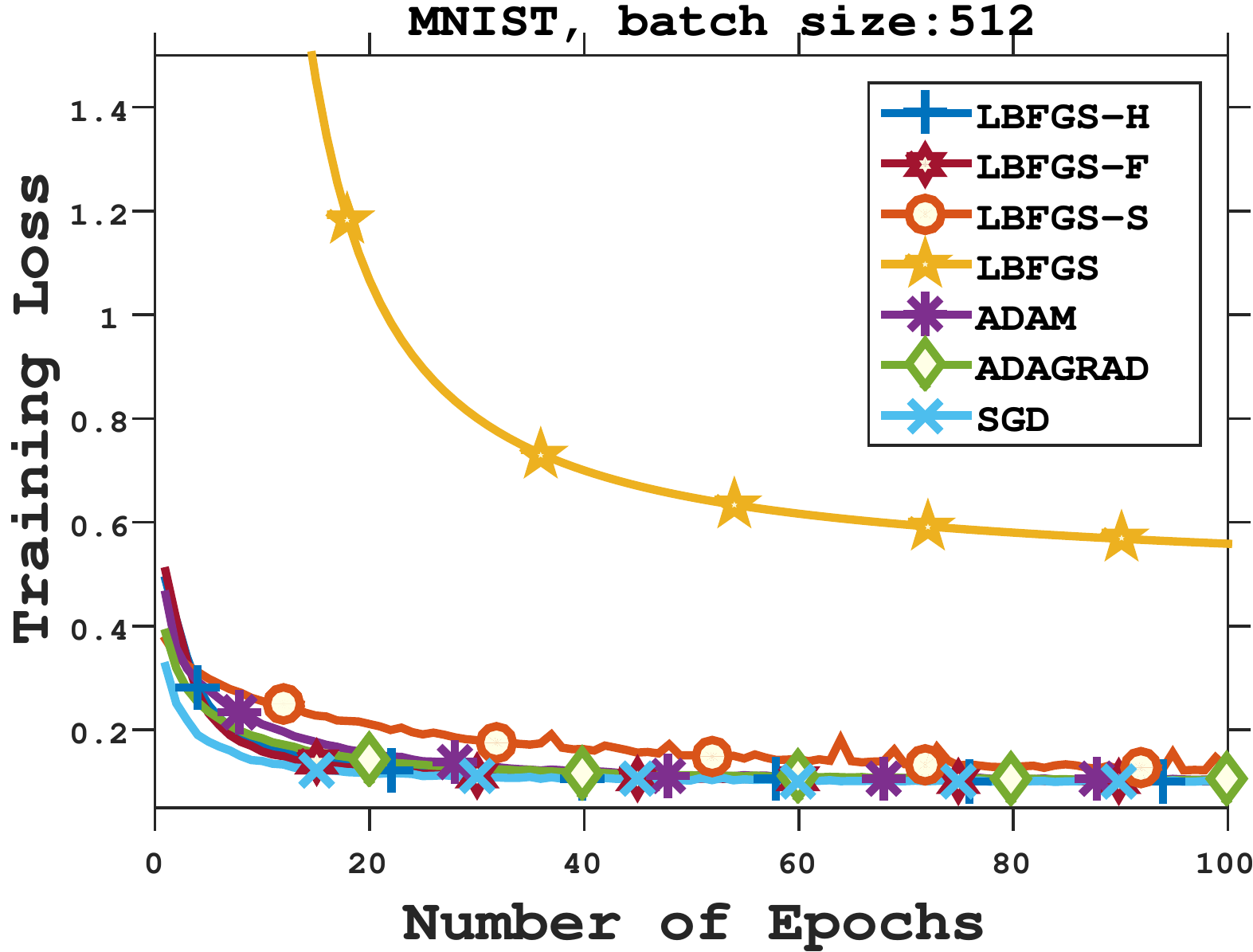,width=0.24\textwidth}
  \epsfig{file=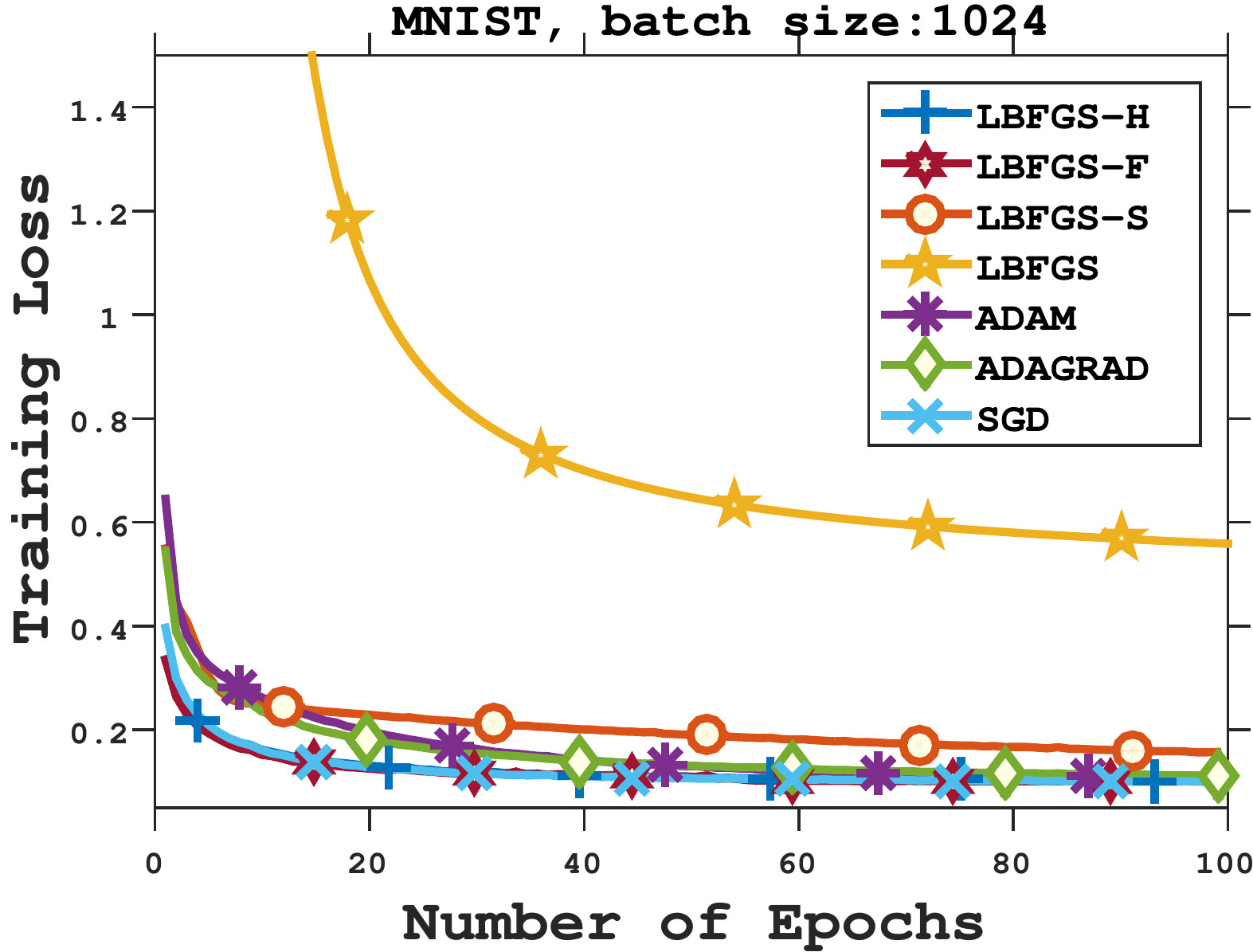,width=0.24\textwidth} 
  \epsfig{file=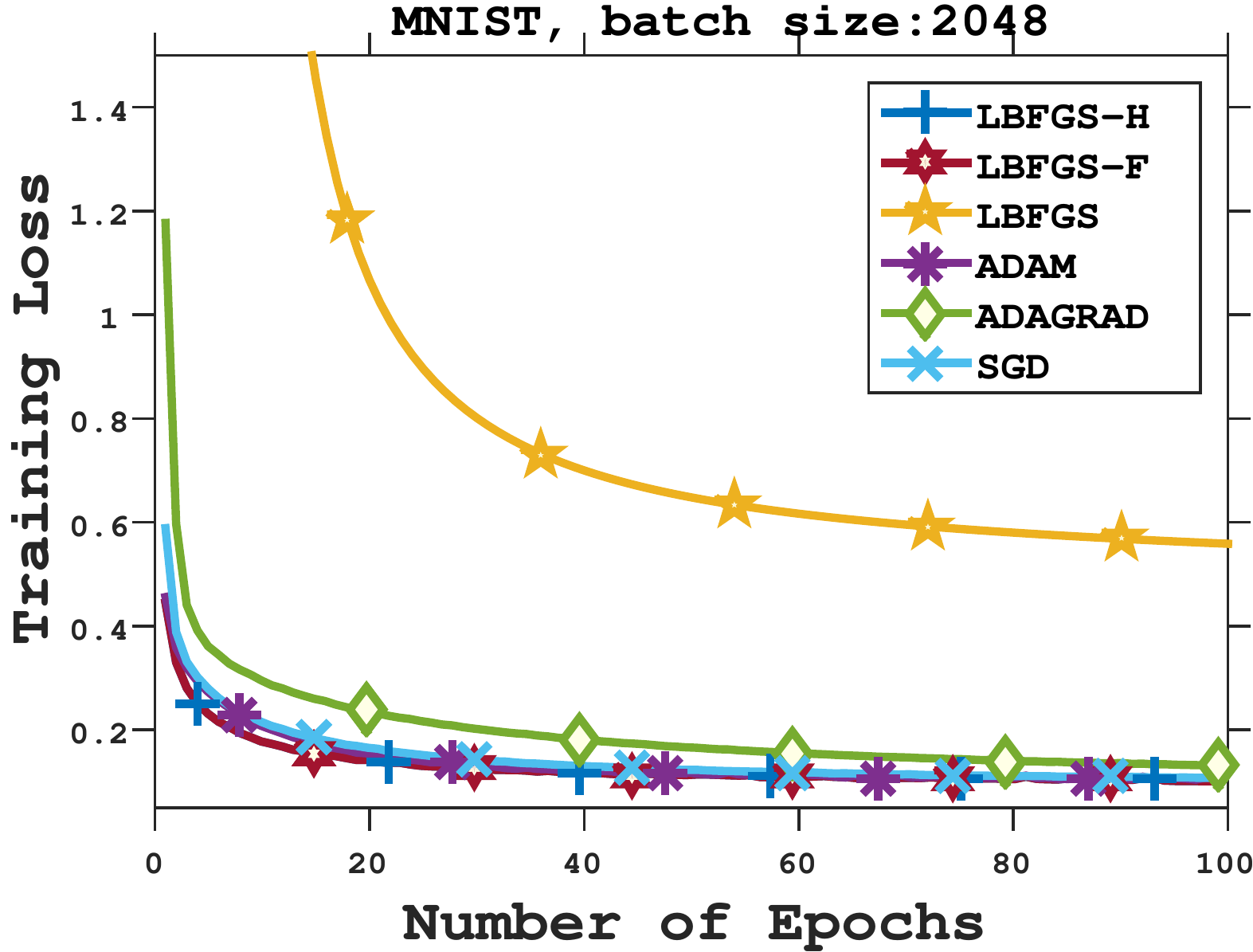,width=0.24\textwidth} 
  \epsfig{file=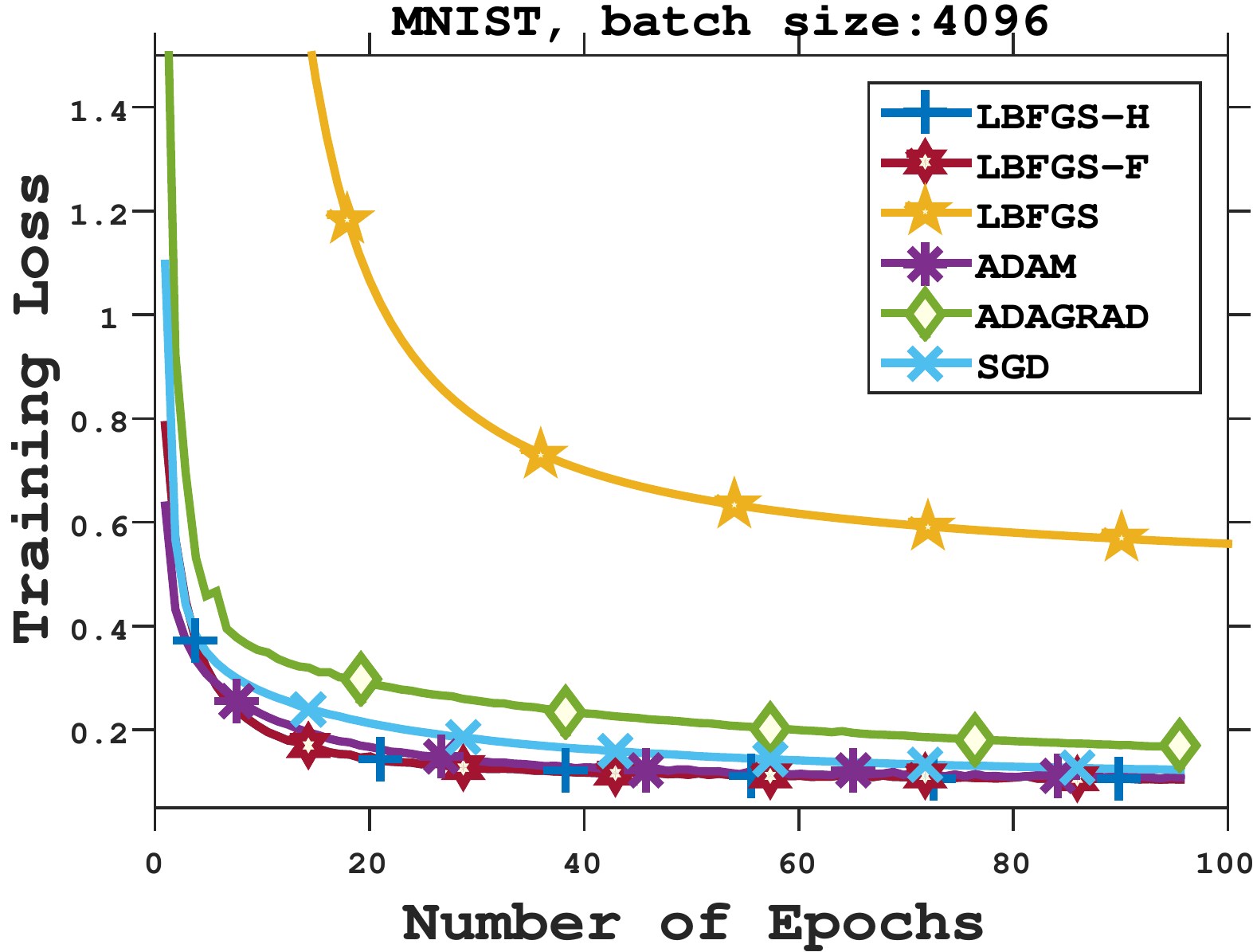,width=0.24\textwidth} 
  
  \epsfig{file=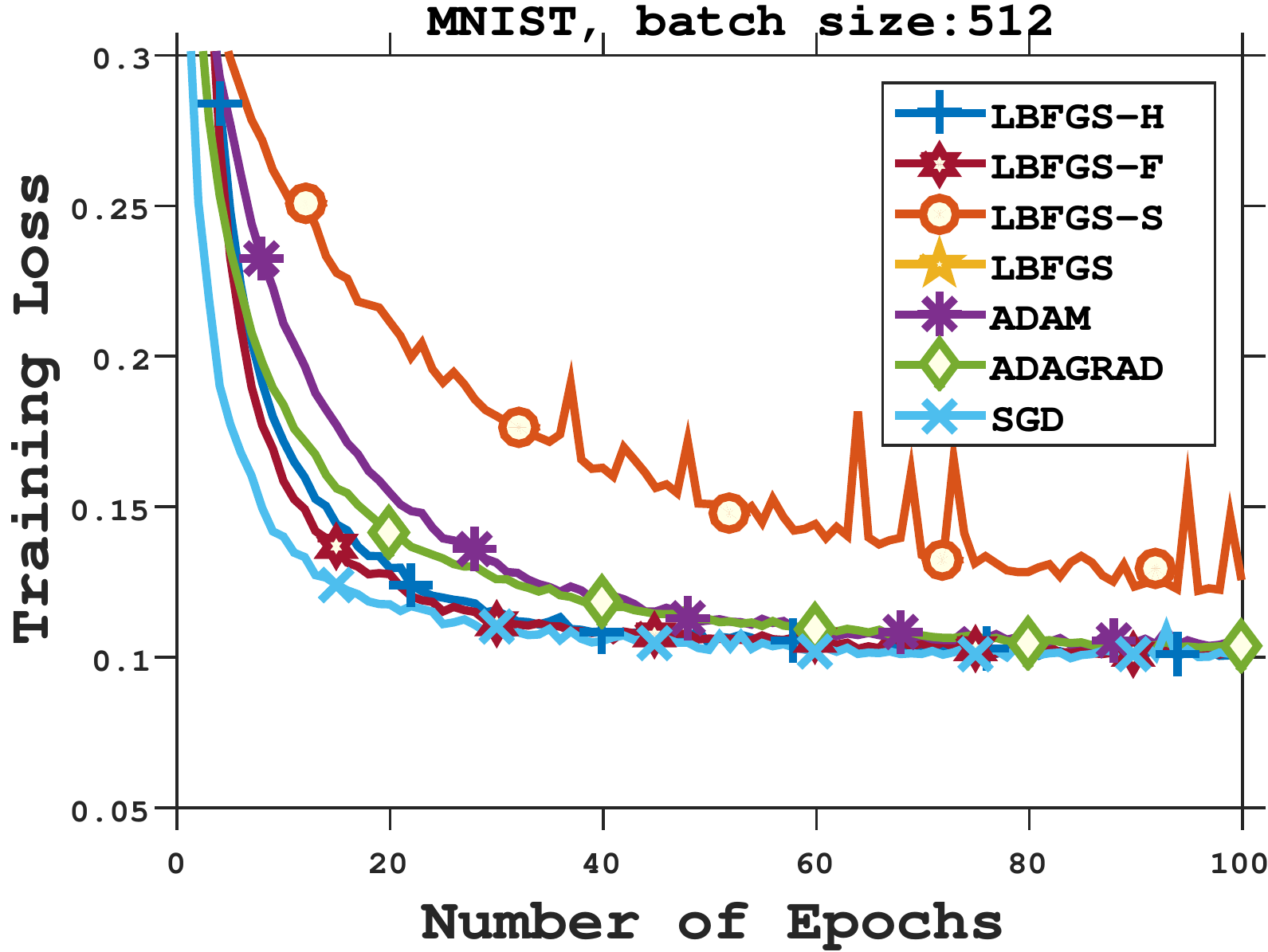,width=0.24\textwidth}
  \epsfig{file=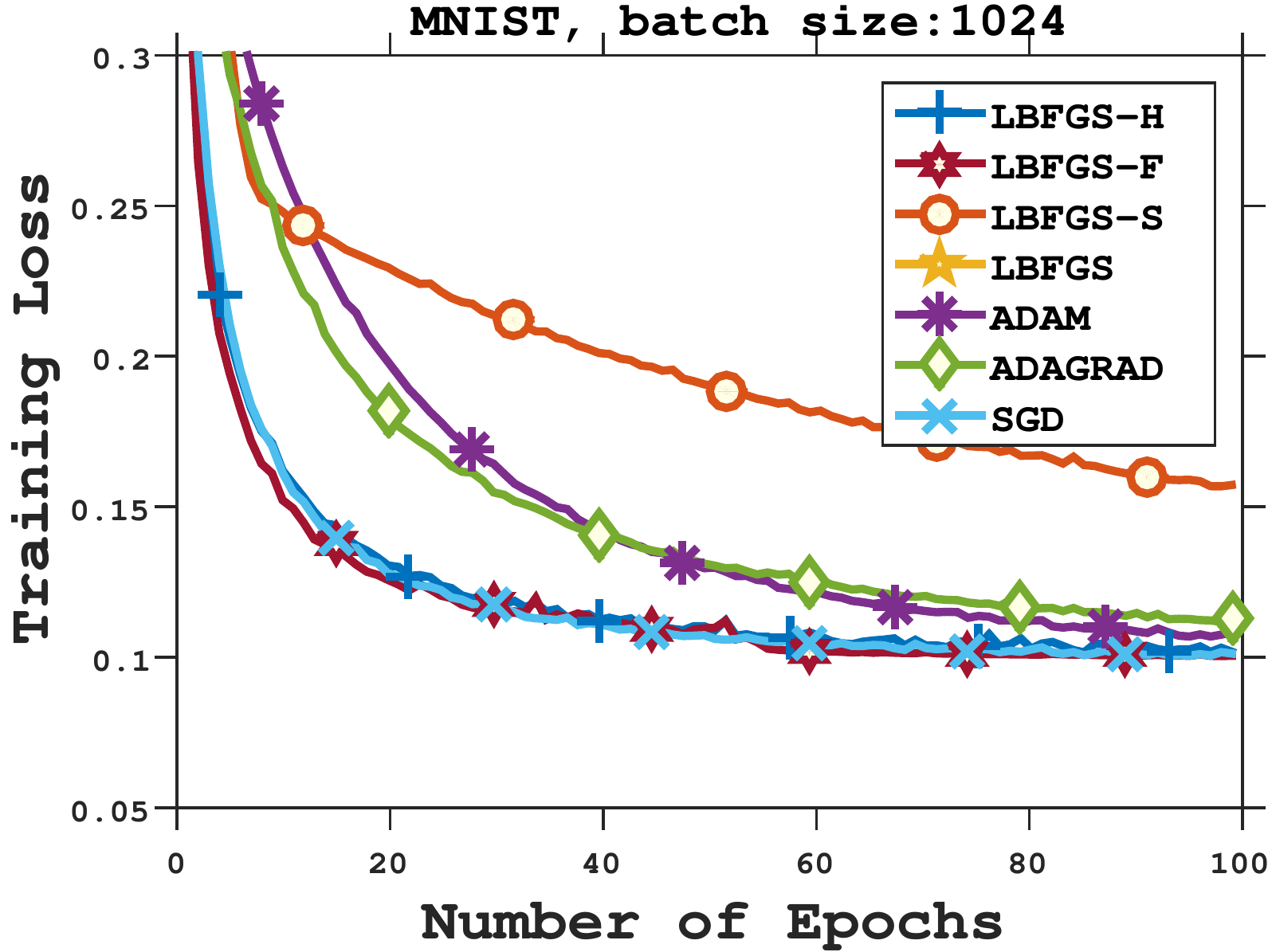,width=0.24\textwidth} 
  \epsfig{file=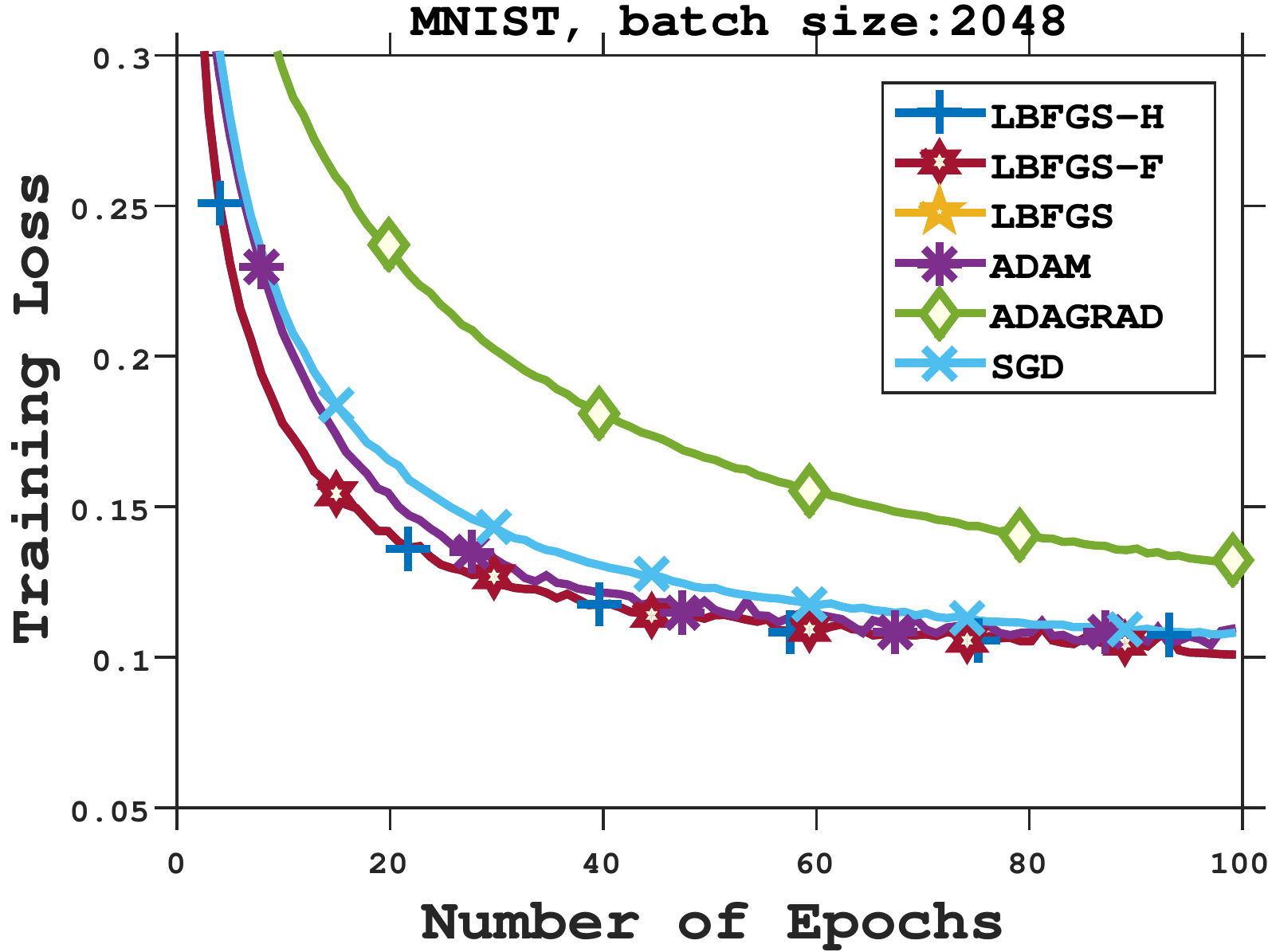,width=0.24\textwidth} 
  \epsfig{file=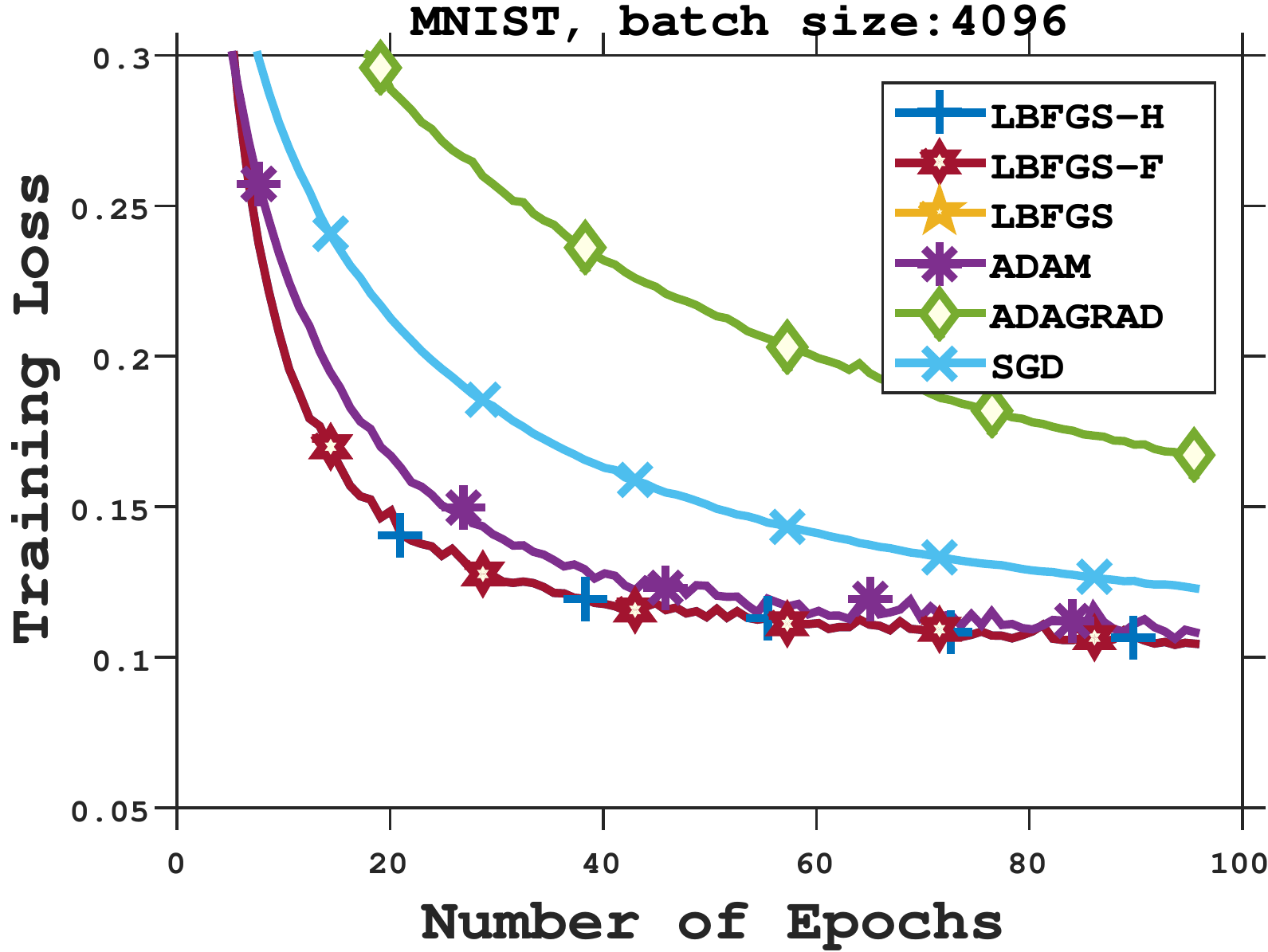,width=0.24\textwidth}

 \epsfig{file=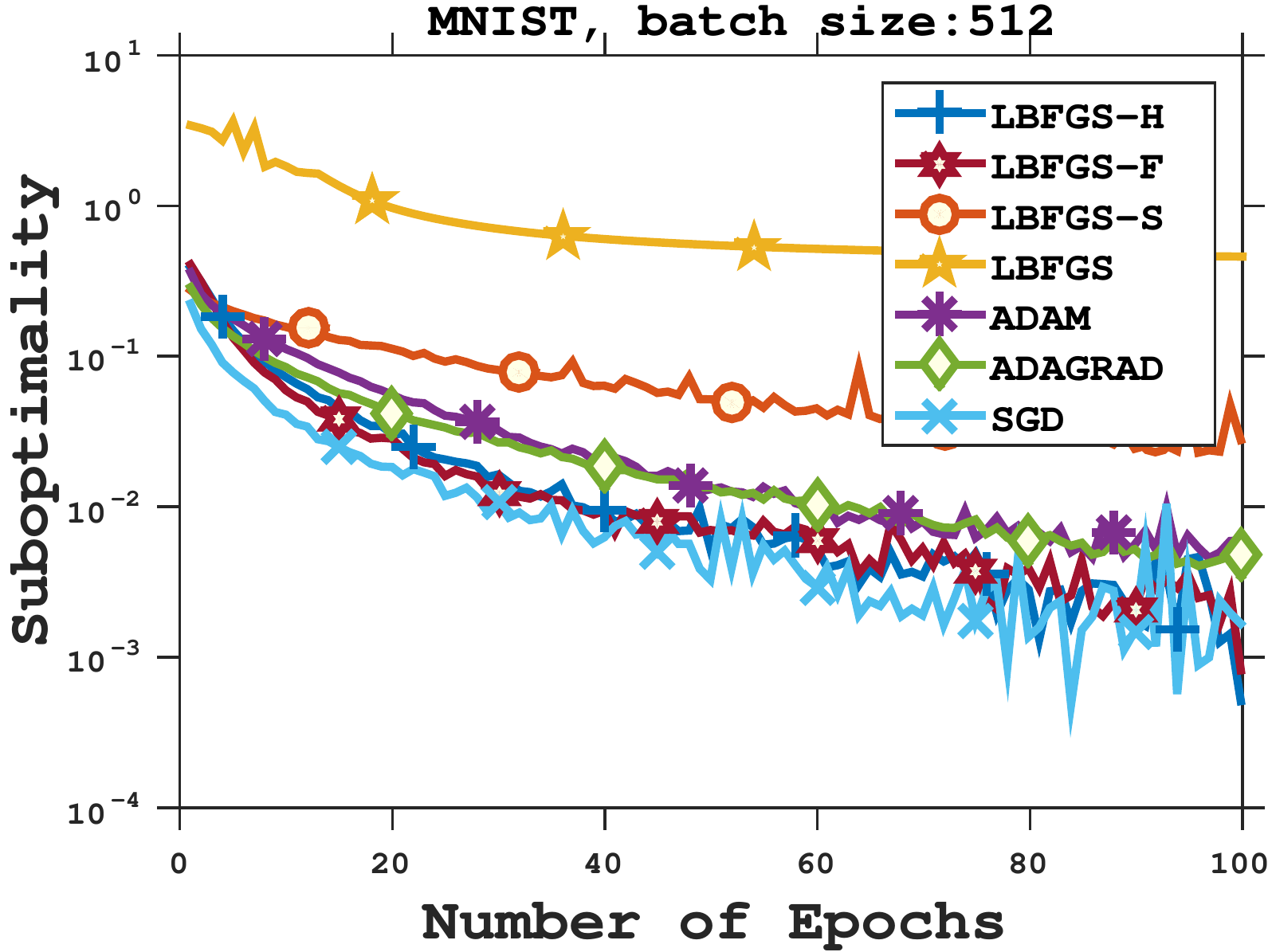,width=0.24\textwidth}
 \epsfig{file=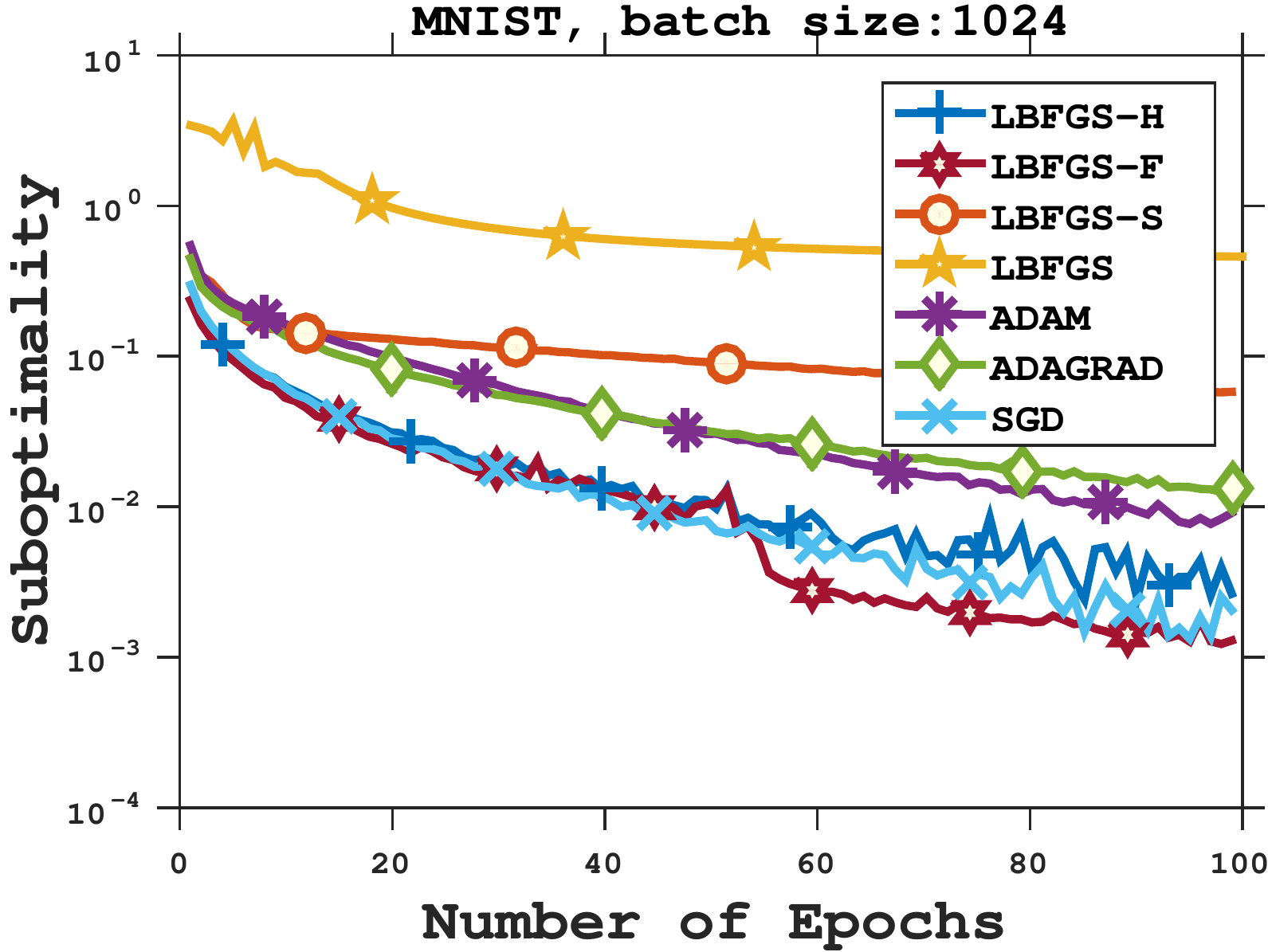,width=0.24\textwidth} 
  \epsfig{file=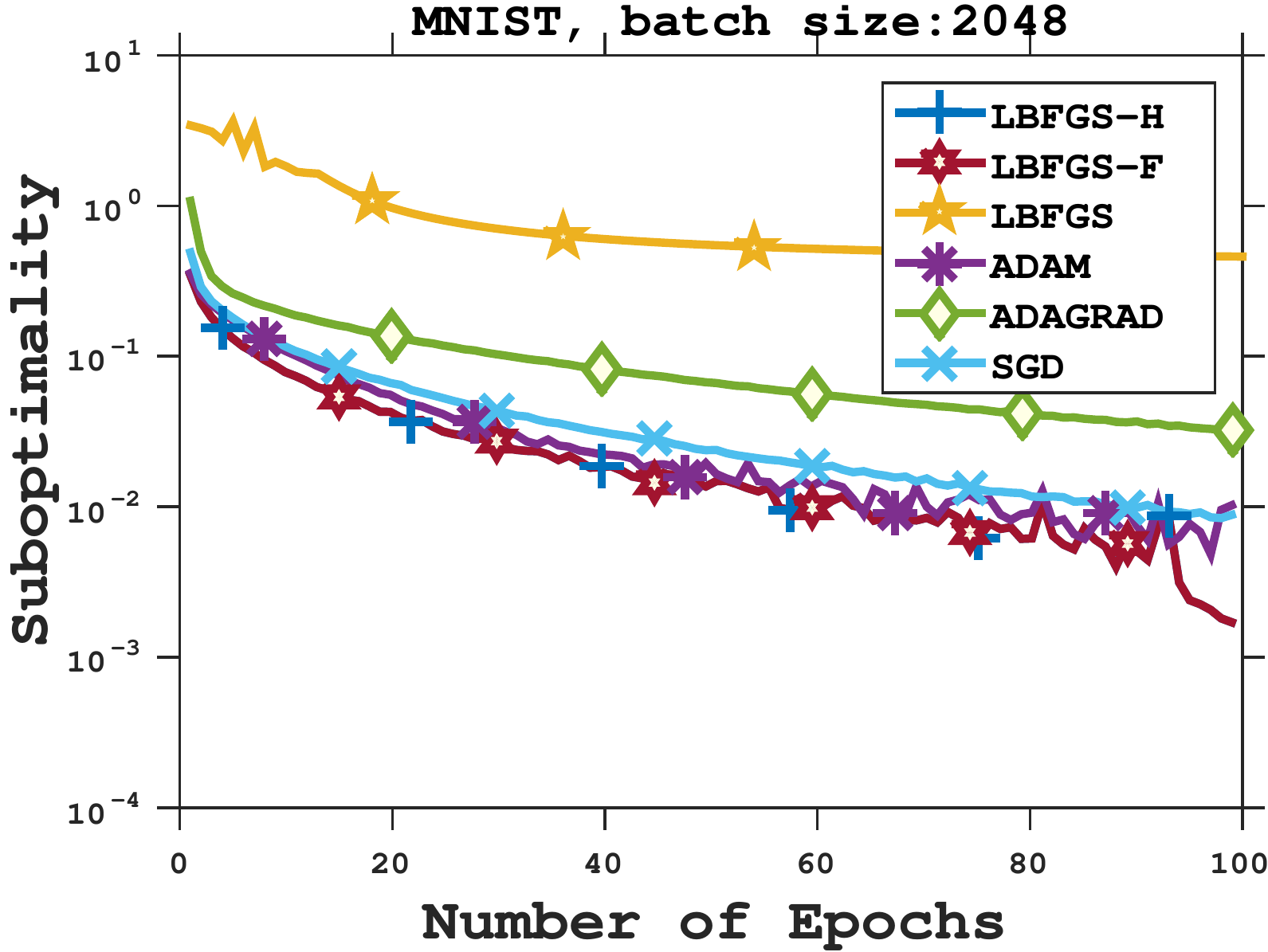,width=0.24\textwidth} 
   \epsfig{file=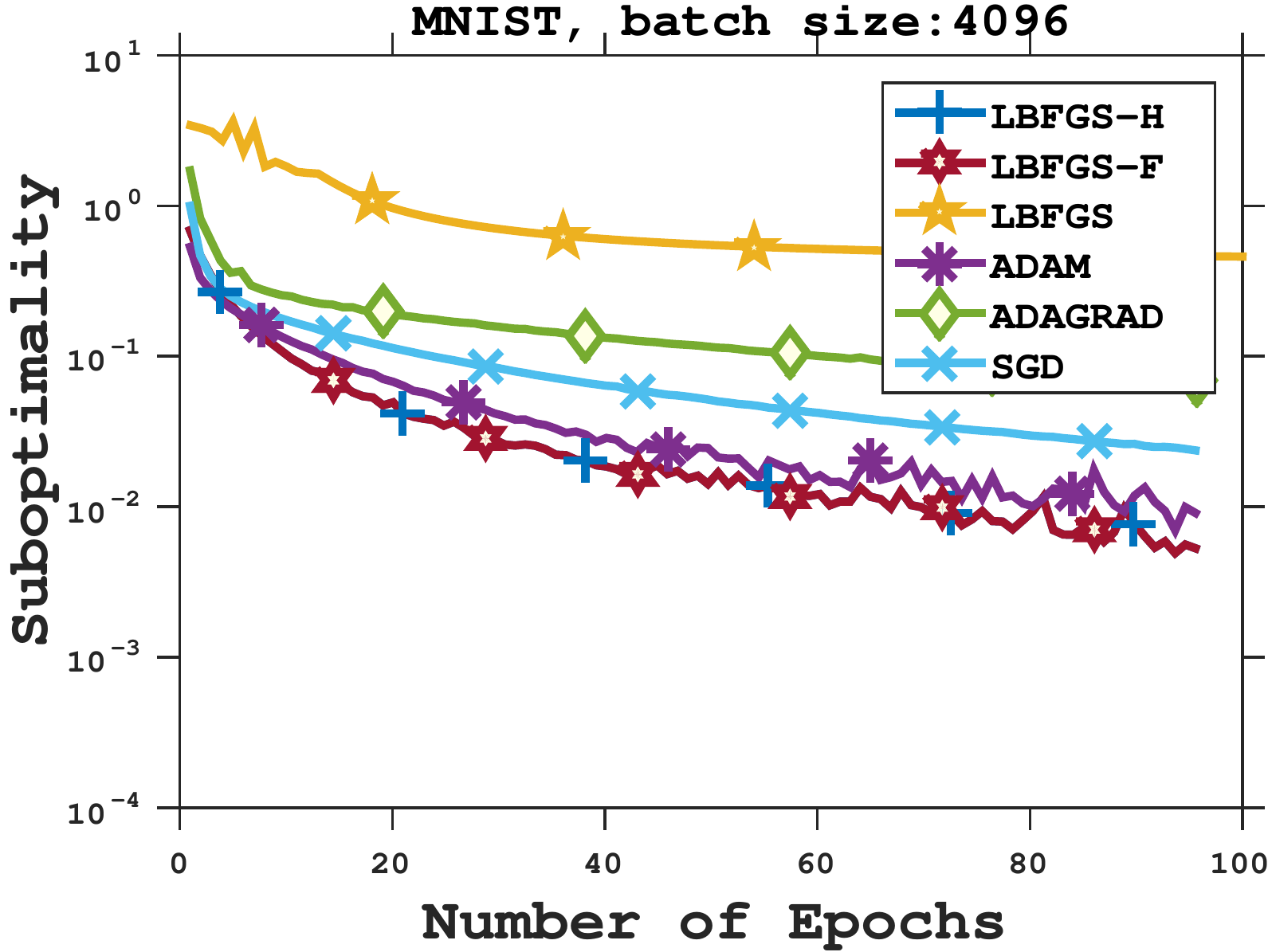,width=0.24\textwidth} 
  
 \epsfig{file=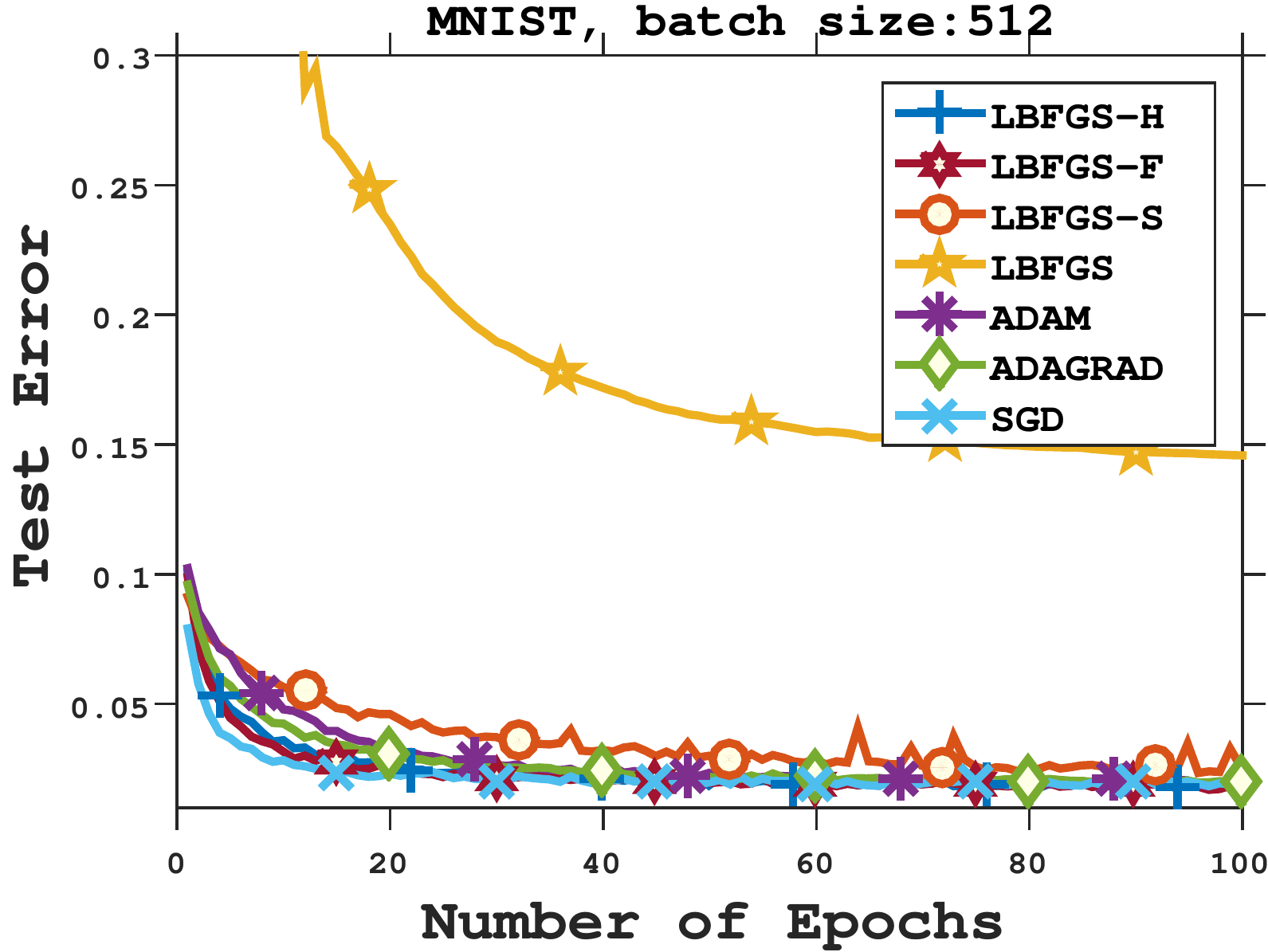,width=0.24\textwidth}
   \epsfig{file=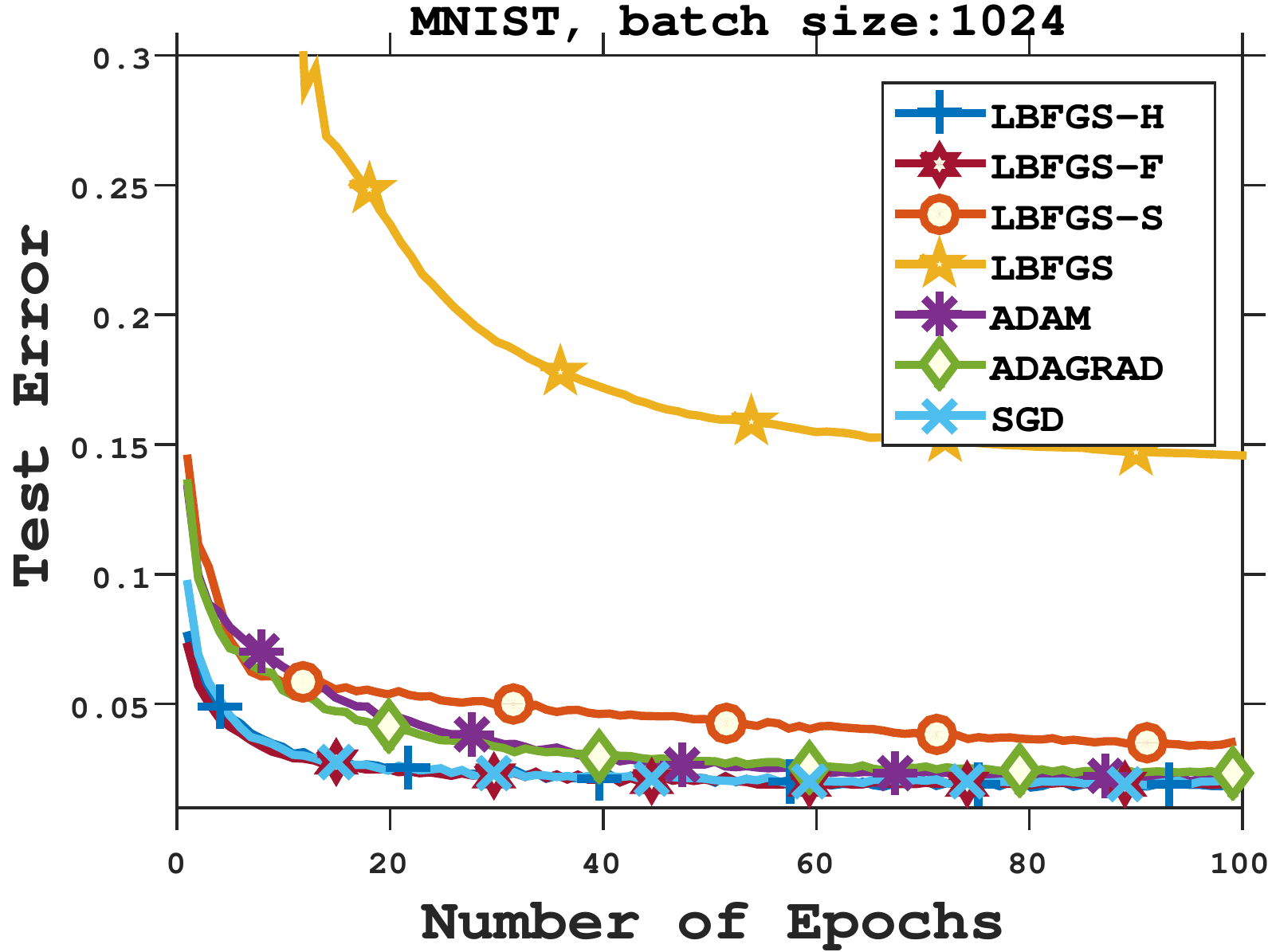,width=0.24\textwidth} 
    \epsfig{file=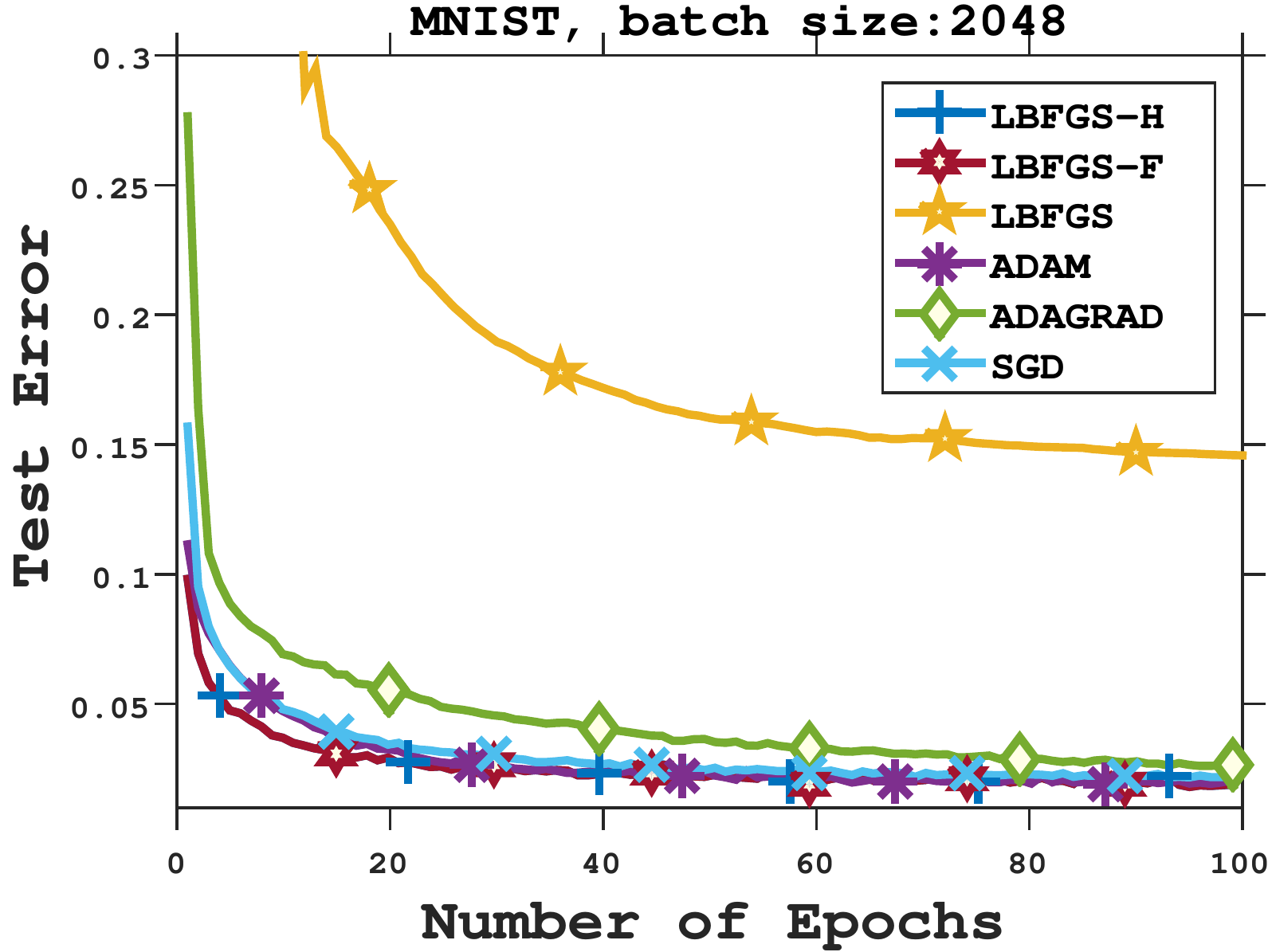,width=0.24\textwidth}
     \epsfig{file=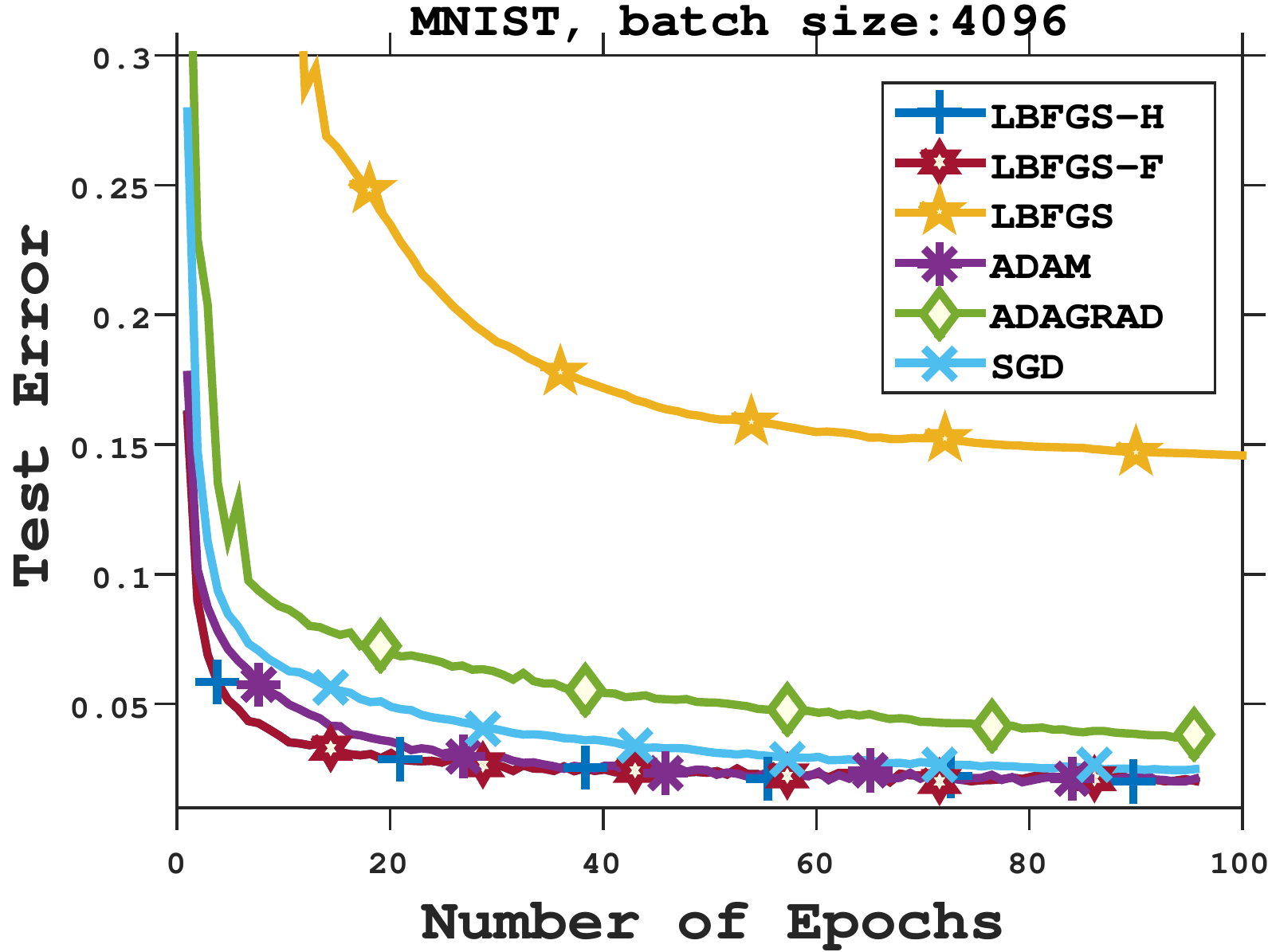,width=0.24\textwidth}
     
    \epsfig{file=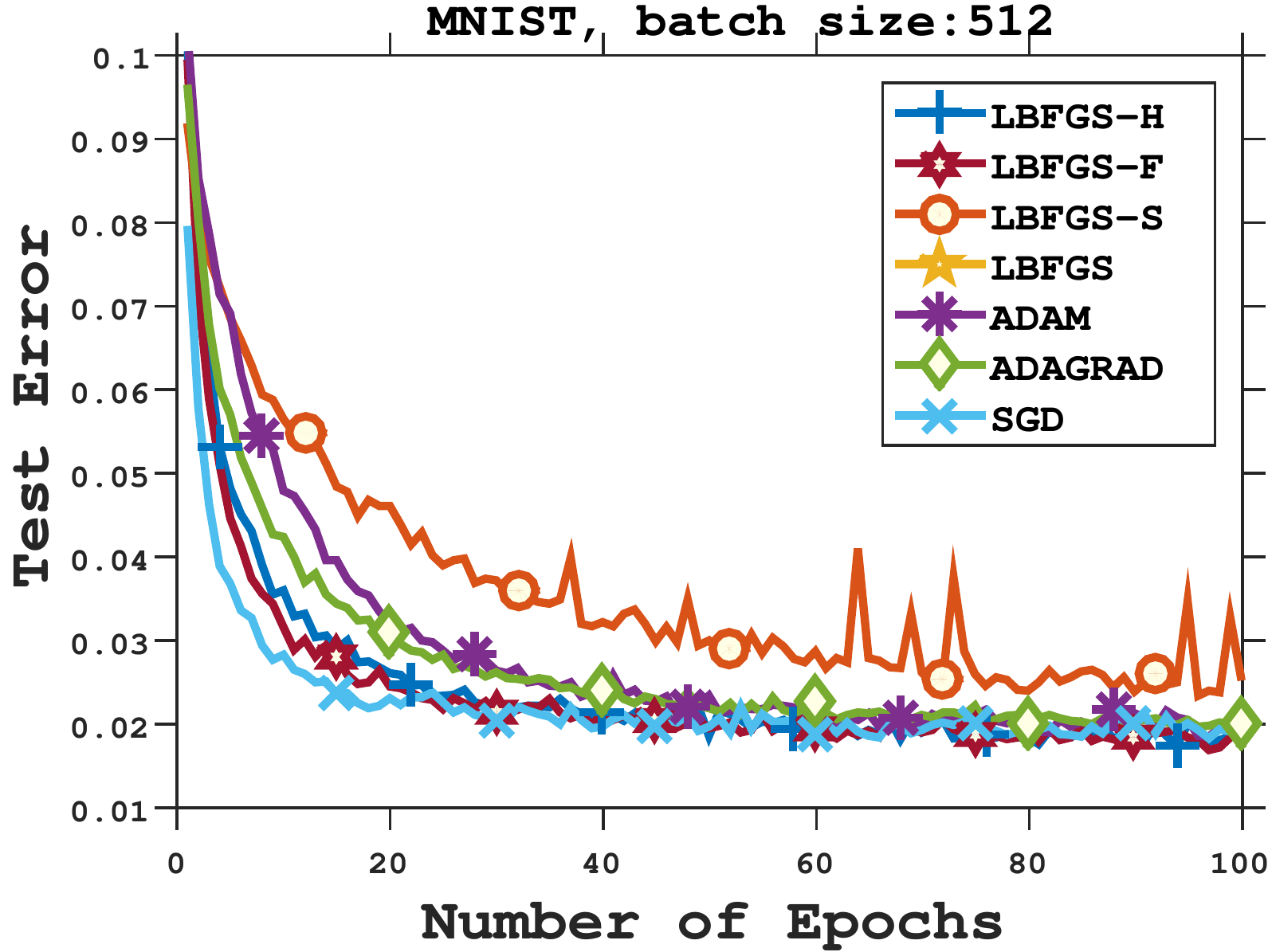,width=0.24\textwidth}
   \epsfig{file=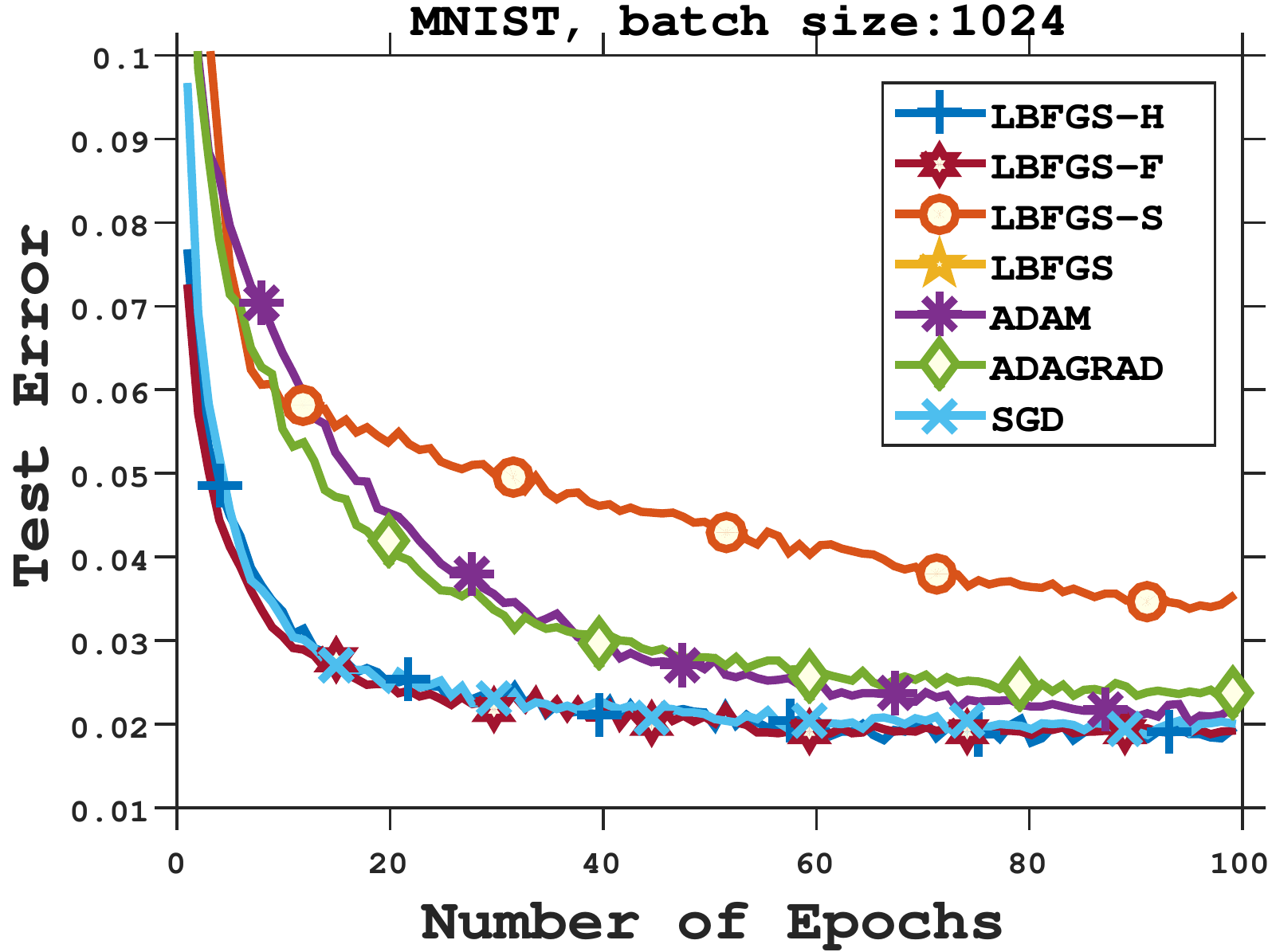,width=0.24\textwidth} 
    \epsfig{file=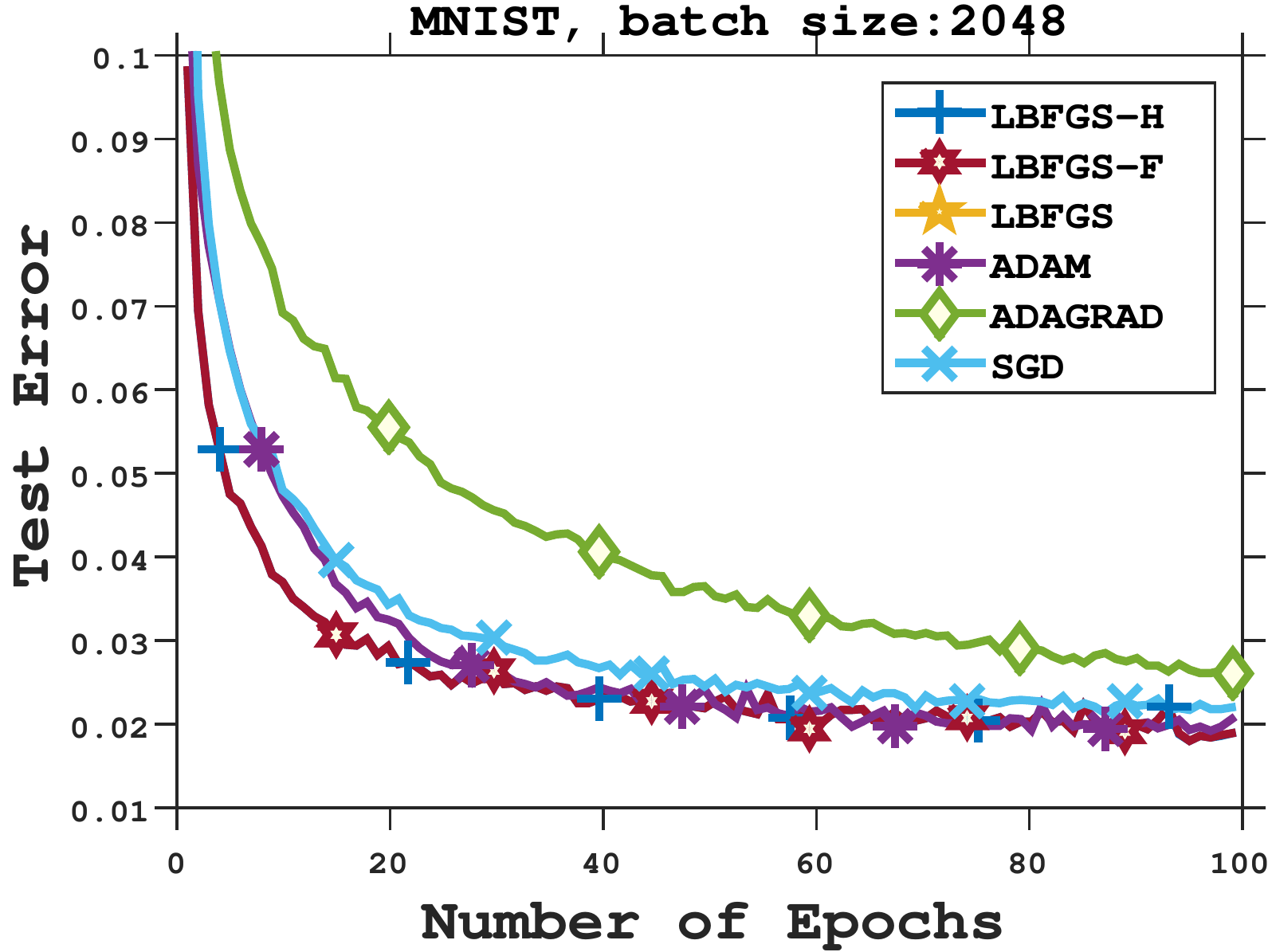,width=0.24\textwidth}
     \epsfig{file=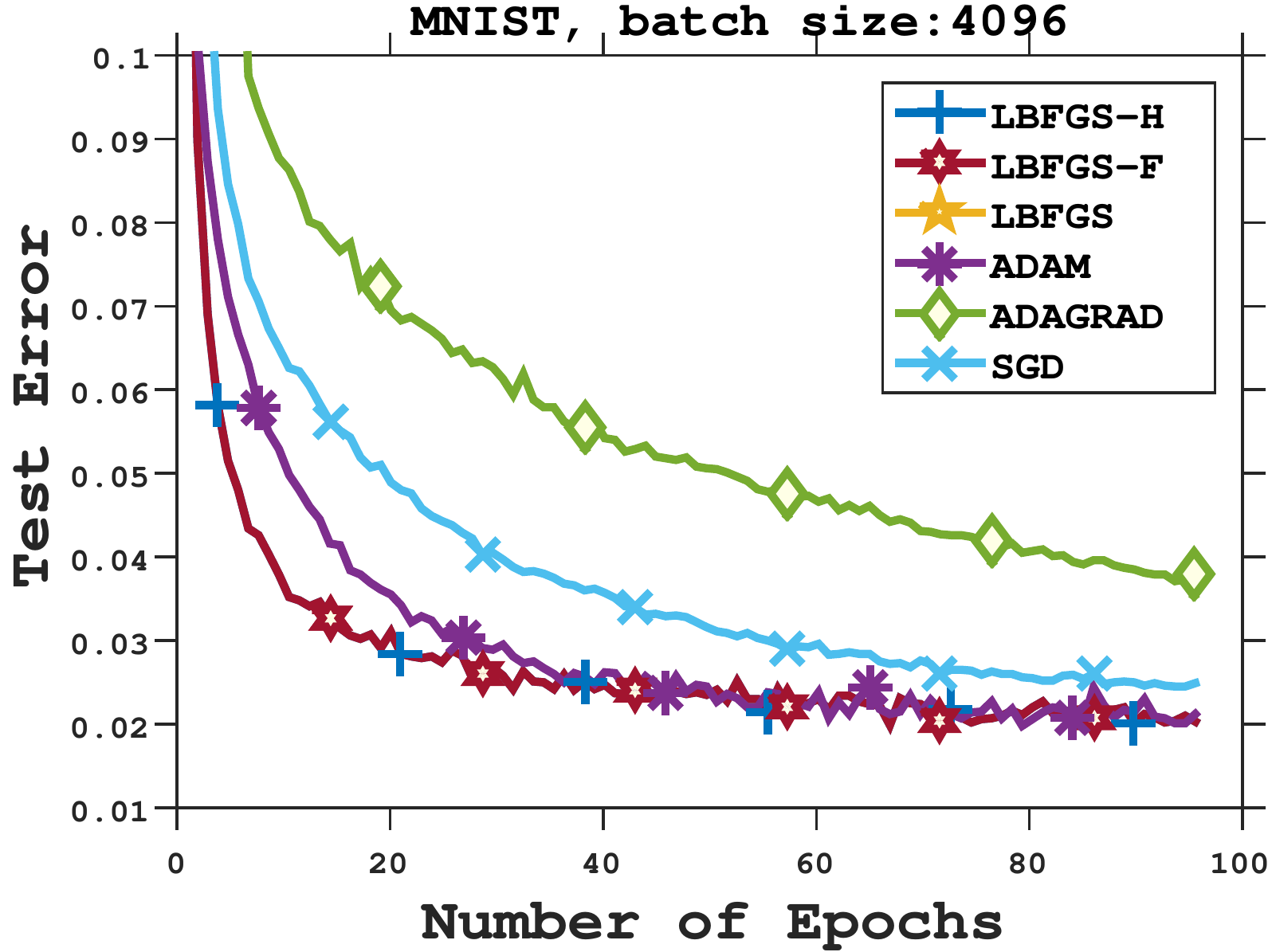,width=0.24\textwidth}
 
 \caption{\footnotesize Comparisons of training loss (top 2 rows), sub-optimality (middle row) and test errors (bottom 2 rows) from different algorithms with batch sizes 512, 1024, 2048 on \emph{MNIST}, nonconvex, neural network with 1 hidden layer.}
   \label{fig:add6}
 \end{figure}

  Figure~\ref{fig:add6} shows the performance of different algorithms on the same experiment with large batch sizes $b=512, 1024, 2048, 4096$. The convergence of LBFGS-H (LBFGS-F) slows down a little but still outperforms that of the other methods, while ADAGRAD and SGD slow down obviously with the increasing batch size. The performance of LBFGS-H approaches that of LBFGS-F with large batch sizes. Note that we did not show performance for LBFGS-S for $b=2048, 4096$ but the trends suggest the performance getting worse.

  \newpage
\subsection{Results on a convolutional neural network -- LeNet-5, \emph{MNIST}}

\begin{figure}[h]
\centering
   \epsfig{file=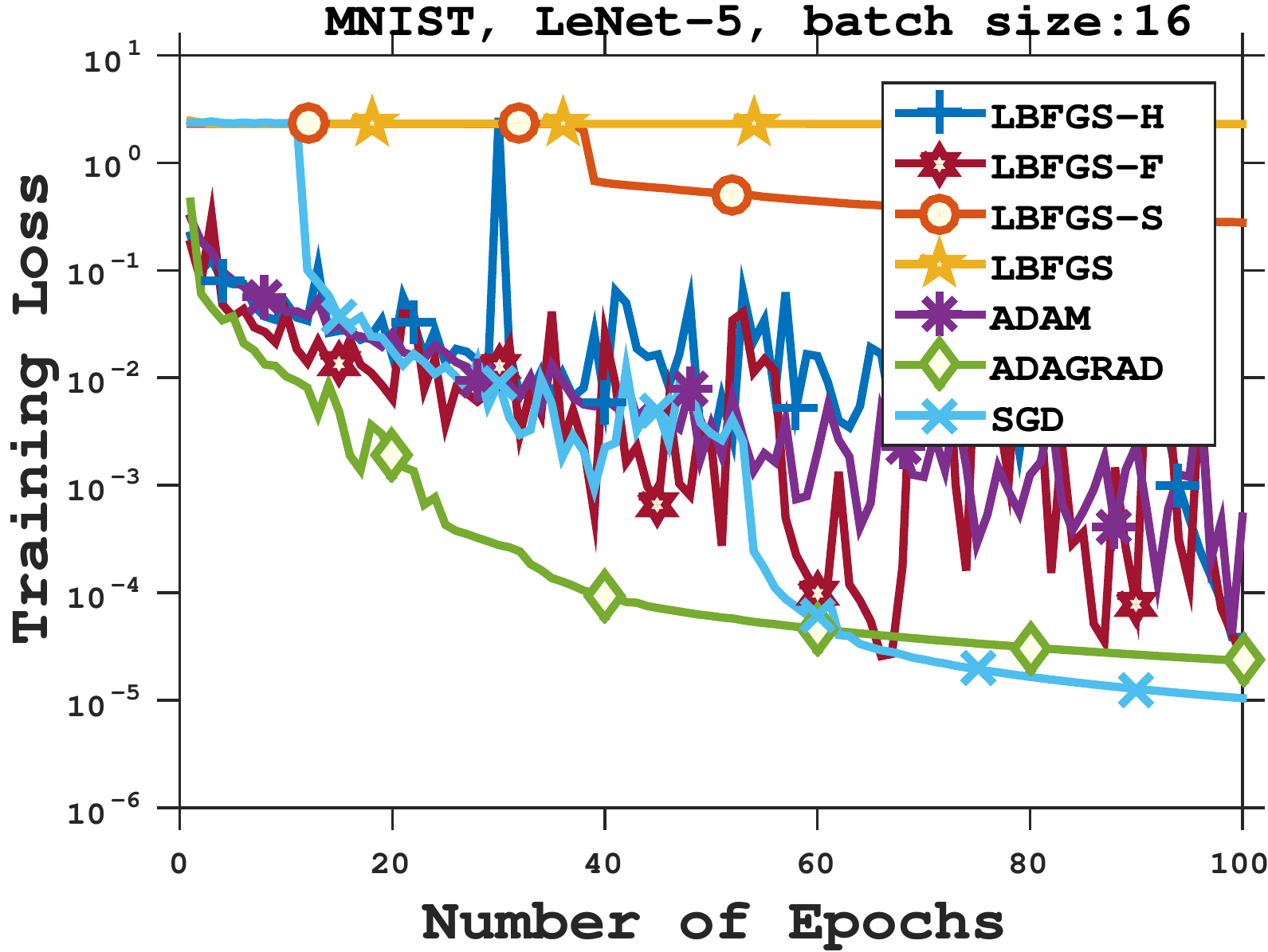,width=0.32\textwidth}
  \epsfig{file=Figs/mnistL_loss64_2_v2.eps,width=0.32\textwidth} 
  \epsfig{file=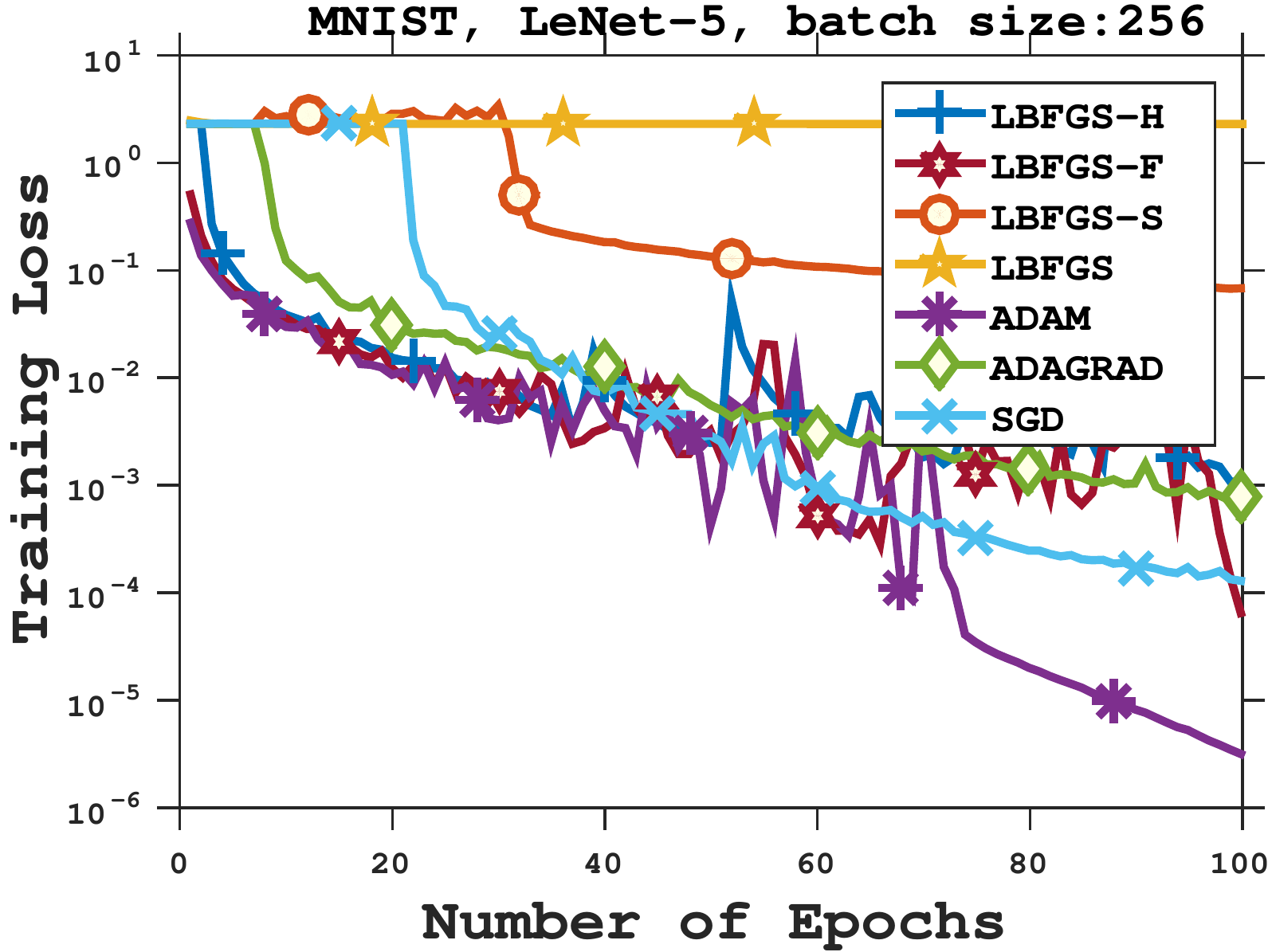,width=0.32\textwidth} 

   \epsfig{file=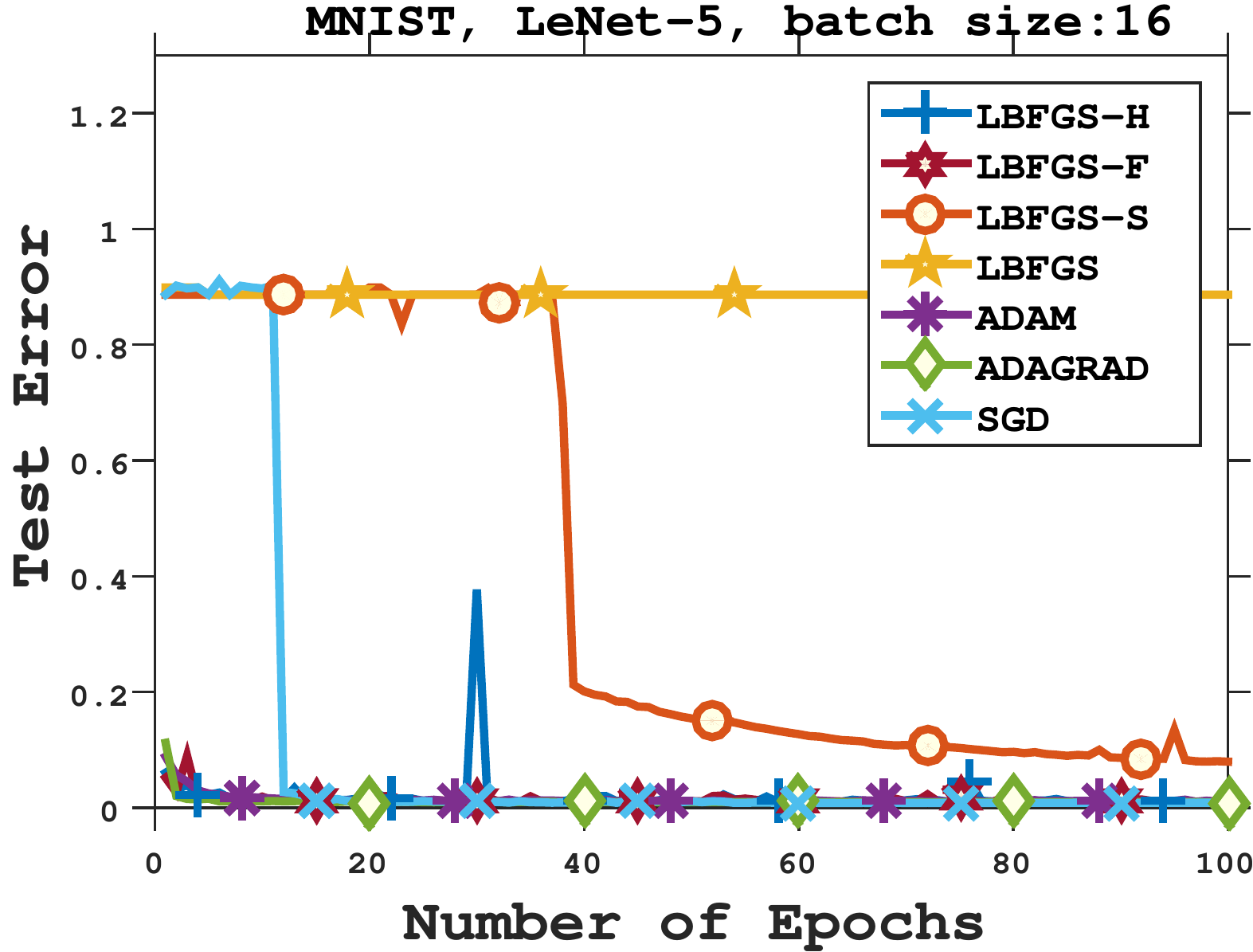,width=0.32\textwidth}
   \epsfig{file=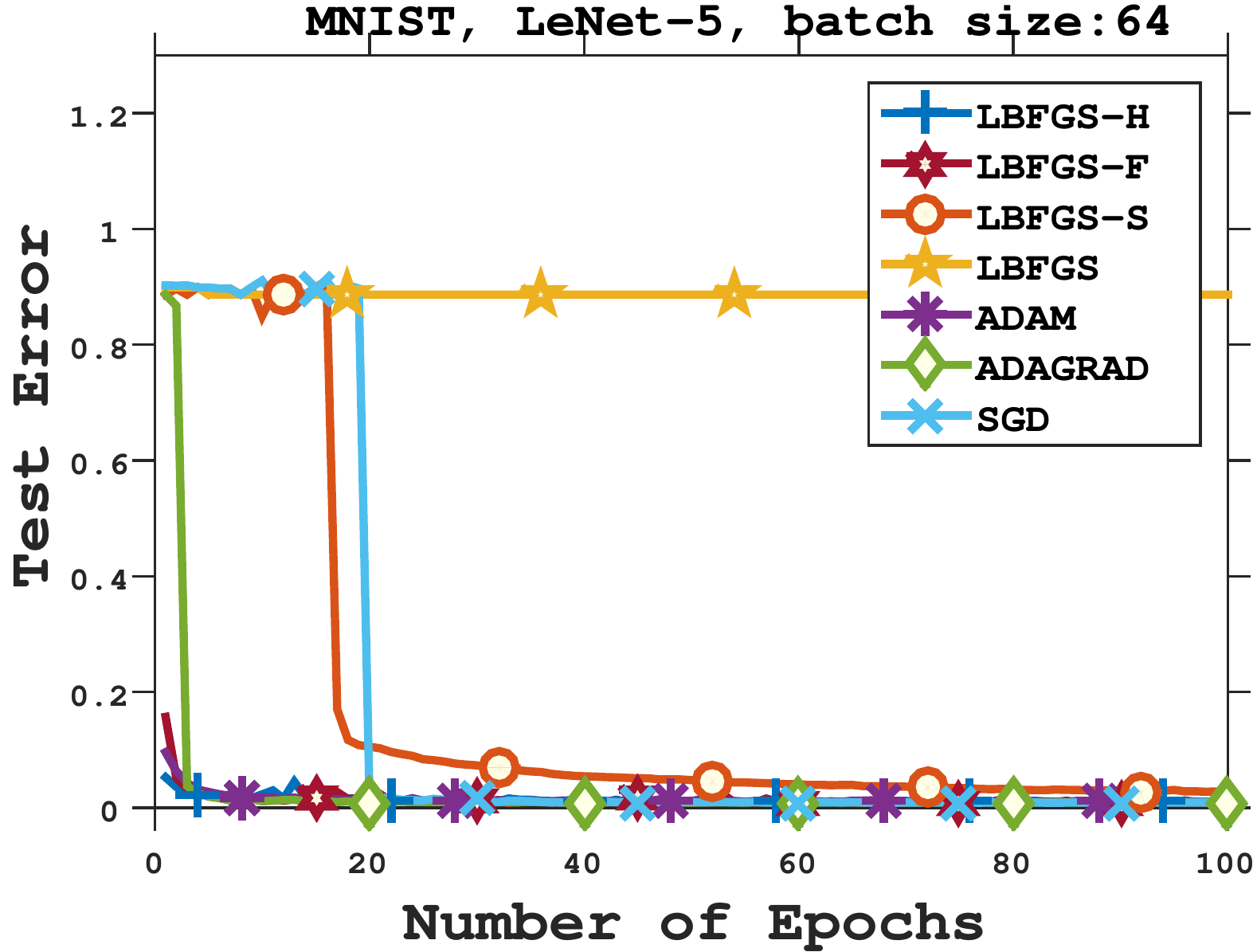,width=0.32\textwidth} 
    \epsfig{file=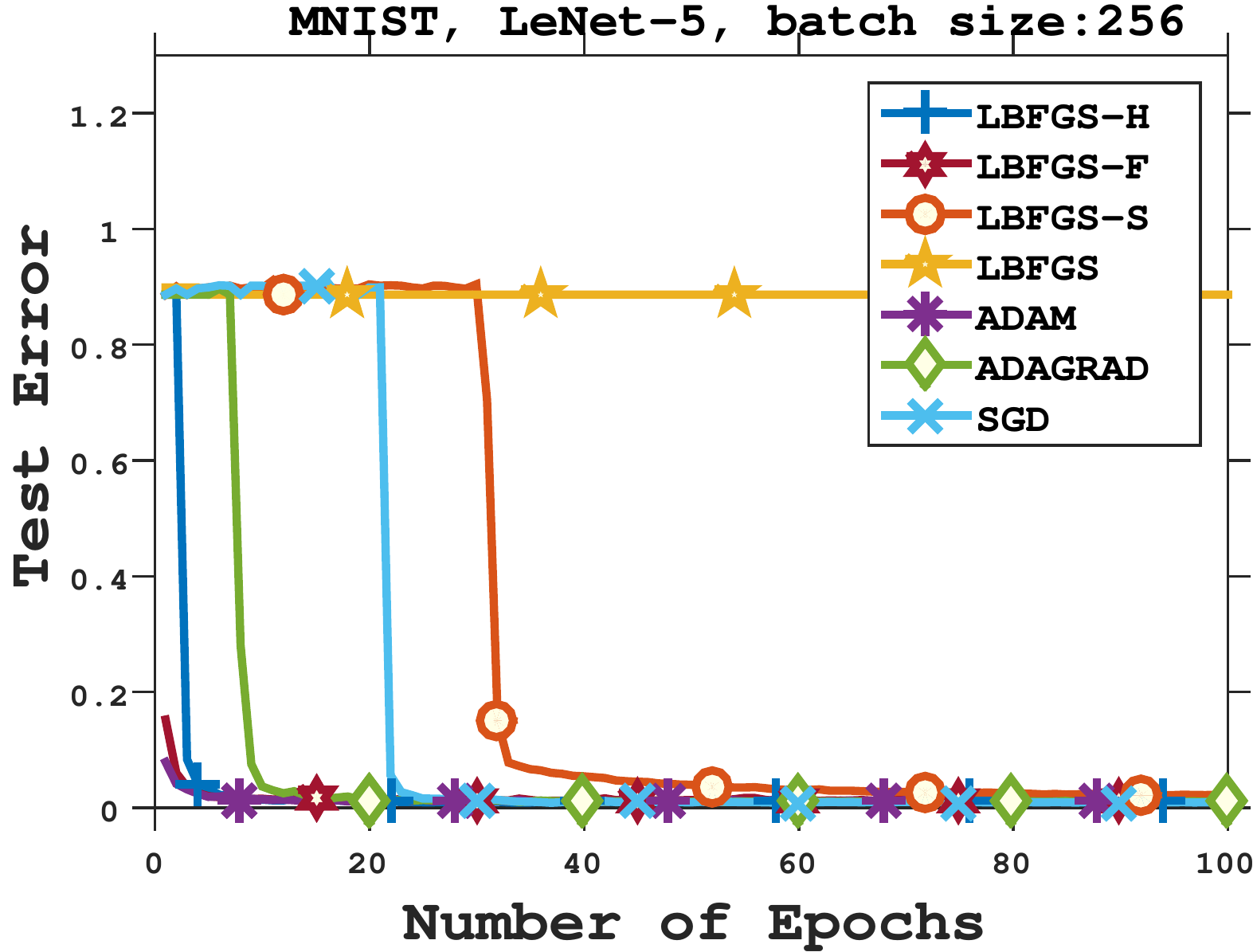,width=0.32\textwidth}
    \epsfig{file=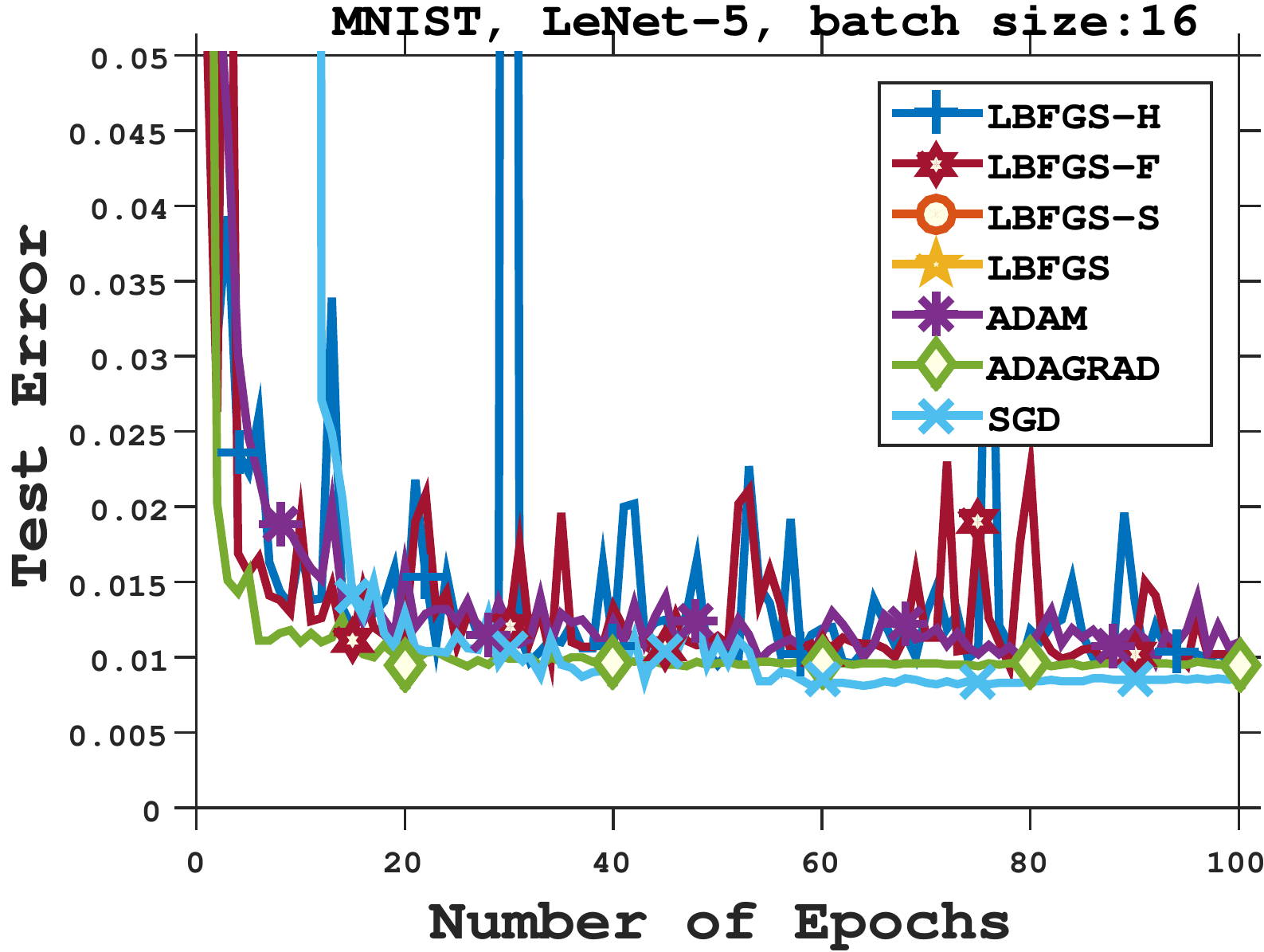,width=0.32\textwidth}
   \epsfig{file=Figs/mnistL_error64closer_v2.eps,width=0.32\textwidth} 
    \epsfig{file=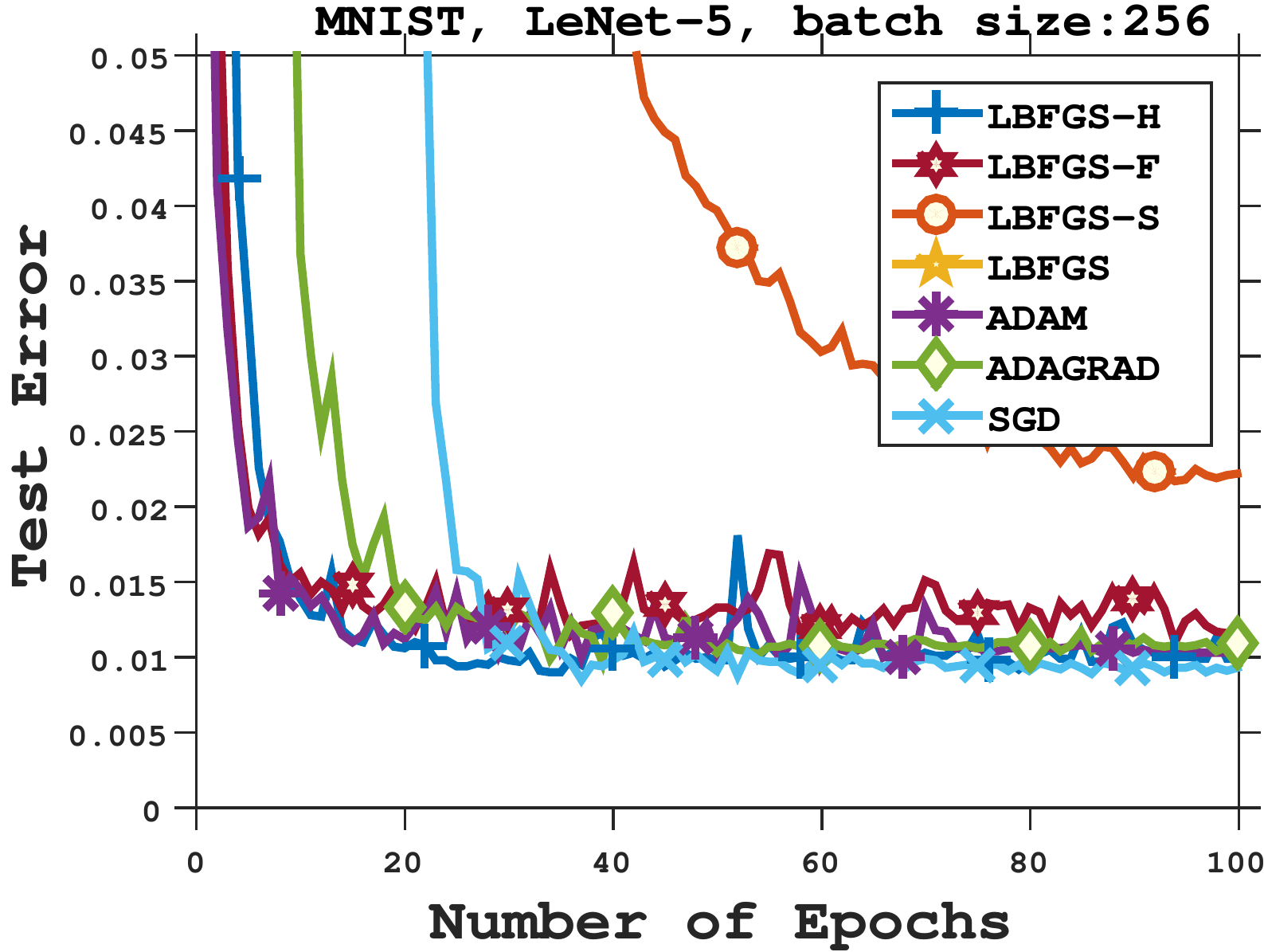,width=0.32\textwidth}
 
 \caption{\footnotesize Comparisons of training loss (top) and test errors (middle and bottom rows) from different algorithms with batch sizes 16, 64, 256 on \emph{mnist}, nonconvex, LeNet-5 (a convolutional neural network).}
   \label{fig:add7}
 \end{figure}

\begin{figure}[h]
\centering
%
  
   \epsfig{file=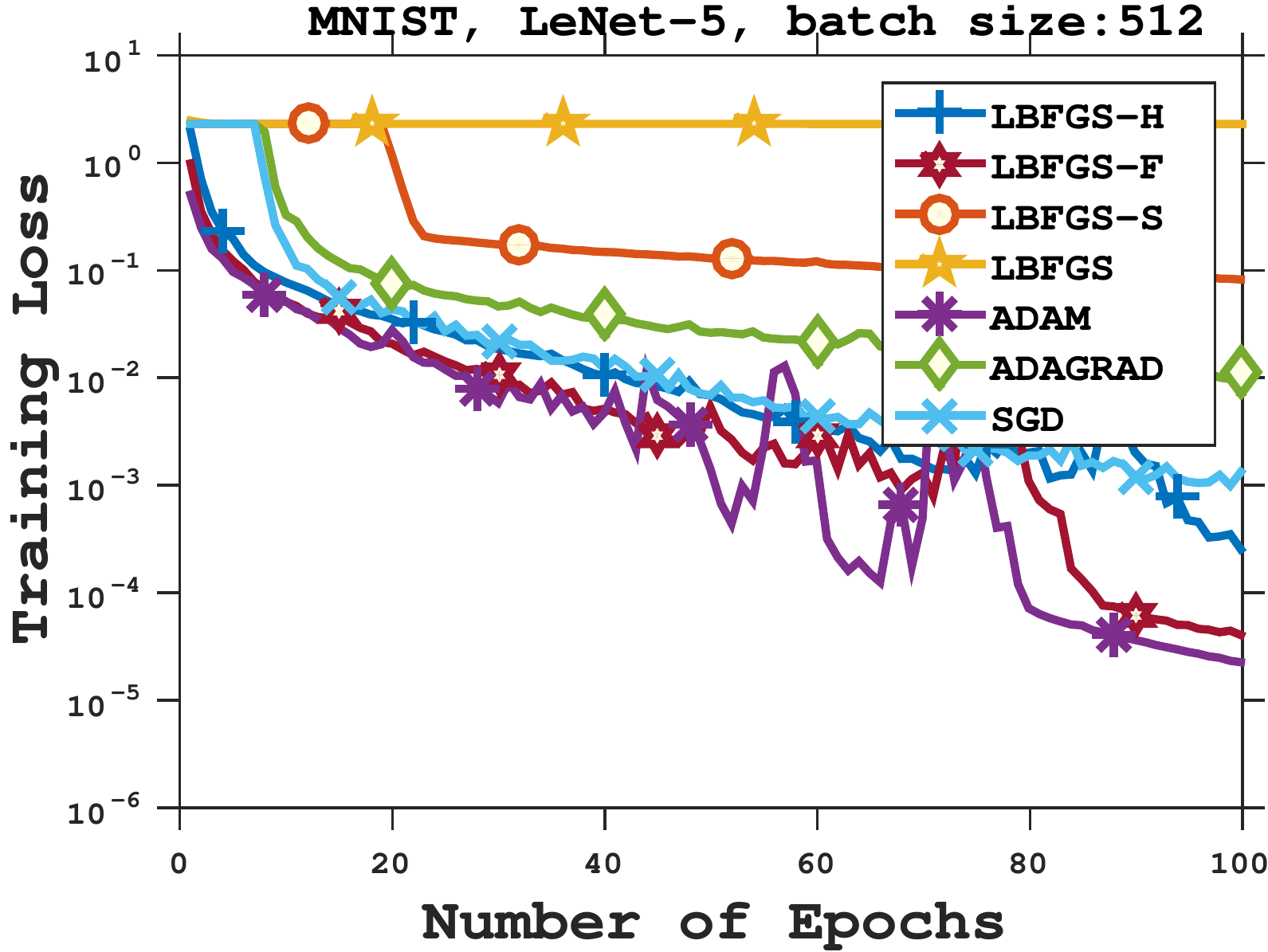,width=0.32\textwidth}
  \epsfig{file=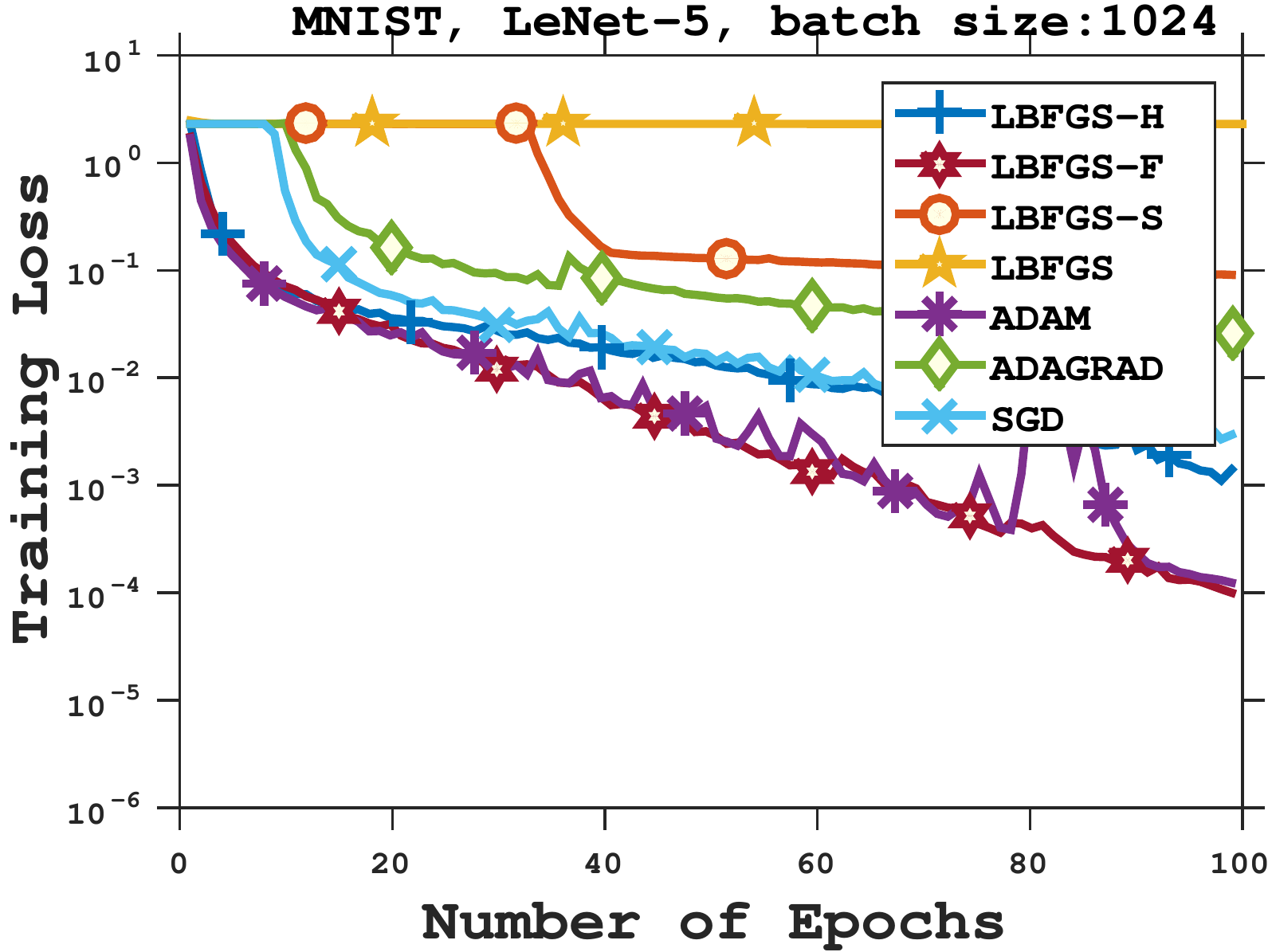,width=0.32\textwidth} 

   \epsfig{file=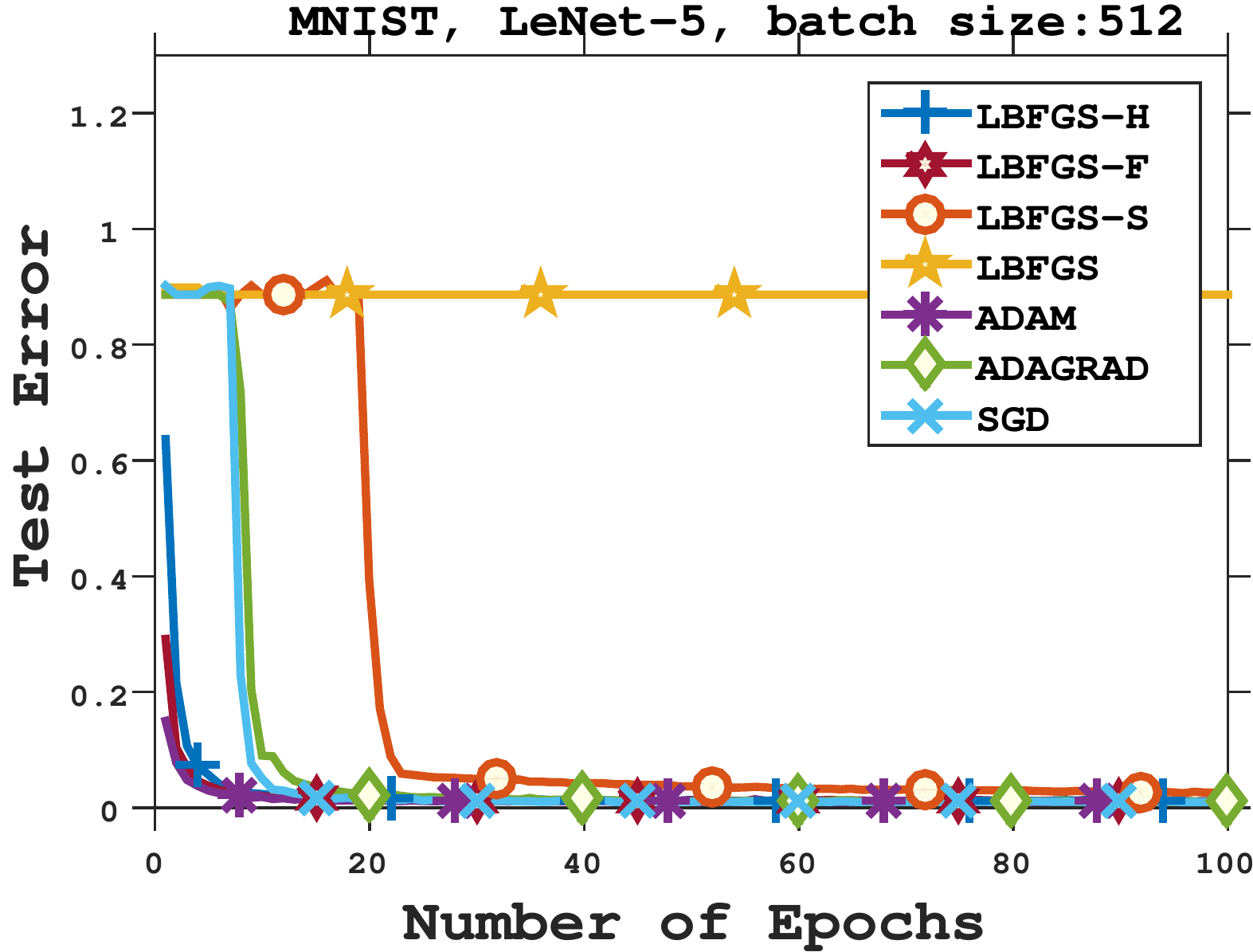,width=0.32\textwidth}
   \epsfig{file=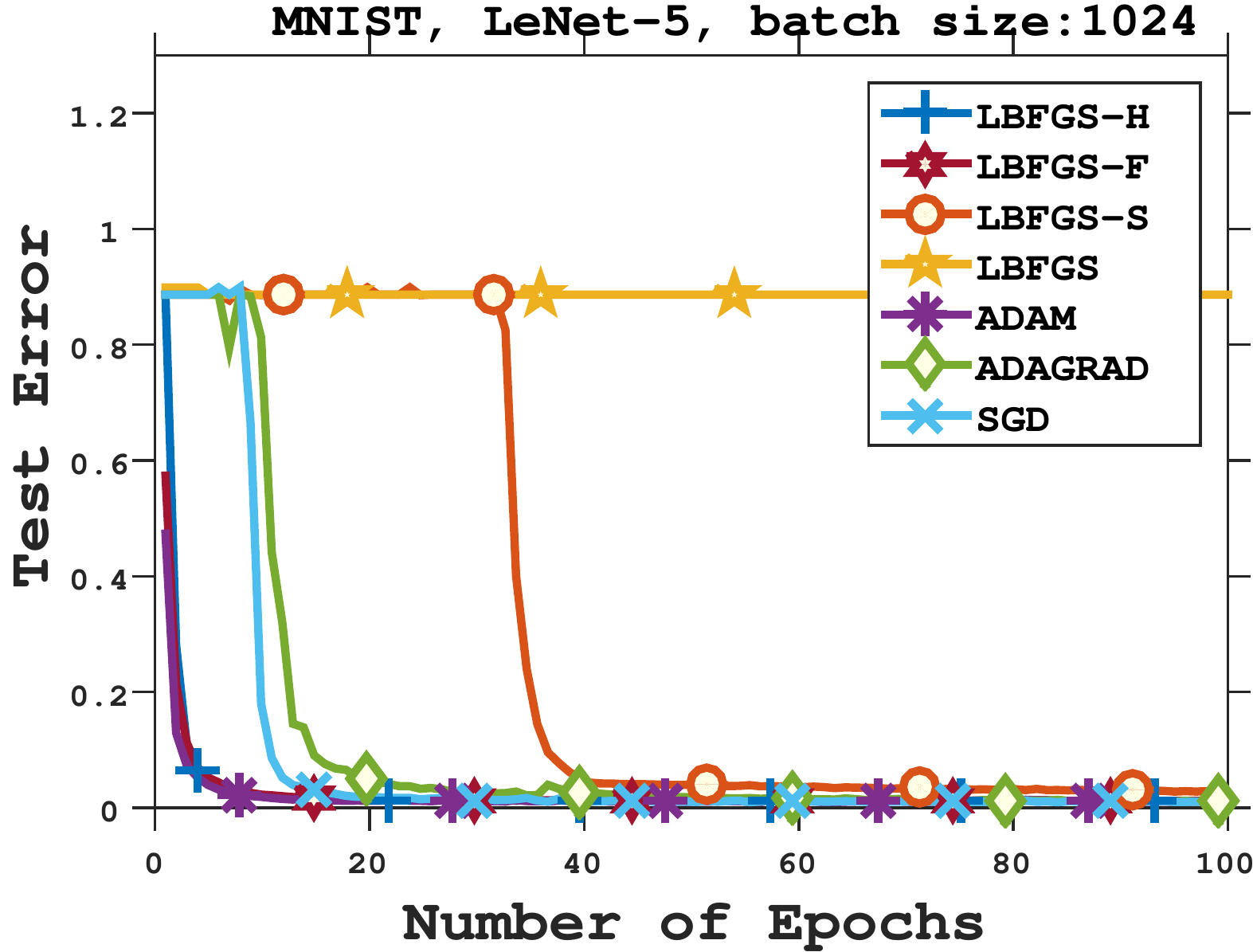,width=0.32\textwidth} 
 
    \epsfig{file=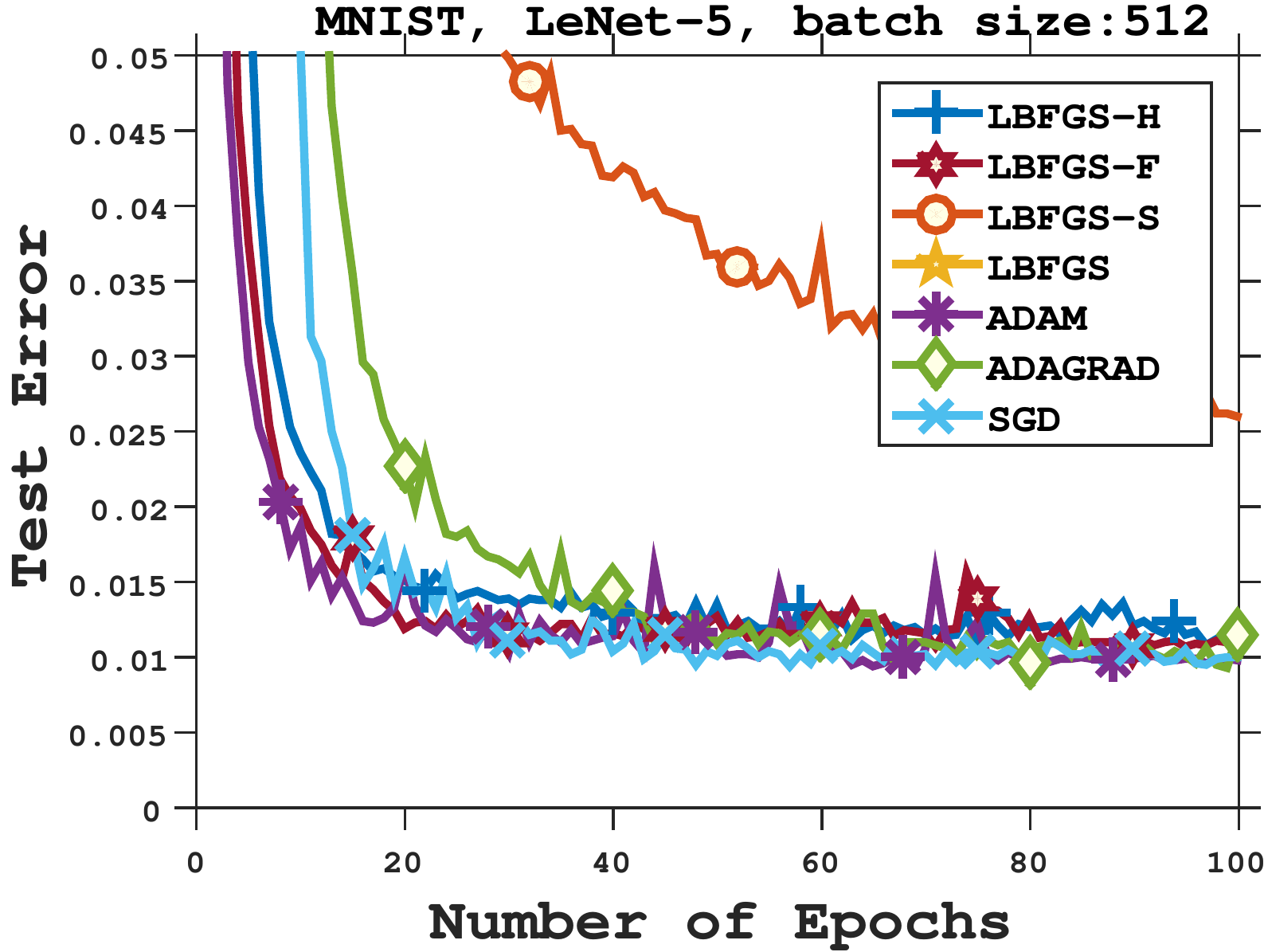,width=0.32\textwidth}
   \epsfig{file=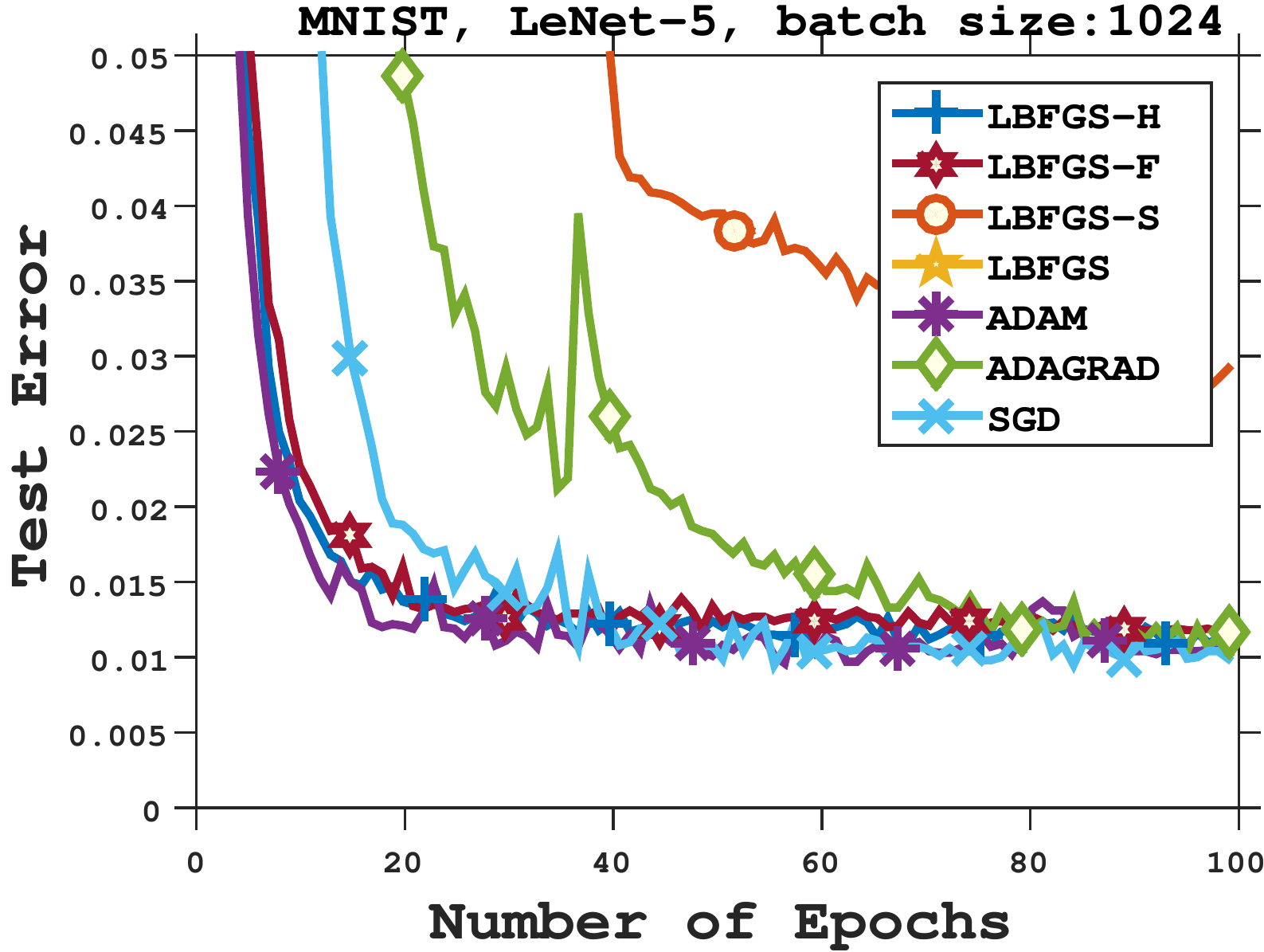,width=0.32\textwidth} 
 
 \caption{\footnotesize Comparisons of training loss (top) and test errors (middle and bottom rows) from different algorithms with batch sizes 512, 1024 on \emph{mnist}, nonconvex, LeNet-5 (a convolutional neural network).}
   \label{fig:add8}
 \end{figure}

 The fourth experiment is conducted for a convolutional neural network LeNet-5 with on \emph{MNIST}. The bottom row in Figure~\ref{fig:add7} is just a zoom-in version of the middle row on the test errors. Combining with the results in Figure~\ref{fig:add8}, the complicated structure of LeNet-5 amplifies the effects of different algorithms. In details, LBFGS-S gets stuck at the beginning and converges slowly while SGD and ADAGRAD continues to worsen much more as the batch size increases in this example. However, ADAM and LBFGS-F outperforms the others apparently with large batch sizes, e.g. $b=512, 1024$.

 \newpage
 
 \newpage

%
%
%

\end{document}